\documentclass[final]{clv2025} % single-spaced
\jvol{vv}
\jnum{nn}
\jyear{yyyy}
\dochead{Long Paper} % or Short Paper, Survey, etc.
\runningtitle{LLM Analysis of 150+ Years of German Parliamentary Debates on Migration}
\runningauthor{Kostikova et al.}

\pageonefooter{Action editor: Saif M. Mohammad. Submission received: 1 April 2026; revised version received: 27 August 2026; accepted for publication: 18 October 2026.}

\usepackage{hyperref}
\usepackage{placeins}
\usepackage[T1]{fontenc}
\usepackage[utf8]{inputenc}
\usepackage[main=english]{babel}
\usepackage{amsmath, amsfonts}
\usepackage{cleveref}
\usepackage[nopatch]{microtype}
\usepackage{booktabs}
\usepackage{graphicx}
\usepackage{subcaption}
\usepackage{tabularx}
\usepackage{multirow}
\usepackage{makecell}
\usepackage{enumitem}
\usepackage[most]{tcolorbox}
\usepackage[table]{xcolor}
\usepackage{csquotes}
\usepackage{lipsum}
\usepackage{xcolor}
\usepackage[size=tiny]{todonotes}
\usepackage{xparse}
\usepackage[most]{tcolorbox}
\usepackage{soul}
\usepackage{float}
\newcounter{examplectr}
\renewcommand{\theexamplectr}{\arabic{examplectr}}

\newcommand{\examplerow}[5]{%
  \refstepcounter{examplectr}%
  (\theexamplectr)\label{ex:#1} \textbf{#2}\newline{}#3
  & #4
  & #5 \\
}

\newcounter{errorexamplectr}
\renewcommand{\theerrorexamplectr}{\arabic{errorexamplectr}}

\newcommand{\errorexamplehead}[4]{%
  \multicolumn{2}{p{\dimexpr0.96\textwidth+2\tabcolsep\relax}}{%
    \vspace{-0.8ex}%
    \raggedright\arraybackslash\ignorespaces
    \refstepcounter{errorexamplectr}\label{errex:#1}%
    \textit{Example (\theerrorexamplectr)}: #2 \textbf{Human reference:} #3 #4%
  }\\%[-0.4em]
}

\newcommand{\exref}[1]{Example~\ref{ex:#1}}
\newcommand{\exreftwo}[2]{Examples~\ref{ex:#1} and~\ref{ex:#2}}

\newcommand{\errexref}[1]{Example~\ref{errex:#1}}
\newcommand{\errexreftwo}[2]{Examples~\ref{errex:#1} and~\ref{errex:#2}}
\newcommand{\errexrefthree}[3]{Examples~\ref{errex:#1}, \ref{errex:#2}, and~\ref{errex:#3}}

\definecolor{bestcolor}{RGB}{199, 233, 192}    % soft green
\definecolor{secondcolor}{RGB}{255, 242, 204}  % soft yellow
\newcommand{\dimrow}{\color{gray}}

\definecolor{pastelgreen}{RGB}{230,245,230}
\definecolor{pastelred}{RGB}{248,220,220}
\definecolor{pastelgrey}{RGB}{220,220,220}
\definecolor{pastellila}{RGB}{235,228,245}

\newcommand{\justgrey}[1]{%
  {\sethlcolor{pastelgrey}\hl{#1}}%
}

\newcommand{\solidarity}[1]{%
  {\setlength{\fboxsep}{0.5pt}\colorbox{pastelgreen}{\strut #1}}%
}
\newcommand{\antisolidarity}[1]{%
  {\setlength{\fboxsep}{0.5pt}\colorbox{pastelred}{\strut #1}}%
}
\newcommand{\none}[1]{%
  {\setlength{\fboxsep}{0.5pt}\colorbox{pastelgrey}{\strut #1}}%
}
\newcommand{\mixed}[1]{%
  {\setlength{\fboxsep}{0.5pt}\colorbox{pastellila}{\strut #1}}%
}

\newtcolorbox{guidelinebox}{
  enhanced,
  breakable,
  colback=gray!8,
  colframe=gray!60,
  boxrule=0.5pt,
  arc=2pt,
  left=8pt,
  right=8pt,
  top=8pt,
  bottom=8pt
}

\newtcbtheorem[number within=section]{definitionbox}{Box}%
{enhanced,
 breakable,
 colback=gray!4,
 colframe=gray!50,
 coltitle=black,
 boxrule=0.4pt,
 fontupper=\small,
 width=0.95\textwidth}{box}

\crefname{section}{section}{sections}
\Crefname{section}{Section}{Sections}
\crefname{subsection}{section}{sections}
\Crefname{subsection}{Section}{Sections}
\crefname{subsubsection}{section}{sections}
\Crefname{subsubsection}{Section}{Sections}
\crefname{tcb@cnt@definitionbox}{box}{boxes}
\Crefname{tcb@cnt@definitionbox}{Box}{Boxes}

\definecolor{akcolor}{RGB}{0,85,170}
\definecolor{bpcolor}{RGB}{140,0,191}
\newcommand{\aknew}[1]{\textcolor{black}{#1}} %akcolor
\newcommand{\ak}[1]{\textcolor{black}{#1}} %akcolor
\newcommand{\bp}[1]{\textcolor{black}{#1}} %bpcolor
\NewDocumentCommand\de{m g}{\IfNoValueTF{#2}{\textit{#1}}{\textit{#1} (\textit{en.}:~#2)}} 
\newcommand{\Frau}[0]{\texttt{Woman}}
\newcommand{\Migrant}[0]{\texttt{Migrant}}

\begin{document}

% === Title and author info ===
\title{LLM Analysis of 150+ Years of German Parliamentary Debates on Migration Reveals a Shift from Post-War Solidarity to Anti-Solidarity in the Last Decade}

\author{Aida Kostikova\thanks{Corresponding author}$^{1}$, Ole Pütz$^{1}$, Steffen Eger$^{2}$, Olga Sabelfeld$^{1}$, Benjamin Paassen$^{1}$}

\affilblock{
    \affil{Bielefeld University\\\quad \email{aida.kostikova@uni-bielefeld.de}}
    \affil{University of Technology Nuremberg}
}

\maketitle

\begin{abstract}
Migration has been a core topic in German political debate, from %the postwar displacement of millions of expellees 
postwar expellee displacement to labor migration and recent refugee movements. %Studying political speech across such wide-ranging phenomena in depth 
Large-scale analysis of such political discourse has traditionally required extensive manual annotation, limiting coverage. %to small subsets of the data. 
Large language models (LLMs) offer a %potential way to overcome this constraint. 
scalable alternative. Using a theory-driven annotation scheme, we examine how well LLMs annotate subtypes of solidarity and anti-solidarity in German parliamentary debates and whether the resulting labels support valid downstream inference. We first %provide a comprehensive evaluation of 
evaluate multiple LLMs %analyzing the effects of 
across model size, prompting strategies, fine-tuning, historical versus contemporary data, and systematic errors. %patterns. %We find that 
The strongest models, especially GPT-5 and gpt-oss-120B, achieve %human-level agreement on this task, 
macro-F1 scores comparable to human %leave-one-out 
agreement, although their %errors remain systematic and bias downstream results. 
systematic errors can bias downstream results. %To address this issue, 
We therefore combine soft-label model outputs with Design-based Supervised Learning (DSL) to reduce bias in long-term trend estimates. %This combination improves recovery of human-derived aggregate trends, with the largest benefits for fine-grained estimates. 
Beyond the methodological evaluation, we interpret the resulting annotations from a social-scientific perspective across the Reichstag (1867-1933), West German Bundestag (1949-1990) and Re-Unified German Bundestag (1990-2025) corpus to trace trends in solidarity and anti-solidarity toward migrants, %in postwar and contemporary Germany. 
with detailed analysis %focusing on 
of postwar Germany (1949-1957) and contemporary Germany (2009-2025). %Our approach shows 
We find relatively high levels of solidarity in the postwar period, especially in group-based and compassionate forms, and a marked rise in anti-solidarity since 2015. %framed through exclusion, undeservingness, and resource burden. 
We argue that LLMs can support large-scale social-scientific text analysis, %but only when their outputs are rigorously validated and statistically corrected.
but their outputs require rigorous validation and, where systematic errors are present, statistical correction.
\end{abstract}

% === Main content ===
\section{Introduction}\label{sec:introduction}

Migration has shaped societies throughout history and remains a major social and political issue in the 21st century
%Migration is a defining feature of the 21st century% and is recognized in the 2030 Agenda for Sustainable Development 
\citep{wilkinson2022climate}. %adger2024migration}. 
Economic imbalances, demographic divergence, and climate change %make migration %increasingly unavoidable, 
increase pressures for migration, especially from the Global South to the Global North \citep{worldbank2023, almulhim2024climate, %korsi2022we, 
marois2020population}. %Yet, 
If governed responsibly, it can %also promote 
contribute to prosperity and support sustainable development \citep{%worldbank2023, 
gavonel2021migration, adger2024migration}.

Within Europe, Germany is a major migration destination \citep{eurostat2024migration}, and migration has profoundly shaped its development, from post-Second World War expellee resettlement 
to labor migration in the 1950s-1970s and refugee movements in %the past decade 
the 2010s-2020s \citep{kaya2017inclusion, frohlich2025migration}. 

As the EU's largest economy, %it 
Germany remains an attractive destination, while demographic decline and skilled-labor shortages make continued immigration economically beneficial \citep{bertoli2016crisis, smith2025demographic, eissel2021economic, angenendt2023germany}. At the same time, anti-migration stances have become more visible in political debate, often in relation to perceived risks such as pressure on public services, labor-market and resource competition, and the rise of right-wing populism \citep{Frey2020, gessler2023politicization, messing2021anti, esguerra2023accommodation}. This raises %a central 
an important substantive question: how does German politics %speak about 
discuss migration, and to what extent does it express solidarity or anti-solidarity toward migrants?

Answering such questions requires systematic analysis of political discourse across long time spans to assess continuity and change in migration discourse.
 
Such analysis has traditionally relied on manual textual analysis and qualitative interpretation \citep{randour2020twenty}. 
However, such methods are labor-intensive \citep{bhattacherjee2012social} and, hence, often restricted to selective subsets of data, which risks sampling bias and leaves large corpora underutilized \citep{babalola2021current}. Computational social science (CSS) emerged partly in response to the scale and complexity of modern data, which exceed the analytic reach of traditional approaches \citep{Lazer2009}. More recently, large language models (LLMs) have been introduced into CSS as a means of scaling annotation and automating parts of the research workflow \citep{pavlovic2024effectiveness}. %and being able to adapt to new topics, languages, and contexts \citep{bang2023multitask, wang2024instruction, karjus2025machine}. 
At the same time, their use raises several methodological challenges. (i) Although LLMs show strong performance on tasks such as ideological stance detection \citep{ziems2024can}, it remains unclear whether they can perform more complex, theory-driven sociopolitical annotation tasks. (ii) Even when LLMs appear to perform well, recent work questions whether their outputs can support valid downstream inference or serve as ground-truth labels, given their sensitivity to model choice, prompt design, and configuration, as well as their potential to introduce systematic bias \citep{baumann2025large, gligoric2025unconfident, krsteski2025valid, egami2023using}. 
(iii) These concerns connect to ongoing debate in CSS emphasizing that computational methods and digital data must be tightly integrated with social scientific theory, concepts, and theoretically meaningful research questions \citep{lazer2024future, jungherr2017empiricist, theocharis2021computational, soemer2025social}.

Accordingly, the overarching goal of this study is to assess whether LLMs can support valid, theory-driven analysis of (anti-)solidarity in political discourse at scale. We address this goal through three research questions: (RQ1) How accurately can LLMs perform a complex annotation task focused on (anti-)solidarity in parliamentary text and grounded in a theoretically informed, fine-grained taxonomy? (RQ2) To what extent do LLM-generated annotations support valid downstream inference about temporal trends in (anti-)solidarity? (RQ3) What does the resulting analysis reveal about solidarity and anti-solidarity toward migrants in German parliamentary discourse over time? %particularly in the postwar period and the recent past?

% In a prior conference paper, 
These questions build on our prior conference paper, in which we addressed %challenge (i) 

RQ1 by showing that LLMs can approximate human performance on %a 
this %complex annotation 
task %focused on (anti-)solidarity in parliamentary text and grounded in a theoretically informed, fine-grained taxonomy 
\citep{kostikova-fine}. In this paper, we extend that analysis in three ways: the first two are methodological, and the third is substantive. First, %to address (i) 
we revisit RQ1 more rigorously %, we evaluate 
by evaluating a broader range of LLMs, %compare 
comparing their predictions against a larger set of human reference annotations, and systematically %assess 
assessing the effects of prompts, model size, and historical versus contemporary data. Second, to address %(ii), 
RQ2, we examine whether strong annotation performance is sufficient for valid downstream inference by analyzing systematic model errors, testing ensemble predictions, and applying statistical bias correction \citep{egami2023using}. Third, to address %(iii), 
RQ3, we apply this validated pipeline to the analysis of solidarity and anti-solidarity toward %women and 
migrants in German parliamentary speeches and interpret the resulting trends between 1949 and 1957 and between 2009 and 2025 in light of German political history.

\begin{figure}%[!htb]
  \centering \includegraphics[width=0.9\linewidth]{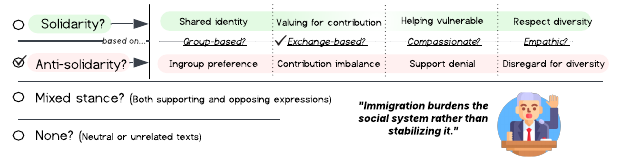}
  \caption{Overview of the annotation scheme adapted from \citet{thijssen2012mechanical}. Statements are first classified at a high level as expressing \emph{solidarity}, \emph{anti-solidarity}, \emph{mixed}, and \emph{none}. \emph{Solidarity} and \emph{anti-solidarity} instances are then further categorized into subtypes based on the underlying rationale: \emph{group-based}, \emph{exchange-based}, \emph{compassionate}, and \emph{empathic}. The example sentence in the figure illustrates exchange-based anti-solidarity.}
  \label{fig:annotation-scheme}
\end{figure}

%We focus on solidarity as a particularly demanding test case for such an approach. 
We focus on solidarity because it is a particularly demanding test case for operationalization and annotation, both for human annotators and for LLMs. Solidarity is an essentially contested concept: it is difficult to define, value-laden, ambiguous, multi-faceted, context-dependent, and open to change over time \citep{kozlowska2024solidarity}. This makes annotation demanding, as coders must map concrete and historically variable political statements onto abstract categories of (anti-)solidarity. %While frequently invoked in political discourse, solidarity has often been treated as a symbolic notion rather than an operational concept, %\citep{kozlowska2024solidarity}, 
%This has limited large-scale empirical research. 
To address this challenge, we approach solidarity as a %socially meaningful 
theoretically informed category and apply solidarity typology of \citet{thijssen2012mechanical}, which offers a structured, annotatable framework. It distinguishes four types of (anti-)solidarity -- \emph{group-based}, \emph{compassionate}, \emph{exchange-based}, and \emph{empathic} (see the annotation scheme in \Cref{fig:annotation-scheme}) -- based on the interplay of structural conditions (e.g., group membership) and intersubjective recognition (e.g., emotional identification). 

We apply this framework to migration because solidarity has become closely associated with both migration and forced displacement in public and academic discourse \citep{drotbohm2022solidaritat,bauder2020solidarity}. Migration debates repeatedly raise questions about belonging, obligations toward vulnerable groups, and the distribution of rights and resources \citep{%drotbohm2022solidaritat,
wallaschek2020discursive,kneuer2025solidarity}, which also correspond closely to the aspects captured by our (anti-)solidarity annotation scheme. 
Germany is a particularly suitable context for such a study since solidarity is deeply rooted in its political vocabulary \citep{fiegle2003solidarite, stjerno2005solidarity, reisz2006solidaritat}, and its multi-party system enables to trace how meanings of (anti-)solidarity vary across parties and over time.

Overall, we find that in our task, the strongest models %approach human agreement, 
achieve macro-F1 scores comparable to human leave-one-out agreement, 
%and in some settings even surpass it, 
with \texttt{GPT-5} %performing best overall 
achieving the highest reported scores and \texttt{gpt-oss-120B} %being the strongest open model. 
showing similarly competitive performance as the strongest open model. 
However, we conclude that %\emph{\textbf{raw model predictions are not sufficient for reliable downstream inference}}: 
\emph{\textbf{strong instance-level performance does not guarantee accurate downstream inference}}: although strong models achieve %near-human agreement 
relatively strong %instance-level 
performance at the instance level, their errors remain structured and correlated, limiting the benefits of simple ensemble methods. At the same time, human agreement itself remains limited, especially at the fine-grained level, indicating substantial label variation. In response, combining soft-label model outputs with statistical correction, such as Design-based Supervised Learning (DSL) \citep{egami2023using}, %achieves more accurate 
improves recovery of human-derived temporal trends, 
with the largest benefits for fine-grained estimates. For example, raw model trends overemphasize \emph{group-based solidarity}, whereas DSL more closely recovers the \emph{compassionate}-dominant pattern observed in human annotations. Substantively, this validated pipeline reveals high levels of migrant-directed \emph{solidarity} in the postwar period, especially forms centered on shared belonging and support for vulnerable groups (\emph{group-based} and \emph{compassionate}), thereby corroborating existing qualitative research on early postwar inclusion politics. For the contemporary period, it reveals a marked shift since 2015 toward \emph{anti-solidarity} in German parliamentary discourse, framed through exclusion, undeservingness, and burdens on collective resources (\emph{group-based}, \emph{compassionate} and \emph{exchange-based}). %}
To our knowledge, this is the first large-scale quantitative evidence of that shift.
\subsection{Related Work}\label{sec:related-work}

Our work builds on and connects the following directions of existing research: (i) sociological and political theories of solidarity and their operationalization in NLP;\footnote{We limit this subsection to solidarity as expressed or constructed in discourse \citep[i.e., \emph{meso-discursive solidarity}, in the terminology of][]{wallaschek2020discursive}, excluding work on individual attitudes toward solidarity (\emph{micro-behavioural}) or on institutionalized solidarity (\emph{macro-structural}).} (ii) %the use of 
NLP and LLMs for political and social science analysis, with a particular focus on annotation; and (iii) valid downstream inference from imperfect machine-generated labels.

\paragraph{Solidarity in %Political 
Discourse and NLP}

%%% 1. THIJSSEN BACKGROUND, RELATED PAPERS ON THE SAME TYPOLOGY %%%
The theoretical foundation for our analysis is the solidarity framework of \citet{thijssen2012mechanical}, which extends Durkheim's distinction between mechanical and organic solidarity by incorporating recognition-based theories of \citet{honneth1996struggle, honneth2014disrespect}. Thijssen distinguishes four subtypes of solidarity---group-based, compassionate, exchange-based, and empathic---applicable to both historical and contemporary discourse.  \citet{thijssen2022s} apply this typology to Belgian party manifestos, showing how parties emphasize different frames in line with ideological divides; later work extends it to voter preferences in Flanders \citep{verheyen2023solidarity} and to party discourse across different welfare regimes \citep{luypaert2024comparative}.

%%% 2. OTHER SOLIDARITY TYPOLOGIES/FRAMEWORKS APPLIED TO DISCOURSE STUDIES %%%

Other work has conceptualized solidarity in discourse through alternative frameworks \citep[e.g.][]{kneuer2022claiming, berrocal2021constructing}, including applications to parliamentary debates \citep{salvati2024debating, sarfo2024discursive, theiss2024narrating},
%%% 2.1. AMONG THOSE IN POINT 2, THESE ARE APPLIED LONGITUDINALLY... WITH THE FOLLOWING METHODS... %%%
as well as longitudinal analyses %of solidarity discourse 
\citep{bartenstein2025variations, arlauskaite2025solidarity}. The empirical analyses above, however, largely rely on \emph{manual} qualitative methods.

%%% 3. AMONG THOSE IN POINT 2.1 AND 2, THESE ARE STUDIED WITH COMP METHODS %%%
%%% 3.1. COMP ANALYSIS: DISCOURSE-NETWORK, DEICTIONARIES... %%%

\emph{Computational approaches} have also been used to analyze %the structure and evolution of 
solidarity discourse. \citet{wallaschek2020discursive, wallaschek2020solidarity} combine human-coded solidarity claims with discourse-network analysis to study German media discourse over time. Other work identifies solidarity-related material through lexical cues, such as explicit mentions of solidarity, and then applies methods such as topic modeling %to trace solidarity themes in social media 
\citep{kajta2025social} and sentiment and network analysis %of solidarity-related Twitter discourse during COVID-19 
%discourse 
\citep{kalaloi2021mediated}. \citet{gao2024reconstructing} use corpus-based discourse analysis to examine solidarity in UN discourse. However, %these approaches generally rely on human coding or %identified through keyword
%solidarity-related lexical cues to identify expressions of solidarity, rather than directly automating solidarity classification.
these approaches do not automate the classification of solidarity expressions in text.

%%% 3.2. AUTOMATED NLP CLASSIFICATION OF SOLIADRITY. LIMITED. CRISIS-FOCUSED, MEDIA-FOCUSED, binary or only supervised classifier. THIS IS WHERE WE ARE SITUATED

Recent NLP %studies have examined expressions of solidarity
work directly addresses this task of automated solidarity annotation. For example, \citet{weber2024social, eger2022measuring} train supervised classifiers to identify expressions of solidarity and anti-solidarity in social-media data on refugee movements and subsequently examine their prevalence over time, while %, the COVID-19 pandemic \citep{kalaloi2021mediated}, 
\citet{santhanam2019stand} use a recurrent neural network to detect solidarity during natural disasters. %These works typically use supervised models to detect solidarity or anti-solidarity in large-scale Twitter data and track how such expressions evolve over short timeframes. While they demonstrate the feasibility of automated solidarity detection, they focus on specific, time-bound crises and treat solidarity as a binary sentiment-like category rather than a differentiated moral frame. In contrast, our work applies a theory-driven typology to political speech, enabling fine-grained analysis of solidarity in parliamentary discourse.
These studies focus on specific, time-bound crises--with \citet{kneuer2025solidarity} as a longitudinal exception, examining German migration discourse from 2005-2019--and %typically treat (anti-)solidarity as a binary classification task, whereas our work applies a differentiated, theory-based typology to parliamentary discourse.
use binary (e.g. \emph{solidarity} vs. \emph{non-solidarity}) categories. %A related longitudinal study by \citet{kneuer2025solidarity} extends the temporal scope to German citizens' migration discourse from 2005-2019, combining keyword retrieval for explicit solidarity with DistilBERT classification for implicit cases. The latter, however, remains binary, while finer-grained frames and stance are analyzed through manual coding or dictionaries.

Our work brings these strands together by using prompted LLMs to automatically apply a fine-grained, theory-based typology of (anti-)solidarity across German parliamentary debates on migration spanning 150 years, %This allows us to combine corpus-scale longitudinal analysis with focused analyses of selected historical periods and differences across migrant groups and political parties, while 
which allows us to distinguish forms of solidarity that prior automated work captures only at a coarser level. In doing so, we combine two directions highlighted by \citet{luypaert2024comparative}: scaling solidarity-frame analysis with LLMs and extending it longitudinally.

\paragraph{%LLMs as Annotation Tools in Political and Social Science
NLP and LLMs for Political and Social Science %Annotation
Analysis} 

A closely related line of research examines how social groups are perceived. The Stereotype Content Model (SCM) characterizes stereotypes of social groups primarily along the dimensions of warmth and competence and links these perceptions to intergroup relations \citep{fiske2002model}. Recent NLP work has operationalized warmth, competence, and related stereotype dimensions to analyze stereotype content in natural-language data \citep[e.g.][]{fraser2021understanding, fraser2024stereotype, mohammad2025words}, while other work has used these dimensions to examine how social groups are represented by LLMs \citep{jeoung2023stereomap}. While this work concerns perceptions and stereotypes of social groups rather than solidarity itself, it is conceptually related through its focus on social-group evaluation and ingroup-outgroup relations.

A related NLP task is \emph{stance detection}, which aims to determine an author's position toward a target, typically using relatively coarse categories such as \emph{favor}, \emph{against}, and \emph{neutral} or \emph{neither}. Much of this literature has focused on social media and political discourse, including the widely used SemEval-2016 Task 6 benchmark on stance detection in tweets \citep{mohammad2016semeval}. Subsequent work has extended stance detection to zero-shot settings with unseen targets \citep[e.g.][]{allaway2020zero} and to multilingual and cross-lingual settings \citep[e.g.][]{vamvas2020xstance}. Methodologically, approaches have ranged from feature-based and neural classifiers to pretrained language models and, more recently, generative LLMs \citep[for recent surveys, see][] {gera2025deep, huang2026survey}. While stance detection captures the direction of a position toward a target, our framework also distinguishes the social and normative grounds of (anti-)solidarity.

More broadly, LLMs are increasingly used as annotation tools in political and social science research. Prior work has considered annotation of political ideology \citep[e.g.][]{heseltine2024large}, sentiment \citep{lashitew2025corporate, fu2024deciphering}, stance \citep[e.g.][]{hamerlik2024chatgpt}, and misinformation \citep[e.g.][]{cao2024can, huang2025unmasking}. For broader surveys of LLM applications %benefits, and challenges 
in social and political science, see \citet{thapa2025large, li2024political, ziems2024can}. %However, conclusions about LLM performance vary: some studies find near-human results \citep{rytting2023towards} or even outperformance of humans and traditional models on short texts and tasks like ideology classification \citep{le2025positioning, heseltine2024large}. In contrast, several studies show that smaller fine-tuned models outperform LLMs, especially when labeled data is available \citep{wang2024selecting, burnham2024political} or when tasks require expert-defined taxonomies \citep{ziems2024can}. 
Conclusions about LLM performance vary: some find near-human \citep{rytting2023towards} or superior results on short texts \citep{le2025positioning, heseltine2024large}, while others show smaller fine-tuned models perform better with labeled data or expert taxonomies \citep{wang2024selecting, burnham2024political, ziems2024can}. Our work provides a systemic evaluation of LLMs for such a complex, theory-informed task, namely (anti-)solidarity detection.

Beyond %standard classification 
these common annotation tasks, a subset of research has applied language models to more complex, theory-driven categories. %For example, 
\citet{erhard2025popbert} develop PopBERT, a fine-tuned transformer for detecting populist rhetoric %and ideological dimensions
in German Bundestag speeches, %using a custom annotation scheme.
while \citet{card2022computational} analyze 140 years of U.S. political speech with fine-tuned RoBERTa and BERT models to classify immigration tone, frames, and dehumanizing metaphors. %Related work has similarly applied fine-tuned models to 
Other work uses fine-tune models for pragmatic and rhetorical categories in German debates, such as speaker attribution, speech acts, and opinion roles \citep{reinig2024politics, rehbein2024out, rehbein2024new}. In contrast to fine-tuned approaches, \citet{ranjit2024oath} use general-purpose LLMs (e.g., GPT-4, Flan-T5) with a custom typology of attitudes toward homelessness %developed from framing theory and grounded theory. Their evaluation focuses on aggregate trends and correlations with external structural indicators, using LLMs 
to annotate large-scale Twitter data and study aggregate trends.
%Our work differs in several important ways. Unlike \citet{erhard2025popbert} and \citet{card2022computational}, we use state-of-the-art LLMs and avoid the need for supervised data sets and fine-tuning. Unlike \citet{ranjit2024oath}, who also use prompted LLMs, we adopt a theory-driven typology, apply it to 150+ years of parliamentary speech rather than social media texts, and compare systematically to human annotations, thus enabling fine-grained, longitudinal insight into political discourse. 

\enlargethispage{1\baselineskip}
\paragraph{Valid Inference from Imperfect LLM Labels} %\ak{In line with concerns that computational text analysis often prioritizes technical performance over measurement validity \citep{baden2022three}, r
Recent work moves beyond raw annotation accuracy to ask whether machine-generated labels actually support valid downstream inference. As \citet{egami2023using} stress, downstream uses of surrogate labels often ignore measurement error, that is, the unknown, heterogeneous mismatch between gold-standard and surrogate labels, which can distort estimates and invalidate inference even when surrogate labels appear highly accurate \citep{egami2023using, knox2022testing}. In CSS, such error is especially difficult to avoid because annotation tasks are often intrinsically challenging \citep{ziems2024can}, and prior work shows that LLM annotations often diverge from expert coding decisions, especially on more complex tasks \citep{lin2025navigating, de2024can}. %LLM-based labels may also reflect social and demographic biases \citep{zhao2018gender, bender2021dangers, weidinger2022taxonomy} and remain sensitive to modeling and prompting choices \citep{perez2021true, bail2024can, abraham2025prompt}.
In response, bias-corrected %\footnote{Here, bias refers to systematic error in a statistical estimate, not to social or demographic bias in model outputs.} 
estimators combine machine-generated labels with smaller sets of gold-standard annotations to support valid downstream inference. These include Prediction-Powered Inference (PPI) \citep{angelopoulos2023ppi, angelopoulos2023ppi+}, %for example, starts from an estimator based on machine predictions and corrects it post hoc using gold-standard labels. 
Design-Based Supervised Learning (DSL) \citep{egami2023using}, 
%constructs bias-corrected pseudo-outcomes that can be used directly in downstream analysis. 
and Confidence-Driven Inference (CDI) \citep{gligoric2025can}. %further uses LLM verbalized confidence to target human annotation where it is most informative. 
We build most directly on DSL because it was developed with CSS applications in mind and fits our design, which combines a fixed decade-stratified human-coded sample with multiple historical trend analyses. Recent benchmarking also suggests that DSL often yields lower bias and more precise estimates than PPI in text classification tasks \citep{de2025benchmarking}.

DSL has been evaluated in a limited number of benchmark and applied settings. \citet{egami2023using} assess DSL in a custom regression task and in class prevalence estimation across CSS text classification datasets from \citet{ziems2024can}. %showing that even accurate surrogate labels can bias inference, whereas DSL restores validity. 
\citet{rister2025correcting} apply DSL to AI-assisted image labeling, and \citet{krsteski2025valid} %study inference from 
to LLM-generated survey responses. %likewise showing that %naive synthetic responses bias population estimates, while DSL and PPI substantially reduce bias.
In settings closer to ours, \citet{stuhler2026time} study U.S. media representations of the timing and urgency of climate change using a hybrid pipeline %that combines 
of rule-based extraction %of temporal horizons with 
and RoBERTa-based classifiers, %for horizon interpretation and urgency detection, 
and apply DSL to debias trend and regression estimates. \citet{liu2025contagion} examines Chinese-language social media discourse around two major 2024 U.S. political events using LLM-based coding of misinformation, partisan stance, affective polarization, and attack language. There, LLM annotation is the main measurement framework, while DSL is presented as a robustness check and is mainly implemented in the user-engagement analysis rather than across all analyses.
%Our study differs from both in three ways. First, we apply DSL to LLM outputs rather than to a pipeline built on rule-based extraction and supervised classifiers such as RoBERTa. Second, unlike \citet{stuhler2026time}, whose design centers on information extraction, we apply DSL to a theory-driven classification task. Third, we study parliamentary discourse over longer time spans and treat LLM validation and DSL-based bias correction as core parts of the inferential workflow across all analyses.

Across work using both raw and bias-corrected model predictions, our study differs in several respects. Unlike %studies that rely on fine-tuned supervised models or hybrid pipelines built on rule-based extraction and classifiers such as RoBERTa 
\citet{erhard2025popbert, card2022computational, stuhler2026time}, we use state-of-the-art prompted LLMs and avoid the need for task-specific fine-tuning, and apply DSL directly to their outputs. Unlike \citet{ranjit2024oath} and \citet{liu2025contagion}, which also use prompted LLMs, we apply a theory-driven typology to %more than 150+ years of parliamentary speech
long-term parliamentary discourse rather than social media text, and we %make LLM validation and DSL-based bias correction central to inference across all analyses 
systematically evaluate how LLM validation and DSL-based bias correction affect downstream inference rather than treating them as robustness checks, %thus enabling 
and apply the resulting corrected estimates to fine-grained, longitudinal %insight into 
analysis of political discourse.
\section{Methodology}\label{sec:methodology}

\subsection{Data}\label{sec:data}

We build on the DeuParl dataset introduced in \citet{DeuParl}, which consists of plenary protocols from the German \de{Reichstag}{imperial diet} (sourced from the \textit{Reichstagsprotokolle}) and the \de{Bundestag}{federal parliament} (via the \textit{Open Data} platform), spanning 1867–2022\footnote{Protocols from the Volkskammer (the GDR parliament) are not included due to lack of availability. As a result, GDR-specific narratives and solidarity frames are absent from the corpus.}. We extend this corpus %to include protocols up 
to June 2025, following the preprocessing steps from \citet{DeuParl} and including metadata such as date, session number, and legislative period.

We reuse the keyword sets (see a full list of keywords in 
\Cref{app:appendix-keywords} in the Appendix) for the category \de{Migrant}{migrant} %and \de{Frau}{woman} 
introduced in \citet{kostikova-fine}, comprising 32 terms. %32 and 18 terms, respectively. 
Based on these, %keywords, 
we extract 63k migrant-related %and 138k women-related 
sentences (hereafter, \emph{instances}) from DeuParl, including the three preceding and three following sentences as context. We enrich each instance with speaker and party information, using \textit{Open Data} metadata for recent sessions (20th-21st periods) and improved XML markup for earlier ones (1st-19th) \citep{poradaautomatische}; see \Cref{app:appendix-parties} in the Appendix for details. %We include women alongside migrants in the evaluation to test model generalizability across distinct marginalized groups, though our substantive analysis focuses on migrants.

\begin{figure}[!htb]
  \centering
  \includegraphics[width=0.96\linewidth]{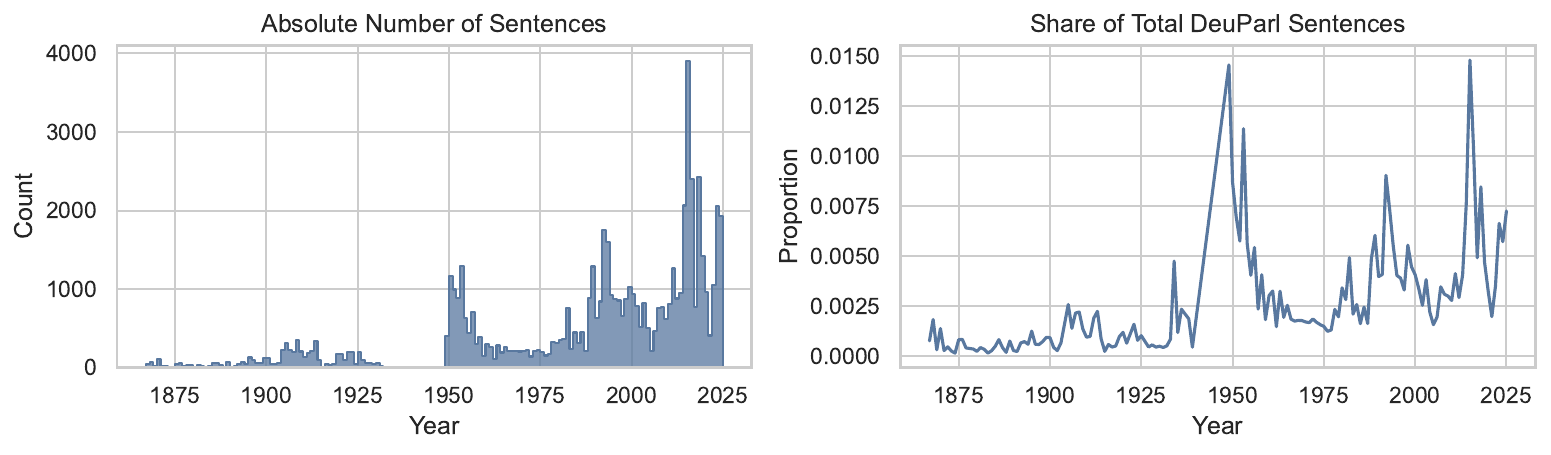}
  \caption{Absolute number of instances (left) and their share relative to all DeuParl sentences (right) for the %\Frau{} and 
  \Migrant{} dataset. %categories
  }
  \label{fig:instances-per-year}
\end{figure}

\Cref{fig:instances-per-year} shows the number of instances over time and their share relative to all DeuParl sentences. Mentions of migrants increase substantially after 1990, with migrant-related speech peaking around 2015 during refugee crisis debates. %Mentions of both migrants and women increase substantially after 1990, with migrant-related speech peaking and briefly overtaking that of women around 2015 during refugee crisis debates.
%Notably, mentions of migrants also surpass women-related discourse in the immediate post-war years.
\aknew{The proportional series also shows pronounced peaks in the immediate postwar period, particularly in 1949 and 1953, likely reflecting debates on expellees and displaced persons. The 1949 value, however, is based on a limited number of observations and covers only September–December, as the Bundestag first convened in September 1949.}

%\vspace{-0.05cm}
%\vspace{-0.35cm}

\subsection{Data Annotation}\label{sec:data-annotation}

To validate our models and enable qualitative analysis, we use a human-annotated dataset created in two batches. The first, introduced in \citet{kostikova-fine}, contained both migrant- and woman-related instances %comprising 2,864 instances (1,428 migrant-related, 1,436 woman-related)
from DeuParl, annotated by five trained student assistants. %through an iterative process, achieving Cohen's $\kappa$ of 0.62 (high-level) and 0.42 (fine-grained). %with final labels assigned by majority vote. 
For the present study, which focuses on migration, we use the 1,428 migrant-related instances from this batch. The second batch adds 756 migrant-related instances with broader historical coverage (pre-1950), annotated by four of the same assistants following the same procedure. Together, the two batches comprise a human-annotated subset of 2,184 Migrant instances. The annotation guidelines were refined over multiple rounds of discussion among the annotators and authors; details of the annotation process and the final guidelines are provided in \Cref{app:annotation-details,app:annotation-guidelines} in the Appendix. %For this batch, inter-annotator agreement reached a Cohen's $\kappa$ of 0.47 for high-level and 0.37 for fine-grained labels, with the lower scores compared to the first batch explained by the inclusion of older historical data (discussed in \Cref{sec:model-performance}).

Each instance is labeled according to a fine-grained solidarity framing scheme distinguishing four high-level stance categories -- \emph{solidarity}, defined as willingness to share resources \citet{Lahusen.2018, TwitterDataset}, \emph{anti-solidarity} (restriction or exclusion), \emph{mixed} (both supportive and opposing elements), and \emph{none} (neutral or unrelated content). Solidarity and anti-solidarity are further assigned one of four subtypes adapted from \citet{thijssen2022s}: \emph{group-based} %(shared identity or exclusion), 
(inclusion in or exclusion from a community of belonging and mutual obligation), \emph{compassionate} %(protection or dismissal of marginalized groups)
(support based on vulnerability and need, or denial of such need), \emph{exchange-based} %(valuing contributions or claiming imbalance), 
(valuing contributions and reciprocity, or emphasizing imbalance and burden), and \emph{empathic} %(respect for or rejection of diversity). 
(acceptance of cultural difference and diversity, or rejection of it). We refer to the subtypes as fine-grained annotation. \Cref{tab:solidarity-frames-examples} provides illustrative examples of how the four subtypes are expressed as solidarity and anti-solidarity in the human-annotated data.

\begin{table*}[t]
    \centering
    \footnotesize
    \renewcommand{\arraystretch}{1.15}
    \setlength{\tabcolsep}{5pt}

    \begin{tabularx}{\textwidth}{
        >{\raggedright\arraybackslash}p{0.13\textwidth}
        >{\raggedright\arraybackslash}X
        >{\raggedright\arraybackslash}X}
        \toprule
        \textbf{Frame} & \textbf{Solidarity example} & \textbf{Anti-solidarity example} \\
        \midrule

        \textbf{Group-based} % 19940428-bt-12-225-01994 and 19131206-rtdkr-13.1-184-00227
        &
        \enquote{[...] No state can accept that a significant part of the population [...] remains outside the political community [...]. \textbf{No immigrant who [...] shares the ways of life and values of the native population, can be expected to remain excluded as a foreigner under the current laws.} [...]%In Germany, foreigners are excluded in many respects from rights and opportunities available to German citizens in the sphere of state. [...]
        }
        &
        \enquote{[...] \textbf{A public administration 
        [must not] employ foreigners to drive down %prevailing 
        wages. 
        [...]} 
        %If the mine owners %truly wish 
        To act in the national interest, [mine owners] have a duty to dismiss the foreign workers whenever suitable German workers remain unemployed. The administration has many means, %at its disposal, 
        including %the power of 
        expulsion, to %give effect to 
        advance the national idea.}
        \\

        %\addlinespace
        \midrule
        \textbf{Exchange-based} % 20150129-bt-18-082-00641 and 20180419-bt-19-026-02457
        &
        \enquote{[...] We must remain open to the world because, economically, our country cannot afford to shut itself off along national lines. \textbf{Our economy is export-oriented, and we need qualified immigration to Germany.} We want to organize this, and we will. [...]}
        &
        \enquote{\textbf{[...] Migration can fail, especially when immigrants have low qualifications.} In 2013 [...] 40 percent of non-EU immigrants had no qualifications. Since the refugee wave, stabbings have increased by 20 percent [...]. Is that successful migration?}
        \\

        %\addlinespace
        \midrule
        \textbf{Compassion.} % 19890914-bt-11-158-00776 and 19921104-bt-12-116-00755
        &
        \enquote{[...] We can support all measures that serve to help people in an extreme situation. \textbf{The immigrants whom we in the Federal Republic 
        refer to as ethnic German resettlers and migrants from East Germany are indeed in such a situation.} [...]}
        &
        \enquote{[...] there is no end in sight to the abuse of asylum. \textbf{There has been a dramatic increase in immigration by people who are not politically persecuted.} [...] The abuse of asylum has developed a dynamic of its own and has long since become a vicious circle.}
        \\

        %\addlinespace
        \midrule
        \textbf{Empathic} % 20020124-bt-14-212-00534 and 20220112-bt-20-010-01011
        &
        \enquote{[...] \textbf{%Because my parliamentary group supports fostering mutual understanding, 
        we are concerned that the immigration law %currently under discussion 
        [...] contradicts the goals of our foreign cultural policy; for it does not promote encounters between cultures on an equal footing, but demands an almost complete willingness to assimilate as the price of immigration.} We are in favor of integration, not assimilation.}
        &
        \enquote{[...] \textbf{an SPD politician warned:} \enquote{\textbf{We have overextended ourselves by taking in people from completely different cultural worlds; 7 million foreigners in Germany represent a misguided development.} We must prevent immigration from foreign cultures; %immigration by people from Anatolia and sub-Saharan Africa 
        [it] only creates an additional problem.} Reasonable views. [...]}
        \\

        \bottomrule
    \end{tabularx}

    \caption{Illustrative examples of the four (anti-)solidarity frames from the human annotated subset. Bold text is the main sentence, the other sentences are for context. Original German texts are provided in \Cref{tab:solidarity-frames-examples-german} in the Appendix.}
    \label{tab:solidarity-frames-examples}
\end{table*}

\begin{figure}[!htb]
  \centering
  \includegraphics[width=0.57
  \linewidth]{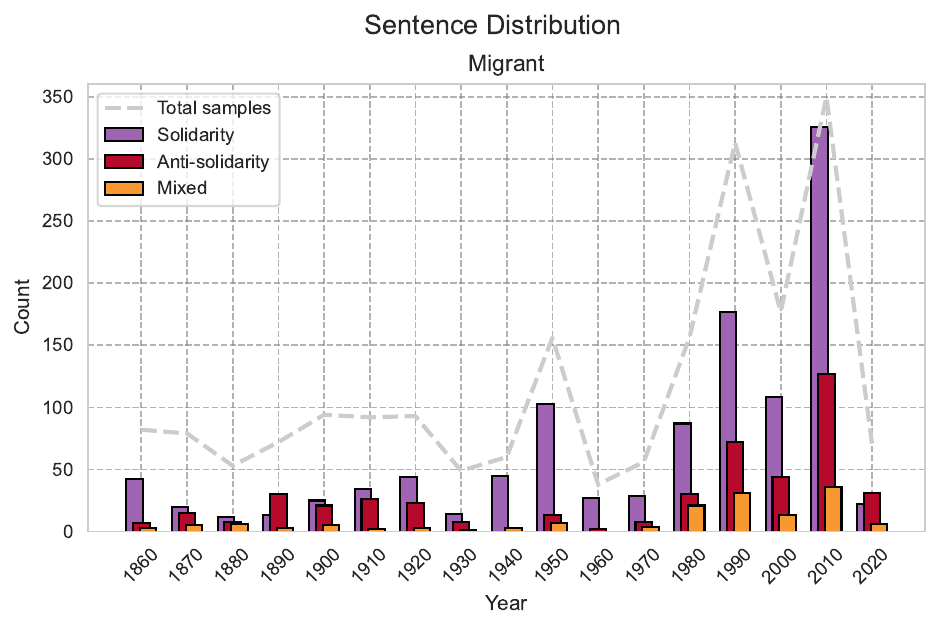}
  \caption{Distribution of human annotated instances by time %and target group. 
  for the \Migrant{} dataset.
  \emph{Total} indicates the number of instances per decade 
  (including \emph{None}); see \Cref{tab:instances-distribution} in the Appendix for exact counts.}
  \label{fig:instances-distribution}
\end{figure}

The final gold-standard label for each instance is determined by majority vote. %and the dataset also includes 368 previously introduced curated instances.
For the first batch, inter-annotator agreement reached a Cohen's $\kappa$ of 0.62 for the high-level labels and 0.42 for the fine-grained labels. For the second batch, agreement reached 0.47 for the high-level labels and 0.37 for the fine-grained labels. We attribute the lower agreement in the second batch to the inclusion of older historical data, which we discuss further in \Cref{sec:model-performance}.
Statistics on the human annotated subset are shown in \Cref{fig:instances-distribution} and \Cref{tab:instances-distribution} in the Appendix. %Label distributions differ by target group: woman-related instances are predominantly solidarity (74.1\%), while migrant-related instances are more diverse, with less solidarity (51.6\%) and more anti-solidarity (21.3\%).

\subsection{Models and Experiments}\label{sec:models-experiments}

We evaluate a diverse set of models to cover variation in model size, provider, and access setting (open vs.\ closed), while also including supervised classifier baselines. We evaluate both closed- and open-weight LLMs; for the large-scale downstream analyses, however, we focus on open-weight models because they can be run locally, improve reproducibility, and reduce annotation costs of the full corpus. Our selection comprises:

\begin{enumerate}[label=(\roman*), align=left, leftmargin=*]
  \item smaller pre-trained LMs \texttt{BERT} and \texttt{SBERT} as supervised baselines. Full model configurations, fine-tuning setups, and inference details are provided in \Cref{app:models-details} in the Appendix;
  \item closed-source LLMs \texttt{gpt-4-1106-preview} and \texttt{gpt-5-chat};
  \item open-weight LLMs, including:
    \begin{itemize}
      \item \texttt{gpt-oss-20B} and \texttt{gpt-oss-120B};
      \item \texttt{Mistral-2-Large} (123b parameters);
      \item \texttt{Llama-4-Scout-17B-16E-Instruct};
      \item \texttt{Llama-3.3-70B-It}, \texttt{Llama-3-8B-It}; %and a fine-tuned \texttt{Llama-3-70B-It};
      \item \texttt{Qwen2.5-72B-Instruct};
      \item \texttt{Gemma-2-9B-It} and \texttt{Gemma-2-27B-It}.
    \end{itemize}
\end{enumerate}

\paragraph{Data and Metrics} We adopt a 70/15/15 train/dev/test split of the migrant-related instances from the first annotated batch \citep{kostikova-fine}, generating three random splits, each comprising about 1,000 training, 215 development, and 215 test (Test 1) instances. %2,000 training, 400 development, and 430 test (Test 1) instances in total, covering both target groups. 
Supervised models are evaluated on these splits,\footnote{The BERT and SBERT baselines retain the original training setup of \citet{kostikova-fine}, in which the training data included both migrant- and woman-related instances; evaluation reported in the present study is restricted to migrant-related instances.} while inference-only experiments use two test sets: Test 1 (original) and Test 2 (756 newly added migrant-related instances, 1860s-1950s). We report Macro F1 %to account for class imbalance, 
to give equal weight to performance across classes despite class imbalance, along with per-class F1 for high-level (four stance categories) and fine-grained (ten labels) classification.\footnote{High-level: solidarity, anti-solidarity, mixed, none. Fine-grained: four solidarity subtypes, four anti-solidarity subtypes, plus mixed and none.}

\paragraph{Statistical Comparison of Models} To assess whether differences in macro-F1 between the strongest models are statistically reliable, we use a paired stratified bootstrap \citep{dror2018hitchhikers}. Within each test split or test set, we resample instances with replacement within gold-label strata, preserving the class composition of the original test data. Both models are evaluated on the same resampled instances, and the test statistic is the difference in macro-F1. We use 10,000 bootstrap samples to estimate percentile-based 95\% confidence intervals and two-sided bootstrap $p$-values. %For \textit{Woman} and \textit{Migrant (T1)}, which each contain three test splits,
For \textit{Migrant (T1)}, which contains three test splits, we compute the macro-F1 difference separately for each split and use the mean difference across the three splits as the test statistic. For \textit{Migrant (T2)}, which contains one test set rather than three splits, the statistic is the macro-F1 difference on that test set. We apply Holm correction across the planned model comparisons separately within each evaluation setting, defined by test set and annotation level. %defined by target group and annotation level.
%Detailed results are reported in \Cref{app:significance-testing} in the Appendix.

\paragraph{Prompting Setup} For the inference-only experiments with LLMs, we use manually designed prompts. %for each target group. 
Each prompt consists of a classification instruction concatenated with the parliamentary instance (the keyword sentence plus three sentences of left/right context as defined in \Cref{sec:data}). 
Speaker and party metadata are available in the corpus but removed from the model input. Instructions include (i) chain-of-thought cues in zero-/few-shot settings (\textit{think step by step}); (ii) label definitions; and (iii) a two-step format: high-level classification followed by subtype prediction.\footnote{For GPT-4, both steps are combined in a single prompt (\Cref{fig:one-step-prompt-migrant} %and \Cref{fig:one-step-prompt-frau} 
in the Appendix); for open-weight models, we use two prompts: one for high-level classification and, conditional on that, a second for subtype selection (\Cref{fig:two-step-prompt-migrant} % and \Cref{fig:two-step-prompt-frau} in the Appendix
in the Appendix).} In few-shot mode, one representative example per label is included.
%Additionally, we fine-tune \texttt{Llama-3-8B-It} and \texttt{Llama-3-70B-It} with a three-model pipeline (high-level plus two subtype classifiers), using human-generated labels and GPT-4 explanations as training data. 

\paragraph{Model Ensemble} We additionally evaluate a model ensemble to test whether combining strong open-weight LLMs improves performance relative to individual models. The ensemble setup also resembles the human annotation procedure in that final labels are derived from multiple judgments rather than a single annotator. The ensemble combines the best-performing open models, each in its strongest configuration (\Cref{tab:models_performance}), using majority vote. The label with the most votes is selected as the final prediction; in tie cases, votes are weighted by each model's macro-F1 score as measured on the validation set. We report ensemble results in \Cref{sec:model-ensemble}.

\paragraph{Design-based Supervised Learning (DSL)} %When employing LLMs as annotators, it may be that single labels are systematically preferred compared to human annotators, resulting in biased estimates and incorrect  conclusions. 
Ultimately, our downstream goal is not to obtain a perfectly correct model label for every instance, but to recover reliable aggregate patterns. Even models with strong instance-level performance may systematically over- or under-predict particular labels, and these errors can accumulate into biased temporal or group-level estimates. We therefore use \bp{a variant of} Design-based Supervised Learning (DSL; \citealt{egami2023using}) to correct such systematic biases. 
Conceptually, DSL combines large-scale %surrogate outputs with a subsample of human-coded data 
model-generated labels (hereafter: \emph{surrogate labels}) with a smaller subsample of human-coded data to construct bias-%corrected 
adjusted outcomes (hereafter: \emph{pseudo-outcomes}) %for 
which are then used directly in downstream inference. In our application, the surrogate labels are soft label distributions derived from a committee of LLMs. The correction additionally accounts for the variables with respect to which we conduct the downstream analysis (hereafter: \emph{covariates}; e.g., decade in the temporal analysis) and for the known probability that an instance was selected for human annotation.

For each parliamentary instance $i$, we observe surrogate labels $Q_i$ and covariates $X_i$. For a (small) subset, we also have human annotations $Y_i$. The surrogate labels $Q_i$ are soft-label distributions over annotation categories, computed as the proportion of committee models assigning each label (e.g., if two of three models predict \emph{solidarity} and one predicts \emph{mixed}, $Q_i$ assigns $2/3$ to \emph{solidarity} and $1/3$ to \emph{mixed}), while the human labels $Y_i$ are distributions over categories constructed from multiple human annotators.\footnote{For comparison, we also evaluate hard-label DSL variants in \Cref{sec:dsl}, where $Q_i$ and $Y_i$ are one-hot vectors corresponding to a single model prediction and the human majority label, respectively.} Human annotations are observed only for a subset of %documents 
instances under a decade-stratified annotation design.\footnote{The human-annotation sample was stratified by decade due to the corpus's uneven temporal distribution, ensuring representation across the full 150+ year period and supporting our goal of tracing changes in (anti-)solidarity over time.} We assign each instance a decade-specific inclusion probability $\pi_i$, computed as the proportion of %documents 
instances in that decade that received human annotation, and treat these as fixed design probabilities in the DSL correction.

DSL corrects downstream estimands by constructing a bias-adjusted pseudo-outcome
\begin{equation}\label{eq:dsl-pseudo-outcome}
\tilde{Y}_i
= \hat g(Q_i, X_i)
+ \frac{R_i}{\pi_i}\big(Y_i - \hat g(Q_i, X_i)\big),
\end{equation}
where $R_i$ indicates whether instance $i$ is human-annotated, and $\hat g(Q_i,X_i)$ estimates %the conditional expectation of human labels 
the expected human label distribution given surrogate labels and %time. 
covariates. We model $\hat g$ using a multinomial regression trained on soft targets and implement $K$-fold cross-fitting. 

Because outcomes are multi-class, DSL is applied in a vector-valued form, giving a vector of DSL-adjusted class scores $\tilde{Y}_i$ for each instance. We use them directly in downstream analyses.

We estimate separate DSL models for the high-level and fine-grained tasks to avoid carrying subtype-level error from fine-grained predictions into aggregated high-level estimates.

\section{Results}\label{sec:results}

In this section, we evaluate the models in two steps. We first assess \emph{instance-level annotation %quality
performance} against human labels, including overall performance, task-specific challenges, and systematic error patterns in \Cref{sec:model-performance}. We then examine whether these model outputs support reliable downstream inference by comparing hard-label, soft-label, and DSL-corrected trend estimates against human-derived reference trends in \Cref{sec:downstream-reliability}.

\subsection{Annotation %Quality 
Performance at the Instance Level}\label{sec:model-performance}

\Cref{tab:models_performance} reports macro F1 scores for each model under inference and fine-tuned settings. We benchmark against the leave-one-out F1 score of human annotators,\footnote{For each annotator, we exclude their labels, compute a consensus from the others, and calculate the macro F1 score between the excluded annotator and the consensus; scores are then averaged across annotators.} and show results for both high-level and fine-grained tasks (the latter in parentheses). \ak{We use leave-one-out F1 here to make human and model classification performance directly comparable using the same metric; inter-annotator agreement is reported separately in \Cref{sec:data-annotation}.} We evaluate instance-level annotation performance along four dimensions: (i) overall performance of individual models; (ii) task- and label-related challenges; (iii) model ensemble performance; and (iv) error analysis.

\begin{table}[t]
\centering
\renewcommand{\arraystretch}{1.15}
\footnotesize

\begin{tabular}{@{}lcc!{\vrule width 0.4pt}cc@{}}
\toprule
& \multicolumn{2}{c!{\vrule width 0.4pt}}{\textbf{Migrant (T1)}} 
& \multicolumn{2}{c}{\textbf{Migrant (T2)}} \\
\cmidrule(lr){2-3}\cmidrule(l){4-5}
\textbf{Model}
& \textbf{Zero} & \textbf{Few}
& \textbf{Zero} & \textbf{Few} \\
\midrule

\multicolumn{5}{@{}l}{\textit{Closed-weight models}} \\
\addlinespace[2pt]

GPT-4
& 0.42 (0.73)
& \dimrow 0.43 (0.63)
& 0.29 (0.62)
& \dimrow 0.26 (0.55) \\ % Woman: zero 0.37 (0.60), few 0.37 (0.54)

GPT-5
& \dimrow 0.43 (0.67)
& \cellcolor{bestcolor}0.50 (0.71)
& \dimrow 0.33 (0.60)
& \cellcolor{bestcolor}0.34 (0.58) \\ % Woman: zero 0.36 (0.53), few 0.41 (0.65)

\midrule

%\addlinespace[4pt]
\multicolumn{5}{@{}l}{\textit{Large open-weight models}} \\
\addlinespace[2pt]

gpt-oss-20B
& \dimrow 0.37 (0.56)
& 0.40 (0.56)
& \dimrow 0.26 (0.49)
& 0.26 (0.49) \\ % Woman: zero 0.26 (0.43), few 0.29 (0.48)

gpt-oss-120B
& \dimrow 0.42 (0.68)
& \cellcolor{secondcolor}0.48 (0.70)
& \dimrow 0.25 (0.55)
& \cellcolor{secondcolor}0.31 (0.57) \\ % Woman: zero 0.30 (0.55), few 0.40 (0.56)

Mistral-2-Large
& \dimrow 0.30 (0.52)
& 0.38 (0.56)
& \dimrow 0.26 (0.49)
& 0.29 (0.46) \\ % Woman: zero 0.23 (0.49), few 0.27 (0.53)

Llama-4-Scout
& \dimrow 0.30 (0.53)
& 0.42 (0.65)
& \dimrow 0.25 (0.50)
& 0.28 (0.50) \\ % Woman: zero 0.26 (0.47), few 0.27 (0.48)

Qwen-2.5
& \dimrow 0.39 (0.61)
& 0.41 (0.61)
& \dimrow 0.26 (0.52)
& 0.28 (0.51) \\ % Woman: zero 0.23 (0.42), few 0.29 (0.47)

Llama-3.3-70B
& \dimrow 0.31 (0.55)
& 0.38 (0.65)
& \dimrow 0.26 (0.49)
& 0.29 (0.57) \\ % Woman: zero 0.23 (0.51), few 0.28 (0.49)

% Llama-3-70B & Fine-tuned
% & \dimrow 0.27 (0.65)
% & \dimrow 0.21 (0.47) \\
% Woman: 0.26 (0.49)

%\addlinespace[4pt]
\midrule
\multicolumn{5}{@{}l}{\textit{Mid- and small-scale open-weight models}} \\
\addlinespace[2pt]

Llama-3-8B
& 0.17 (0.42)
& \dimrow 0.18 (0.35)
& 0.15 (0.40)
& \dimrow 0.12 (0.42) \\ % Woman: zero 0.13 (0.34), few 0.14 (0.35)

% & Fine-tuned
% & 0.20 (0.40)
% & \dimrow 0.13 (0.27) \\
% Woman: 0.16 (0.38)

Gemma-2-27B
& \dimrow 0.30 (0.52)
& 0.32 (0.54)
& \dimrow 0.21 (0.47)
& 0.26 (0.53) \\ % Woman: zero 0.20 (0.44), few 0.24 (0.52)

Gemma-2-9B
& \dimrow 0.30 (0.48)
& 0.31 (0.55)
& \dimrow 0.21 (0.47)
& 0.24 (0.43) \\ % Woman: zero 0.22 (0.40), few 0.23 (0.44)

%\addlinespace[4pt]
\midrule
\multicolumn{5}{@{}l}{\textit{Classical baselines}} \\
\addlinespace[2pt]

BERT (2000 train instances)
& \multicolumn{2}{c!{\vrule width 0.4pt}}{0.28 (0.49)}
& \multicolumn{2}{c}{0.24 (0.48)} \\ % Woman: 0.19 (0.29)

SBERT (2000 train instances)
& \multicolumn{2}{c!{\vrule width 0.4pt}}{0.23 (0.43)}
& \multicolumn{2}{c}{0.19 (0.37)} \\ % Woman: 0.16 (0.33)

%\addlinespace[3pt]
\midrule
\textit{Human reference}
& \multicolumn{2}{c!{\vrule width 0.4pt}}{0.48 (0.78)}
& \multicolumn{2}{c}{0.30 (0.58)} \\ % Woman: 0.32 (0.67)

\bottomrule
\end{tabular}

\caption{
Comparative performance (macro F1). Scores are reported as fine-grained (high-level). \enquote{Migrant (T1)} is the original benchmark; \enquote{Migrant (T2)} is newly introduced in this paper. Green cells mark the best score per test set (model's best setting); yellow cells mark the best score among open models (second best overall). Gray text indicates suboptimal settings for a given model. \ak{Pairwise statistical comparisons of selected few-shot models are reported in \Cref{app:significance-testing} in the Appendix.}
}
\label{tab:models_performance}
\end{table}

\subsubsection{Overall model performance}\label{sec:overall-model-performance}
  \begin{itemize}[leftmargin=2em]
    \item \textbf{\ak{GPT-5 achieves the highest reported scores,} but some open models offer a strong alternative.} Across %\textit{Woman} and 
    both \textit{Migrant} test sets, \texttt{GPT-5} (few-shot) %approaches human annotators on the high-level task and surpasses them on the fine-grained task. 
    \ak{achieves macro-F1 scores comparable to the human leave-one-out reference, with scores slightly above the reference on the fine-grained task.} Among open models, \texttt{gpt-oss-120B} likewise %matches 
    achieves scores close to the human reference value in fine-grained performance and comes close to \texttt{GPT-4} and \texttt{GPT-5} at the high level.\footnote{\ak{We also assess the within-model decoding stability of gpt-oss-120B across five decoding seeds and find limited variation in macro-F1 across the two test sets; see \Cref{app:decoding-stability} in the Appendix.}} \ak{However, the pairwise significance tests (\Cref{tab:model-significance-tests} in the Appendix) show %that \texttt{GPT-5} significantly outperforms \texttt{gpt-oss-120B} only on the high-level \textit{Woman} task; their differences are not significant in the other five evaluation settings
    \texttt{GPT-5} and \texttt{gpt-oss-120B} do not differ significantly on any of the evaluation settings.}
    Among the remaining open-weight models, \texttt{Llama-4-Scout}, \texttt{Qwen-2.5}, and \texttt{Llama-3.3-70B} show the strongest performance. These models achieve comparable results across %targets and 
    evaluation levels (e.g., on Migrant T1, fine-grained F1 scores range from 0.38 to 0.42 and high-level scores from 0.61 to 0.65). While their relative strengths vary by setting, where \texttt{Qwen-2.5} and \texttt{Llama-4-Scout} achieve the strongest fine-grained scores, and \texttt{Llama-3.3-70B} the highest high-level scores, their overall performance remains closely clustered.

    \begin{figure}%[htb]
        \centering
        \includegraphics[width=\linewidth]{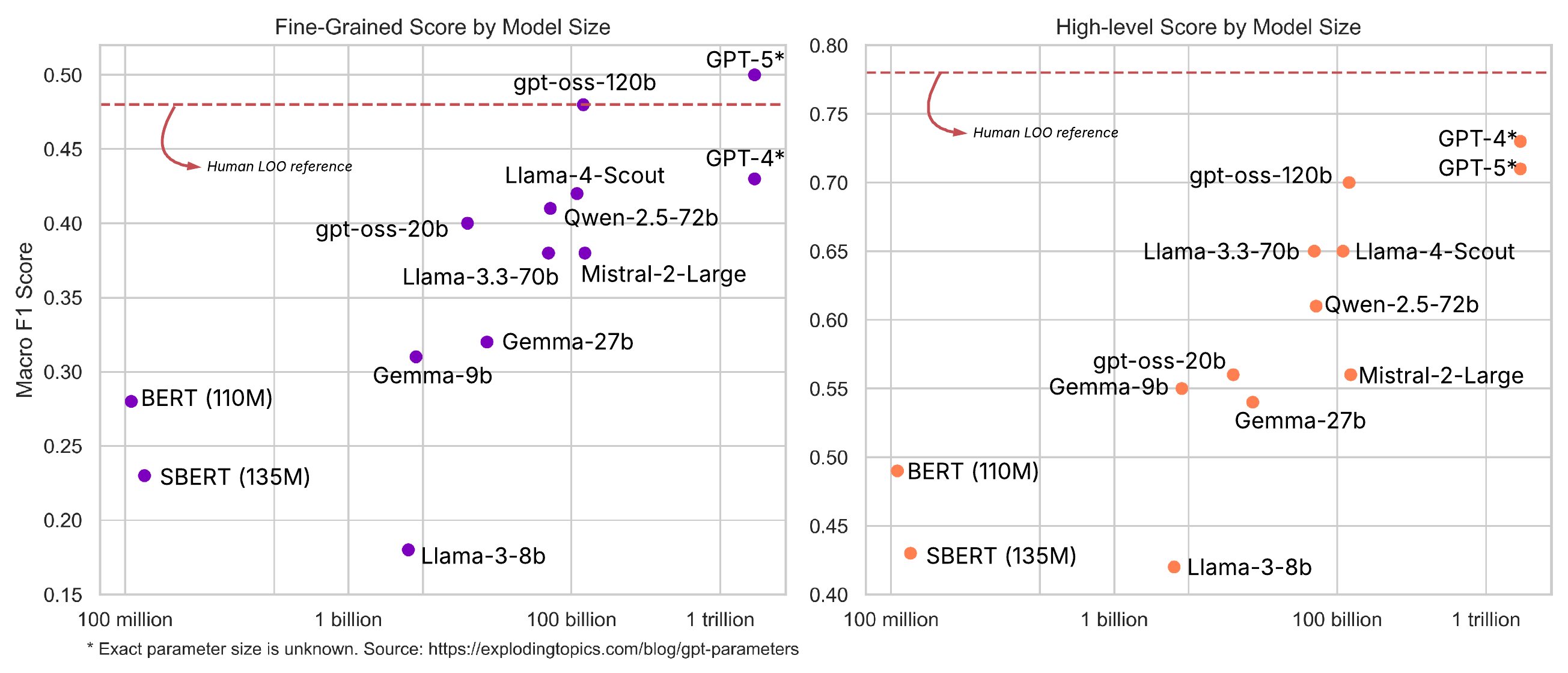}
        \caption{Model size vs. performance on high-level and fine-grained classification in \textit{Migrant (T1)}. Each model is plotted with its best-performing configuration (zero- or few-shot). The x-axis is logarithmic, so spatial distances do not represent linear differences in parameter count.}
        \label{fig:size-performance}
        \end{figure}
    
    \item %\textbf{Model size correlates with performance, but not perfectly.} 
    \textbf{Performance increases with model size, but the relationship is non-monotonic.} Larger models generally outperform smaller ones on both high-level and fine-grained tasks (see \Cref{fig:size-performance}). However, this relationship is non-monotonic: moving from small to mid-sized models brings clear benefits, with notable exceptions such as \texttt{Llama-3-8B}, which underperforms relative to smaller baselines like \texttt{BERT} and the similarly sized \texttt{Gemma-2-9B}. Among large models ($\approx$70B-120B+ parameters), performance gains diminish. Moreover, mid-sized models can match or outperform larger systems; for example, \texttt{gpt-oss-20B} clusters with \texttt{Qwen-2.5} and \texttt{Mistral-2-Large}, while \texttt{gpt-oss-120B} approaches \texttt{GPT-4/5} on both the fine-grained and high-level benchmarks. These results suggest that factors beyond parameter count influence model performance. %including training data, instruction tuning, and architectural choices.

    %\ak{The pairwise significance tests show a model-size advantage most clearly on %\textit{Woman} and 
    %\textit{Migrant (T1)}. \texttt{gpt-oss-120B} significantly outperforms \texttt{Qwen-2.5} at both annotation levels on this test set, \texttt{Llama-3.3-70B} on the fine-grained tasks, and \texttt{gpt-oss-20B} at both annotation levels. On the historical \textit{Migrant (T2)} set, only the high-level advantage over \texttt{gpt-oss-20B} remains significant. \texttt{Llama-3.3-70B} and \texttt{Qwen-2.5} also significantly outperform \texttt{Gemma-2-27B} at both levels on \textit{Migrant (T1)}, but not on \textit{Migrant (T2)} (\Cref{tab:model-significance-tests,tab:model-significance-tests-additional} in the Appendix).}

    \item \textbf{Examples help mainly mid-sized models.
    } 
    Most models benefit from few-shot prompting, with the most consistent improvements observed for mid-sized models such as \texttt{Gemma} models, \texttt{Mistral-2-Large}, \texttt{Llama-3.3-70B} and \texttt{Llama-4-Scout}. Smaller models like \texttt{Llama-3-8B} show only marginal gains, while \texttt{GPT-4} demonstrates minimal improvement. %likely due to its strong zero-shot performance. 
    Overall, the effect of few-shot prompting is widespread but not uniform across model scales.
    
    %\item \textbf{Fine-tuning brings little benefit.} Fine-tuned variants (e.g., \texttt{Llama-3.3-70B-\\it}, \texttt{Llama-3-8B-it}) often show no improvement over inference-only models.

\end{itemize}
  
\subsubsection{Task and label-related challenges}

\begin{itemize}[leftmargin=2em]
  \item \textbf{All models (and human annotators) perform worse on the fine-grained task}, which indicates its complexity.
  \item \textbf{Solidarity is easier to detect than anti-solidarity or mixed stance} (\Cref{fig:f1-scores-plot}). Even smaller models perform well on solidarity %especially in the Woman category 
  (e.g., \texttt{BERT} 0.67, \texttt{Gemma-2-9B} 0.56). In contrast, \textbf{anti-solidarity and mixed stances remain challenging}. %with anti-solidarity being most difficult for Woman. 
  This pattern is likely due in part to class imbalance (\Cref{tab:instances-distribution} in the Appendix), especially because %anti-solidarity is very rare for \de{Woman} and 
  mixed is a less common label. %in both target groups. 
  That said, GPT models (both closed-source models and \texttt{gpt-oss-120B}) achieve the strongest anti-solidarity performance (0.67-0.70), % for Migrant; 0.47-0.65 for Woman), 
  while mid-sized models reach 0.56-0.59. Only \texttt{GPT-4/5}, \texttt{gpt-oss-120B} and \texttt{Llama-3.3-70B} consistently capture mixed stances, with F1 scores between 0.32 and 0.48. \texttt{Llama-4-Scout} shows some asymmetry for Migrant: strong performance on solidarity (0.78) and anti-solidarity (0.62) but near failure on mixed stance (0.04).

  \item \textbf{On the fine-grained level, GPT-5, gpt-oss-120B, Llama-3.3-70B and Llama-4-Scout offer the most balanced performance on the fine-grained level} (see \Cref{tab:f1_solidarity} in the Appendix). Best fine-grained macro F1 is achieved by GPT-5. Among non-GPT-4/5 models, Llama-3.3-70B and Llama-4-Scout follow. %with gpt-oss-120B showing the best overall balance for the 10-label setup. 
  Nevertheless, all models perform poorly on specific subtypes, including empathic solidarity (e.g., 0.11 for Migrant with Llama-3.3-70B) and compassionate anti-solidarity (e.g., 0.06 for Migrant with gpt-oss-120B). Empathic anti-solidarity fails across all models. %For Woman, all models consistently struggle with fine-grained anti-solidarity.

  \begin{figure}%[!htb]
    \centering
    \includegraphics[width=0.9\linewidth]{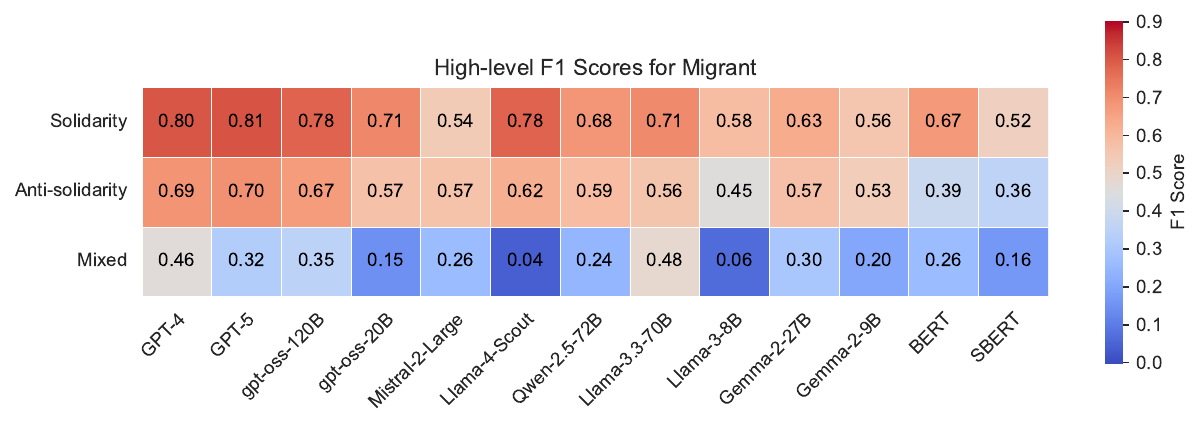}
    \caption{F1 scores for high-level labels on the combined Migrant test sets (Test 1 and Test 2), based on the best-performing configuration for each model.
    }
    \label{fig:f1-scores-plot}
    \end{figure}

  \item \textbf{Performance drops on historical data.} All models, including the strongest, perform worse on the historical \textit{Migrant (T2)} set. This reflects a broader trend: performance declines on older data (see \Cref{fig:macro-f1-kappa-over-time} in the Appendix), as does human agreement (Cohen's Kappa). We examine errors on the historical data in more detail in \Cref{sec:error-analysis}.
\end{itemize}

\subsubsection{Model Ensemble Performance}\label{sec:model-ensemble} % 

We construct a %weighted 
majority-vote ensemble from the best-performing open models in \Cref{tab:models_performance}: Llama-3.3-70B, Qwen-2.5, and gpt-oss-120B. We exclude GPT-5 due to its higher cost and because the subsequent social-scientific analysis focuses on open models.

We find that the ensemble does not outperform the strongest individual model (\texttt{gpt-oss-120B}), indicating that simple %weighted 
majority voting does not consistently improve performance in this setting (see \Cref{tab:ensemble_performance}). This is in line with prior work showing that ensemble methods are most effective when individual models achieve comparable performance while making at least partly uncorrelated errors \citep{trad2024ensemble}.

\begin{comment}
\begin{table}[h]
\centering
\renewcommand{\arraystretch}{1.2}
\footnotesize
\begin{tabular}{@{}lccc@{}}
\toprule
\textbf{Model} & \textbf{Woman} & \textbf{Migrant (T1)} & \textbf{Migrant (T2)} \\
\midrule
gpt-oss-120B (Few-shot)      & 0.40 (0.56) & 0.48 (0.70) & 0.31 (0.57) \\
Llama-3.3-70B (Few-shot)     & 0.28 (0.49) & 0.38 (0.65) & 0.29 (0.57) \\
Qwen-2.5 (Few-shot)          & 0.29 (0.47) & 0.41 (0.61) & 0.28 (0.51) \\
\midrule
\textbf{Ensemble} & 0.38 (0.56) & 0.48 (0.72) & 0.31 (0.59) \\
\bottomrule
\end{tabular}
\caption{
Performance (macro F1) of models included in the ensemble and the resulting majority-weighted ensemble on fine-grained and high-level tasks (in parentheses). In tie cases, votes are weighted by validation-set macro-F1. Tie-breaking weights are based on validation-set performance, ranking gpt-oss-120B $>$ Llama-3.3-70B $>$ Qwen-2.5 for the migrant setting and Llama-3.3-70B $>$ gpt-oss-120B $>$ Qwen-2.5 for the woman setting.
}
\label{tab:ensemble_performance}
\end{table}
\end{comment}

\begin{table}[h]
\centering
\renewcommand{\arraystretch}{1.2}
\footnotesize
\begin{tabular}{@{}lcccc@{}}
\toprule
\textbf{Test set} &
\textbf{gpt-oss-120B} &
\textbf{Llama-3.3-70B} &
\textbf{Qwen-2.5} &
\textbf{Ensemble} \\
\midrule
Migrant (T1) &
0.48 (0.70) &
0.38 (0.65) &
0.41 (0.61) &
\textbf{0.48 (0.72)} \\

Migrant (T2) &
0.31 (0.57) &
0.29 (0.57) &
0.28 (0.51) &
\textbf{0.31 (0.59)} \\

% Woman &
% 0.40 (0.56) &
% 0.28 (0.49) &
% 0.29 (0.47) &
% 0.38 (0.56) \\

\bottomrule
\end{tabular}
\caption{
Performance (macro F1) of models included in the ensemble and the resulting majority-weighted ensemble on fine-grained and high-level tasks (in parentheses). All individual models use their few-shot configuration. In tie cases, votes are weighted by validation-set macro-F1. Tie-breaking weights are based on validation-set performance, ranking gpt-oss-120B $>$ Llama-3.3-70B $>$ Qwen-2.5.
% For the woman setting, the ranking is Llama-3.3-70B $>$ gpt-oss-120B $>$ Qwen-2.5.
}
\label{tab:ensemble_performance}
\end{table}

%In summary, GPT-5 achieves the highest overall scores, exceeding the human-reference F1 on the fine-grained task and approaching it on the high-level task. However, its closed-weight nature and high cost make it less suitable for our large-scale social-scientific analysis. Among open-weight models, gpt-oss-120B performs comparably to GPT-5 and approaches the human-reference performance, even matching it on the fine-grained task, making it the strongest open-weight alternative. 
In summary, GPT-5 achieves the highest overall scores, %exceeding the human-reference F1 on the fine-grained task and approaching it on the high-level task. 
\ak{with macro-F1 comparable to the human leave-one-out reference.} Among open-weight models, gpt-oss-120B %performs comparably to GPT-5 and 
\ak{achieves similarly competitive scores.} %approaches the human-reference performance, even matching it on the fine-grained task. %making it the strongest open-weight alternative. 
However, GPT-5's closed-weight nature and high cost make it less suitable for our large-scale social-scientific analysis.  
\ak{Moreover, the pairwise significance tests show that %GPT-5 significantly outperforms gpt-oss-120B in only one of the six evaluation settings, while the differences are not significant in the remaining five. 
there is no significant difference between GPT-5 and gpt-oss-120B. Thus, the additional resources associated with \texttt{GPT-5} are not consistently associated with statistically reliable gains over the strongest open-weight model. Among open-weight models, benefits from greater model scale are clearest on %\textit{Woman} and 
\textit{Migrant (T1)}, but are substantially weaker on the historical \textit{Migrant (T2)} set (\Cref{app:significance-testing} in the Appendix), which is consistent with the broader performance decline observed on historical data.}

That said, %the fact that these models reach or exceed 
\ak{comparability to} human reference values does not imply that they can be used as direct substitutes of human annotators. %In our setup, the human reference approximates maximum attainable agreement under current annotation conditions, estimated through a LOO procedure and therefore representing a practical ceiling on inter-annotator consistency, %not an absolute measure of correctness. 
First, the observed macro-F1 scores remain low enough that substantial and systematic model errors can still occur, including errors that human annotators are less likely to make; moreover, recent work shows that even high agreement with human consensus can coexist with systematic misclassification patterns \citep{bavaresco2025llms}. We therefore examine these patterns through error analysis (\Cref{sec:error-analysis}).

Second, in our setup, the human reference is derived from leave-one-out agreement and therefore inherits the uncertainty present in the human annotations. As a result, perfect F1 may itself be unattainable because some disagreement among human annotators likely reflects genuine label variation. In a subjective task like ours, this motivates moving beyond hard labels to consider label distributions instead (\Cref{sec:label-distrib-similarity}). We address both issues next.

\subsubsection{Error Analysis}\label{sec:error-analysis}

\begin{figure}%[!htb]
    \centering

    \begin{subfigure}[t]{\linewidth}
        \centering
        \includegraphics[width=\linewidth]{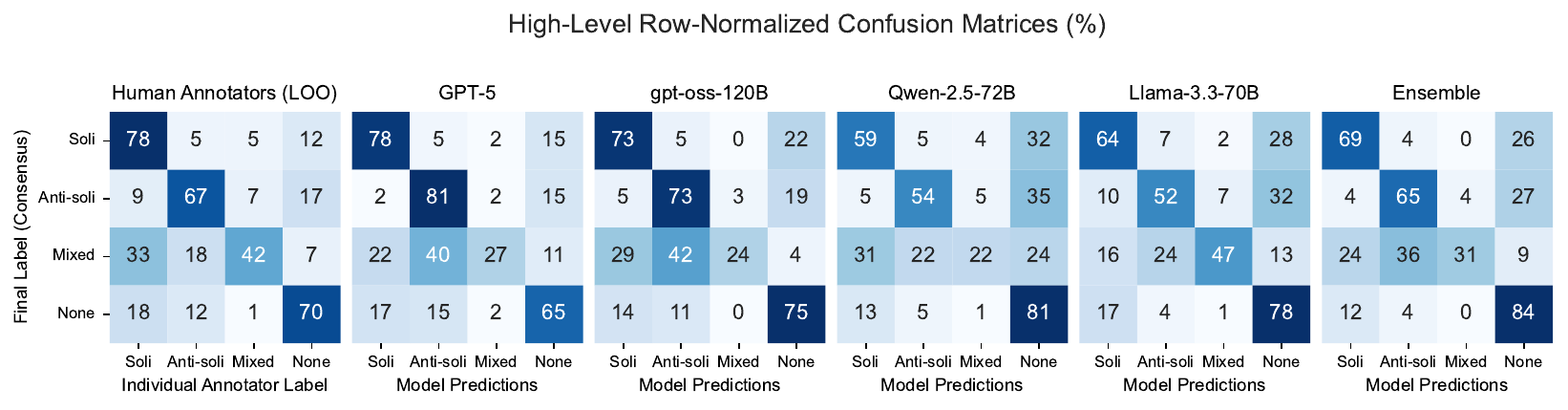}
        \caption{High-level confusion matrices.}
        \label{fig:confusion-matrices-top-high}
    \end{subfigure}

    %\vspace{0.8em}

    \begin{subfigure}[t]{\linewidth}
        \centering
        \includegraphics[width=\linewidth]{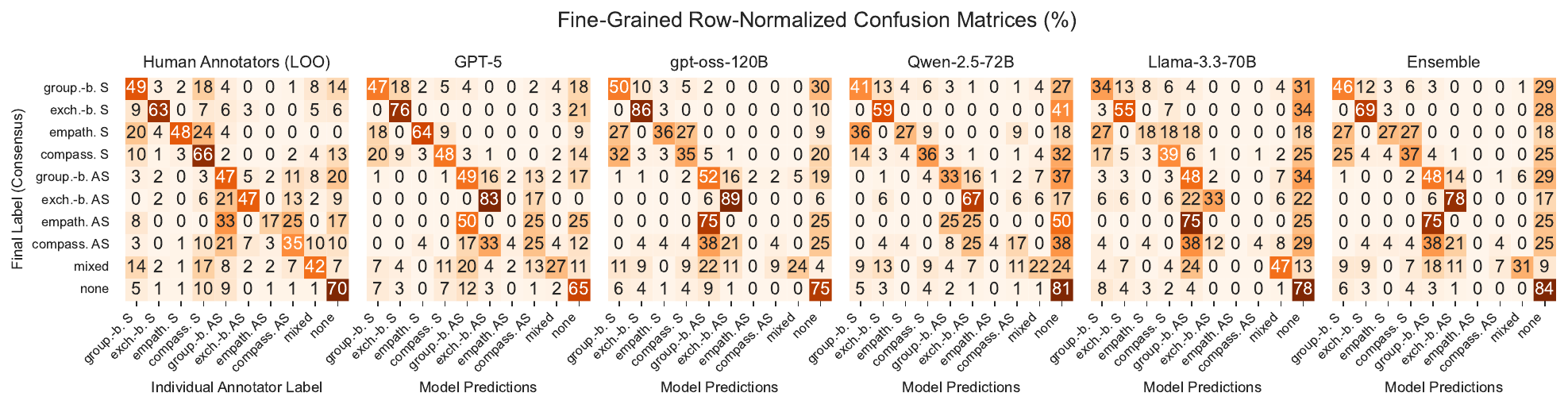}
        \caption{Fine-grained confusion matrices.}
        \label{fig:confusion-matrices-top-fine}
    \end{subfigure}

    \caption{
    Row-normalized confusion matrices (\%) for the best-performing configuration of each of the top models by macro-F1, as reported in \Cref{tab:models_performance}, on the two Migrant test sets combined (Test 1 and Test 2). Each row sums to 100, showing how items with a given reference label are distributed across predicted labels. \emph{Human Annotators (LOO)} shows aggregated leave-one-out annotator comparisons, where each annotator is compared to the consensus of the remaining annotators. The \emph{Ensemble} model aggregates predictions from Llama-3.3-70B, Qwen-2.5-72B, and gpt-oss-120B.
    }
    \label{fig:confusion-matrices-top}
\end{figure}

We analyze errors of the strongest models through their confusion matrices on the full \textit{Migrant} test set, comparing them with a human reference aggregated from leave-one-out (LOO) annotator comparisons.\footnote{Because the human panel compares each annotator to a LOO consensus, whereas the model panels compare predictions to gold labels, we use row-normalized confusion matrices so that error patterns for each reference label are more directly comparable.} \Cref{fig:confusion-matrices-top} shows the top models; full matrices appear in \Cref{fig:confusion-matrices-highlevel-full,fig:confusion-matrices-finegrained-full} in the Appendix. We consider four aspects of model error: (i) high-level confusion; (ii) confusion within (anti-)solidarity subtypes; (iii) errors in historical instances (1867-1900); and (iv) ensemble failure analysis, with particular attention to error correlation across models. %Since our substantive analysis focuses on migrant-related discourse, we restrict the error analysis to the \de{Migrant} category.

\paragraph{\justgrey{\textbf{(i) High-level confusion}}} Most models recover the broad high-level structure of the task. GPT-5 and gpt-oss-120B show the clearest separation between \emph{solidarity}, \emph{anti-solidarity}, and \emph{none}, but even these stronger models do not fully match human error patterns. Relative to the human LOO panel, models systematically overassign cases to \emph{none}, suggesting that they often fail to detect low-intensity or indirect stance expressions, despite the prompt instruction to consider even slight expressions of solidarity regardless of the main topic of the text (see \errexref{19131205-rtdkr-13.1-183-00456} %(example 1) %previously 0; 19131205-rtdkr-13.1-183-00456
in \Cref{tab:error-analysis-examples} in the Appendix).

The models also preserve \emph{mixed} much less often: for reference-\emph{mixed} cases, the share predicted as \emph{mixed} is 42\% for humans, compared with 27\% for GPT-5, 24\% for gpt-oss-120B, 22\% for Qwen-2.5, and 31\% for the ensemble; Llama-3.3-70B is the only exception, reaching 47\%. Instead, models resolve these cases into one-sided solidarity or anti-solidarity labels, suggesting that models overlook weaker cues of the opposing stance, as in \errexref{20180615-bt-19-040-03954}.%example 2. %20180615-bt-19-040-03954

%\ak{EXTEND AND QUANTIFY THIS ERROR HERE}. 
A smaller but qualitatively important error type arises when %speakers from party A quote or paraphrase anti-migrant rhetoric from party B to criticize party B and its stance. 
\ak{models attribute an anti-migrant position to the current speaker even though the speaker is criticizing %or distancing themselves from 
it.} In such cases, models sometimes misclassify the speaker's stance as anti-solidarity (e.g. Llama-3.3-70B and Qwen-2.5 in \errexreftwo{19950119-bt-13-012-04686}{20240410-bt-20-162-03269}). % ex. 3-4 previously; 19950119-bt-13-012-04686 and 20240410-bt-20-162-03269
\ak{We further refer to such cases as %\emph{misattributed anti-solidarity errors}. 
\emph{anti-solidarity errors due to misattribution} (hereafter: \emph{misattribution errors}).}

\ak{To quantify this failure, we manually audit cases from 1990 onward\footnote{\ak{We extend the audit to 1990 to obtain sufficient human-annotated data for the party-conditioned analysis; German reunification also provides a natural historical boundary.}} in the 2,184-instance human-annotated subset where the human reference is not \emph{anti-solidarity} but at least one of gpt-oss-120B, Llama-3.3-70B, or Qwen-2.5 predicts \emph{anti-solidarity}. We manually classify each candidate model prediction as (i) a misattributed anti-solidarity error, (ii) another clear anti-solidarity false positive, or (iii) a plausible disagreement with the human reference; categories (i) and (ii) together constitute all clear false positives (see \Cref{app:appendix-attribution-audit} in the Appendix for annotation details and examples). %We use this audit for two analyses: here, we focus on the subset from 2015 onward to test whether the failure could affect the recent increase in \emph{anti-solidarity} after 2015 examined in \Cref{sec:recent-overall-solidarity-distrib}. 
%Across the broader audit from 1990 onward, misattribution errors account for 15.2\% of the 912 \emph{anti-solidarity} predictions made by the three models, and all clear false positives for 19.4\%. For the robustness analysis of the recent increase in \emph{anti-solidarity} (\Cref{sec:recent-overall-solidarity-distrib}), we focus on the subset from 2015 onward, where misattribution errors account for 7.9\% of the 479 \emph{anti-solidarity} predictions and all clear false positives for 11.3\%. %We examine whether these errors affect the trend from 2015 onward in \Cref{sec:recent-overall-solidarity-distrib}.
Because our substantive analysis finds a marked recent increase in \emph{anti-solidarity} (\Cref{sec:recent-overall-solidarity-distrib}), we separately examine cases from 2015 onward: misattribution errors account for 7.9\% of all 479 \emph{anti-solidarity} predictions in this period, and all clear false positives for 11.3\%. %The full 1990-onward audit, used for the party-conditioned analysis below, yields corresponding rates of 15.2\% and 19.4\%.
}

\ak{Because our substantive analysis compares (anti-)solidarity across parties (\Cref{sec:recent-distribution-across-parties}), we additionally test whether model errors vary by party on the human-annotated subset from 1990 onward. Qwen-2.5 and Llama-3.3-70B most clearly overpredict \emph{anti-solidarity} for Grüne and Linke speakers, and a follow-up manual audit shows that these errors are largely driven by the misattribution error identified above; Qwen-2.5 shows a related skew for FDP. Full party-conditioned results are reported in \Cref{app:party-conditioned-model-performance} in the Appendix. We examine whether these skews affect the substantive 2009-2025 party patterns in \Cref{sec:recent-distribution-across-parties}.}

\paragraph{\justgrey{\textbf{(ii) Confusion within (anti-)solidarity subtypes}}} Within \solidarity{solidarity}, and focusing on the better-represented labels (\Cref{tab:instances-distribution} in the Appendix), human disagreement is greatest between \emph{group-based} and \emph{compassionate solidarity}: in the human LOO matrix, 18\% of \emph{group-based solidarity} cases are labeled as \emph{compassionate solidarity} by the held-out annotator. Models show less confusion in this direction (5-6\%), but more often predict \emph{group-based solidarity} for human-labeled \emph{compassionate solidarity} instead (e.g.\ 32\% for gpt-oss-120B). Humans also identify \emph{compassionate solidarity} more consistently overall (66\% vs.\ roughly 40\% for models).

The two subtypes can co-occur in migration debates. Human annotators reflect this ambiguity, whereas models tend to shift such cases toward \emph{group-based solidarity}, apparently relying more heavily on cues such as integration, rights, and legal inclusion (\errexrefthree{19600505-bt-03-112-00165}{20101028-bt-17-068-06571}{20200116-bt-19-140-01684}). %5-7 previously: 19600505-bt-03-112-00165, 20101028-bt-17-068-06571, 20200116-bt-19-140-01684  
However, rights language alone does not imply \emph{group-based solidarity}. When rights claims are framed in terms of vulnerability, deprivation, or need, \emph{compassionate solidarity} is the stronger reading, as it aims to place the other \enquote{in an equal position as I am}. When such claims are framed in terms of shared membership or common rights and duties, \emph{group-based solidarity} is more appropriate \citep{thijssen2012mechanical, thijssen2022s}.

A related pattern is observed for other subtypes: confusion occurs when broader integration language associated with \emph{group-based solidarity} overlaps with characteristics of other frames, as in \errexref{19650212-bt-04-163-01657}. % was 8 before, 19650212-bt-04-163-01657
Since \citet{thijssen2022s} describe the categories as conceptually close and the coding as interpretive, we treat such variation as meaningful uncertainty rather than noise.
Within \antisolidarity{anti-solidarity}, the sharpest subtype confusions appear in the rarer categories, especially \emph{empathic} and \emph{compassionate anti-solidarity}. These cases are often collapsed into \emph{group-based anti-solidarity}, suggesting that models over-rely on broad cues of protecting the interests of the native in-group (see \errexref{20190517-bt-19-102-03948}). % was 9 before; 20190517-bt-19-102-03948 
Nevertheless, some models also show confusion among the better-represented anti-solidarity categories such as \emph{exchange-based anti-solidarity} and \emph{group-based anti-solidarity} (22\% cases for Llama-3.3-70B, 21\% for humans, see \errexref{20170601-bt-18-237-02406}). %was 10 before; 20170601-bt-18-237-02406

\begin{figure}[t]
    \centering
    \begin{subfigure}[t]{\linewidth}
        \centering
        \includegraphics[width=\linewidth]{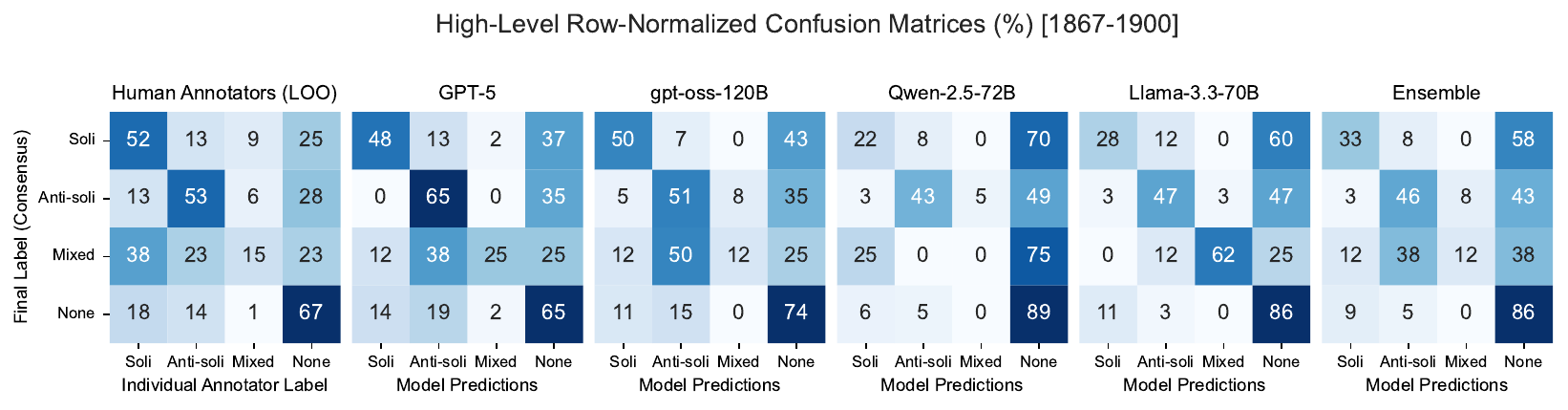}
        \caption{High-level confusion matrices.}
        \label{fig:confusion-matrices-top-high-1867-1900}
    \end{subfigure}

    \begin{subfigure}[t]{\linewidth}
        \centering
        \includegraphics[width=\linewidth]{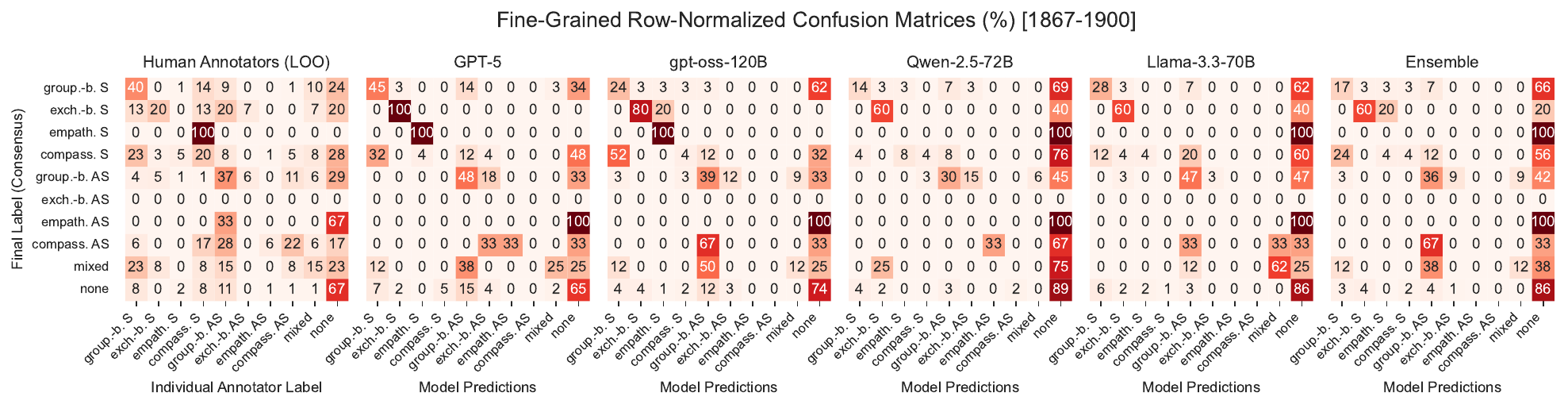}
        \caption{Fine-grained confusion matrices.}
        \label{fig:confusion-matrices-top-fine-1867-1900}
    \end{subfigure}

    \caption{Row-normalized high-level and fine-grained confusion matrices (\%) for the best-performing configurations of the top models on the combined Migrant test sets, \textbf{restricted to 1867-1900}. As in \Cref{fig:confusion-matrices-top}, rows sum to 100, \emph{Human Annotators (LOO)} denotes aggregated leave-one-out annotator comparisons, and the \emph{Ensemble} aggregates Llama-3.3-70B, Qwen-2.5-72B, and gpt-oss-120B.}
    \label{fig:confusion-matrices-top-1867-1900}
\end{figure}

\paragraph{\justgrey{\textbf{(iii) Confusion for historical instances}}} The historical sample (restricted to 1867-1900, \Cref{fig:confusion-matrices-top-1867-1900}) shows an even stronger overprediction of \emph{none}. Across nearly all models, errors for many true classes collapse into \emph{none}, more than in the human LOO panel; this is particularly visible for Qwen-2.5, Llama-3.3-70B, and the ensemble. Human disagreement also increases on the historical subset, %Human disagreement is less pronounced: annotators still disagree on labels, 
but annotators more often distribute errors across neighboring stance categories. \errexreftwo{18680416-rtndb-02-008-00361}{18860115-rtdkr-06.2-025-00082} %were 11 and 12 before, 18680416-rtndb-02-008-00361 and 18860115-rtdkr-06.2-025-00082
%indicate why this matters for historical inference. 
\ak{illustrate this challenge. Historical texts in our dataset can be lengthy and highly context-dependent, and} solidarity \ak{may be} expressed through legal-constitutional reasoning rather than overt emotional language. %and models %often 
\ak{This can make it difficult for both humans and LLMs to distinguish expressions of (anti-)solidarity from descriptive statements of laws or policies; however, models are more likely than humans to} fail to recognize such low-intensity stance at all. 
\ak{These errors may partly reflect temporal domain shift, as German parliamentary discourse has changed over time in its vocabulary and linguistic and communicative conventions, %including a shift from earlier discussion-oriented parliamentary debate toward more media-oriented contemporary discourse 
with contemporary debates also using more explicit evaluative and confrontational language \citep{burkhardt2016german}.
}

Overall, strong models reproduce the broad label structure of the task, but their errors remain systematic rather than random. As a result, trends in (anti-)solidarity and %mixed stance 
none are likely to be underestimated, especially in older instances. %samples. 
Within subtypes, \emph{group-based solidarity} is likely to absorb other solidarity claims, while rarer categories are often collapsed or disappear entirely. The ensemble mitigates these biases only modestly, if at all (e.g. it still overpredicts \emph{none} and underpredicts \emph{mixed} relative to humans). %We next examine why this is the case.
This highlights the need for correction techniques such as DSL, which we analyze in \Cref{sec:dsl}.

\paragraph{\justgrey{\textbf{(iv) Ensemble failure analysis}}} 
We begin by asking how often ensemble errors contain any useful signal at all. %This pattern is also evident in the ensemble's errors 
On the combined Test 1 and Test 2 data, the ensemble makes 1,112 errors. Of these, 743 (66.8\%) occur on instances where all models are wrong, whereas only 369 (33.2\%) occur when at least one model predicts the correct label. Thus, most ensemble failures occur on examples where none of the component models predicts the correct label. 

The remaining 369 cases suggest that complementary signal exists in principle. To %assess whether ensembling could, in principle, improve performance, 
quantify how much performance could be recovered from this signal, we construct an oracle that selects a correct model prediction whenever at least one model is correct. The oracle reaches markedly higher macro-F1 than the ensemble--0.67 (0.86) on Test 1 and 0.49 (0.74) on Test 2, compared with 0.48 (0.72) and 0.31 (0.59), respectively. This indicates that although useful complementary signal is present in the model pool, simple ensembling fails to recover much of it in practice, which explains the ensemble's substantially lower empirical F1.

A closer look at model overlap helps explain why this signal is difficult to exploit. Across the pooled Test 1 and Test 2 data, pairwise error Jaccard scores are consistently high (0.57-0.64; see \Cref{app:error-metrics} in the Appendix and \Cref{tab:pairwise_error_overlap}), indicating that models often fail on the same instances. Moreover, conditional on two models both being wrong, they predict the same wrong label in roughly half of cases (45-49\%), which means that shared errors are often reinforced rather than corrected.

Taken together, these results suggest that the main limitation is \emph{limited error diversity across models on difficult instances}. \Cref{sec:downstream-reliability} examines how the biases discussed in this section affect downstream trend estimates. %by comparing trend correlations across models and methods.
\subsection{Improving Downstream Reliability}\label{sec:downstream-reliability}

The preceding analyses identify two challenges for using LLM annotations in downstream inference. First, disagreement between human annotators does not necessarily reflect error, but may indicate intrinsic ambiguity and multiple defensible interpretations. Following work that discusses such cases under the umbrella term of label variation \citep{plank2022problem}, we therefore move beyond hard labels and compare label distributions. Second, the error analysis shows that model predictions are systematically biased, with some labels over- or under-predicted relative to human annotations. Such bias can distort downstream analyses when predictions are aggregated into temporal or group-level trends. We address both issues in this section: we first assess the similarity of model and human label distributions (\Cref{sec:label-distrib-similarity}), and then test whether debiasing with DSL improves recovery of human-derived trends (\Cref{sec:dsl}).

\subsubsection{Label distribution similarity}\label{sec:label-distrib-similarity}

%The preceding instance-level evaluation shows that %strong annotation performance does not by itself guarantee reliable downstream inference. 
%hard-label predictions might discard variation that can reflect meaningful uncertainty. Following work that treats label variation as signal rather than error \citep{plank2022problem}, we compare model and human label distributions instead of only hard predictions.

For each instance with at least three human annotations, we derive a human label distribution from the annotators' votes. For the model ensemble, we analogously construct a label distribution from the votes of the three models (Llama-3.3-70B, Qwen-2.5, and gpt-oss-120B). For individual models, we use sharp distributions that assign 100\% probability mass on the predicted label. We compare these distributions using Jensen-Shannon divergence (JSD) and total variation distance (TV), where lower values indicate higher similarity (see \Cref{app:label-distance} in the Appendix for definitions). The analysis covers both the high- and fine-grained levels. We compare model-human distances against three baselines: (i) human-human distance, averaged over ten random annotator splits; (ii) a randomized model baseline, with predictions shuffled across instances; and (iii) single-model baselines comparing each model to the human distributions.

\Cref{tab:distribution-distances} in the Appendix shows that the ensemble (TV 0.49; JSD 0.39) is closer to human label distributions (TV 0.45; JSD 0.35) than any individual model (TV 0.57-0.59; JSD 0.48-0.50) and far closer than the randomized baseline (TV 0.75; JSD 0.69), with the same pattern observed at the high level. These results indicate that strong LLMs reproduce human-like patterns of disagreement. At the same time, the ensemble-human distance does not surpass the human-human baseline. %consistent with earlier observation that model errors are correlated rather than complementary.

These results have the following implications: First, since model disagreement parallels human uncertainty, this variation should be preserved through soft labels rather than reduced to single hard predictions. Second, because models do not surpass human-human similarity and their errors remain correlated, %these soft labels still contain structured bias. As a result, model predictions cannot be used directly without correcting for bias. We address this issue in the next section.
\ak{directly aggregating these soft labels can still result in biased estimates. We therefore evaluate bias correction in the next section.}

\subsubsection{Bias Correction with DSL}\label{sec:dsl}

We validate all trend estimators considered below using the complete human-annotated subset of 2,184 Migrant instances % that were annotated by humans% (hereafter the \textit{evaluation subset} 
(see \Cref{tab:instances-distribution} in the Appendix). Trends computed from these annotations serve as the reference. %representing the best available expert-based estimate of temporal patterns. 
Because %the evaluation subset is fully human-annotated, 
human annotations are available for every instance in this subset, we simulate a partial-label setting for DSL evaluation. Specifically, %we randomly retain human labels for only a subset of instances within each decade and treat the remaining labels as unobserved. 
\ak{we repeat the partial-label simulation for 300 random seeds, with each seed defining a different sample of the human-annotated data. In each repetition, we retain approximately 20\% of the human-annotated instances overall, drawing the same number of instances without replacement from each of the 14 decades. The remaining human labels are treated as unobserved.} We then apply DSL using the corresponding decade-specific sampling probabilities and compare the resulting trends with those computed from the full human annotations. \bp{Note that this setup yields a lower bound on the realistic performance of DSL because only 20\% of the available human-annotated data is used for DSL.} 

We compare the human-derived reference trends against several alternative estimators. We first evaluate \emph{uncorrected estimators} based on raw model outputs: (i) single-model hard-label predictions from the strongest models (Llama-3.3-70B, Llama-4-Scout, Qwen-2.5, gpt-oss-120B, GPT-5), (ii) a majority-vote ensemble of the best open models (Llama-3.3-70B, Qwen-2.5, gpt-oss-120B), and (iii) soft-label distributions obtained from these three open models' outputs. We then evaluate \emph{debiased estimators} obtained by applying DSL \ak{to (iv) hard-label predictions from Llama-3.3-70B, (v) hard-label predictions from gpt-oss-120B, and (vi) the soft-label ensemble.} For each estimator, we compute temporal trends by grouping %documents 
instances by decade and computing the share of %documents 
instances assigned to each label in each decade. For DSL-based approaches, we use the bias-corrected pseudo-outcomes $\tilde{Y}_i$ directly in these aggregations. We then quantify agreement %with the reference model 
with the human-derived reference trends by computing Pearson and Spearman correlations, as well as root mean squared error (RMSE), between model-based and human-based time series. We compute correlations separately for each label and report the average across labels. \ak{For the DSL estimators, we compute these metrics separately for each of the 300 random seeds and report their mean and standard deviation (SD) across seeds. The uncorrected estimators are fixed and therefore have no seed-to-seed variation.}

%\Cref{tab:validation-dsl-full} shows that the soft-label ensemble corrected with DSL recovers human-derived temporal trends more accurately than other approaches, including a single-model DSL estimator, at both the high-level and fine-grained levels.

\ak{\Cref{tab:validation-dsl-full-humansoft} summarizes performance against human soft-label trends (hereafter: \textit{HumanSoft}), while \Cref{tab:validation-dsl-full-humanmaj} in the Appendix reports the corresponding results against human majority-vote trends (hereafter: \textit{HumanMaj}) as a robustness check.}

\begin{comment}
\begin{table}%[ht]
\centering
\small
\begin{tabular}{lrrr}
\toprule
\textbf{Model / Estimator} & \textbf{Pearson} & \textbf{Spearman} & \textbf{RMSE} \\
\midrule
\multicolumn{4}{l}{\textbf{High-level trends}} \\
\midrule
Llama-3.3-70B                     & 0.87 & 0.86 & 0.08 \\
Llama-4-Scout                     & 0.74 & 0.76 & 0.08 \\
Qwen-2.5                          & 0.72 & 0.71 & 0.11 \\
gpt-oss-120B                      & 0.84 & 0.80 & 0.07 \\
GPT-5                             & 0.82 & 0.82 & 0.07 \\
Majority vote ensemble            & 0.88 & 0.87 & 0.08 \\
DSL (Llama-3.3-70B)               & 0.85 & 0.83 & 0.06 \\
DSL (soft-label ensemble)         & 0.89 & 0.91 & 0.06 \\
\midrule
\multicolumn{4}{l}{\textbf{Fine-grained trends}} \\
\midrule
Llama-3.3-70B                     & 0.64 & 0.58 & 0.09 \\
Llama-4-Scout                     & 0.58 & 0.58 & 0.07 \\
Qwen-2.5                          & 0.36 & 0.38 & 0.11 \\
gpt-oss-120B                      & 0.44 & 0.52 & 0.10 \\
GPT-5                             & 0.56 & 0.50 & 0.08 \\
Majority vote ensemble            & 0.64 & \textbf{0.67} & 0.10 \\
DSL (Llama-3.3-70B)               & 0.50 & 0.49 & 0.07 \\
DSL (soft-label ensemble)         & \textbf{0.68} & \textbf{0.67} & \textbf{0.04} \\
\bottomrule
\end{tabular}
\caption{Comparison with human-based decade-level trends across estimators. The table reports average Pearson and Spearman correlations and RMSE (\textit{reference:} human majority vote).}
\label{tab:validation-dsl-full}
\end{table}

\end{comment}

\begin{table}[t]
\centering
\small
\begin{tabular}{lccc}
\toprule
\ak{\textbf{Model / Estimator}} & \ak{\textbf{Pearson}} & \ak{\textbf{Spearman}} & \ak{\textbf{RMSE}} \\
\midrule
\multicolumn{4}{l}{\ak{\textbf{High-level trends}}} \\
\midrule
\ak{Llama-3.3-70B} & \ak{0.899} & \ak{0.881} & \ak{0.082} \\
\ak{Llama-4-Scout} & \ak{0.777} & \ak{0.802} & \ak{0.082} \\
\ak{Qwen-2.5} & \ak{0.757} & \ak{0.744} & \ak{0.118} \\
\ak{gpt-oss-120B} & \ak{0.881} & \ak{0.822} & \ak{0.068} \\
\ak{GPT-5} & \ak{0.855} & \ak{0.829} & \ak{0.070} \\
\ak{Majority-vote ensemble} & \ak{\textbf{0.913}} & \ak{\textbf{0.908}} & \ak{0.082} \\
\ak{Soft-label ensemble} & \ak{0.879} & \ak{0.839} & \ak{0.077} \\
\ak{DSL (Llama-3.3-70B)} & \ak{0.848 $\pm$ 0.041} & \ak{0.820 $\pm$ 0.046} & \ak{0.057 $\pm$ 0.006} \\
\ak{DSL (gpt-oss-120B)} & \ak{0.845 $\pm$ 0.044} & \ak{0.817 $\pm$ 0.047} & \ak{0.056 $\pm$ 0.007} \\
\ak{DSL (soft-label ensemble)} & \ak{0.898 $\pm$ 0.030} & \ak{0.858 $\pm$ 0.039} & \ak{\textbf{0.042 $\pm$ 0.005}} \\
\midrule
\multicolumn{4}{l}{\ak{\textbf{Fine-grained trends}}} \\
\midrule
\ak{Llama-3.3-70B} & \ak{0.478} & \ak{0.488} & \ak{0.091} \\
\ak{Llama-4-Scout} & \ak{0.483} & \ak{0.538} & \ak{0.065} \\
\ak{Qwen-2.5} & \ak{0.392} & \ak{0.446} & \ak{0.104} \\
\ak{gpt-oss-120B} & \ak{0.491} & \ak{0.481} & \ak{0.098} \\
\ak{GPT-5} & \ak{0.494} & \ak{0.486} & \ak{0.076} \\
\ak{Majority-vote ensemble} & \ak{0.557} & \ak{0.530} & \ak{0.089} \\
\ak{Soft-label ensemble} & \ak{0.537} & \ak{0.575} & \ak{0.086} \\
\ak{DSL (Llama-3.3-70B)} & \ak{0.601 $\pm$ 0.069} & \ak{0.546 $\pm$ 0.064} & \ak{0.059 $\pm$ 0.005} \\
\ak{DSL (gpt-oss-120B)} & \ak{0.605 $\pm$ 0.067} & \ak{0.557 $\pm$ 0.061} & \ak{0.059 $\pm$ 0.005} \\
\ak{DSL (soft-label ensemble)} & \ak{\textbf{0.756 $\pm$ 0.045}} & \ak{\textbf{0.720 $\pm$ 0.047}} & \ak{\textbf{0.040 $\pm$ 0.004}} \\
\bottomrule
\end{tabular}
\caption{\ak{Comparison with human-based decade-level trends across estimators. For DSL estimators, values are reported as mean $\pm$ standard deviation (SD) across 300 random seeds. %uncorrected estimators are fixed and therefore have no seed-to-seed variation.
\textit{Reference:} human soft labels. Results against human majority vote are reported in \Cref{tab:validation-dsl-full-humanmaj} in the Appendix.}}
\label{tab:validation-dsl-full-humansoft}
\end{table}

%\textbf{At the high level}, the strongest single models and ensemble variants already achieve high agreement with human trends, with Pearson and Spearman correlations above 0.80 and low RMSE. In this setting, DSL shows moderate but consistent improvements: DSL applied to the soft-label ensemble achieves the highest correlations (Pearson 0.89; Spearman 0.91) and the lowest RMSE (0.06).

\ak{\textbf{At the high level}, the strongest single models and ensemble variants already show high agreement with HumanSoft trends, with Pearson and Spearman correlations above 0.75 and low RMSE. The majority-vote ensemble achieves the highest average correlations (Pearson 0.913; Spearman 0.908). Soft-label DSL shows comparable correlations (Pearson $0.898\pm0.030$; Spearman $0.858\pm0.039$) but substantially lowers RMSE to $0.042\pm0.005$, compared with 0.077 for the raw soft-label ensemble and 0.082 for both Llama-3.3-70B and the majority-vote ensemble. The single-model DSL variants also reduce RMSE substantially (0.057 for Llama-3.3-70B and 0.056 for gpt-oss-120B), though less than soft-label DSL. Thus, in this setting, DSL shows moderate but consistent improvements, mainly in RMSE.}

\begin{figure}%[!htb]
    \centering

    % =====================================================
    % High-level
    % =====================================================
    {High-level trends}\\[0.6em]

    \begin{subfigure}[t]{0.32\linewidth}
        \centering
        \includegraphics[width=\linewidth]{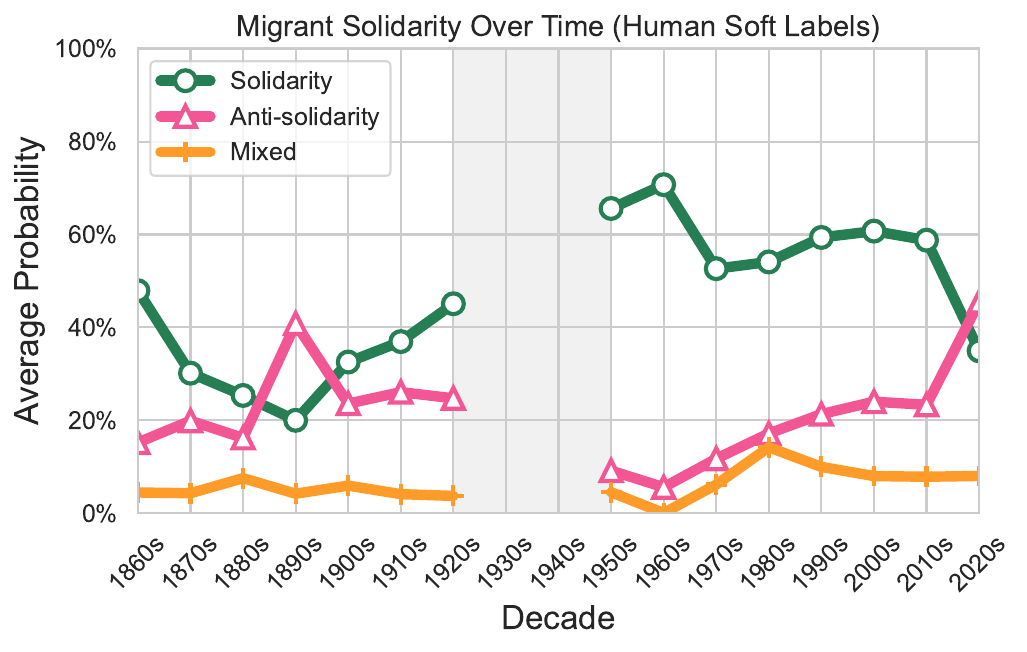}
        \caption{Human soft labels}
    \end{subfigure}
    \hfill
    \begin{subfigure}[t]{0.32\linewidth}
        \centering
        \includegraphics[width=\linewidth]{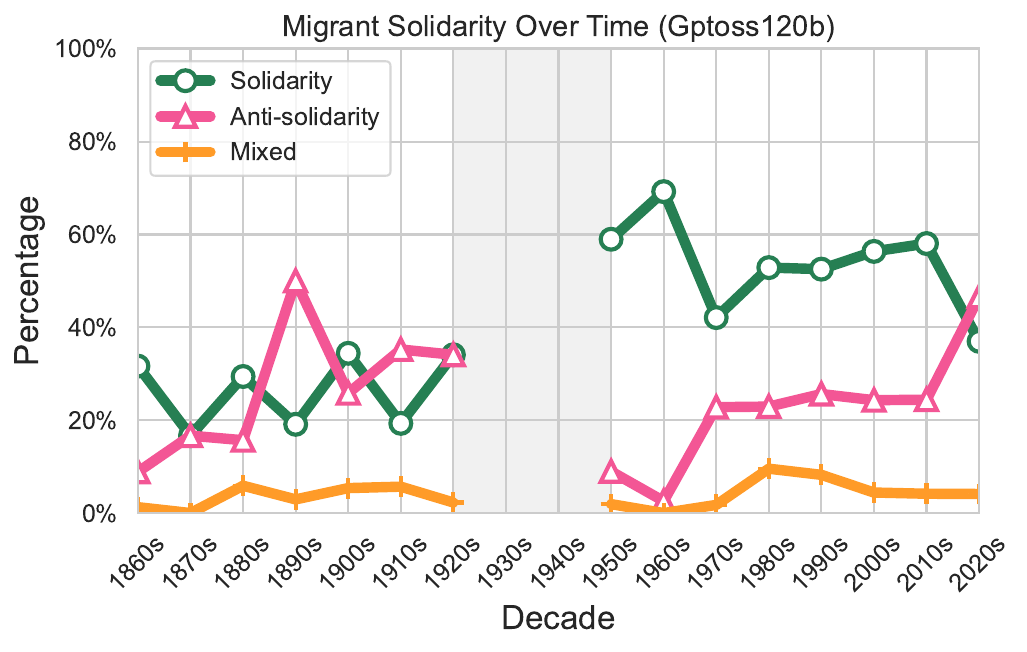}
        \caption{gpt-oss-120B}
    \end{subfigure}
    \hfill
    \begin{subfigure}[t]{0.32\linewidth}
        \centering
        \includegraphics[width=\linewidth]{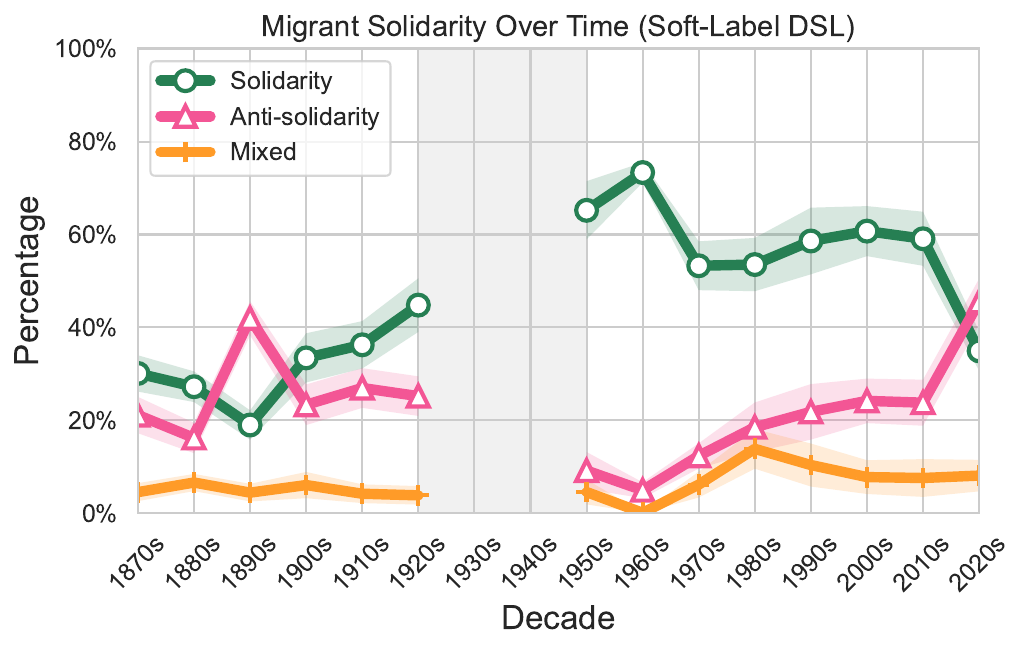}
        \caption{DSL (soft labels)}
    \end{subfigure}

    \vspace{0.6em}

    % =====================================================
    % Fine-grained
    % =====================================================
    {Fine-grained trends}\\[0.6em]

    \begin{subfigure}[t]{0.49\linewidth}
        \centering
        \includegraphics[width=\linewidth]{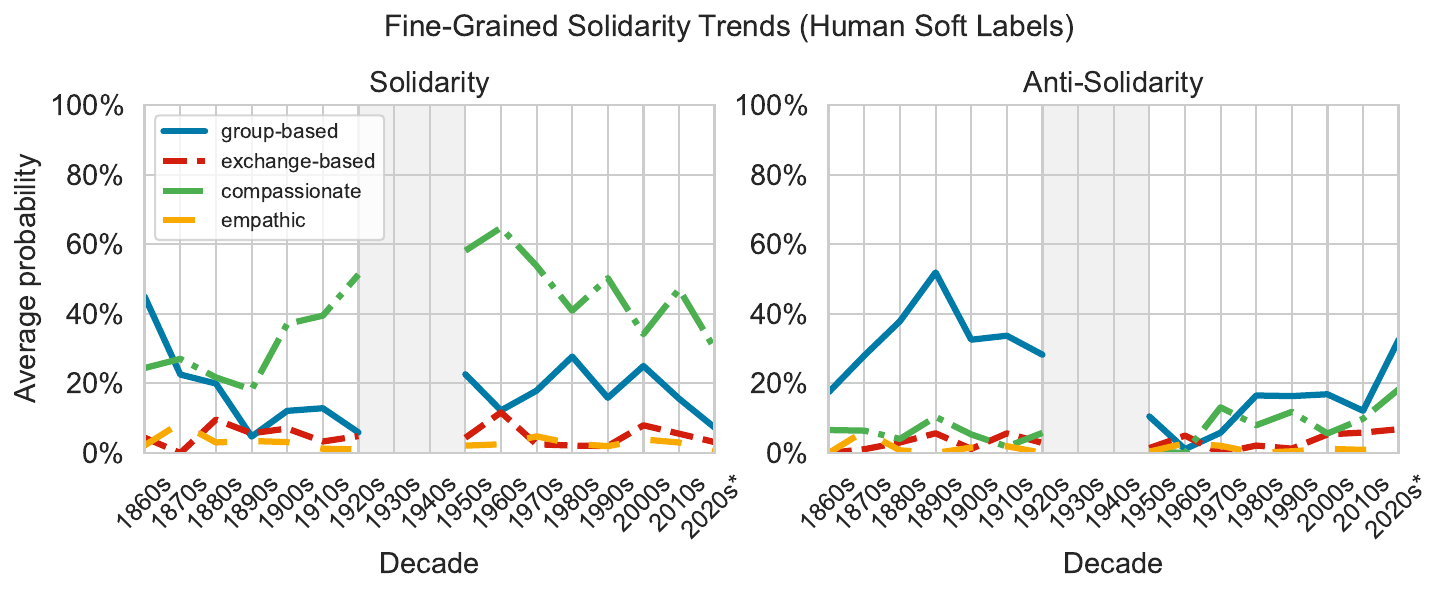}
        \caption{Human soft labels}
    \end{subfigure}
    \hfill
    \begin{subfigure}[t]{0.49\linewidth}
        \centering
        \includegraphics[width=\linewidth]{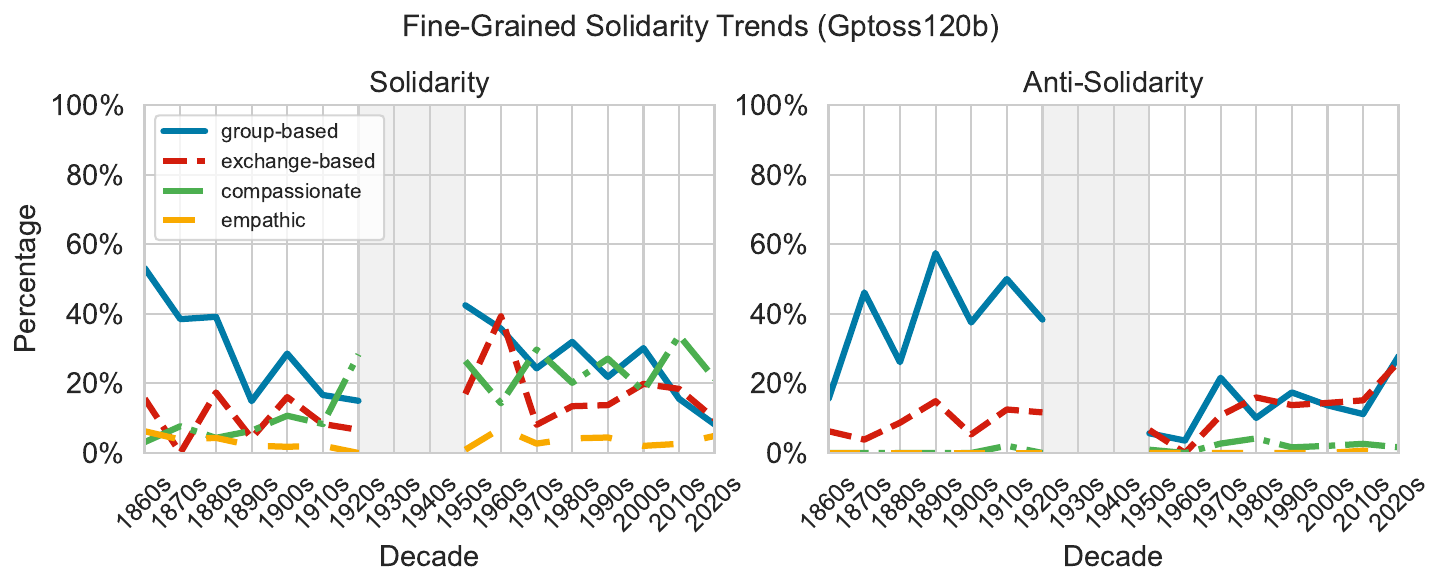}
        \caption{gpt-oss-120B}
    \end{subfigure}

    \begin{subfigure}[t]{0.49\linewidth}
        \centering
        \includegraphics[width=\linewidth]{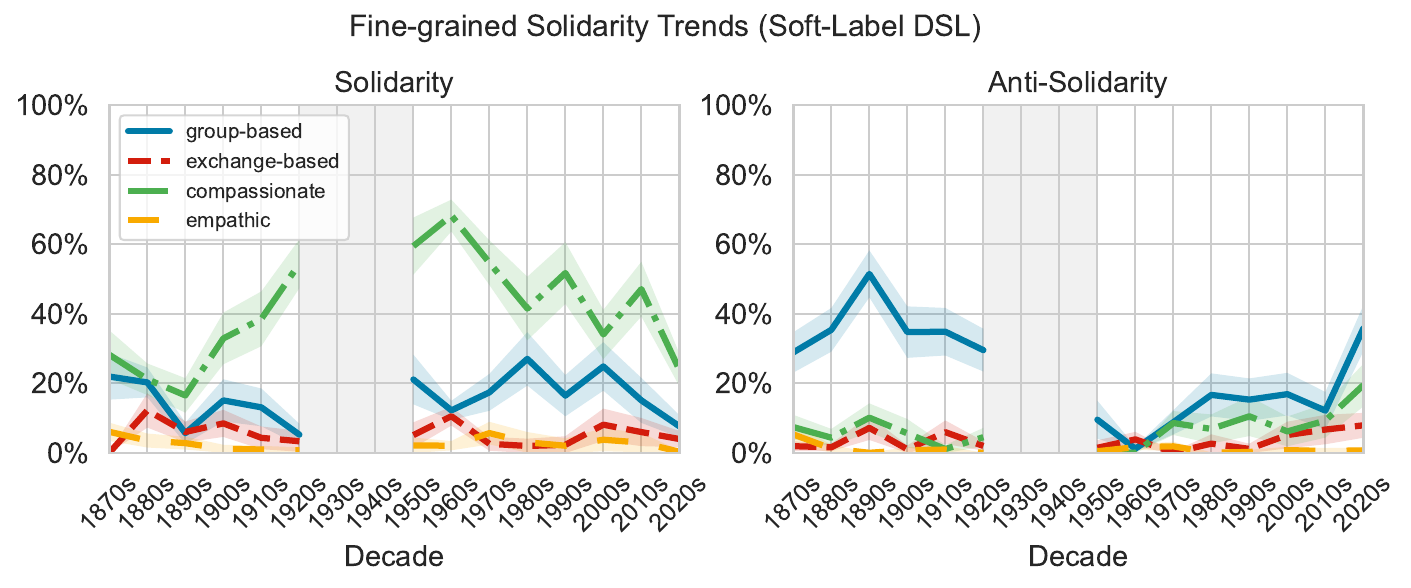}
        \caption{DSL (soft labels)}
    \end{subfigure}

    \caption{
    High-level and fine-grained (anti-)solidarity trends for the \textit{Migrant} category under selected labeling and correction strategies discussed in \Cref{sec:dsl}, based on the %\textbf{2k evaluation subset}. 
    \textbf{2,184-instance human-annotated subset}. Trends are computed as label shares within each decade (see definitions in \Cref{box:trend-definitions}). The period from 1930 to 1949 in each panel (grey shaded area) is excluded due to limited data availability during and immediately after the NS dictatorship. \ak{For DSL, lines show the mean estimate across 300 simulated runs. In each run, approximately 20\% of the human-annotated instances are retained as labeled data, while the remaining human labels are treated as unobserved. Colored shaded areas show the mean $\pm$ the decade-specific SD across runs.} \Cref{fig:highlevel-dsl-all,fig:finegrained-dsl-all} in the Appendix show the corresponding trends for the remaining configurations.
    }
    \label{fig:dsl-trends}
\end{figure}

%In contrast, \textbf{fine-grained trends} are substantially harder to approximate from raw model predictions alone. Single-model and ensemble estimators show lower correlations and higher error: notably, even models that achieve the highest macro F1 scores in \Cref{sec:overall-model-performance} show only moderate agreement with human-derived fine-grained trends (for example, gpt-oss-120B reaches a Pearson correlation of 0.44 with an RMSE of 0.10, and the majority-vote ensemble reaches 0.64 with an RMSE of 0.10). This indicates that strong instance-level performance does not guarantee accurate recovery of fine-grained temporal trends. Here, DSL has a clear effect, particularly in reducing error: DSL applied to soft-label ensembles achieves the lowest RMSE (0.04) and higher correlations (Pearson 0.68) than all uncorrected approaches.

In contrast, \textbf{fine-grained trends} are substantially harder to recover from raw model predictions alone. \ak{Even models that achieve strong instance-level performance in \Cref{sec:overall-model-performance} show only moderate agreement with HumanSoft trends; for example, gpt-oss-120B reaches a Pearson correlation of 0.491 with an RMSE of 0.098,} which confirms that strong instance-level performance does not necessarily translate into accurate recovery of fine-grained temporal trends. \ak{Here, soft-label DSL has a much clearer effect, increasing average Pearson to $0.756\pm0.045$ and Spearman to $0.720\pm0.047$, while reducing RMSE to $0.040\pm0.004$, compared with 0.086 for the raw soft-label ensemble. The RMSE reduction is observed for all eight fine-grained labels (\Cref{tab:per-label-fine-humansoft} in the Appendix).}

\ak{The HumanMaj results in \Cref{tab:validation-dsl-full-humanmaj} in the Appendix support the same overall conclusion: DSL consistently reduces RMSE at both levels, with soft-label DSL again achieving the lowest RMSE. At the high level, the three DSL variants show comparable correlations, whereas at the fine-grained level the hard-label DSL variants based on Llama-3.3-70B and gpt-oss-120B achieve higher correlations with HumanMaj than soft-label DSL. This pattern is consistent with the respective training targets: the hard-label DSL variants use human majority labels, whereas soft-label DSL is trained on human label distributions and correspondingly performs better against HumanSoft. We use HumanSoft as the primary reference because it preserves variation in annotator judgments rather than collapsing disagreement into a single majority label, as motivated by our discussion %of label variation 
in \Cref{sec:label-distrib-similarity}. %while HumanMaj serves as a robustness check.
}

\Cref{fig:dsl-trends} %and \Cref{fig:finegrained-dsl} 
illustrates these results. First, DSL reduces overuse of the None category and correspondingly increases estimated trends for solidarity and anti-solidarity. %for example for Llama-3.3-70B. This pattern is also reflected in \Cref{tab:per_label_trend_comparison} in the Appendix, which compares per-label correlations with human-annotated decade-level trends for Llama-3.3-70B and soft-label DSL. There, DSL reduces the RMSE for None from 0.13 to 0.07.
\ak{The per-label results in \Cref{tab:per-label-high-humansoft} in the Appendix show, for example, that RMSE for None decreases from 0.138 for Llama-3.3-70B and 0.124 for the raw soft-label ensemble to $0.044\pm0.009$ with soft-label DSL. The main exception is Mixed, for which the raw estimators already recover the human trend well and DSL does not improve performance.}

\begin{center}
\begin{definitionbox}{Label Shares Over Time}{trend-definitions}

Each speech belongs to a time period $t$ (e.g.\ a decade or a year) with $N_t$ speeches and receives a high-level or fine-grained label $\ell$. 
The estimated label share $\hat p_{t,\ell}$ in period $t$ is defined as follows:

\begin{enumerate}[label=(\alph*)]

\item \textbf{Hard-label trends} (e.g.\ human majority vote\footnotemark, gpt-oss-120B):  
$\hat p_{t,\ell} = \frac{N_{t,\ell}}{N_t}$,
where $N_{t,\ell}$ denotes the number of speeches in period $t$ assigned label $\ell$.

\item \textbf{Soft-label trends}:  
$\hat p^{\text{soft}}_{t,\ell} = \frac{1}{N_t}\sum_{i\in t} Y_{i,\ell}$,
where $Y_{i,\ell}$ denotes the probability of label $\ell$ for speech $i$, obtained from the distribution of annotator or model labels.

\item \textbf{DSL (soft-label) trends:}  
$\hat p^{DSL}_{t,\ell} = \frac{1}{N_t}\sum_{i\in t}\tilde{Y}_{i,\ell}$,
where $\tilde{Y}_{i,\ell}$ denotes the DSL-adjusted class score for label $\ell$ in speech $i$, as defined in \Cref{eq:dsl-pseudo-outcome} in \Cref{sec:methodology}.

\end{enumerate}

\end{definitionbox}
\end{center}

\footnotetext{Reported in \Cref{fig:finegrained-dsl-all} in the Appendix.}

At the fine-grained level, raw model outputs, including gpt-oss-120B, Llama-3.3-70B, Qwen-2.5, and their ensemble variants, show unstable trends, with dominance shifting between \textit{group-based}, \textit{exchange-based}, and \textit{compassionate} categories across decades. %DSL achieves more stable trends that follows with the human reference more closely, particularly within solidarity. 
Most notably, all raw model trends overemphasize \textit{group-based solidarity} at the expense of \textit{compassionate solidarity}. \ak{Soft-label DSL corrects much of this imbalance and more closely recovers the dominance of \textit{compassionate solidarity} observed in the human annotations.} %\Cref{tab:per_label_trend_comparison} in the Appendix shows that these corrections correspond to substantial error reductions, with RMSE for \textit{group-based solidarity} decreasing from 0.20 to 0.06 and for \textit{compassionate solidarity} from 0.20 to 0.08. A similar pattern is observed for fine-grained anti-solidarity. In raw model outputs, anti-solidarity is often limited to a single dominant subtype or disappears entirely; for example, \textit{empathic} and \textit{compassionate anti-solidarity} are absent from trends derived from Llama-3.3-70B predictions. DSL restores these subtypes and recovers differentiation between \textit{group-based} and \textit{exchange-based anti-solidarity}.
\ak{These changes correspond to substantial reductions in RMSE (\Cref{tab:per-label-fine-humansoft} in the Appendix): for \textit{group-based solidarity}, RMSE decreases from 0.180 for Llama-3.3-70B and 0.117 for the raw soft-label ensemble to $0.055\pm0.011$ with DSL; for \textit{compassionate solidarity}, it decreases from 0.197 and 0.219, respectively, to $0.073\pm0.014$.} In raw model outputs, anti-solidarity is often %limited to 
concentrated in a single dominant subtype, %or disappears entirely; 
while other subtypes are scarcely represented or absent. For example, in Llama-3.3-70B predictions, anti-solidarity is dominated by \textit{group-based anti-solidarity}, whereas \textit{empathic} and \textit{compassionate anti-solidarity} are essentially absent across decades. DSL %restores 
assigns non-zero mass to these subtypes and recovers differentiation between \textit{group-based} and \textit{exchange-based anti-solidarity}.

\ak{However, soft DSL (\Cref{tab:per-label-high-humansoft,tab:per-label-fine-humansoft} in the Appendix) shows substantial seed-to-seed variation for some labels; for example, Pearson is $0.747\pm0.113$ for Mixed and $0.730\pm0.230$ for Empathic AS. The repeated-sampling diagnostics (see \Cref{tab:dsl-coverage-diagnostics-high,tab:dsl-coverage-diagnostics-fine} in the Appendix) indicate that DSL instability appears to be mainly driven by data sparsity. Lower-prevalence labels are more likely to be missed or poorly represented in a sample, making DSL estimates more sensitive to the sampled seed. This problem becomes more pronounced when the label shows little temporal variation in the human reference data (column \emph{HumanSoft trend SD}) because variation due to random seed is then large relative to the underlying human signal (column \emph{Seed variation / human signal}). For example, group-based anti-solidarity and group-based solidarity have similar prevalence in the complete human-annotated subset (13.4\% and 12.5\%), but group-based solidarity has a weaker HumanSoft trend SD (0.069 vs. 0.136) and a higher Pearson SD (0.095 vs. 0.034).%This is also visible at the extremes: compassionate solidarity has 32.7\% prevalence, a 0.0\% sampling-miss rate (share of seed-decade samples in which the label is missed despite being present in the full human data), and a Pearson SD of 0.040, whereas empathic anti-solidarity has only 0.5\% prevalence, a 47.1\% sampling-miss rate, and a Pearson SD of 0.230.
}

\ak{Despite this instability, soft DSL improves most rare labels over the raw soft-label ensemble, %(\Cref{tab:per-label-high-humansoft,tab:per-label-fine-humansoft}), 
especially when the raw model performs poorly on a given label. For example, empathic solidarity has a prevalence of only 1.9\%, yet soft DSL improves Pearson from 0.036 to $0.556\pm0.195$ and reduces RMSE from 0.029 to $0.023\pm0.005$. Mixed is the main exception. Here, the raw soft-label ensemble already recovers the label reasonably well (Pearson $=0.757$, RMSE $=0.024$),\footnote{The raw soft-label ensemble is weaker than Llama-3.3-70B for Mixed %A separate comparison shows that this is mainly due to adding 
because of Qwen-2.5: the gpt-oss-120B $+$ Llama-3.3-70B ensemble reaches Pearson $=0.914$ and RMSE $=0.018$, % whereas the three-model ensemble including Qwen-2.5 reaches Pearson $=0.757$ and RMSE $=0.024$. 
compared with 0.757 and 0.024 for the three-model ensemble.
This %however, 
does not explain the further deterioration on Mixed after DSL.%since soft DSL is slightly worse than the raw three-model ensemble on Pearson and RMSE.
} while soft DSL achieves Pearson $0.747\pm0.113$ and RMSE $0.029\pm0.006$. Thus, DSL is most useful when the raw model poorly recovers a rare label, but can add noise when the true temporal signal %is weak 
has little variation and the raw model already captures it well.}

In sum, these results suggest that combining soft-label model outputs with DSL is a viable strategy for studying long-run trends when full human annotation is not feasible, %especially on the fine-grained level. 
\ak{with the clearest benefits for fine-grained inference; at the high level, the broad temporal pattern is already recovered well by uncorrected estimates, while DSL's main advantage is lower absolute estimation error.} 
\ak{The repeated-sampling analysis shows that DSL becomes less stable for rare labels with small temporal variation. Accordingly, trends for the rarest labels should be interpreted with caution.} \bp{Still, DSL improved average trend recovery for most rare labels and subsequent analyses are based on DSL using not only 20\% but all of the available human labels, thus reducing the sparsity underlying this instability.} %\citet{egami2023using} also note that optimal sampling strategies for gold-standard labels remain an open question.
\section{Temporal Trends in Solidarity Towards Migrants}\label{sec:sociological-analysis}

In this section, we analyze temporal trends in the full \textit{Migrant} dataset (63k instances) using DSL-corrected soft labels. We relate the resulting patterns to findings from political science, sociology, and history on migration to assess whether the temporal trends recovered by the DSL-corrected estimates are consistent with established interpretations.

For these analyses, DSL uses the decade-stratified human-annotated subset of 2,184 \textit{Migrant} instances (3.5\% of the full corpus) to estimate bias-corrected label distributions for the full corpus of 63k speeches. Since human annotations were obtained under a decade-stratified sampling design (see \Cref{sec:data-annotation}), we estimate DSL with decade covariates and decade-specific labeling probabilities $\pi_i$, and reuse the resulting pseudo-outcomes $\tilde{Y}_i$ for all analyses that condition only on time, including the sub-period analyses introduced in \Cref{sec:notable-trends}. For analyses that condition on additional variables, such as party or keyword, we re-estimate DSL on the relevant analysis sample, since the prediction function $\hat g(Q_i, X_i)$ changes when additional covariates are included. In these analyses, party or keyword indicators enter the DSL prediction step as additional covariates alongside decade indicators, while labeling probabilities $\pi_i$ remain defined at the decade level, consistent with the annotation design.

We first present the DSL-corrected trends for the full dataset, summarizing the broad long-term pattern in parliamentary discourse about migrants and briefly assessing whether the aggregate trends are robust to variation in keyword composition. We then identify two periods with particularly notable shifts in solidarity--the post-war period and the recent past--which we analyze in more detail in \Cref{sec:postwar} and \Cref{sec:recent}.

\begin{figure}[!htb]
    \centering
    \subfloat[High-level categories. Percentages may not sum to 100\% as \enquote{None} instances are omitted but included in calculations.]{
        \includegraphics[width=0.37\linewidth]{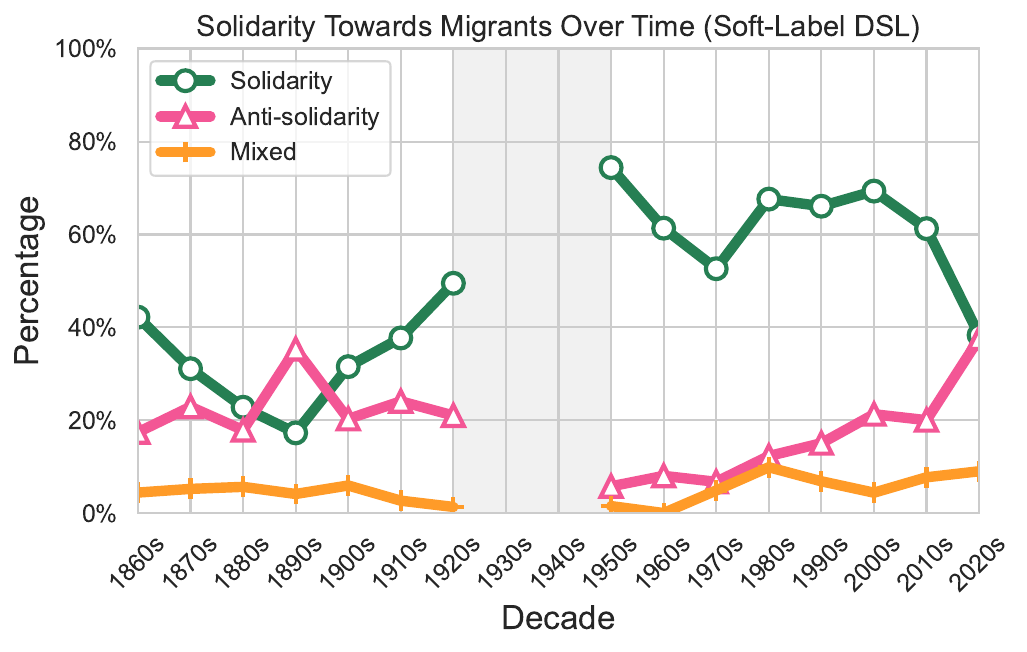}
        \label{fig:dsl-fullscale-highlevel}
    }
    \hfill
    \subfloat[Fine-grained subcategories.]{
        \includegraphics[width=0.57\linewidth]{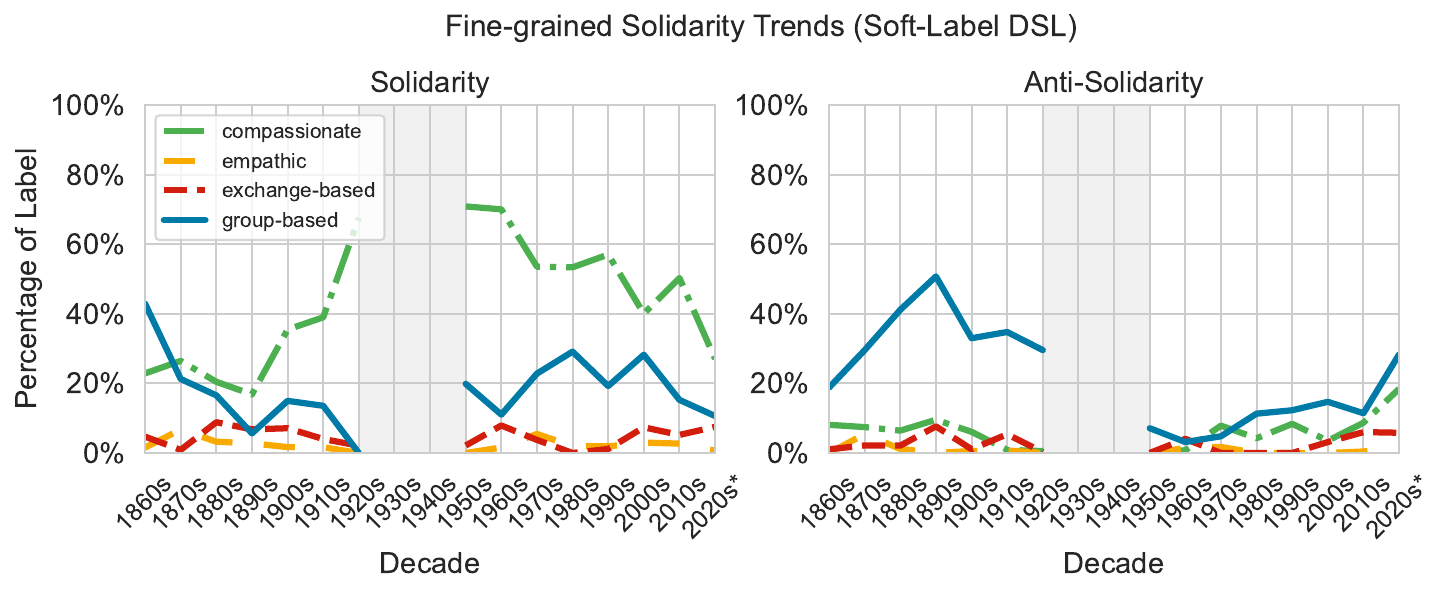}
        \label{fig:dsl-fullscale-finegrained}
    }
    %\vspace{-0.1cm}
    \caption{Solidarity trends for \textit{Migrant} speeches computed using DSL-corrected soft labels (see definition (c) in \Cref{box:trend-definitions}) on the \textbf{full 63K dataset}. \Cref{fig:dsl-fullscale-highlevel} shows the high-level (anti-)solidarity trends, while \Cref{fig:dsl-fullscale-finegrained} shows distribution of fine-grained (anti-)solidarity subtypes over time. The period 1930–1949 (grey area) is excluded due to limited data during and after the Nazi dictatorship.}
    \label{fig:dsl-full-dataset}
\end{figure}

\subsection{Overall Solidarity Trends with DSL, 1867-2025}

The overall trends from 1867 to 2025 are shown in \Cref{fig:dsl-full-dataset}. Across most of the observable period, and especially from the beginning of the 20th century through the 2010s, solidarity exceeds anti-solidarity in parliamentary discourse about migrants. This pattern is driven mainly by \emph{compassionate solidarity}, with \emph{group-based solidarity} playing a less prominent but persistent role. Solidarity peaks in the postwar decades, while anti-solidarity remains comparatively low until the late twentieth century. From the 1980s onward, anti-solidarity rises (especially its \emph{group-based}, \emph{compassionate}, and \emph{exchange-based} subtypes), with the sharpest jump occurring between the 2010s and 2020s. %Whereas anti-solidarity before 1900 is dominated mainly by \emph{group-based anti-solidarity}, the recent increase is spread more evenly across \emph{group-based}, \emph{compassionate}, and \emph{exchange-based anti-solidarity}.

Before turning to specific historical periods, we assess whether these aggregate trends are robust to changes in keyword composition over time.

\subsubsection{%Trend Stability Across Keywords}
Robustness to Keyword Composition}\label{sec:trend-stability} % comments: Aida: maybe even move up here to the main text? 
% Benjamin: Maybe. I think in general this stability analysis seems more like a method point that could be included in the previous section whereas the historic contextualization and discussion should remain here (and maybe be slightly expanded, in line with Ole's suggestion).
As \Cref{fig:keywords-distrib-over-time,fig:1867-2025-keywords} in the Appendix show, individual keywords vary widely in temporal distribution in the balance of solidarity vs.\ anti-solidarity associated with them. To ensure our long-term trends are not driven by such effects, we conduct two robustness tests.

First, we conduct a stability test by randomly sampling 200 subsets of at least five keyword groups (grouping inflectional variants). For each subset, we re-estimate the high-level DSL trends (as defined in \Cref{sec:dsl}) and compare the resulting time series to the full-sample trend using Pearson's correlation coefficient. The trends remain broadly stable, with mean correlations of 0.93 for solidarity, 0.88 for anti-solidarity, and 0.82 for mixed; for more detail on keywords and the stability test, see \Cref{app:appendix-keywords} in the Appendix. Second, we restrict the corpus to the two most frequent and temporally widespread terms, \de{Ausländer}{foreigners} (19.7k mentions) and \de{Flüchtlinge}{refugees} (12k), and re-estimate DSL on this reduced sample. Both checks confirm that overall solidarity and anti-solidarity trends are preserved both under this restricted vocabulary (see \Cref{fig:1867-2022-general-keywords} in the Appendix) and across different keyword selections.

\enlargethispage{2\baselineskip}

\subsubsection{Notable Trends in 1867-2025}\label{sec:notable-trends} Inspecting \Cref{fig:dsl-full-dataset}, we identify three periods with notable shifts in solidarity: 1890-1900s, the post-war period, and the recent period after 2010. We note that the 1930s and 1940s are outliers, as the parliament was swiftly sidelined by the Nazi regime and the corpus contains only 109 available %samples 
instances from 1933 to 1949. %mostly concern inter-state relations (mentioning, e.g., \enquote{foreign creditors}) that fall into the \enquote{None} category. 
Hence, we exclude these data points from further analysis.

The first of these, 1890-1900, is difficult to interpret with confidence. In this early part of the series, even the strongest model in our ensemble performs worse than in later decades in terms of macro-F1, and agreement between human annotators is also lower (see \Cref{fig:macro-f1-kappa-over-time} in the Appendix). Although DSL corrects the resulting estimates, we do not consider this early trend sufficiently reliable for detailed substantive analysis. This period also contains fewer instances than the post-war and recent periods that we analyze in detail below (see \Cref{fig:instances-per-year}).

For the two remaining periods, 1949–1957 and 2009–2025, we take a more detailed look.

\subsection{Germany after the Second World War, 1949-1957}
\label{sec:postwar}

The Second World War triggered the largest wave of forced migration in European history, with %millions of Jews murdered and 
tens of millions of people %both non-Germans and ethnic Germans, 
displaced or resettled across Central and Eastern Europe \citep{schulze2017commemoration}. %culminating in the mass expulsion of Germans at the war's end 
The affected populations %displaced by the war 
were heterogeneous---refugees, displaced persons (DPs), forced laborers---and---forming the largest group of expellees after 1945 in both Europe as a whole and Germany---ethnic Germans. For those who settled in postwar Germany, integration posed political and societal challenges shaped by regional disparities, economic hardship, and tensions with locals \citep{chevalier2024forced}.

This subsection examines the early postwar period (1949-1957), when the Federal Republic institutionalized solidarity as part of nation-building. This period is analytically important because solidarity %was no longer merely a moral gesture, but 
became a tool of governance, embedded in laws like the \de{Lastenausgleich (LAG)}{Equalization of Burdens}, which redistributed resources between the war-damaged and the less affected to promote social cohesion and democratic legitimacy \citep{kittel2020stiefkinder}. %from Ole: Based on my reading of the later sections, I think we should stress here that the LAG and the BVFG apply to ethnic Germans and are beneficial for them (and only them, I think). This is an important difference to other migrants and also imporant for the keyword analysis.
These measures applied specifically to \de{Volkszugehörige}{ethnic Germans}, excluding other migrant groups. %from the benefits provided by these laws
Since solidarity was neither static nor uniform, shaped by diverse migrant groups, competing ideologies, and contested notions of deservingness \citep{ther1996integration}, we employ Thijssen's typology of solidarity \citep{thijssen2012mechanical} to guide our analysis. By distinguishing identity-, need-, and contribution-based forms, we analyze how solidarity was constructed and sustained amid institutional strain.

\subsubsection{Overall solidarity distribution}

\begin{figure}[!htbp]
    \centering
    \includegraphics[width=0.85\linewidth]{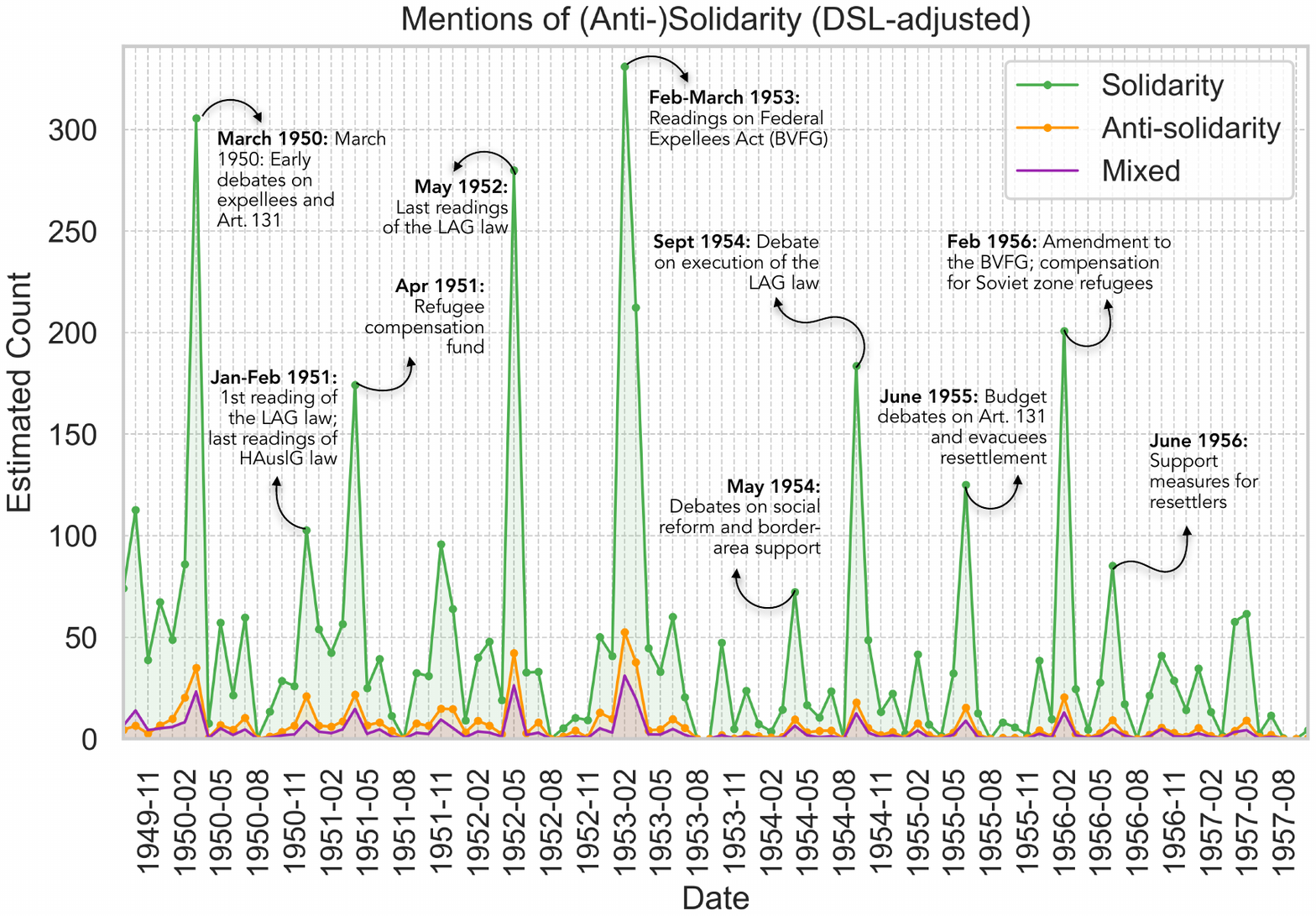}
    \caption{Estimated counts of (anti-)solidarity mentions in Bundestag debates, 1949–1957 (DSL-adjusted, see definition (d) in \Cref{box:monthly-count-definitions}). LAG $=$ \de{Lastenausgleichsgesetz}{Equalisation of Burdens Law}; BVFG $=$ \de{Bundesvertriebenengesetz}{Federal Expellees Act}; HAuslG $=$ \de{Gesetz über die Rechtsstellung heimatloser Ausländer im Bundesgebiet}{Law on the Legal Status of Stateless Foreigners in the Federal Territory}.}
    \label{fig:1949-1957-highlevel}
\end{figure}

During 1949–1957, solidarity dominated parliamentary discourse on postwar population movements. As \Cref{fig:1949-1957-highlevel} shows,\footnote{We report (DSL-adjusted) counts rather than proportions (see definition (d) in \Cref{box:monthly-count-definitions}) %for interpretability (see definition (d) in \Cref{box:monthly-count-definitions})}, given solidarity's overwhelming dominance and the low frequency of anti-solidarity
because counts make spikes visible whenever migration-related laws are debated and enable comparison of debate volume across months.} mentions of solidarity consistently exceeded anti-solidarity and mixed statements, at times by a factor of ten.

Mentions of \emph{solidarity} closely track key legislative moments. A major spike in March 1950 aligns with early discussions on the integration of expellees and Article 131, which addressed the rehabilitation of former public servants of the Nazi regime, many of them displaced. A second cluster in early 1952 reflects debates on the LAG, preceding its adoption in May (see \exref{19520514-bt-01-211-00608} %previously Example 1, 19520514-bt-01-211-00608
in \Cref{tab:postwar-predictions-examples} in the Appendix). Another sharp rise in February–March 1953 corresponds to readings of the \de{Bundesvertriebenengesetz (BVFG)}{Federal Expellee Law},  which defined the legal status of expellees and refugees and codified integration measures (\exref{19530225-bt-01-250-01231}%previously: Example 2,
). Later peaks in September 1954 and June 1955 reflect %continued 
debate over LAG implementation and budgets. %issues. 
Smaller but sustained increases in 1956 relate to new support measures for \de{Sowjetzonenflüchtlinge}{Soviet-zone refugees} and \de{Aussiedler}{resettlers}.

\begin{center}
\begin{definitionbox}{Estimated Monthly Counts of (Anti-)Solidarity}{monthly-count-definitions}

Speeches are grouped by month $t$, with $N_t$ speeches in month $t$. We define the estimated number of mentions of label $\ell$ in month $t$ as:

\begin{enumerate}[label=(\alph*), start=4]
\item \textbf{DSL-adjusted monthly counts:} 
$\hat C^{DSL}_{t,\ell} = \sum_{i=1}^{N_t} \tilde{Y}_{i,\ell}$,
where $\tilde{Y}_{i,\ell}$ denotes the DSL-adjusted class score for label $\ell$ in speech $i$, as defined in \Cref{eq:dsl-pseudo-outcome} in \Cref{sec:methodology}.
\end{enumerate}

\end{definitionbox}
\end{center}

\begin{figure}%[!htb]
    \centering
    \includegraphics[width=\linewidth]{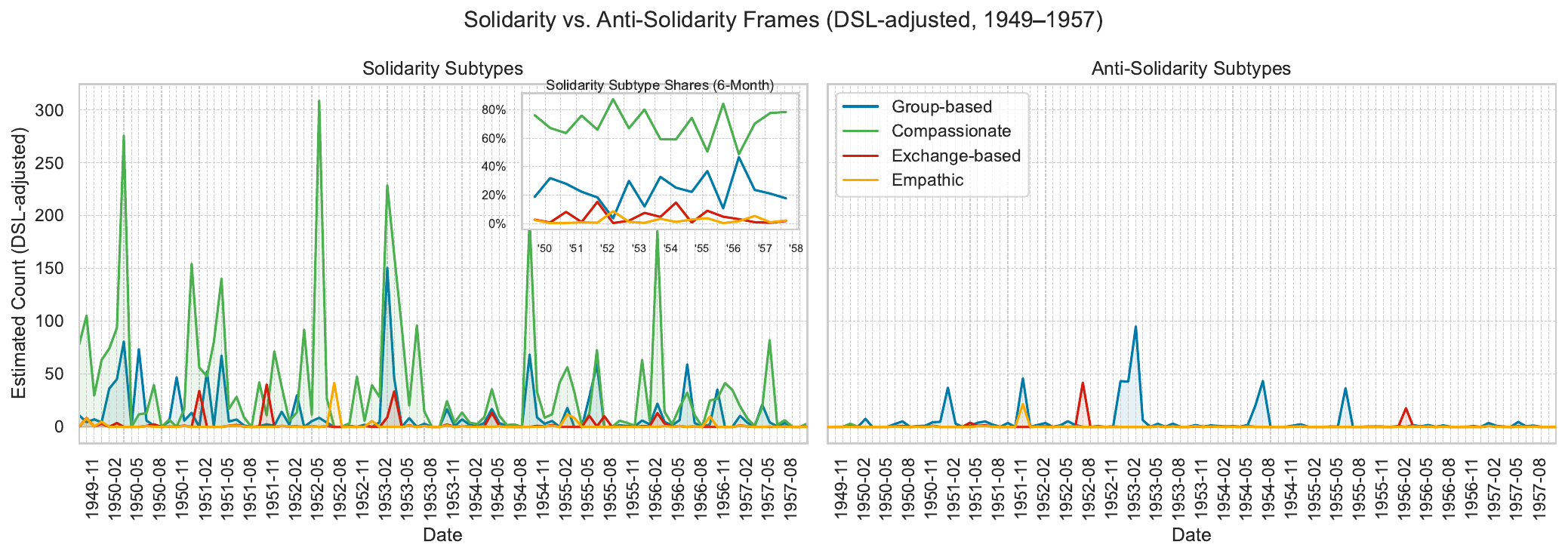}
    \caption{Distribution of (anti-)solidarity subtypes, 1949–1957 (DSL-adjusted expected counts, see definition (d) in \Cref{box:monthly-count-definitions}). Inset shows the evolving composition of solidarity subtypes, normalized within each six-month period as shares of all solidarity framing (see definition (c) in \Cref{box:trend-definitions}).}
    \label{fig:1949-1957-finegrained}
\end{figure}

The prominence of solidarity in this period %is unsurprising. 
can be explained by the fact that debates used it not only to justify compensation but also to manage postwar social and economic fragmentation and competing claims to belonging \citep{hughes1999shouldering}. The distribution of all four subtypes across key legislative moments (\Cref{fig:1949-1957-finegrained}) reflects this %complexity. 
mix of solidarity rationales.
\emph{Compassionate} and \emph{group-based} frames dominate throughout the period (see inset of \Cref{fig:1949-1957-finegrained}). By emphasizing shared national belonging (group-based) and the suffering of affected populations (compassionate), these frames construct a victim-centered \de{Schicksalsgemeinschaft}{community of fate}, uniting Germans through shared loss \citep{gengler2020new} (see \exref{19520514-bt-01-211-00608} %previously: Example 1, 19520514-bt-01-211-00608
in \Cref{tab:postwar-predictions-examples} in the Appendix\footnote{Example labels are based on the dominant DSL subtype, that is, the subtype with the highest DSL-adjusted score. This is useful for illustrating which subtype is most strongly reflected in a given speech within the DSL-based analysis; however, since DSL is designed for aggregate inference, these labels do not guarantee corrected true labels for individual %documents
instances.}). This coexistence of the two subtypes is consistent with \citet{thijssen2012mechanical}, who argues that compassionate and group-based solidarity form a dialectical pair: identification with a collective (\enquote{we}) generates obligations toward its members, while emphasizing their suffering reinforces that collective bond.

Additionally, a small but persistent share of \emph{exchange-based solidarity} appears throughout the period. It reflects debates over efficient allocation and the reintegration of displaced populations into labor and housing markets, with measures such as the LAG framed as tools to rebuild productive capacity (\exreftwo{19530225-bt-01-250-01231}{19510131-bt-01-115-00271}).%previously (Examples 2 and 15).

These patterns show that solidarity in parliamentary debates %managed 
mediated the challenges of postwar reconstruction through key legislative reforms. However, since these reforms differentiated between migrant groups, we examine how \emph{(anti-)solidarity} subtypes were expressed across key migrant-related terms in this period.

\subsubsection{Solidarity Frames Across Keywords in 1949-1957}

The analysis includes migrant-related keywords for which the DSL-adjusted expected number of stanced instances (i.e., \emph{solidarity}, \emph{anti-solidarity}, or \emph{mixed} rather than \emph{none}) is at least 150. Keywords dominated by \emph{none} or with low DSL-adjusted expected counts (e.g., \de{Asylsuchende}{asylum seekers}, \de{Einwanderer}{immigrants}, \de{Emigranten}{emigrants}) are excluded; see \Cref{tab:keyword_distribution_postwar} in the Appendix for full distributions. \Cref{fig:1949-1957-keywords} shows the distribution of (anti-)solidarity frames across these terms. 

\textbf{\de{Vertriebene}{expellees}}, a category reserved primarily for \de{Volkszugehörige}{ethnic Germans}, are most often associated with \emph{compassionate solidarity} (about half of all expressions with a stance), %while also receiving a substantial share of 
and also with \emph{group-based solidarity} (nearly 25\%). This reflects their dual position: as key representatives of German victimhood %(alongside prisoners of war) 
during and after the World War II \citep{moeller2001war}, they were both central to West Germany's nation-building narrative and portrayed as victims of Allied or foreign injustice.

\begin{center}
\begin{definitionbox}{Distribution of (Anti-)Solidarity Subtypes Across Groups}{group-distribution-definitions}

Speeches are grouped by a categorical variable $g$ (a keyword or a political party). $I_g$ is the set of speeches belonging to group $g$. The share of subtype $\ell$ within group $g$ is defined as:

\begin{enumerate}[label=(\alph*), start=5]

\item \textbf{Subtype shares within groups:}  
$\hat s_{g,\ell} = \frac{\hat M_{g,\ell}}{\hat M_g}$, where

\begin{itemize}
\item $\hat M_{g,\ell} = \sum_{i \in I_g} \tilde{Y}_{i,\ell}$ is the DSL-adjusted (anti-)solidarity amount of subtype $\ell$ in group $g$, where $\tilde{Y}_{i,\ell}$ denotes the DSL-adjusted class score for label $\ell$ in speech $i$, as defined in \Cref{eq:dsl-pseudo-outcome} in \Cref{sec:methodology}.
\item $\hat M_g = \sum_{\ell \in L} \hat M_{g,\ell}$ is the total (anti)-solidarity of group $g$ across all labels $L$ (four solidarity subtypes, four anti-solidarity subtypes, and mixed stance).
\end{itemize}

Shares sum to one within each group $g$.

\end{enumerate}

\end{definitionbox}
\end{center}

%The term \emph{Heimatvertriebener} carried strong symbolic meaning: it evoked both the loss of a homeland and the injustice of displacement and, when paired with the political idea of a \enquote{right to the homeland}, reinforced a collective sense of belonging and entitlement to return \citep{lubbe1967streit}. 
Still, around 10\% of stance expressions toward \emph{Heimatvertriebene} were \emph{anti-solidarity}, particularly \emph{group-based} (solidarity limited to \enquote{us}, the in-group), reflecting concerns that expellees strained local infrastructures or were unfairly privileged over equally suffering \de{Einheimische}{locals}, mirroring broader public resentment \citep{kossert2008kalte}. Perceptions of injustice grew with the uneven distribution of expellees across West German states \citep{braun2014employment}, especially where their prominence in institutions and politics sparked fears of \de{Überfremdung}{overforeignization} and unequal burden-sharing \citep{levy2005memories} (see \exref{19501213-bt-01-106-00058} %previously Example 4 % 19501213-bt-01-106-00058
in \Cref{tab:postwar-predictions-examples} in the Appendix). 

\begin{figure}%[!htb]
    \centering
    \includegraphics[width=\linewidth]{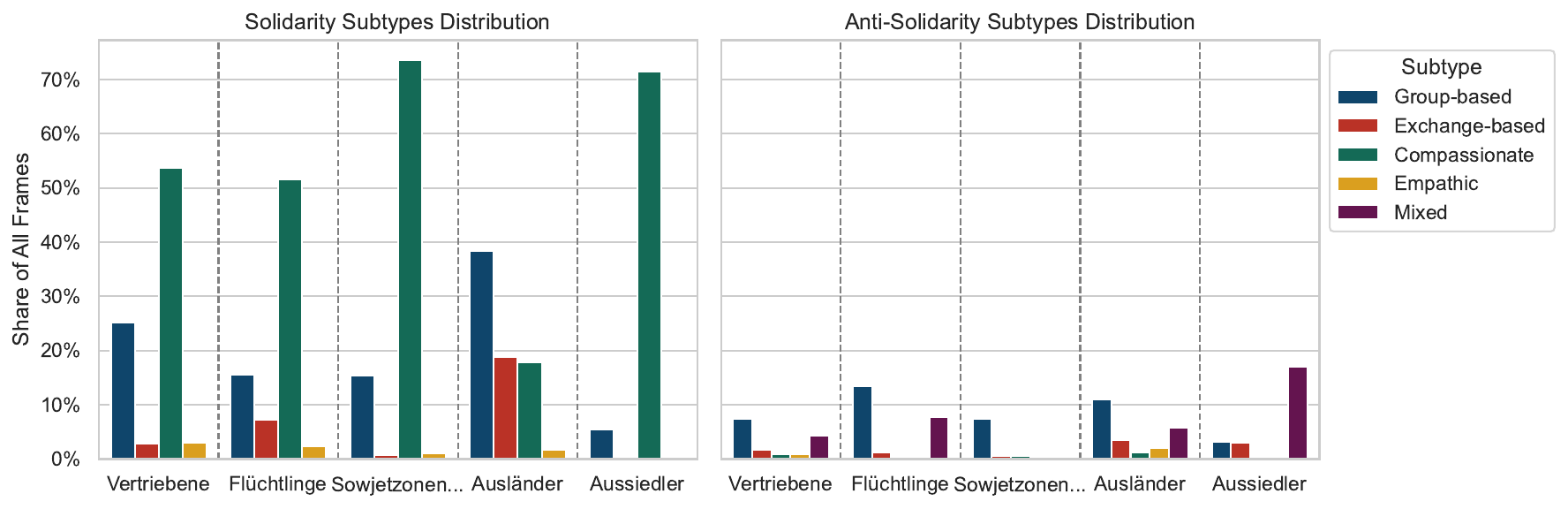}
    \caption{Relative distribution of (anti-)solidarity subtypes across selected migrant-related keywords, 1949–1957 (DSL-adjusted estimates; see definition (e) in \Cref{box:group-distribution-definitions}). Bars show each subtype's share within a keyword group; shares sum to 100\%. Full label distributions appear in \Cref{tab:keyword_distribution_postwar} in the Appendix.}
    \label{fig:1949-1957-keywords}
\end{figure}

\textbf{\de{Flüchtlinge}{refugees}, \de{Sowjetzonenflüchtlinge}{Soviet-zone refugees}, and \de{Aussiedler}{resettlers}} are also primarily framed through \emph{compassionate solidarity} (50–72\%), followed by \emph{group-based solidarity} (15\%). 
The label \emph{Flüchtlinge} began as a broad catch-all in the late 1940s but, even after the 1953 BVFG differentiated it into \emph{Vertriebene}, \emph{Sowjetzonenflüchtlinge}, and \emph{Aussiedler}, it retained connotations of marginality and foreignness \citep{nachum2018semantics}.
This helps explain %both the higher share of \emph{compassionate solidarity} and 
a more notable share of \emph{anti-solidarity} (around 10\%). \emph{Sowjetzonenflüchtlinge}, framed as Cold War refugees from communist oppression, were less integrated into the nation-building narrative than \emph{Vertriebene}, prompting more \emph{compassionate} than \emph{group-based solidarity} \citep{nachum2018semantics, boke1996fluchtlinge} (see \exref{19540923-bt-02-044-01194}%previously Example 5
).

\textbf{Among the migrant categories, \de{Ausländer}{foreigners}} shows one of the highest relative shares of anti-solidarity. A discursively constructed \enquote{other}, including DPs\footnote{With the 1951 HAuslG, authority over the remaining DPs was transferred from Allied administration to the Federal Republic, and they were reclassified as \de{heimatlose Ausländer}{stateless foreigners}, a noncitizen status that granted them most civil rights without granting citizenship \citep{holian2011between}.}, Eastern European border crossers, wartime laborers, and early foreign workers \citep{smith1994germany}, they were contrasted with %German 
groups like Vertriebene or Einheimische and framed as non-belonging (evident in group-based anti-solidarity share), economically burdensome (exchange-based anti-solidarity), or even criminalized \citep{holian1945missing} (see \exref{19560621-bt-02-151-01596}). %previously Example 6, 19560621-bt-02-151-01596. 
Yet solidarity, especially \emph{group-based}, still exceeded anti-solidarity, partly due to the category's fluidity: it sometimes included ethnic Germans not yet formally recognized as such in the early postwar years \citep{ruhkopf2023institutionalisierte}. Speakers %Politicians such as Lukaschek 
reinforced this dynamic by linking DPs and expellees through shared narratives of lost \de{Heimat}{homeland}, enabling symbolic inclusion into the national victim community \citep{ruhkopf2023institutionalisierte} (\exref{19501018-bt-01-092-00330}). %previously (Example 7). %Their position was also reflected in the 1951 HAuslG, which granted them certain civic and social rights while maintaining their formal exclusion from the national community \citep{holian2011between}.

The second-highest subtype, \emph{exchange-based solidarity}, reflects the perceived political utility of \textit{Ausländer} in strengthening West Germany's international legitimacy %within the Western alliance 
\citep{ruhkopf2023institutionalisierte}, for example through the 1951 Homeless Foreigners Law (HAuslG), passed under Allied pressure %granted DPs civic and social rights largely to meet conditions for sovereignty and improve the country's international standing
\citep{holian1945missing} (\exref{19510419-bt-01-136-01759}). %previously(Example 8). 
However, %instances of genuine solidarity 
solidarity also extended to non-ethnic German foreigners, particularly in CDU and SPD statements, showing that solidarity with \textit{Ausländer} was not limited to Germanness or foreign policy utility but also reflected broader humanitarian commitments (see \exreftwo{19501018-bt-01-092-00327}{19550615-bt-02-087-03279}). %7.1 and 9). %%% “Resettlers, refugees, expellees—we may call them what we want, they are victims of Hitler’s war.” in the first post-war years (1945-1947). (“New Citizens” or “Community of Fate”? Early Discourses and Policies on “Flight and Expulsion” in the Two Postwar Germanys)% cannot use this quote because it comes from a Soviet zone newspaper and refers only to 1945-1947

In sum, compassionate framing dominated across most categories, reflecting a broader discourse of shared suffering in which displaced populations were positioned as victims of the war and its consequences. Nevertheless, solidarity was still shaped by perceived \enquote{deservingness} and ethnic identity. While \emph{Vertriebene} were more firmly incorporated into the national community as ethnic Germans, others, especially \emph{Ausländer}, %were treated as outsiders and 
received comparatively less support and conditional inclusion \citep{nachum2013reconstructing}. These differences reflected both group characteristics and the ideological strategies of political parties. The next subsection examines how different parties in the early Bundestag framed (anti-)solidarity in their parliamentary speech.

\subsubsection{Distribution across parties}

\begin{figure}%[!htb]
    \centering
    \includegraphics[width=\linewidth]{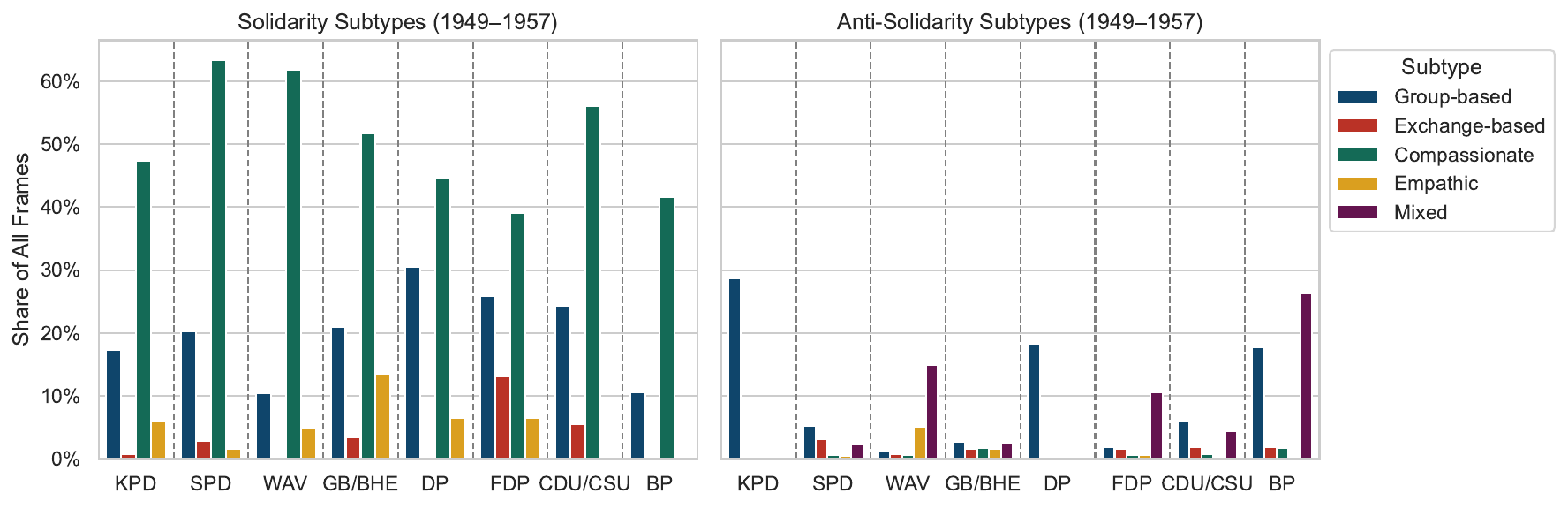}
   \caption{Distribution of (anti-)solidarity subtypes across Bundestag parties, 1949–1957 (DSL-adjusted estimates; see definition (e) in \Cref{box:group-distribution-definitions}). Bars show each subtype's share within a party, summing to 100\%. The analysis includes parties with at least 150 DSL-adjusted stanced statements. Full label distributions appear in \Cref{tab:party_distribution_postwar} in the Appendix.}
    \label{fig:1949-1957-parties}
\end{figure}

\Cref{fig:1949-1957-parties} shows party distributions of \emph{solidarity} and \emph{anti-solidarity} in 1949-1957, \ak{based on the 8,057 of 9,643 instances for which party information was available. Of these, 111 also had human reference labels.} \emph{Solidarity} dominated overall, but parties framed it differently, shaped by ideology and strategy.

\textbf{The main parties -- governing CDU/CSU and FDP, as well as opposition SPD --} supported expellee and refugee integration through legal and institutional measures, primarily expressing \emph{compassionate} or \emph{group-based} solidarity (see \Cref{fig:1949-1957-parties}). The CDU/CSU led in proposing the LAG, framing solidarity as a national duty toward fellow ethnic Germans, and even included expellee representatives in the party ranks \citep{hughes1999shouldering}. The SPD also supported the law but voiced concerns about former Nazis and communist agents among the newcomers \citep{connor2018refugees}, resulting in occasional anti-solidarity statements. The FDP, while formally aligned with CDU/CSU and SPD on refugee issues, framed solidarity in more individualistic, \emph{exchange-based} terms, opposing large-scale redistribution and favoring market-based solutions consistent with its liberal economic orientation \citep{connor2018refugees} (see \exref{19491020-bt-01-012-01359} %previous: Example 10 %19491020-bt-01-012-01359
in \Cref{tab:postwar-predictions-examples} in the Appendix). %where FDP speakers argued that expellees should build their own homes through self-help and cooperative effort, with state support limited to loans and low-interest credit).
%The FDP's resistance to preferential or structural redistribution while formally supporting refugee compensation is also reflected in its comparatively higher proportion of mixed stance (10\%).

\textbf{The expellee advocacy party GB/BHE}%and, in a more limited sense,} DP\footnote{While the DP %adopted right-to-homeland rhetoric and 
%addressed expellee demands, they were not a core part of the party's structure, and the DP did not primarily define itself as a refugee party \citep{aschoff2014regionalparteien}}.)
\footnote{\de{Gesamtdeutscher Block/Bund der Heimatvertriebenen und Entrechteten (GB/BHE)}{All-German Bloc/League of Expellees and Deprived of Rights}. The DP (\de{Deutsche Partei}{German Party}) also addressed expellee demands, especially in Lower Saxony because of their electoral weight there, but it began as a \de{Heimatpartei}{\enquote{homeland party}} rather than a refugee party \citep{aschoff2014regionalparteien}. This is visible in the party's share of \emph{group-based anti-solidarity}, reflected in its defense of local interests (\exref{19530318-bt-01-254-02259}).} positioned itself as a political representative of displaced Germans. %though on different bases. 
Following multiple successes in state elections, GB/BHE, founded by and for expellees, entered the federal coalition in 1953 and pressured other parties to address expellee demands \citep{connor2018refugees}. %The DP, by contrast, began as a \de{Heimatpartei}{\enquote{homeland party}} and responded to expellee concerns mainly because of their electoral weight in Lower Saxony \citep{aschoff2014regionalparteien}. Both of these parties 
It expressed \emph{compassionate} and \emph{group-based solidarity}, portraying expellees as full members of the national community entitled to compensation and integration (see \exref{19531029-bt-02-005-01150}%Example 11%19531029-bt-02-005-01150
). Additionally, the GB/BHE shows the highest share of \emph{empathic solidarity} (15\%), consistent with its role as an expellee party defining expellees as a distinct community. %deserving recognition.
%By contrast, around 18\% of DP statements express \emph{group-based anti-solidarity}, reflected in its defense of local interests (\exref{19530318-bt-01-254-02259}), %previously A 
%which led expellee organizations to regard the DP as a particularistic party of locals rather than a genuine refugee party \citep{aschoff2014regionalparteien}. 
Over time, CDU and SPD managed to absorb major expellee demands, which made GB/BHE %and DP 
less successful \citep{connor2018refugees, kittel2020stiefkinder}.

\textbf{Regionalist parties such as WAV and BP\footnote{\de{Wirtschaftliche Aufbau-Vereinigung}{Economic Reconstruction Union} and \de{Bayernpartei}{Bavarian Party}}} emphasized \emph{compassionate solidarity}, but restricted it to ethnic Germans. The short-lived WAV appealed to expellees through anti-communist rhetoric and demands to reclaim eastern territories \citep{kranenpohl1998zwischen} (\exref{19500323-bt-01-050-00140}). %previous (Example 12). 
The BP, responding to Bavaria's disproportionate refugee burden, called for invoked collective responsibility, fairer redistribution and favored groups like the Sudeten Germans \citep{braun2020101330}, which corresponds to its high share of \emph{group-based anti-solidarity} (see \exref{19491020-bt-01-012-01314}). %previous Example 13
%where the BP argued that Sudeten and Southeast Germans should be concentrated in Bavaria because they were \enquote{closest to the Bavarian cultural sphere} and thus easier to integrate
Both parties also expressed the highest levels of \emph{mixed stance}, either by excluding non-German migrants (WAV) \citep{connor2018refugees} or opposing specific groups such as Silesians (BP).

\textbf{The KPD\footnote{\de{Kommunistische Partei Deutschlands}{Communist Party of Germany}}} also addressed displaced populations, framing them through socialist integration and emphasizing \emph{compassionate and} \emph{group-based solidarity}, though in an assimilationist rather than inclusive sense. Avoiding the label \textit{Vertriebene} in line with Soviet policy, the party instead spoke of \enquote{refugees}, portraying them as victims of fascism and capitalism %and linking their integration to broader projects such as land reform 
\citep{heidemeyer2007deutsche, connor2018refugees}.

%shaped early postwar debates and helped define solidarity and its boundaries of deservingness. 

Parties like KPD, WAV, and BP did not stay in parliament for long, but nonetheless shaped early postwar debates %. Rather than merely reflecting %existing 
%social tensions, as suggested by \citet{lipset1967cleavage}, %these parties actively contributed to defining what solidarity meant 
and helped define solidarity and %and who was considered deserving of support. 
its boundaries of deservingness. As suggested by \citet{thijssen2022s}, such framings were conditioned by the period's political and social challenges and reflected broader efforts to maintain or renegotiate societal cohesion.

%\begin{figure}[!htb] %%% OLD FIGURE %%%
%    \centering
%    \includegraphics[width=\linewidth]{plots/analysis/1949-1969/KeywordsFinegrained.pdf}
%    \caption{Relative distribution of (Anti-)solidarity subtypes across selected migrant-related keywords, 1949–1957. Bars show each subtype's share within a keyword group (summing to 100\%).}
%    \label{fig:1949-1957-keywords}
%\end{figure}

%\begin{figure}[!htb] %%% OLD FIGURE %%%
%    \centering
%    \includegraphics[width=\linewidth]{plots/analysis/1949-1969/Old/1949–1957_Parties.pdf}
%    \caption{(Anti-)solidarity frame distribution across major Bundestag parties, 1949–1957, by election period. Percentages are normalized within each party, with all subtypes summing to 100\%. Only parties with sufficient stance instances $(N>100)$ are included. Full label distributions appear in \Cref{tab:party_distribution_postwar} in the Appendix.}
%    \label{fig:1949-1957-parties}
%\end{figure}

\subsection{Recent German history 2009-2025}
\label{sec:recent}

Between 2009 and 2015, annual migration to Germany rose from just under 800,000 to more than 2 million people, peaking in 2015 in the context of the Syrian war \citep{gessler2023politicization}. This period has since been associated with the label \enquote{migration crisis} \citep{gessler2023politicization, frohlich2025migration}.
Since then, the right-wing AfD has made electoral gains and entered the German federal parliament in 2017. Another wave of migration followed the 2022 Russian invasion of Ukraine \citep{bamf2025}. Alongside these %historical 
events in the last four parliamentary periods (2009-2025), substantial shifts in solidarity occurred.

\subsubsection{Overall solidarity distribution}\label{sec:recent-overall-solidarity-distrib}

\begin{figure}%[!htb]
    \centering

    \begin{subfigure}[t]{0.37\linewidth}
        \centering
        \includegraphics[width=\linewidth]{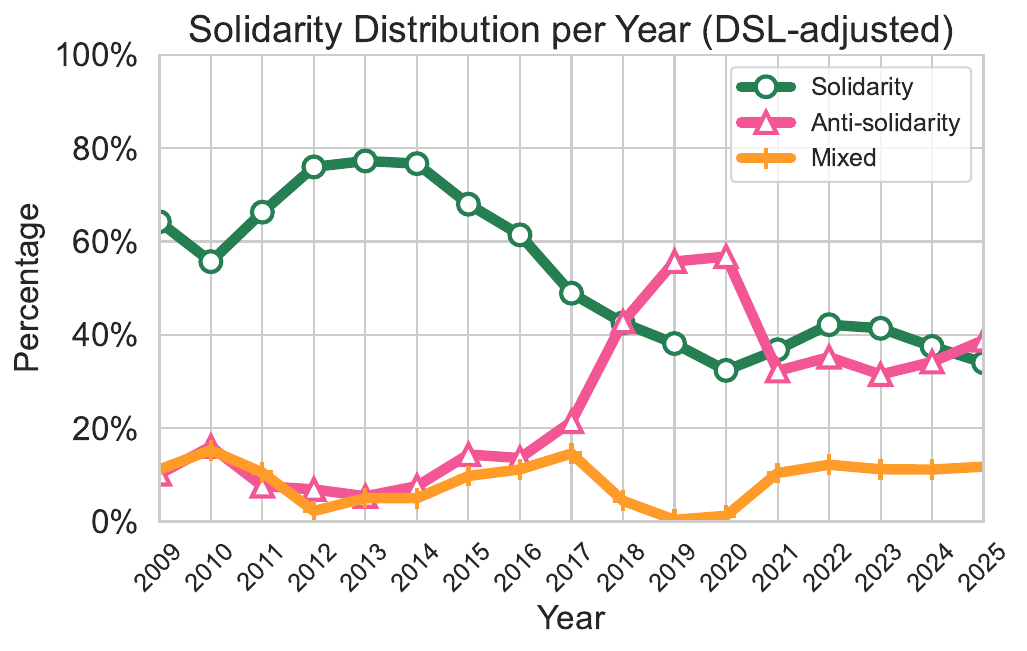}
        \caption{Solidarity and anti-solidarity distribution}
        \label{fig:highlevel-2009-2025}
    \end{subfigure}
    \hfill
    \begin{subfigure}[t]{0.60\linewidth}
        \centering
        \includegraphics[width=\linewidth]{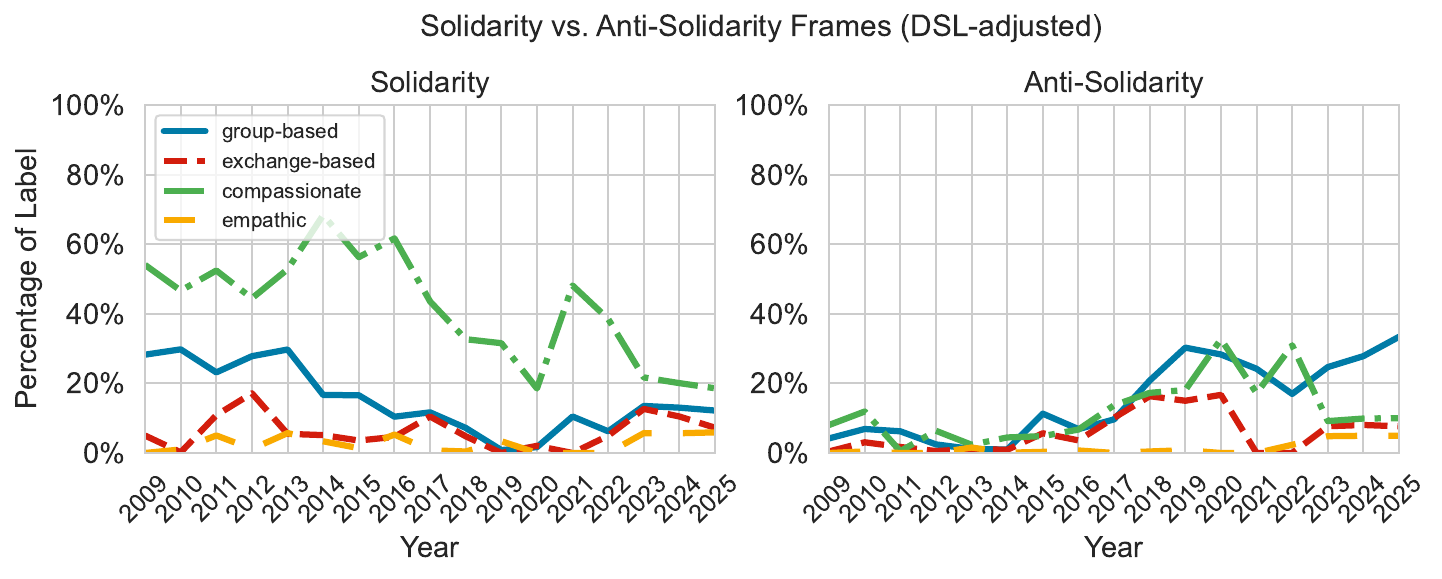}
        \caption{Solidarity and anti-solidarity frames distribution}
        \label{fig:finegrained-2009-2025}
    \end{subfigure}
    \caption{(Anti-)solidarity trends for the \textit{Migrant} category, 2009–2025. \Cref{fig:highlevel-2009-2025}: Yearly DSL-adjusted shares of solidarity, anti-solidarity, and mixed stance (see definition (c) in \Cref{box:trend-definitions}). The \textit{none} category is included in the normalization but omitted from the visualization. \Cref{fig:finegrained-2009-2025}: Yearly DSL-adjusted shares of solidarity (left) and anti-solidarity (right) subtypes (see definition (c) in \Cref{box:trend-definitions}). Within each year, subtype shares are renormalized across the eight solidarity and anti-solidarity categories (excluding mixed and none), summing to 100\%.}
    \label{fig:highlevel-finegrained-2009-2025}
\end{figure}

\Cref{fig:highlevel-finegrained-2009-2025} shows (anti-)solidarity trends from 2009–2025 (DSL-adjusted). Solidarity declined from 2014 to 2019, increased modestly between 2020 and 2022, and then saw a slight decline after 2022. This decrease in solidarity is associated with an almost corresponding increase in anti-solidarity. %(except for some fluctuation in mixed-solidarity instances). 
In terms of (anti-)solidarity subtypes, compassionate solidarity peaks around 2014–2016, drops sharply, briefly rebounds in 2021,\footnote{This rebound coincides with the COVID-19 context, rising asylum applications, humanitarian responses to the Taliban takeover in Afghanistan, and the formation of the SPD–Greens–FDP coalition government \citep{Brandt2024}.} and declines again. The decrease in solidarity since 2022 corresponds to a decrease across solidarity subtypes, especially in \emph{compassionate} and \emph{exchange-based} solidarity. In terms of anti-solidarity, \emph{group-based anti-solidarity} rose strongly after 2017, accounting for over one third of all stanced migrant-related statements. %Exchange-based anti-solidarity rose slightly between 2015-2020 but dropped again with the beginning of the last parliamentary period in 2021.
\emph{Compassionate anti-solidarity} spikes around 2020 and then declines after 2022.

\textbf{Overall, the period since 2019 marks the first time in post-war history that the fraction of anti-solidarity statements exceeds that of solidarity statements in the German parliament.} This key finding is visible both in the DSL-adjusted estimates but also in human annotations (\Cref{fig:instances-distribution})\bp{, and remains robust even under error analysis: using the manual error audit data from \Cref{sec:error-analysis}, we estimate false positive rates for anti-solidarity of at most 11.3\%, which reduces but does not fully close the gap to solidarity (refer to \Cref{app:appendix-attribution-sensitivity} in the Appendix for further details).}

We next contextualize it with recent Germany history and a deeper look into party distribution. The 2014 spike in \emph{compassionate solidarity} coincided with heightened public support for Syrian refugees: 15.4\% of German survey respondents named solidarity their most important value in 2014, up from 12.3\% in 2013 \citep[p. 38]{DeLeur2023}. Refugees were commonly framed as needing help, motivating calls for a \enquote{German welcome culture} \citep{Gebauer2023} (see \exref{20141106-bt-18-063-04755} %previous Example 15 
in \Cref{tab:recent-predictions-examples} in the Appendix). By early 2016, a counter-narrative emerged portraying refugees as overburdening German capacities and as contributing to a \enquote{refugee crisis} \citep{Gebauer2023} (see \exref{20151001-bt-18-127-00708}). %Example 16 previously. %Another turning point was new year's eve 2015/2016, when criminal activities at the Cologne main train station were framed as driven by foreigners, followed by a sharp increase in anti-refugee violent crime \citep{Frey2020}. 
During the same period, the party AfD---founded in 2013 with a focus on EU criticism---aligned itself with anti-migrant protests, shifted further towards the right of the political spectrum, and entered parliament in 2017 after major electoral gains \citep{Frey2020,Gebauer2023}. Regarding the broader European context, \citet{DeLeur2023} diagnose a post-2016 \enquote{solidarity fatigue}, driven by an inability to sustain prolonged compassion on the individual level and dwindling resources on the institutional level. 
These counter-narratives match several notions of anti-solidarity according to \citet{thijssen2022s}: refugees as an outgroup competing with German population (\emph{group-based}, see \exref{20250514-bt-21-003-01509}%previous Example 21
), as undeserving of compassion (\emph{compassionate}, see \exref{20160219-bt-18-156-00811}%before: Example 28
), and as straining resources without contributing (\emph{exchange-based}, see \exref{20180516-bt-19-032-01219}%before:Example 17
). We also note a general shift towards right-wing sentiments in the general German population \citep{Zick2023}, which may be associated with heightened anti-solidarity.
%The brief rebound in compassionate solidarity around 2021 coincides with the COVID-19 context, rising asylum applications, humanitarian responses to the Taliban takeover in Afghanistan, and the formation of the SPD–Greens–FDP coalition government \citep{Brandt2024}.

More recently, the Russian invasion of Ukraine and the resulting arrival of Ukrainian refugees in Germany, marking the largest net migration since 1950 \citep{Atanisev2024}, triggered another, though less prominent, spike in public solidarity \citep{weber2024social}. In our data, solidarity increases slightly in 2022, followed by a continued %rapid 
decline. \citet{Bojarskich2025} identified several factors promoting anti-solidarity towards Ukraine among the German public, such as conspiracy mentality (i.e.\, receptivity to contrarian narratives and disinformation emphasizing the risks of supporting Ukraine) and support for populist parties, but also fears about economic and societal consequences, indicating \emph{group-based} (competition between groups) or \emph{exchange-based} (insufficient contribution) anti-solidarity (\exref{20221019-bt-20-062-02988}). %before (Example 18). 
Anti-Ukrainian framing may also have been reinforced by the AfD's pro-Russian stance, including opposition to sanctions and military aid \citep{arzheimer2023russia}.
%In contrast to the postwar period, (anti-)solidarity shifts in the German parliament between 2009 and 2025 appear less attached to specific legislative initiatives. Because of Germany's membership in the European Union, migration policy has become more complex and is negotiated  between European institutions and member states \citep {geddes2020migration}. 
In contrast to the postwar period, (anti-)solidarity shifts between 2009 and 2025 appear less tied to specific legislation, likely because much of German migration policy is negotiated at the EU level \citep{geddes2020migration}. Additionally, the AfD integrated anti-migrant talking points across a wide range of legislative topics. %As such, we do not look deeper into specific laws passed during this time frame but rather focus on the trends for specific parties.
As such, we focus less on legislation and more on party-specific trends.

\begin{figure}[!htb]
    \centering
    \includegraphics[width=0.6\linewidth]{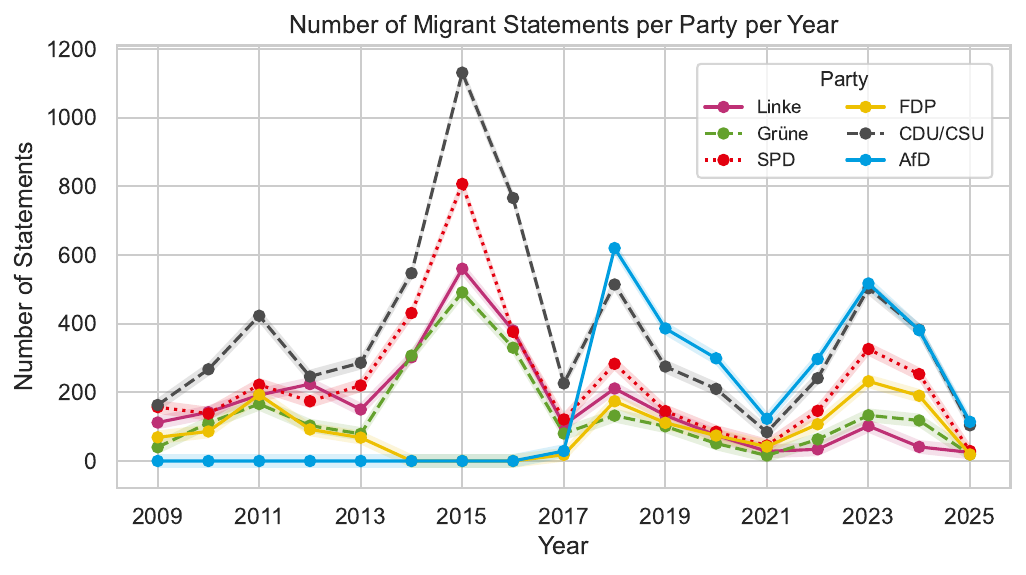}
    \caption{Raw yearly counts of parliamentary statements relating to migrants by selected political parties, 2009-2025.}
    \label{fig:party-raw-utterances}
\end{figure}

\subsubsection{Distribution across parties}\label{sec:recent-distribution-across-parties}

\begin{figure*}[!htb]
    \centering

    \begin{subfigure}[t]{0.49\linewidth}
        \centering
        \includegraphics[width=\linewidth]{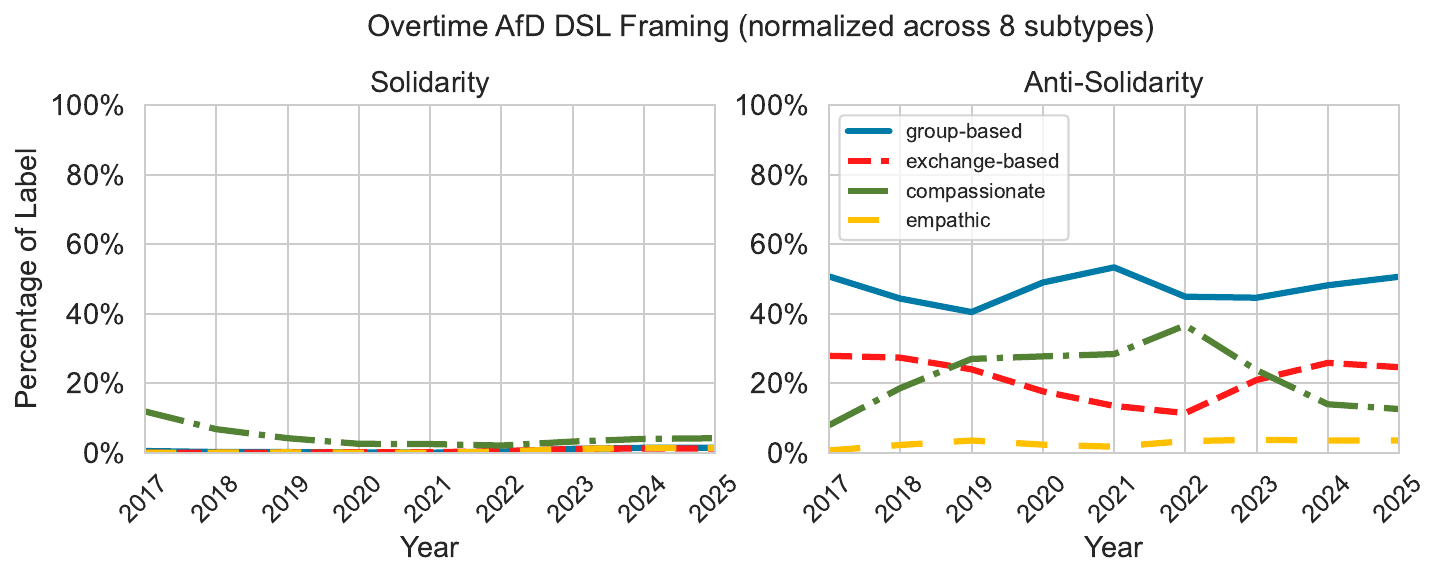}
        \caption{AfD}
        \label{fig:finegrained-afd}
    \end{subfigure}
    \hfill
    \begin{subfigure}[t]{0.49\linewidth}
        \centering
        \includegraphics[width=\linewidth]{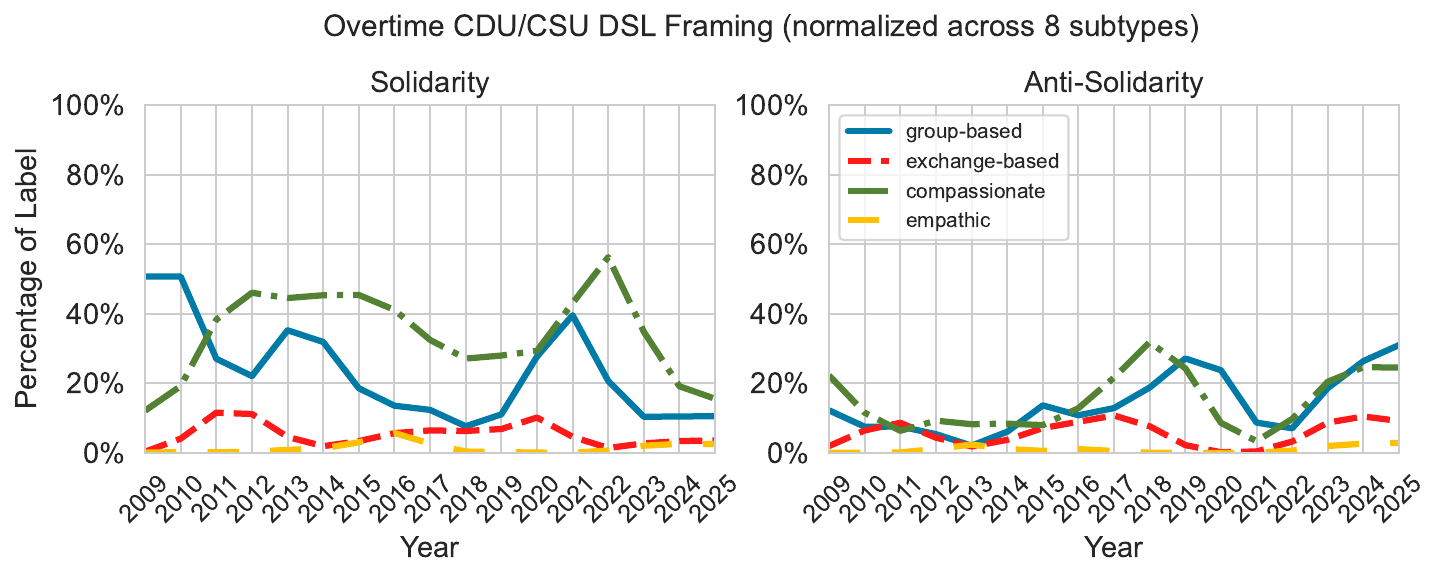}
        \caption{CDU/CSU}
        \label{fig:finegrained-cdu}
    \end{subfigure}

    \vspace{1em}

    \begin{subfigure}[t]{0.49\linewidth}
        \centering
        \includegraphics[width=\linewidth]{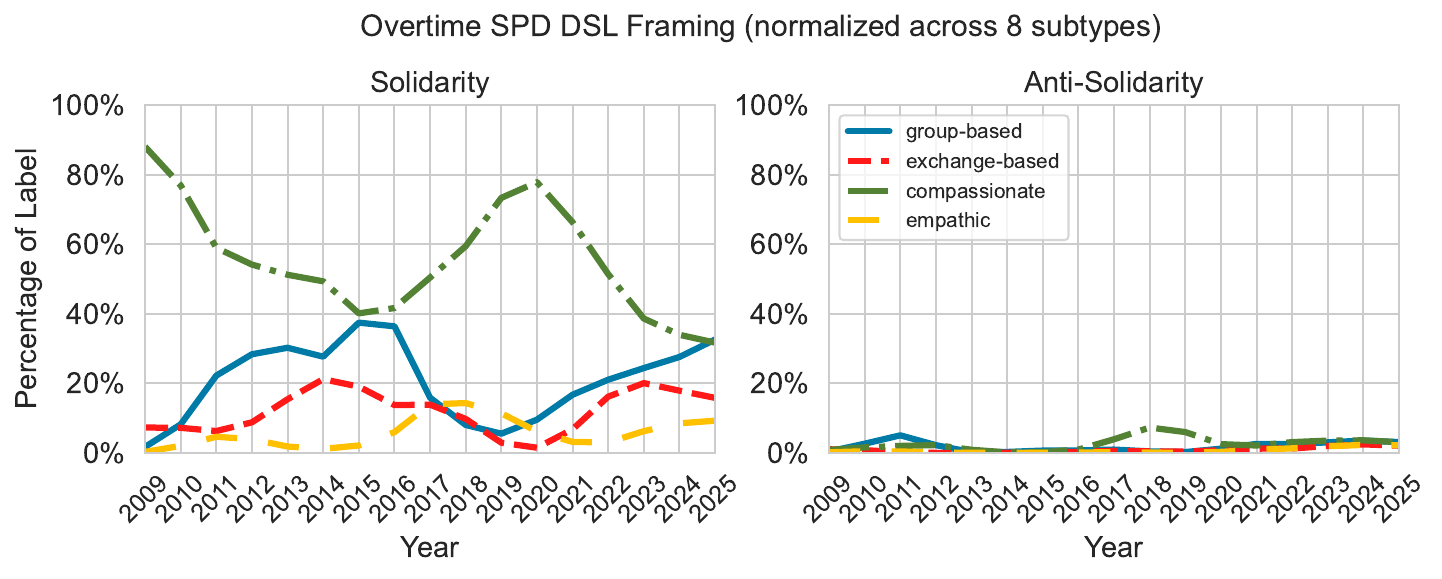}
        \caption{SPD}
        \label{fig:finegrained-spd}
    \end{subfigure}
    \hfill
    \begin{subfigure}[t]{0.49\linewidth}
        \centering
        \includegraphics[width=\linewidth]{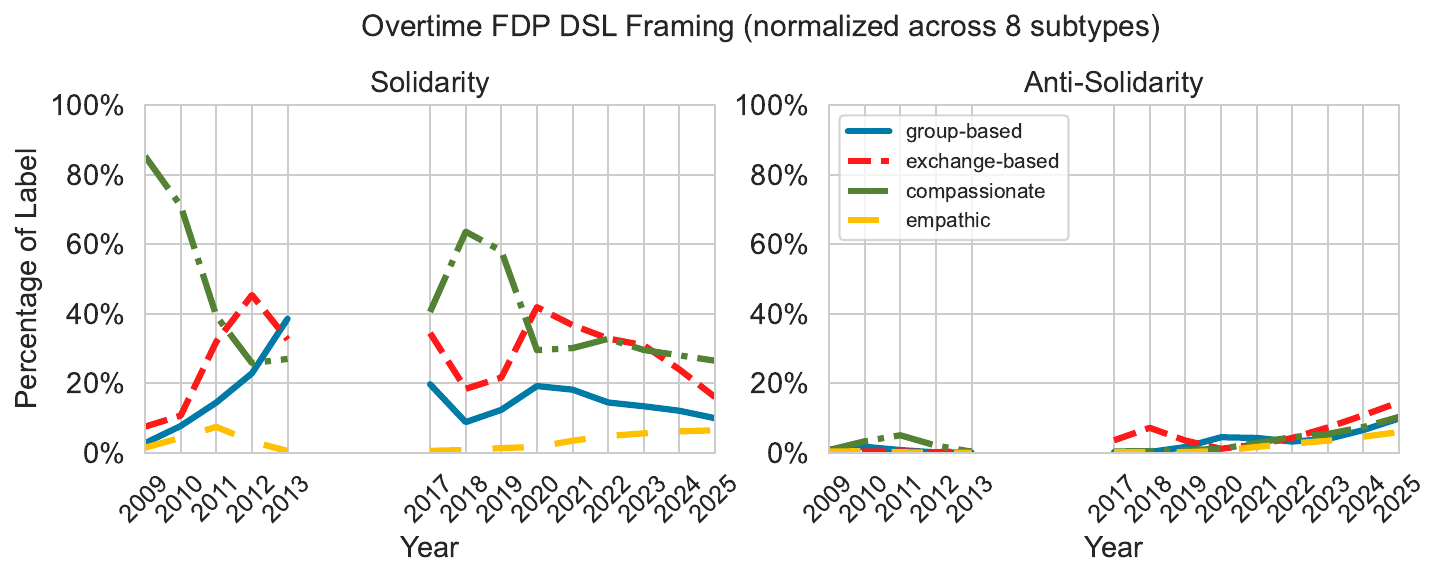}
        \caption{FDP}
        \label{fig:finegrained-fdp}
    \end{subfigure}
    \begin{subfigure}[t]{0.49\linewidth}
        \centering
        \includegraphics[width=\linewidth]{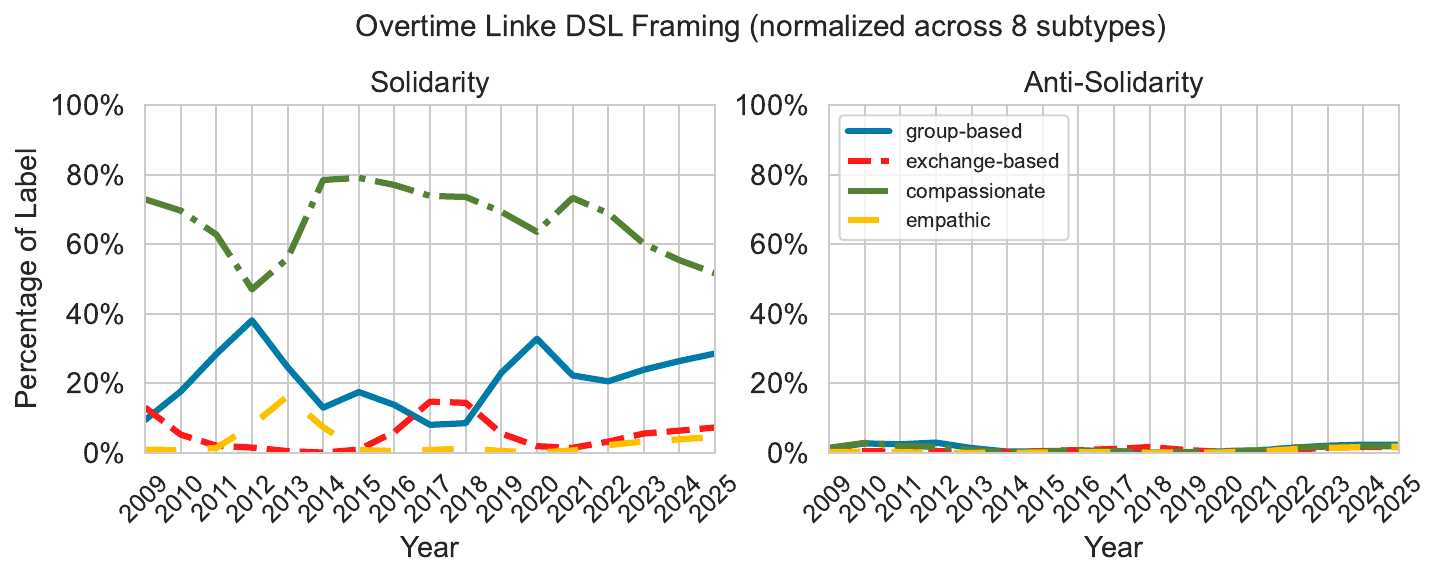}
        \caption{Die Linke}
        \label{fig:finegrained-linke}
    \end{subfigure}
    \begin{subfigure}[t]{0.49\linewidth}
        \centering
        \includegraphics[width=\linewidth]{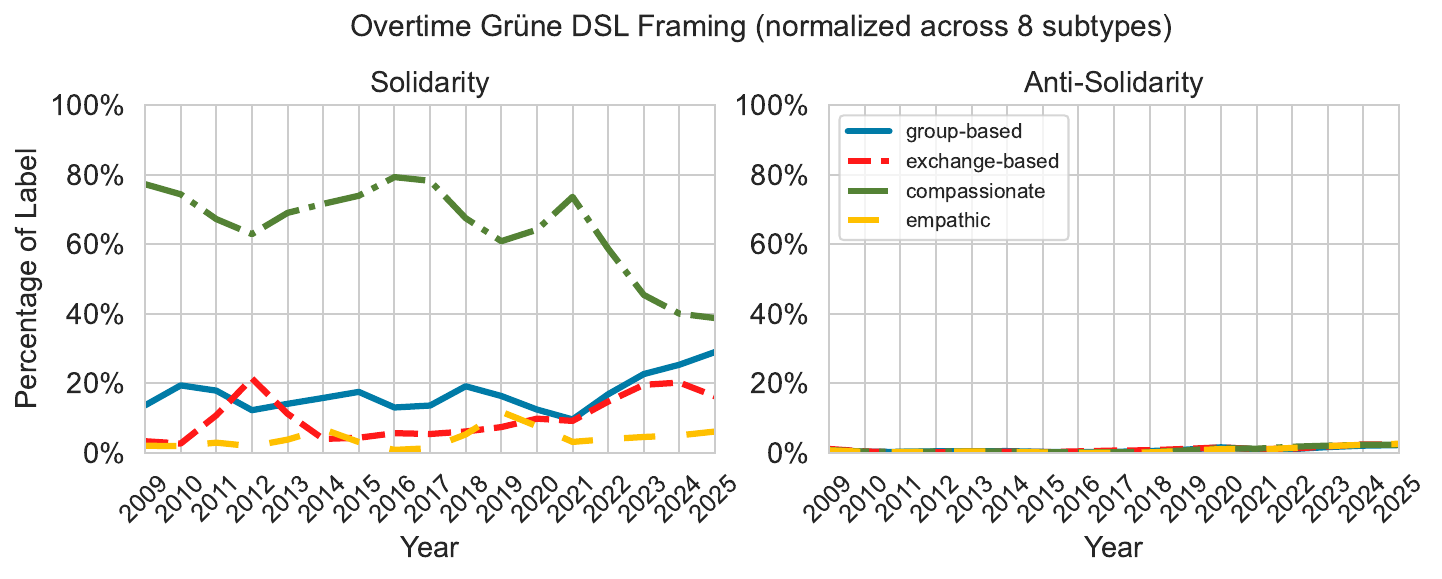}
        \caption{Die Grünen}
        \label{fig:finegrained-grünen}
    \end{subfigure}

    \caption{DSL-adjusted yearly shares of (anti-)solidarity subtypes across Bundestag parties, 2009-2025 (see definition (c) in \Cref{box:trend-definitions}). Within each year and party, subtype shares are renormalized across the eight solidarity and anti-solidarity categories so that they sum to 100\%. Note: the AfD entered the Bundestag only in 2017, while the FDP was absent from 2013–2017; accordingly, no observations appear for these periods. \ak{See larger versions of the party-specific panels in \Cref{fig:finegrained-per-party-enlarged} in the Appendix.}}
    \label{fig:finegrained-per-party-2x2}
\end{figure*}

\Cref{fig:party-raw-utterances} shows the raw numbers of migration-related statements per party and year, and \Cref{fig:finegrained-per-party-2x2} shows the fine-grained solidarity trends. \ak{For this analysis, party information was available for 19,759 of 23,897 instances; 543 of these also had human reference labels.} % Broadly, we divide the parties into four blocks: The right-wing AfD, the center-right CDU/CSU and FDP, the center-left SPD, and the center-left to left Green party, Linke, and BSW.

\begin{description}
\item[The right-wing AfD] expresses almost no solidarity, with \emph{group-based} anti-solidarity being most common (ca.\ 50\% throughout 2017-2025), followed by  \emph{compassionate} and \emph{exchange-based} anti-solidarity (each ca.\ 25\% with some fluctuation). %in the parliamentary period 2017-2021. 
Though one of the smaller factions in parliament, the AfD accounts for most anti-solidarity instances overall (53.3\% over the whole period). The sharp rise of anti-solidarity after 2017 (\Cref{fig:highlevel-2009-2025}) aligns with the AfD's entry into parliament and their rapid increase in migration-related utterances (\Cref{fig:party-raw-utterances}). %This can be explained by the fact that 
Many AfD speeches, across topics, %contain references to 
link migrants to illegal border crossings, criminality, or 
%migrants imposing 
additional burdens on the German society, without necessarily addressing the actual topic of legislative debate (see \exref{20230512-bt-20-104-02937} %before Example 19 
in \Cref{tab:recent-predictions-examples} in the Appendix%where an AfD member argues that refugees displace elderly people from care homes
). 
% For example, multiple AfD instances in early 2023 reference elderly care homes being closed down to make space for refugees. We note that some of these examples are annotated as group-based anti-solidarity by Llama-3.3 70B (emphasizing the otherness of the refugee group) but exchange-based solidarity would also be a plausible annotation, such that the exact distribution of solidarity subtypes is attached with some uncertainty.
%
\item[The center-right CDU/CSU and liberal FDP] changed the most in our data. In 2009-2015, most statements expressed solidarity, %with a spike of compassionate solidarity for the CDU/CSU in 2014 (the FDP was not in parliament at that time). 
with both parties focusing mostly on \emph{compassionate solidarity}, followed by group-based solidarity for CDU/CSU and exchange-based solidarity for FDP, for instance by stressing the contributions of incoming skilled workers to the German economy, motivating a new \de{Fachkräfteeinwanderungsgesetz}{Skilled Labor Migration Law} in 2023 (see \exref{20230525-bt-20-106-05636}). %before Example 20). 
From the late 2010s onward, both parties shift toward anti-solidarity, most visibly in the CDU/CSU, where \emph{group-based anti-solidarity} and \emph{compassionate anti-solidarity} frames rise after the 2015 refugee crisis, decline during the Ukrainian refugee inflow in 2022, and resurge by 2025, reaching 33\% and 24\% of all frames, respectively. For the FDP, a comparable shift appears by 2025, where \emph{exchange-based}, \emph{group-based}, and \emph{compassionate anti-solidarity} account for 16\%, 12\%, and 12\%, respectively. When inspecting the instances, the framing of migration is mostly in legalistic terms, such as regulating and controlling migration and reducing numbers, marking migrants as an outgroup, %which is classified as group-based anti-solidarity by Llama-3.3-70B 
for which \emph{group-based anti-solidarity} has the highest DSL estimate among the anti-solidarity subtypes (see \exref{20250514-bt-21-003-01509}) %before Example 21) %20250514-bt-21-003-01509
%where a CDU/CSU speaker frames migration as a threat to political stability and an excessive burden on local infrastructure, with calls for deportations, faster procedures, and rejections at the border
and portraying migration as misuse of welfare benefits, corresponding to \emph{exchange-based anti-solidarity} (see \exref{20250130-bt-20-210-02826}%before Example 22 %20250130-bt-20-210-02826
%where an FDP speaker portrays asylum as being misused for state-subsidized residence
).

The %drastic 
shift toward \emph{compassionate and} \emph{group-based anti-solidarity} over the last three years is insufficiently explained by recent literature. One possibility is that CDU/CSU and FDP adopted populist rhetoric from AfD %politicians who entered parliament since then 
after 2017 \citep{esguerra2023accommodation}. However, \citet{Lewandowsky2022} and \citet{kesting2018discourse} did not observe effects of the AfD entering parliament on polarization, the use of populist language, or discourse quality. Notably, the first shifts occurred even \emph{before} the AfD entered parliament. Alternatively, it may be that CDU/CSU and FDP respond to an overall right-wing shift in the German population \citep{Zick2023}, appealing to voters with anti-migration rhetoric and thereby providing a more moderate (but still anti-solidary) alternative to the AfD. %The stance toward cooperating with AfD was a particularly prominent issue in the 2025 federal election with CDU/CSU facing criticism for voting with AfD on a (non-binding) resolution \citep{Martin2025}.
Overall, our findings encourage further research into this striking and rapid trend.

\item[The center-left SPD] shows predominantly \emph{compassionate solidarity}, with a spike in 2020, and is closely followed by \emph{group-based solidarity}, which gains prominence post-2019, and almost taking over in 2025. This is consistent with the party's humanitarian commitments and stronger emphasis on integration \citep{orhan2025intra}. Additionally, exchange-based solidarity is present, at times reaching around 20\%. In line with this perspective, SPD has repeatedly framed Germany as an \de{Einwanderungsland}{country of immigration} and supported immigration policies aimed at attracting qualified workers to meet labor market demands \citep{SPD2021}.
%exhibits a spike in compassionate solidarity in 2014, followed by a gradual shift toward group-based anti-solidarity (ca.\ 20\% in 2024). %Part of this is explained by misclassifications of Llama-3.3-70B, as SPD speakers often cite anti-solidarity stances of other parties (especially AfD) and these citations are misclassified as anti-solidarity by the speaker (see Example 23). %20240410-bt-20-162-03269
%where an SPD speaker cites AfD proposals putting foreigners under general suspicion in order to criticize them

However, there are also instances where SPD speakers express \emph{anti-solidarity} by referring positively to their own policies having enabled deportations of declined asylum seekers or the difficulties associated with managing migration, similar to the frames of CDU/CSU and FDP speakers (see \exref{20240912-bt-20-185-00868}%before Example 24%20240912-bt-20-185-00868
).

\item[The left Green and Linke parties] only display very few instances classified as anti-solidarity. %and almost all of them are misclassifications where they cite others' anti-solidarity statements %or parliamentary motions by other parties 
%(see Example 25%20250605-bt-21-010-05076
%). These instances are nonetheless interesting because they express a polarizing influence of the AfD entering parliament, leading left-wing parties to positioning themselves in contrast \citep{Atzpodien2022}. 
Overall, since 2021, Die Grünen moved from \emph{compassionate solidarity} to \emph{group-based} appeals, stressing integration and rights. In the 2024/25 campaign, and in the 2024/25 campaign strongly emphasized \emph{exchange-based} arguments about migrants' economic and social contributions \citep{Gruene2025} (see \exref{20240119-bt-20-148-01053}%before: Example 26
). Die Linke %likewise began with 
likewise emphasized \emph{compassionate solidarity} but since 2021 started increasingly framing migration as integral to society, advancing a detailed pro-immigration agenda and strengthening \emph{group-based solidarity}. In 2024/25, %they stressed \emph{empathic solidarity}, 
the party continued to oppose stereotyping and position themselves against exclusionary discourses \citep{DieLinke2021, DieLinke2025} (see \exref{20250130-bt-20-210-02938}%before Example 27%20250130-bt-20-210-02942
). A special case is the BSW, which split from Linke party in 2024, partially due to explicit anti-migration positions. However, our data set contains only very few instances from this party (38 statements), so we cannot provide a reliable quantitative analysis.

\end{description}

\ak{As a robustness check on these party-level findings, we examine whether the per-party errors identified in \Cref{sec:error-analysis} (most notably the \emph{anti-solidarity} overprediction by Qwen-2.5 and Llama-3.3-70B for Grüne and Linke, and by Qwen-2.5 for FDP) carry over into the estimates discussed above. The corresponding \emph{anti-solidarity} skews are substantially reduced by the soft-label DSL correction, while two alternative DSL configurations---soft-label DSL without Qwen-2.5 and hard-label DSL using only gpt-oss-120B, thereby excluding both Qwen-2.5 and Llama-3.3-70B---produce only minor changes in these \emph{anti-solidarity} estimates. This suggests that these model-specific errors do not drive the substantive party patterns (see the full analysis in \Cref{app:party-conditioned-dsl-robustness} in the Appendix).}

\section{Discussion and Conclusion}\label{sec:discussion-conclusion}

In this paper, we have evaluated LLMs on the complex, theory-driven task of annotating (anti-)solidarity in German parliamentary speeches and assessed whether their outputs support valid downstream social-scientific inference. We then use a validated and statistically corrected annotation pipeline to derive new insights into solidarity and anti-solidarity toward migrants in the postwar period as well as the recent past. \ak{These analyses correspond to RQ1-RQ3, addressing annotation accuracy, downstream inference, and substantive temporal patterns, respectively.} In more detail, we provide both methodological and social scientific insights:

\paragraph{Methodological perspective} %We find that 
%Instruction-tuned LLMs like \textbf{Llama-3.3-70B and GPT-4 are effective tools for annotating solidarity frames in parliamentary debates}, with open models offering a more cost-effective alternative. In our experiments, Llama-3.3-70B performed competitively with GPT-4 (especially in few-shot), yet no model surpassed GPT-4, which itself remains below the human upper bound. 

\ak{Regarding RQ1}, our results show that strong \textbf{instruction-tuned LLMs %can approximate%--or even surpass--
%can reach human agreement in annotating solidarity frames in parliamentary debates}, 
\ak{can achieve macro-F1 scores comparable to human leave-one-out agreement when annotating solidarity frames in parliamentary debates}}, %where 
with \texttt{GPT-5} achieving the highest overall scores, while open models such as \texttt{gpt-oss-120B} and \texttt{Llama-3.3-70B} provide competitive and more cost-effective alternatives. At the same time, \textbf{performance varies systematically} by %target group, 
historical period, prompting strategy, and model family, with older data and fine-grained labels proving substantially harder. Generally, existing research presents mixed evidence on the reliability of LLMs for annotation tasks: some studies mention limitations in handling subjective or expert-defined taxonomies \citep{ziems2024can} or find that smaller fine-tuned models are more reliable in political science applications \citep{wang2024selecting}, while others show that LLMs can substitute for or outperform human coders \citep{heseltine2024large, tornberg2024large}. %This confirms that LLM-based annotation cannot be treated as a one-size-fits-all solution, but must be validated for the specific task.
Our findings are in line with a more cautious interpretation: \textbf{LLMs are useful annotation tools, but they should not be viewed as neutral coders, nor should their outputs be treated as \enquote{ground truth} data; rather, they are \enquote{measurement tools that require rigorous validation, calibration, and documentation}} \citep{baumann2025large}.
This idea is reinforced by our error analysis, which shows that even the strongest models in our study make systematic mistakes. For example, \texttt{Llama-3.3-70B} occasionally misclassifies anti-solidarity, mistaking criticism of anti-migrant rhetoric for anti-solidarity itself, and all models show a general tendency to overpredict the \emph{None} label. More broadly, they underestimate \emph{Mixed} cases and compress fine-grained subtype distinctions into broader categories, especially \emph{group-based} subtypes of (anti-)solidarity. These errors become more severe on historical instances, where less explicit forms of solidarity are often missed. More importantly, these errors are not independent across models: they are often correlated, which limits the benefit of ensemble methods. This matters for downstream analysis because small but systematic misclassifications can accumulate into biased temporal trends. 

\ak{Turning to RQ2, }recent work highlights this problem: high agreement with human annotators can coexist with systematic misclassifications, meaning that benchmark performance alone does not guarantee valid downstream inference \citep{bavaresco2025llms, gligoric2025unconfident}. Even highly accurate LLM-based surrogates can therefore result in misleading substantive conclusions when used directly in downstream analyses, because doing so ignores measurement error between gold-standard and surrogate labels \citep{egami2023using}.
Consistent with this, our study finds that \textbf{DSL-corrected soft-label outputs %best 
substantially %recover 
improve recovery of human-derived trends, especially in fine-grained categories}. \ak{At the high level, although uncorrected ensemble trends already show strong correlations with the human reference, DSL notably reduces RMSE.} This suggests that large-scale LLM annotation, combined with a limited set of expert annotations (around 2k in our case), is a viable strategy for downstream inference. The benefits are greatest when soft labels are aggregated across multiple strong models and corrected with DSL, rather than treated as hard predictions from a single model. Recent benchmarking further suggests that DSL performs especially well in low-sample settings, often outperforming alternative correction methods \citep{de2025benchmarking}. 

From a resource-allocation perspective, this pipeline combines expensive expert labels with low-cost model labels in a way that improves inference. In the broader annotation project, expert annotation cost over €29,000 for approximately 3,500 migrant- and woman-related instances, whereas model annotation scaled much more cheaply, whether via open-weight models such as \texttt{gpt-oss-120B}, \texttt{Llama-3.3-70B}, and \texttt{Qwen-2.5} on in-house CPU and NVIDIA A40 GPU infrastructure\footnote{\ak{These computational savings, however, also have environmental costs. We retrospectively estimate approximately 232kg CO$_2$ from GPU-board electricity use for the successful large-scale Llama-3.3-70B and Qwen-2.5 runs (see \Cref{app:environmental-cost} in the Appendix).}} or via GPT-4, which annotated one third of the dataset for about €500. A smaller set of costly human labels provides the reference basis for validation and correction, while a much larger set of low-cost model labels extends coverage to corpus scale.

\paragraph{Social science perspective} \ak{Turning to RQ3}, %beyond methodological contributions, 
our study shows how an LLM-based pipeline can support theoretically informed quantitative analysis through two case studies: one historical (1949–1957) and one contemporary (2009–2025). 

\textbf{For the early postwar period (1949–1957), we showed how solidarity was shaped and institutionalized in national reconstruction.} While this era has been examined in previous historical research, our LLM-based, fine-grained analysis offers additional insights. %for historical and social science research. 
First, it moves beyond the assumption that solidarity during this period was simply \enquote{high}: while historical consensus holds that it largely benefited ethnic Germans, \textbf{our data reveal coexisting forms of (anti-)solidarity and shifting rhetorical strategies across migrant groups}. Second, \textbf{our approach integrates migration histories that are often treated separately (expellees, %DPs, 
displaced persons, and refugees) into a shared analytical framework}, enabling direct comparison of how solidarity was articulated. This also sheds light on the emerging distinction between early DPs and the broader category of \textit{Ausländer}, including those later identified as \textit{Gastarbeiter}, showing that solidarity was not uniformly absent but selectively expressed. %Third, by providing a quantifiable and fine-grained view of solidarity, distinguishing subtypes across ideologies and migrant groups, our analysis challenges the zero-sum view of \de{Opferkonkurrenz}{competition of victimhood} and shows that multiple forms of solidarity could coexist in political discourse, shaped by group, context, and strategy.}

\textbf{For the recent past (2009-2025), we see a striking shift from solidarity to anti-solidarity following the 2015 migration crisis.} This may %be partially explained by 
reflect solidarity fatigue \citep{DeLeur2023} or %may be part of 
a larger societal right-wing shift \citep{Zick2023} on migration. While stricter migration stances across Europe have been studied, our study is the first to quantify this shift %from solidarity to anti-solidarity 
in German parliamentary debates. Our data also revealed that \textbf{the shift is not only explained by the right-wing AfD party entering parliament, but it also visible in the conservative CDU/CSU party and the liberal FDP party}, motivating further research. 

This development resonates with broader European trends, where populist radical right parties have increasingly mobilized around anti-immigration narratives \citep{aktas2024rise}, and even mainstream conservative parties have shifted their positions on migration in response to the electoral pressure of right-wing populists \citep{bayerlein2021chasing}. Comparative surveys in Europe likewise show anti-immigrant attitudes remain a central driver of right-wing populist support %across the continent 
\citep{messing2021anti}.

%Such fine-grained, longitudinal analysis of solidarity would hardly be feasible at such scale using manual annotation alone which, in our project, cost over €29,000 for only 3,500 instances. By contrast, with access to computing resources (in our case, in-house hardware with NVIDIA A40 GPUs), open-weight models like \texttt{gpt-oss-120B} and \texttt{Llama-3.3-70B} can annotate tens of thousands of instances at minimal cost. Even closed models like GPT-4 are cheaper, with one third of our dataset annotated for about €500. Machine annotation is thus not \enquote{free} but vastly more scalable, enabling analyses otherwise financially prohibitive. In this sense, our method bridges \enquote{distant reading} approaches common in large-scale text analysis with elements of \enquote{close reading}, enabling theoretically grounded systematic interpretation of parliamentary speech while retaining full corpus coverage. 

Taken together, our findings address a broader question in CSS: whether LLMs can support valid, theory-driven analysis of complex political concepts at scale. Our answer is yes, but only under clear conditions. LLMs are not neutral replacements for expert coders, and their outputs cannot be used uncritically as if they were ground-truth labels, particularly for theory-grounded annotation tasks. Yet when combined with human reference annotation, systematic validation, error analysis, and DSL-based correction, they can support serious social-scientific research by revealing patterns at a scale that would otherwise be accessible only through qualitative analysis. In this sense, our pipeline bridges \enquote{distant reading} approaches common in large-scale text analysis with elements of \enquote{close reading}, enabling theoretically grounded and systematic interpretation of parliamentary speech while retaining full corpus coverage.
%\clearpage
\section{Limitations and Future Work}\label{sec:limitations-future-work}

While our analysis was performed carefully to avoid methodological pitfalls, we nonetheless acknowledge remaining limitations of our study.

%\paragraph{Human Label Variation} 
\paragraph{Human Coded Reference} Our evaluation and DSL correction rely on human annotations as the best available expert-coded reference. %but these annotations still reflect some degree of interpretive uncertainty, 
At the same time, some degree of disagreement remains, especially at the fine-grained level, %Despite 
even after extensive training and refinement of the annotation manual. %substantial disagreement persisted, especially at the fine-grained level. 
This reflects the intrinsic difficulty of %annotating (anti-)solidarity 
interpreting theoretically informed categories in parliamentary speech and the possibility of reasonable disagreement (e.g.\, over whether competition over resources is \emph{group-} or \emph{exchange-based anti-solidarity}, or whether a text is better read as \emph{compassionate} or \emph{group-based solidarity}.). %We rely on majority labels here, but future work should explicitly model and incorporate human label variation \citep{plank2022problem}.
DSL improves %correlation with human annotations, 
recovery of human-derived trends, but it cannot remove %ambiguity when the human reference itself is uncertain, especially where subtype boundaries are unstable.
uncertainty where the reference itself remains interpretive.

\paragraph{Limits of Correction with DSL} DSL improves recovery of human-derived trends, especially relative to raw hard-label predictions, but it does not remove all sources of error. Its performance depends on the quality of the surrogate model outputs, the coverage of the human-annotated subset, and the sampling design. In settings with rare labels or sparse historical coverage, correction may remain unstable.

%\paragraph{Model Variations} We tested multiple models, prompts, and fine-tuning approaches (Llama, BERT), but many variations remain, including ensembles, higher-precision settings, and newer models. With refinement, machine performance may eventually match human annotation quality.

\paragraph{Temporal Reliability} Both model performance and human agreement fluctuate over time (e.g.\, the 1960s for several models; \Cref{fig:macro-f1-kappa-over-time} in the Appendix), reflecting the challenge of historical language. Although our correction framework mitigates part of this problem, historical variation remains a challenge for both annotation and downstream inference. Still, %correlations between GPT-4, Llama-3.3-70B, and human annotations remain high, and 
because our focus is on broad temporal trends rather than individual years, these fluctuations %do not affect our conclusions.
are less consequential for our main substantive conclusions.

\paragraph{(Anti-)Solidarity towards Women} In our previous work, we introduced initial annotations of (anti-)solidarity toward women alongside migrants \citep{kostikova-fine}. Extending this resource to a longitudinal analysis of solidarity toward women constitutes a promising direction for future work. \citet{Diabate2023} notes a leftward shift in Germany on gender issues, which may provide an interesting contrast to the migration trends identified in the present study.

\paragraph{Political Bias} %It may be that the 
LLMs compared in this paper may be subject to political bias as investigated by several prior works \citep{motoki2024more,rettenberger2025assessing}. %with left-wing political bias potentially leading to more anti-solidarity being annotated. Yet our error analysis (\Cref{fig:confusion-matrices-top}) shows no systematic skew toward anti-solidarity; models more often mislabel instances as \enquote{None}. 
\aknew{In our analysis, model errors vary across parties rather than showing a uniform directional skew: Qwen-2.5 and Llama-3.3-70B most clearly overpredict anti-solidarity for Grüne and Linke speakers, while Qwen-2.5 shows a related skew for FDP; errors for other parties are smaller or vary in direction across models (\Cref{sec:error-analysis}; Section \ref{app:party-conditioned-model-performance} in the Appendix). Our manual audit suggests that these discrepancies are often driven by stance misattribution rather than ideological bias. While such party-conditioned errors could affect between-party comparisons, DSL substantially reduces the identified skews.} Annotator bias towards certain labels is also possible but would need to be uniform across annotators, which we do not observe. 
%Even if present, bias would affect only overall levels of (anti-)solidarity, not the differences and trends central to our analysis.

\paragraph{Institutional Scope} Our analysis is limited to the federal parliamentary level. Patterns of (anti-)solidarity may differ across German state parliaments and at the European level, including in how the same parties frame migration across institutional contexts. Existing resources such as SpeakGer \citep{lange2023speakger}, which covers all 16 German state parliaments and the Bundestag, make such comparisons feasible and provide a promising direction for future work.

\paragraph{\aknew{Speaker-Level Variation}} \aknew{Our analysis focuses on aggregate temporal and party-level patterns and does not examine variation between individual speakers. Parliamentary debates can reveal intra-party variation in political positions \citep{fernandes2021unpacking}; future work could therefore use the available speaker metadata to investigate such variation in (anti-)solidarity framing.}

\paragraph{\aknew{Linguistic Realization}} \aknew{Future work could also examine how different forms of (anti-)solidarity are linguistically realized, for example through lexical and rhetorical cues and, where audio data are available, prosodic features.}
\section*{\aknew{Ethics Statement}}

\aknew{The human annotations were produced by student assistants with backgrounds in social science and computer science, who were employed under student assistant contracts and compensated at a rate of €14 per hour. Consent for the annotation work was obtained through these contracts. Formal ethics approval was not required because the task involved annotating publicly available German parliamentary transcripts and did not involve experimental interventions or the collection of sensitive or personal data from the annotators.}

\section*{\aknew{Data and Code Availability}}

\aknew{Human annotations, model outputs, and code used in this study are publicly available on \href{https://github.com/a-kostikova/LLM-Analysis-of-Solidarity}{GitHub}.}

\section*{\aknew{Acknowledgements}}

\aknew{A.K., O.P., and B.P. gratefully acknowledge funding for the project KI-Akademie OWL, financed by the Federal Ministry of Research, Technology and Space (BMFTR) and supported by the VDI/VDE Innovation + Technik GmbH under grant no. 16IS24057, and by the project SAIL, funded by the Ministry of Culture and Science of the State of North Rhine-Westphalia under grant no. NW21-059A. The NLLG Lab (Steffen Eger) gratefully acknowledges support from the German Research Foundation (DFG) through the Heisenberg Grant EG 375/5-1. Olga Sabelfeld gratefully acknowledges support from the German Research Foundation (DFG) through funding within the Collaborative Research Centre 1288 \enquote{Practices of Comparing}.}

% === Bibliography ===
\bibliographystyle{compling}
\bibliography{bibliography}

% === Appendix ===
\clearpage
\appendix
\renewcommand{\theHsection}{app.\Alph{section}}
\renewcommand{\theHsubsection}{app.\Alph{section}.\arabic{subsection}}
\renewcommand{\theHsubsubsection}{app.\Alph{section}.\arabic{subsection}.\arabic{subsubsection}}
\clearpage % temporary

\section*{Appendix}
\raggedbottom
\section{List of Keywords}
\label{app:appendix-keywords}

%For \de{Frau}{woman}, we use \textit{Frauen} (and \textit{Frau}), \textit{Mütter} (and \textit{Mutter}, \textit{Müttern}), \textit{Großmutter}, \textit{Hausfrauen} (and \textit{Hausfrau}), \textit{Ehefrauen} (and \textit{Ehefrau}), \textit{Mädchen}, \textit{Frauenförderung}, \textit{Frauenquote}, \textit{Dienstmädchen}, \textit{Fräulein}, \textit{Kriegerfrauen}, \textit{Arbeiterfrauen}, and \textit{Trümmerfrauen}.

%Since \de{Frau} is the German word for \textit{woman} but also for \textit{Mrs./Ms.}, we only include occurrences of the word \de{Frau} that are not followed by a capitalized word (which are probably surnames).

We use the following migration-related keywords: \textit{Flüchtlinge} (and \textit{Flüchtling}, \textit{Flüchtlingen}), \textit{Ausländer} (and \textit{Ausländern}, \textit{Ausländerinnen}), \textit{Asylsuchende} (and \textit{Asylsuchenden}), \textit{Asylbewerber} (and \textit{Asylbewerbern}), \textit{Heimatvertriebene} (and \textit{Heimatvertriebenen}), \textit{Vertriebene} (and \textit{Vertriebenen}, \textit{Vertriebener}), \textit{Aussiedler} (and \textit{Aussiedlern}), \textit{Einwanderer} (and \textit{Einwanderung}), \textit{Ansiedler} (and \textit{Ansiedlern}), \textit{Zuwanderer} (and \textit{Zuwanderern}, \textit{Zuwanderung}), \textit{Migranten} (and \textit{Migrantinnen}, \textit{Migration}), \textit{Sowjetzonenflüchtlinge}, \textit{Bürgerkriegsflüchtlinge}, \textit{Kriegsflüchtlinge}, \textit{Emigranten}, and \textit{Immigranten}.

For robustness, we conducted a stability test by randomly sampling 200 subsets of at least 5 keyword groups (grouping inflectional variants) each, ensuring that each instance covered at least 10\% of the dataset and spanned at least 75\% of the timeline (such that enough data is present to create the plots). %For each subset we computed Pearson's correlation coefficients between the resulting time-series trends, yielding 19,900 pairwise comparisons. The correlations are highly consistent, with averages of 0.92 for solidarity, 0.81 for anti-solidarity, and 0.94 for mixed. The 25\% quantiles are 0.90, 0.70, and 0.92, respectively. 
For each subset, we re-estimated the high-level DSL trends (as defined in \Cref{sec:dsl}) and compared the resulting time series to the full-sample trend using Pearson's correlation coefficient. The correlations remain high on average, with mean values of 0.93 for solidarity, 0.88 for anti-solidarity, and 0.82 for mixed; the 25\% quantiles are 0.89, 0.85, and 0.75, respectively. This indicates that the main aggregate trends are broadly robust to keyword composition, although robustness is weaker for mixed.

For the analysis of frequency of keywords over time, we calculate the percentages normalized for each keyword, i.e., a value of $p$\% in year $y$ implies that in year $y$ $p$\% of all sentences with this keyword occurred. The trends are shown in the \Cref{fig:keywords-distrib-over-time} in the Appendix. 

\section{Parties}
\label{app:appendix-parties}

For speaker and party extraction, we parse the XML protocols of the Bundestag and Reichstag. For modern sessions (20th–21st terms), we use the official Open Data XML, which already encodes speaker and party information; for earlier sessions (1st–19th terms), we rely on the improved markup provided by \citet{poradaautomatische}. 

Each new intervention is marked in the XML by a \texttt{<redner>} element containing both the speaker's name and their party. During preprocessing (XML parsing, transcript cleaning, and sentence segmentation), we convert these into standardized \texttt{<SPEAKER>} tags and extract the associated party information. During extraction of migrant-% or women-
related sentences, the most recent \texttt{<SPEAKER>} marker (with party) is assigned to each sentence until a new one appears.

List of the parties included in the dataset, along with the variations of their names or abbreviations as they have been recorded:
AfD (Alternative for Germany); Die Linke (The Left) with variations such as PDS, Gruppe der PDS; Bündnis 90/Die Grünen (Alliance 90/The Greens); CDU/CSU (Christian Democratic Union/Christian Social Union); SPD (Social Democratic Party of Germany); FDP (Free Democratic Party); DP (German Party) with variations such as DP/DPB, DP/FVP, FVP; GB/BHE (All-German Bloc/League of Expellees and Deprived of Rights); KPD (Communist Party of Germany); BP (Bavarian Party); WAV (Economic Reconstruction Union).

\section{Models Inference and Training Details}
\label{app:models-details}

\paragraph{Hardware and Inference Settings} Inference was conducted on mixed CPU and GPU environments. Closed-source models (\texttt{gpt-4-1106-preview}, \texttt{gpt-5-chat}) were accessed via API.

Open-weight models were run locally:

\begin{itemize}
    \item \texttt{gpt-oss-20B} on two NVIDIA A40 (80 GB) GPUs in bfloat16 precision;
    \item \texttt{Llama-3-8B-Instruct}, \texttt{Gemma-2-9b-it}, and \texttt{Gemma-2-27b-it} in full precision on a single A40 GPU;
    \item \texttt{Mistral-Large-Instruct-2407} on a rented NVIDIA A100 in 4-bit quantization to accommodate memory requirements;
    \item \texttt{Llama-3.3-70B-Instruct} in 4-bit quantization and \texttt{Qwen2.5-72B-Instruct} on two A40 GPUs in 8-bit quantization;
    \item and \texttt{gpt-oss-120B} and \texttt{Llama-4-Scout-17B-16E-Instruct} in 4-bit quantized form (\texttt{Q4\_K\_M} and \texttt{Q4\_K\_XL}, respectively) on CPU nodes (32 threads, 128 GB RAM) via \texttt{llama.cpp}\footnote{https://github.com/ggml-org/llama.cpp}.
\end{itemize}

All generations used temperature $= 0.6$ and top-p $= 0.9$.

\paragraph{Prompt Format} Each prompt consists of: (i) clear label definitions; (ii) chain-of-thought cues (e.g., \textit{think step by step});  (iii) a two-step classification format (high-level → subtype) (see the prompt text in \Cref{fig:two-step-prompt-migrant}% for \emph{Migrant} and in \Cref{fig:two-step-prompt-frau} in the Appendix for \emph{Woman}
).

\begin{figure*}[t]
  \centering
  \includegraphics[width=0.8\linewidth]{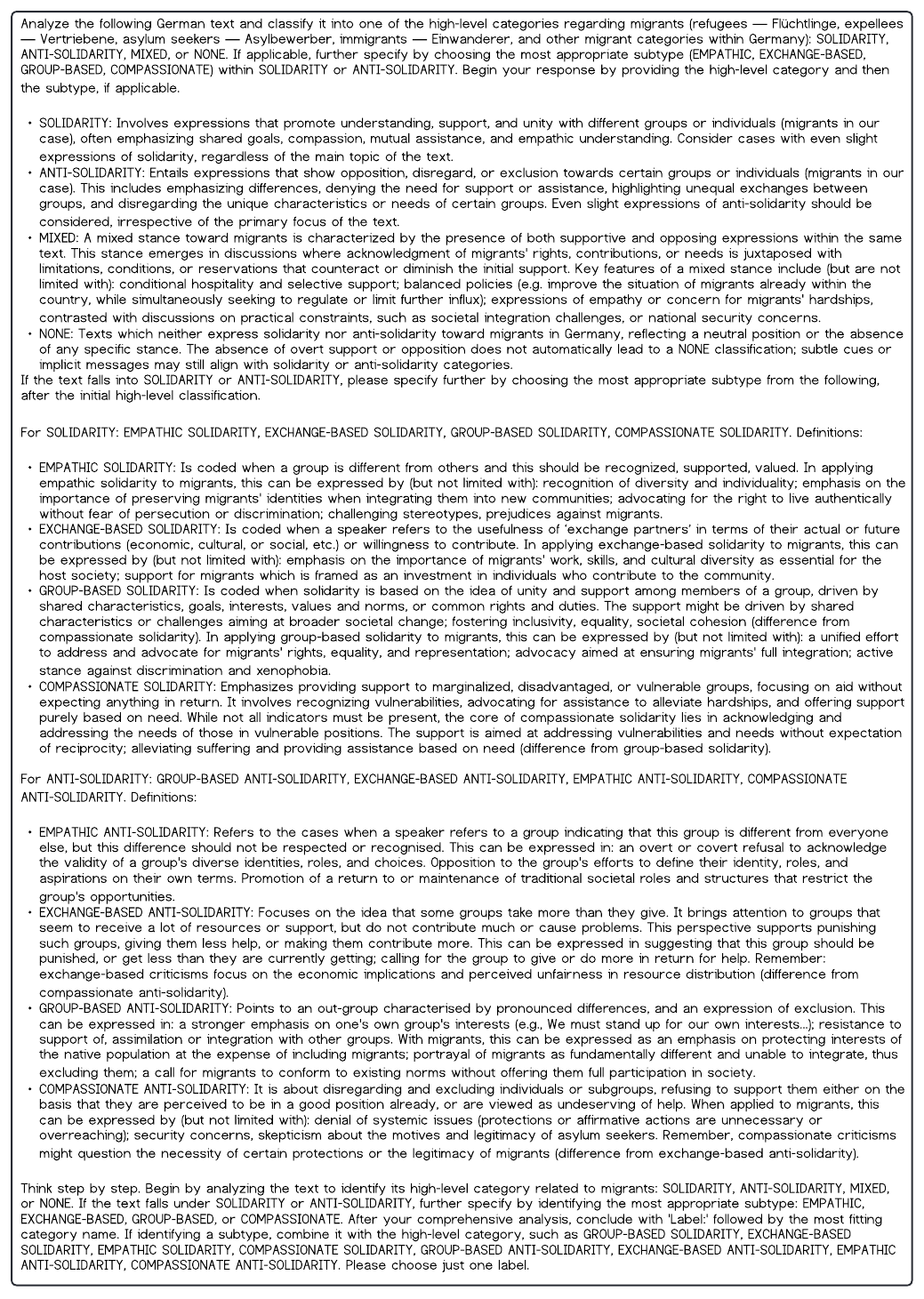}
  \caption{One-step zero-shot prompt set used for the evaluation of GPT-4 on the \emph{Migrant} dataset.}
  \label{fig:one-step-prompt-migrant}
\end{figure*}

In few-shot settings, one representative example is included for each class label. GPT-4 receives all instructions in a single prompt (see the prompt text in \Cref{fig:one-step-prompt-migrant}% for \emph{Migrant} and \Cref{fig:one-step-prompt-frau} in the Appendix for \emph{Woman}
). Open-weight models receive separate prompts: one for high-level classification, and one for subtype prediction if the label is solidarity or anti-solidarity.

\paragraph{Baseline fine-tuning details} For our baseline, we use a BERT-based pipeline with 110M parameters, comprising a high-level category classifier and two subtype models for \emph{solidarity} and \emph{anti-solidarity}. The baseline retains the original training setup of \citet{kostikova-fine}, in which the training data included both migrant- and woman-related instances. Accordingly, the models process inputs with a target indicator (\de{Frau} or \de{Migrant}) and the full text comprising the focus sentence and its left and right context. Evaluation in the present study is restricted to migrant-related instances. We add a fully connected layer with softmax activation atop the pooled output of the BERT-based models, with 4 output units for each model. To counter class imbalance, minority classes are oversampled to parity with the majority class. We finetune for 20 epochs with a learning rate of 4e-4, a warmup ratio of 0.05, linear decay, AdamW optimizer \citep{AdamW}, and categorical crossentropy loss.

Additionally, we use the multilingual SentenceBERT model \textit{distiluse-base-multilingual-cased-v2} \citep{reimers2019sentence}. We modify the network by expanding the embeddings from 512 to 1024 dimensions, then reducing them back to 512 dimensions. The model incorporates ReLU activation and dropout (0.1), and maps the embeddings to four and ten target classes for high-level and fine-grained tasks, respectively. To manage class imbalance, we use oversampling. The finetuning is conducted over 20 epochs with a batch size of 8, starting with a learning rate of 5e-4, a warmup ratio of 0.03, and using AdamW for optimization with learning rate decay and categorical crossentropy.

%\paragraph{Fine-Tuned Llama Models}

%We fine-tune \texttt{Llama-3-8B-It} on a single A40 GPU. \texttt{Llama-3-70B-It} is trained across two A40s using PyTorch FSDP and QLoRA\footnote{\url{https://www.philschmid.de/fsdp-qlora-llama3}}. Training spans 3 epochs with a constant learning rate of 2e-4, AdamW optimizer, and max gradient norm of 0.3.

% Auto-generated significance-testing appendix section.
% Requires: \usepackage{booktabs} and the \ak{...} revision macro

\section{\ak{Significance Testing}}
\label{app:significance-testing}

\ak{\Cref{tab:model-significance-tests} reports the planned pairwise comparisons discussed in \Cref{sec:overall-model-performance}. After Holm correction, %GPT-5 significantly outperforms gpt-oss-120B only for the high-level \textit{Woman} task
GPT-5 and gpt-oss-120B do not differ significantly in %the other five 
any of the evaluation settings. In contrast, gpt-oss-120B shows clearer advantages over smaller or lower-performing open-weight models, particularly for %fine-grained \textit{Woman} and 
both levels of \textit{Migrant (T1)}. These differences are substantially less consistent for the historical \textit{Migrant (T2)} set.}

\begin{table*}[h]
\centering
\scriptsize
\setlength{\tabcolsep}{4pt}
\renewcommand{\arraystretch}{1.08}
\begin{tabular}{@{}p{0.31\textwidth}rrrrrr@{}}
\toprule
\ak{\textbf{Comparison}} & \ak{\textbf{F1 A}} & \ak{\textbf{F1 B}} & \ak{\boldmath$\Delta$\textbf{F1}} & \ak{\textbf{95\% CI}} & \ak{\boldmath$p_{\mathrm{raw}}$} & \ak{\boldmath$p_{\mathrm{Holm}}$} \\
\midrule
\multicolumn{7}{@{}l}{\ak{\textit{Migrant (T1) --- Fine-grained}}} \\
\addlinespace[0.15em]
\ak{\texttt{GPT-5} -- \texttt{gpt-oss-120B}} & \ak{0.500} & \ak{0.485} & \ak{0.015} & \ak{[-0.029, 0.059]} & \ak{0.5215} & \ak{0.5215} \\
\ak{\texttt{gpt-oss-120B} -- \texttt{Llama-3.3-70B}} & \ak{0.485} & \ak{0.376} & \ak{0.109} & \ak{[0.066, 0.154]} & \ak{0.0002} & \ak{\textbf{0.0010}} \\
\ak{\texttt{gpt-oss-120B} -- \texttt{Qwen-2.5}} & \ak{0.485} & \ak{0.407} & \ak{0.078} & \ak{[0.036, 0.121]} & \ak{0.0004} & \ak{\textbf{0.0012}} \\
\ak{\texttt{gpt-oss-120B} -- \texttt{Gemma-2-27B}} & \ak{0.485} & \ak{0.325} & \ak{0.161} & \ak{[0.118, 0.205]} & \ak{0.0002} & \ak{\textbf{0.0010}} \\
\ak{\texttt{gpt-oss-120B} -- \texttt{gpt-oss-20B}} & \ak{0.485} & \ak{0.400} & \ak{0.085} & \ak{[0.041, 0.130]} & \ak{0.0008} & \ak{\textbf{0.0016}} \\
\midrule
\multicolumn{7}{@{}l}{\ak{\textit{Migrant (T1) --- High-level}}} \\
\addlinespace[0.15em]
\ak{\texttt{GPT-5} -- \texttt{gpt-oss-120B}} & \ak{0.714} & \ak{0.700} & \ak{0.014} & \ak{[-0.042, 0.066]} & \ak{0.6269} & \ak{0.6269} \\
\ak{\texttt{gpt-oss-120B} -- \texttt{Llama-3.3-70B}} & \ak{0.700} & \ak{0.647} & \ak{0.053} & \ak{[-0.003, 0.109]} & \ak{0.0664} & \ak{0.1328} \\
\ak{\texttt{gpt-oss-120B} -- \texttt{Qwen-2.5}} & \ak{0.700} & \ak{0.611} & \ak{0.090} & \ak{[0.037, 0.142]} & \ak{0.0020} & \ak{\textbf{0.0060}} \\
\ak{\texttt{gpt-oss-120B} -- \texttt{Gemma-2-27B}} & \ak{0.700} & \ak{0.538} & \ak{0.163} & \ak{[0.115, 0.210]} & \ak{0.0002} & \ak{\textbf{0.0010}} \\
\ak{\texttt{gpt-oss-120B} -- \texttt{gpt-oss-20B}} & \ak{0.700} & \ak{0.563} & \ak{0.138} & \ak{[0.088, 0.191]} & \ak{0.0002} & \ak{\textbf{0.0010}} \\
\midrule
\multicolumn{7}{@{}l}{\ak{\textit{Migrant (T2) --- Fine-grained}}} \\
\addlinespace[0.15em]
\ak{\texttt{GPT-5} -- \texttt{gpt-oss-120B}} & \ak{0.338} & \ak{0.307} & \ak{0.031} & \ak{[-0.016, 0.079]} & \ak{0.1908} & \ak{0.5723} \\
\ak{\texttt{gpt-oss-120B} -- \texttt{Llama-3.3-70B}} & \ak{0.307} & \ak{0.288} & \ak{0.020} & \ak{[-0.028, 0.070]} & \ak{0.4176} & \ak{0.5723} \\
\ak{\texttt{gpt-oss-120B} -- \texttt{Qwen-2.5}} & \ak{0.307} & \ak{0.284} & \ak{0.024} & \ak{[-0.019, 0.066]} & \ak{0.2734} & \ak{0.5723} \\
\ak{\texttt{gpt-oss-120B} -- \texttt{Gemma-2-27B}} & \ak{0.307} & \ak{0.256} & \ak{0.052} & \ak{[0.007, 0.097]} & \ak{0.0234} & \ak{0.1170} \\
\ak{\texttt{gpt-oss-120B} -- \texttt{gpt-oss-20B}} & \ak{0.307} & \ak{0.262} & \ak{0.045} & \ak{[0.004, 0.088]} & \ak{0.0316} & \ak{0.1264} \\
\midrule
\multicolumn{7}{@{}l}{\ak{\textit{Migrant (T2) --- High-level}}} \\
\addlinespace[0.15em]
\ak{\texttt{GPT-5} -- \texttt{gpt-oss-120B}} & \ak{0.579} & \ak{0.569} & \ak{0.010} & \ak{[-0.051, 0.073]} & \ak{0.7311} & \ak{1.0000} \\
\ak{\texttt{gpt-oss-120B} -- \texttt{Llama-3.3-70B}} & \ak{0.569} & \ak{0.567} & \ak{0.001} & \ak{[-0.060, 0.061]} & \ak{0.9689} & \ak{1.0000} \\
\ak{\texttt{gpt-oss-120B} -- \texttt{Qwen-2.5}} & \ak{0.569} & \ak{0.507} & \ak{0.062} & \ak{[0.006, 0.117]} & \ak{0.0306} & \ak{0.1224} \\
\ak{\texttt{gpt-oss-120B} -- \texttt{Gemma-2-27B}} & \ak{0.569} & \ak{0.528} & \ak{0.040} & \ak{[-0.022, 0.103]} & \ak{0.2210} & \ak{0.6629} \\
\ak{\texttt{gpt-oss-120B} -- \texttt{gpt-oss-20B}} & \ak{0.569} & \ak{0.492} & \ak{0.077} & \ak{[0.024, 0.130]} & \ak{0.0062} & \ak{\textbf{0.0310}} \\
\bottomrule
\end{tabular}
\caption{\ak{Paired stratified bootstrap comparisons of the planned few-shot model pairs. Positive differences favor the first model. Unadjusted 95\% percentile bootstrap confidence intervals and two-sided bootstrap $p$-values are based on 10,000 resamples. Holm correction is applied to the five $p$-values within each test-set and annotation-level combination. Bold values indicate $p_{\mathrm{Holm}}<.05$.}}
\label{tab:model-significance-tests}
\end{table*}

\begin{table*}[h]
\centering
\scriptsize
\setlength{\tabcolsep}{4pt}
\renewcommand{\arraystretch}{1.08}
\begin{tabular}{@{}p{0.31\textwidth}rrrrrr@{}}
\toprule
\ak{\textbf{Comparison}} & \ak{\textbf{F1 A}} & \ak{\textbf{F1 B}} & \ak{\boldmath$\Delta$\textbf{F1}} & \ak{\textbf{95\% CI}} & \ak{\boldmath$p_{\mathrm{raw}}$} & \ak{\boldmath$p_{\mathrm{Holm}}$} \\
\midrule
%\multicolumn{7}{@{}l}{\ak{\textit{Woman --- Fine-grained}}} \\
%\addlinespace[0.15em]
%\ak{\texttt{Llama-3.3-70B} -- \texttt{Gemma-2-27B}} & \ak{0.278} & \ak{0.242} & \ak{0.036} & \ak{[-0.019, 0.085]} & \ak{0.2198} & \ak{0.2198} \\
%\ak{\texttt{Qwen-2.5} -- \texttt{Gemma-2-27B}} & \ak{0.294} & \ak{0.242} & \ak{0.052} & \ak{[0.007, 0.097]} & \ak{0.0244} & \ak{\textbf{0.0488}} \\
%\midrule
%\multicolumn{7}{@{}l}{\ak{\textit{Woman --- High-level}}} \\
%\addlinespace[0.15em]
%\ak{\texttt{Llama-3.3-70B} -- \texttt{Gemma-2-27B}} & \ak{0.491} & \ak{0.524} & \ak{-0.033} & \ak{[-0.092, 0.024]} & \ak{0.2572} & \ak{0.2572} \\
%\ak{\texttt{Qwen-2.5} -- \texttt{Gemma-2-27B}} & \ak{0.470} & \ak{0.524} & \ak{-0.054} & \ak{[-0.105, -0.004]} & \ak{0.0356} & \ak{0.0712} \\
%\midrule
\multicolumn{7}{@{}l}{\ak{\textit{Migrant (T1) --- Fine-grained}}} \\
\addlinespace[0.15em]
\ak{\texttt{Llama-3.3-70B} -- \texttt{Gemma-2-27B}} & \ak{0.376} & \ak{0.325} & \ak{0.051} & \ak{[0.009, 0.095]} & \ak{0.0164} & \ak{\textbf{0.0164}} \\
\ak{\texttt{Qwen-2.5} -- \texttt{Gemma-2-27B}} & \ak{0.407} & \ak{0.325} & \ak{0.083} & \ak{[0.043, 0.125]} & \ak{0.0002} & \ak{\textbf{0.0004}} \\
\midrule
\multicolumn{7}{@{}l}{\ak{\textit{Migrant (T1) --- High-level}}} \\
\addlinespace[0.15em]
\ak{\texttt{Llama-3.3-70B} -- \texttt{Gemma-2-27B}} & \ak{0.647} & \ak{0.538} & \ak{0.109} & \ak{[0.062, 0.159]} & \ak{0.0002} & \ak{\textbf{0.0004}} \\
\ak{\texttt{Qwen-2.5} -- \texttt{Gemma-2-27B}} & \ak{0.611} & \ak{0.538} & \ak{0.073} & \ak{[0.028, 0.119]} & \ak{0.0026} & \ak{\textbf{0.0026}} \\
\midrule
\multicolumn{7}{@{}l}{\ak{\textit{Migrant (T2) --- Fine-grained}}} \\
\addlinespace[0.15em]
\ak{\texttt{Llama-3.3-70B} -- \texttt{Gemma-2-27B}} & \ak{0.288} & \ak{0.256} & \ak{0.032} & \ak{[-0.011, 0.074]} & \ak{0.1428} & \ak{0.2856} \\
\ak{\texttt{Qwen-2.5} -- \texttt{Gemma-2-27B}} & \ak{0.284} & \ak{0.256} & \ak{0.028} & \ak{[-0.010, 0.067]} & \ak{0.1566} & \ak{0.2856} \\
\midrule
\multicolumn{7}{@{}l}{\ak{\textit{Migrant (T2) --- High-level}}} \\
\addlinespace[0.15em]
\ak{\texttt{Llama-3.3-70B} -- \texttt{Gemma-2-27B}} & \ak{0.567} & \ak{0.528} & \ak{0.039} & \ak{[-0.021, 0.098]} & \ak{0.2070} & \ak{0.4140} \\
\ak{\texttt{Qwen-2.5} -- \texttt{Gemma-2-27B}} & \ak{0.507} & \ak{0.528} & \ak{-0.022} & \ak{[-0.080, 0.036]} & \ak{0.4434} & \ak{0.4434} \\
\bottomrule
\end{tabular}
\caption{\ak{Additional paired stratified bootstrap comparisons of Llama-3.3-70B and Qwen-2.5 against Gemma-2-27B. Positive differences favor the first model. Unadjusted 95\% percentile bootstrap confidence intervals and two-sided bootstrap $p$-values are based on 10,000 resamples. Holm correction is applied to the two $p$-values within each test-set and annotation-level combination. Bold values indicate $p_{\mathrm{Holm}}<.05$.}}
\label{tab:model-significance-tests-additional}
\end{table*}

\ak{\Cref{tab:model-significance-tests-additional} reports additional comparisons of Llama-3.3-70B and Qwen-2.5 with Gemma-2-27B. Both models significantly outperform Gemma-2-27B on \textit{Migrant (T1)}, whereas the corresponding differences are not significant on \textit{Migrant (T2)}% and are mixed for \textit{Woman}
.}

\section{Error Correlation}
\label{app:error-metrics}

For two models $m_i$ and $m_j$, let $\mathcal{E}_i$ and $\mathcal{E}_j$ denote the sets of instances misclassified by each model, and let $\mathcal{E}^{\text{same}}_{ij}$ be the subset where both models predict the same wrong label. 
We compute two measures of pairwise error overlap, reported in \Cref{tab:pairwise_error_overlap}.

\paragraph{Error Jaccard}
\[
J_{ij} = \frac{|\mathcal{E}_i \cap \mathcal{E}_j|}{|\mathcal{E}_i \cup \mathcal{E}_j|}
\]
This measures how often two models make errors on the same instances relative to their total error sets.

\paragraph{Same-Wrong Rate}
\[
P_{\text{same-wrong} \mid \text{both-wrong}} = 
\frac{|\mathcal{E}^{\text{same}}_{ij}|}{|\mathcal{E}_i \cap \mathcal{E}_j|}
\]
This measures the proportion of jointly misclassified instances on which two models predict the same incorrect label. \Cref{tab:pairwise_error_overlap} reports the resulting values for all model pairs.

\begin{table}[h!]
\centering
\renewcommand{\arraystretch}{1.15}
\footnotesize
\begin{tabular}{@{}l l c c@{}}
\toprule
\textbf{Model 1} & \textbf{Model 2} & \textbf{Error Jaccard} & \textbf{P(Same Wrong $\mid$ Both Wrong)} \\
\midrule
gpt-oss-120B & Qwen-2.5        & 0.638 & 0.493 \\
Llama-3.3-70B & Qwen-2.5       & 0.609 & 0.453 \\
Llama-3.3-70B & gpt-oss-120B   & 0.573 & 0.461 \\
\bottomrule
\end{tabular}
\caption{
Pairwise overlap in model prediction errors.
\textit{Error Jaccard} measures the proportion of shared errors relative to the union of all errors between two models.
\textit{P(Same Wrong $\mid$ Both Wrong)} measures how often two models predict the same wrong label when both are incorrect.
}
\label{tab:pairwise_error_overlap}
\end{table}

\section{Label Distribution Distance Metrics}
\label{app:label-distance}

We compare model and human label distributions (\Cref{sec:label-distrib-similarity}) using Jensen--Shannon Divergence and total variation distance.

\paragraph{Jensen--Shannon Divergence.}
Let $p$ and $q$ denote discrete probability distributions over the same label set, and let
$m = \frac{1}{2}(p + q)$.
The Jensen--Shannon divergence is defined as
\[
\mathrm{JSD}(p \,\|\, q)
= \frac{1}{2}\mathrm{KL}(p \,\|\, m)
+ \frac{1}{2}\mathrm{KL}(q \,\|\, m),
\]
where $\mathrm{KL}(\cdot\|\cdot)$ denotes the Kullback--Leibler divergence.
Lower values indicate more similar label distributions.

\paragraph{Total Variation Distance.}
Total variation distance measures absolute differences in probability mass:
\[
\mathrm{TV}(p, q)
= \frac{1}{2} \sum_i |p_i - q_i|.
\]
It ranges from 0 (identical distributions) to 1 (disjoint support).

\begin{table}[ht]
\centering
\begin{tabular}{lcc}
\toprule
\textbf{Comparison} & \textbf{Total Variation} & \textbf{Jensen–Shannon Divergence} \\
\midrule
Model ensemble vs humans & 0.49 (0.32) & 0.39 (0.22) \\
Human vs human (avg 10 splits) & 0.45 (0.19) & 0.35 (0.12) \\
Randomized model baseline & 0.75 (0.59) & 0.69 (0.47) \\
\textit{gpt-oss-120B} vs humans & 0.58 (0.36) & 0.48 (0.29) \\
\textit{Qwen-2.5} vs humans & 0.59 (0.42) & 0.50 (0.34) \\
\textit{Llama-3.3-70B} vs humans & 0.57 (0.40) & 0.48 (0.32) \\
\textit{GPT-5} vs humans & 0.57 (0.33) & 0.47 (0.25) \\
\bottomrule
\end{tabular}
\caption{Distribution distances between human and model label distributions (lower = closer alignment). 
Values are shown as fine-grained (high-level).}
\label{tab:distribution-distances}
\end{table}

\section{\ak{Environmental Cost of Large-Scale Inference}}\label{app:environmental-cost}

\ak{Following \citet{lacoste2019quantifying}, we retrospectively estimate GPU-board electricity use and CO$_2$ emissions from hardware, runtime, and grid carbon intensity, excluding CPU/RAM use and facility overhead.}

\begin{table}[ht]
\centering
\small
\begin{tabular}{lrrr}
\toprule
\ak{\textbf{Model}} & \ak{\textbf{A40 GPU-h}} & \ak{\textbf{kWh}} & \ak{\textbf{CO$_2$ (kg)}} \\
\midrule
\ak{Llama-3.3-70B} & \ak{1,035.9} & \ak{310.8} & \ak{106.9} \\
\ak{Qwen-2.5}      & \ak{1,211.8} & \ak{363.5} & \ak{125.1} \\
\midrule
\ak{\textbf{Total}} & \ak{\textbf{2,247.7}} & \ak{\textbf{674.3}} & \ak{\textbf{232.0}} \\
\bottomrule
\end{tabular}
\caption{\ak{Estimated GPU-board energy use and CO$_2$ emissions, assuming 300~W per A40 and 344~g CO$_2$/kWh.}}
\label{tab:inference-environmental-cost}
\end{table}

\ak{Using the A40 maximum board power of 300~W and the German Environment Agency's estimated 2025 grid factor of 344 g CO$_2$/kWh \citep{icha2026strommix}, the successful Llama-3.3-70B and Qwen-2.5 runs correspond to approximately 232 kg CO$_2$. We do not estimate emissions for gpt-oss-120B because the physical hardware and power consumption of its virtualized CPU infrastructure were unavailable.}

\section{\ak{Audit of misattributed anti-solidarity errors}} \label{app:appendix-attribution-audit} 

\ak{To quantify the misattributed anti-solidarity error discussed in \Cref{sec:error-analysis}, we conduct a targeted audit of human-annotated instances from 1990 onward for which the human reference is not \emph{anti-solidarity} but at least one of Llama-3.3-70B, Qwen-2.5, or gpt-oss-120B predicts \emph{anti-solidarity}. This results in 203 candidate instances containing 319 individual model \emph{anti-solidarity} predictions (\Cref{tab:attribution-audit-sample}). For the robustness analysis of the recent increase in \emph{anti-solidarity} (\Cref{sec:recent-overall-solidarity-distrib}), we focus on the subset from 2015 onward, comprising 62 candidate instances and 111 individual model \emph{anti-solidarity} predictions.}

\begin{table}[htb]
\centering
\small
\renewcommand{\arraystretch}{1.15}
\setlength{\tabcolsep}{6pt}
\begin{tabular}{lrr}
\toprule
\ak{\textbf{Audit sample construction}} &
\ak{\textbf{1990-onward}} &
\ak{\textbf{2015-onward}} \\
\midrule
\ak{Human-annotated instances} &
\ak{1123} &
\ak{441} \\
\ak{All model-AS predictions} &
\ak{912} &
\ak{479} \\
\ak{Candidate instances: human $\neq$ AS, at least one model $=$ AS} &
\ak{203} &
\ak{62} \\
\ak{Individual model-AS predictions in candidate instances} &
\ak{319} &
\ak{111} \\
\bottomrule
\end{tabular}
\caption{\ak{Construction of the anti-solidarity false-positive audit sample. The broader audit covers cases from 1990 onward; the 2015-onward subset is used in a robustness check for the recent period in \Cref{sec:recent-overall-solidarity-distrib}.}}
\label{tab:attribution-audit-sample}
\end{table}

\ak{Each prediction is manually assigned to one of three categories (see examples of each in \Cref{tab:attribution-audit-examples}):} 

\begin{enumerate} 

\item \ak{\textbf{Misattributed anti-solidarity.} The model assigns \emph{anti-solidarity} to the current speaker based on anti-migrant rhetoric or policies that the speaker is criticizing rather than expressing as their own stance.} 
\item \ak{\textbf{Other clear anti-solidarity false positive.} The model predicts \emph{anti-solidarity}, but the error has a different source, e.g. other misinterpretations of the speaker's stance.%for example when a pro-migrant policy demand is interpreted as anti-migrant.
} 
\item \ak{\textbf{Plausible disagreement with the human reference.} The model predicts \emph{anti-solidarity} although the human reference does not, but the model's interpretation is sufficiently plausible that we do not treat the prediction as a clear false positive. %This does not mean that the model prediction is necessarily correct; rather, the instance allows a reasonable \emph{anti-solidarity} interpretation despite disagreement with the human annotation.
}

\end{enumerate}

\begin{table}[htb]
\centering
\small
\renewcommand{\arraystretch}{1.15}
\setlength{\tabcolsep}{6pt}
\begin{tabular}{lcc}
\toprule
\ak{\textbf{Audit outcome}} &
\ak{\shortstack{\textbf{1990 onward}\\($N_{\mathrm{AS}}=912$)}} &
\ak{\shortstack{\textbf{2015 onward}\\($N_{\mathrm{AS}}=479$)}} \\
\midrule
\ak{(1) Misattributed anti-solidarity errors} &
\ak{139 (15.2\%)} &
\ak{38 (7.9\%)} \\
\ak{(2) Other clear AS false positives} &
\ak{38 (4.2\%)} &
\ak{16 (3.3\%)} \\
\ak{(1+2) All clear AS false positives} &
\ak{177 (19.4\%)} &
\ak{54 (11.3\%)} \\
\ak{(3) Plausible human-model disagreements} &
\ak{142 (15.6\%)} &
\ak{57 (11.9\%)} \\
\bottomrule
\end{tabular}
\caption{\ak{Manual coding results for candidate model-AS predictions. Entries report the number of predictions assigned to each audit outcome, with the percentage of all model anti-solidarity predictions in the corresponding period in parentheses. %The denominators are 912 model anti-solidarity predictions from 1990 onward and 479 from 2015 onward.
}}
\label{tab:attribution-audit-results}
\end{table}

\ak{As shown in \Cref{tab:attribution-audit-results}, across the broader 1990-onward audit, misattributed anti-solidarity errors account for 15.2\% of all model anti-solidarity predictions and all clear anti-solidarity false positives for 19.4\%. In the 2015-onward subset relevant to the recent trend, the corresponding rates are 7.9\% and 11.3\%, respectively; we use these rates in the sensitivity analysis in \Cref{sec:recent-overall-solidarity-distrib}.}

%\ak{Examples: 
%
%1) 20151125-bt-18-139-04137 
%2) 
%3) 
%}

\begin{table*}[ht]
\centering
\footnotesize
\renewcommand{\arraystretch}{1.12}
\setlength{\tabcolsep}{5pt}

\begin{tabular}{
    >{\raggedright\arraybackslash}p{0.31\textwidth}
    >{\raggedright\arraybackslash}p{0.31\textwidth}
    >{\raggedright\arraybackslash}p{0.31\textwidth}
}
\toprule

\ak{\textbf{Original text}} &
\ak{\textbf{Translation}} &
\ak{\textbf{Example model response}} \\

% ============================================================
% EXAMPLE 1
% ============================================================

\specialrule{\lightrulewidth}{2pt}{2pt}

\multicolumn{3}{
    >{\raggedright\arraybackslash}p{\dimexpr\textwidth-2\tabcolsep\relax}
}{
    \ak{\textbf{Audit category (1): Misattributed anti-solidarity}} % 20180628-bt-19-042-06729
} \\

\cmidrule(lr){1-3}

\ak{\enquote{
[...] Die AfD, also die mit dem ganz kleinen braunen Karo, selbst wenn sie blaue Jeansjacketts tragen, hat mal wieder einen Skandal aufgedeckt, der aber gar keiner ist: Kinder von Ausländern erhalten gutes deutsches Kindergeld, sogar in Polen oder Rumänien. \enquote{Florida-Rolf} ist jetzt zehn Jahre alt, heißt \enquote{Andrzej} oder \enquote{Bogdana} und lässt es sich in Warschau oder Bukarest auf deutsche Kosten gut gehen. Welche Ungerechtigkeit! [...]
}}
&
\ak{\enquote{
[...] The AfD, the party with the little brown check pattern even when its members wear blue denim jackets, has once again uncovered a scandal that is not actually a scandal at all: children of foreigners receive good German child benefits, even in Poland or Romania. \enquote{Florida Rolf} is now ten years old, is called \enquote{Andrzej} or \enquote{Bogdana,} and is supposedly living it up in Warsaw or Bucharest at German taxpayers' expense. What an injustice! [...]
}}
&
\ak{
\texttt{gpt-oss-120B}: \antisolidarity{anti-solidarity}; \texttt{human reference}: \solidarity{solidarity}}

\smallskip

\ak{\emph{Label: ANTI‑SOLIDARITY; Explanation: The speaker denounces the fact that children of foreigners receive German child benefits, calling it an injustice and a scandal. This frames migrants as undeserving recipients of public resources and expresses opposition to their inclusion, without any supportive remarks.}
}
\\

% ============================================================
% EXAMPLE 2
% ============================================================

\specialrule{\lightrulewidth}{4pt}{2pt}

\multicolumn{3}{
    >{\raggedright\arraybackslash}p{\dimexpr\textwidth-2\tabcolsep\relax}
}{
    \ak{\textbf{Audit category (2): Other clear anti-solidarity false positive}} % 20210324-bt-19-217-01256
} \\

\cmidrule(lr){1-3}

\ak{\enquote{
[...] Das türkische Regime verletzt [gemeinsame] Werte auf offener Weltbühne, und das seit Jahren. Die türkische Regierung erpresst Europa. Sie droht, mal offen und mal weniger offen, europäische NATO-Partner mit unkontrolliertem Zustrom von Migration und Flüchtlingen zu destabilisieren. [...] Sie will auch unser Schweigen angesichts einer immer schlimmer werdenden Verfolgung unschuldiger Bürgerinnen und Bürger der Türkei im Inneren. [...]
}}
&
\ak{\enquote{
[...] The Turkish regime has been violating [shared] values openly on the world stage for years. The Turkish government is blackmailing Europe. It threatens, sometimes openly and sometimes less openly, to destabilize its European NATO partners through an uncontrolled influx of migrants and refugees. [...] It also wants to secure our silence in the face of the increasingly severe persecution of innocent citizens within Turkey. [...]
}}
&
\ak{
\texttt{Qwen-2.5}: \antisolidarity{anti-solidarity}, \texttt{human reference}: \none{none}}

\smallskip

\ak{\emph{Label: ANTI-SOLIDARITY; Explanation: The text criticizes Turkey's government for using the threat of uncontrolled migration and refugees to extort Europe, framing migrants as a tool of political leverage and a potential source of destabilization. This portrayal of migrants as a threat aligns with an anti-solidarity stance.}
}
\\

% ============================================================
% EXAMPLE 3
% ============================================================

\specialrule{\lightrulewidth}{4pt}{2pt}

\multicolumn{3}{
    >{\raggedright\arraybackslash}p{\dimexpr\textwidth-2\tabcolsep\relax}
}{
    \ak{\textbf{Audit category (3): Plausible disagreement with the human reference}} % 20200422-bt-19-155-02439
} \\

\cmidrule(lr){1-3}

\ak{\enquote{
[...] Der Schleppermafia ist das Handwerk gelegt worden, die Grenzen werden kontrolliert. Australien hilft jetzt jährlich 20 000 echten Flüchtlingen und nicht 10 000 anderen, die den Schleppern die allerhöchsten Preise zahlen können.	[...] jeder vernünftige Neuanfang in der Asylpolitik muss sich an solchen modernen Erfolgsmodellen orientieren. Wenn nicht, wird der linksgrüne Traum von einer Welt ohne Grenzen zu einem grenzenlosen Albtraum für unser Land und für Europa. Und das müssen und werden wir verhindern, meine Damen und Herren.
}}
&
\ak{\enquote{
[...] The smuggling networks have been put out of business, and the borders are being controlled. Australia now helps 20,000 genuine refugees each year instead of 10,000 others who can afford to pay smugglers the highest prices. [...] Any sensible new beginning in asylum policy must follow such modern models of success. Otherwise, the left-green dream of a world without borders will turn into a boundless nightmare for our country and for Europe. And that is something we must and will prevent.
}}
&
\ak{
\texttt{Llama-3.3-70B}: \antisolidarity{anti-solidarity}, \texttt{human reference}: \mixed{mixed}}

\smallskip

\ak{\emph{Label: ANTI-SOLIDARITY Explanation: The text expresses opposition towards a liberal migration policy, emphasizing the need for control and the preference for helping \enquote{echten Flüchtlingen} (genuine refugees) over those who can pay the highest prices, implying a critique of asylum seekers who are not considered genuine. The text also warns of a \enquote{grenzenlosen Albtraum} (boundless nightmare) for the country and Europe if a strict migration policy is not implemented, indicating a stance against open borders and unregulated migration. [...]}
}
\\

\bottomrule
\end{tabular}

\caption{\ak{Examples from the manual audit of anti-solidarity predictions. For each audit category, we show the original German passage, its English translation, and one representative model response.}}
\label{tab:attribution-audit-examples}
\end{table*}

\clearpage

\subsection{\ak{Sensitivity to misattributed anti-solidarity errors}}\label{app:appendix-attribution-sensitivity}

\ak{We conduct two complementary checks to assess the robustness of the recent increase in anti-solidarity discussed in \Cref{sec:recent-overall-solidarity-distrib} to the misattributed anti-solidarity errors identified in the manual audit (\Cref{app:appendix-attribution-audit} in the Appendix).}

\ak{\textit{First-stage calibration diagnostic.} We use the 2,184-instance human-annotated subset to draw the human subsamples used for DSL calibration and examine whether the first-stage calibration component lowers anti-solidarity on the manually audited predictions from 2015 onward. We repeat this diagnostic across the same 300 sampling seeds used in the DSL stability analysis (\Cref{sec:dsl}). For each seed, we re-estimate $\hat g(Q_i,X_i)$ and compare its anti-solidarity component $\hat g_{\mathrm{AS}}(Q_i,X_i)$ with the corresponding anti-solidarity component of the raw surrogate distribution $Q_i$. We then summarize the mean calibrated anti-solidarity within each audit category across the 300 seeds.\footnote{\ak{Although this diagnostic does not establish that the calibrated prediction is correct for any individual case, it allows us to assess whether the first-stage calibration consistently moves the audited cases in the expected direction.}}}

\begin{table*}[ht]
\centering
\small
\begin{tabular}{lrrrrr}
\toprule
\ak{\textbf{Audit outcome}} &
\ak{\textbf{N}} &
\ak{\textbf{Raw AS}} &
\ak{\shortstack{\textbf{$\hat g_{\mathrm{AS}}$}\\\textbf{mean $\pm$ SD}}} &
\ak{\shortstack{\textbf{95\% seed}\\\textbf{interval}}} &
\ak{\shortstack{\textbf{Seeds with}\\\textbf{$\hat g_{\mathrm{AS}}<$ Raw AS}}} \\
\midrule
\ak{(1) Misattribution} & \ak{38} & \ak{0.579} & \ak{0.374 $\pm$ 0.068} & \ak{[0.254, 0.521]} & \ak{100.0\%} \\
\ak{(2) Other AS FPs} & \ak{16} & \ak{0.771} & \ak{0.587 $\pm$ 0.066} & \ak{[0.448, 0.718]} & \ak{100.0\%} \\
\ak{(3) Plausible disagreements} & \ak{57} & \ak{0.789} & \ak{0.641 $\pm$ 0.057} & \ak{[0.521, 0.753]} & \ak{99.7\%} \\
\bottomrule
\end{tabular}
\caption{\ak{First-stage DSL calibration for the manually audited anti-solidarity predictions across 300 seeds. \textit{Raw AS} is the mean anti-solidarity in the raw ensemble. Calibrated $\hat g_{\mathrm{AS}}$ is reported as mean $\pm$ SD across seeds with the 2.5th-97.5th percentile interval. The final column gives the percentage of seeds for which calibrated anti-solidarity is lower than raw anti-solidarity. $N$ is the number of audited model anti-solidarity predictions in each category.
}}
\label{tab:attribution-dsl-calibration}
\end{table*}

\ak{As shown in \Cref{tab:attribution-dsl-calibration}, the calibration stage consistently lowers anti-solidarity in the audited categories across repeated samples of the human-labeled data. For the 38 misattributed anti-solidarity errors, mean anti-solidarity decreases from 0.579 in the raw ensemble to 0.374 on averagae across the 300 DSL seeds (SD $=0.068$; 95\% seed interval: 0.254-0.521), with a reduction in all 300 seeds. For the other clear false positives, the corresponding mean decreases from 0.771 to 0.587 (SD $=0.066$), again in all 300 seeds. %For plausible human-model disagreements, it decreases from 0.789 to 0.641 (SD $=0.057$), with a reduction in 99.7\% of seeds. 
The largest average reduction occurs for the misattributed anti-solidarity errors. Thus, the first-stage calibration consistently moves these audited errors away from anti-solidarity. %although this diagnostic does not establish that any individual prediction is corrected.
}

\ak{\textit{Audit-based trend sensitivity.}}
\ak{We assess whether the post-2015 anti-solidarity trend persists under sensitivity adjustments based on the error rates observed in the manual audit. Because the full corpus is not human annotated, we cannot identify and remove individual false-positive instances from the social-scientific analysis. We therefore conduct an audit-based sensitivity analysis. Among all 479 model anti-solidarity predictions in the human-annotated subset from 2015 onward, 38 (7.9\%) were identified as misattribution errors and 54 (11.3\%) as clear anti-solidarity false positives of either type (see \Cref{tab:attribution-audit-results} in the Appendix). We use these two observed rates as sensitivity scenarios for the yearly DSL anti-solidarity estimates.}

\ak{For each year $t$ from 2015 onward, we proportionally reduce the original DSL anti-solidarity estimate as $\widehat{AS}^{\,\mathrm{sens}}_t=\widehat{AS}^{\,\mathrm{DSL}}_t(1-r)$, where $\widehat{AS}^{\,\mathrm{DSL}}_t$ is the original yearly DSL anti-solidarity estimate, $\widehat{AS}^{\,\mathrm{sens}}_t$ is the sensitivity-adjusted estimate, and $r$ is the audit-based error rate ($r=0.079$ for the misattribution-error scenario and $r=0.113$ for the broader all-clear-false-positive scenario). The DSL solidarity estimate is left unchanged.}\footnote{\ak{These adjustments should not be interpreted as identifying or correcting particular errors in the full corpus; rather, they show how the substantive comparison would change if the error rates observed in the manual audit generalized to the post-2015 predictions.}}

\begin{table}[ht]
\centering
\small
\begin{tabular}{lrrrr}
\toprule
\ak{\textbf{Year}} &
\ak{\shortstack{\textbf{DSL}\\\textbf{S}}} &
\ak{\shortstack{\textbf{DSL}\\\textbf{AS}}} &
\ak{\shortstack{\textbf{AS after}\\\textbf{7.9\% adjustment}}} &
\ak{\shortstack{\textbf{AS after}\\\textbf{11.3\% adjustment}}} \\
\midrule
\ak{2015} & \ak{0.674} & \ak{0.156} & \ak{0.144} & \ak{0.139} \\
\ak{2016} & \ak{0.624} & \ak{0.125} & \ak{0.115} & \ak{0.111} \\
\ak{2017} & \ak{0.480} & \ak{0.195} & \ak{0.180} & \ak{0.173} \\
\ak{2018} & \ak{0.424} & \ak{\textbf{0.441}} & \ak{0.406} & \ak{0.391} \\
\ak{2019} & \ak{0.385} & \ak{\textbf{0.566}} & \ak{\textbf{0.521}} & \ak{\textbf{0.503}} \\
\ak{2020} & \ak{0.310} & \ak{\textbf{0.607}} & \ak{\textbf{0.559}} & \ak{\textbf{0.539}} \\
\ak{2021} & \ak{0.368} & \ak{0.278} & \ak{0.256} & \ak{0.247} \\
\ak{2022} & \ak{0.430} & \ak{0.369} & \ak{0.339} & \ak{0.327} \\
\ak{2023} & \ak{0.417} & \ak{0.304} & \ak{0.280} & \ak{0.270} \\
\ak{2024} & \ak{0.375} & \ak{0.340} & \ak{0.313} & \ak{0.302} \\
\ak{2025} & \ak{0.336} & \ak{\textbf{0.394}} & \ak{\textbf{0.363}} & \ak{\textbf{0.350}} \\
\bottomrule
\end{tabular}
\caption{\ak{Yearly solidarity and anti-solidarity estimates from 2015 onward under the original DSL estimates and the two audit-based sensitivity adjustments. Bold anti-solidarity values indicate years in which the corresponding anti-solidarity estimate exceeds the solidarity estimate.}}
\label{tab:attribution-sensitivity-crossovers}
\end{table}

\begin{figure*}[!htb]
    \centering
    \includegraphics[width=0.45\linewidth]{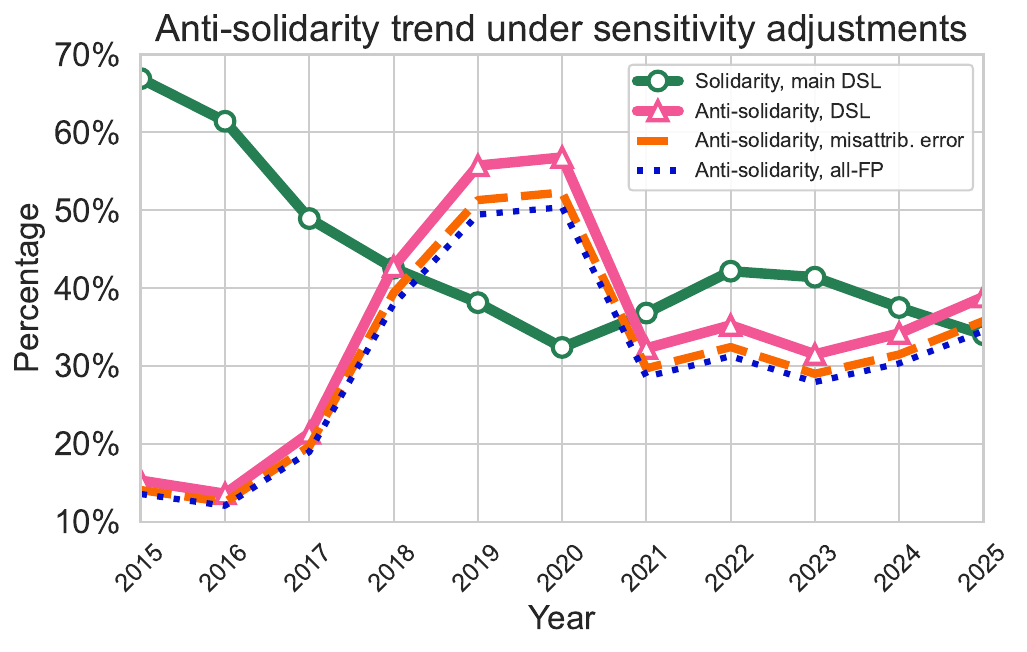}
    \caption{\ak{Sensitivity of the post-2015 anti-solidarity trend to false positives identified in the manual audit. Original DSL-adjusted estimates are compared with two scenarios that reduce anti-solidarity proportionally by the estimated rate of misattribution errors (7.9\%) and all clear anti-solidarity false positives (11.3\%).}}
    \label{fig:attribution-sensitivity-trend}
\end{figure*}

\ak{\Cref{tab:attribution-sensitivity-crossovers} and \Cref{fig:attribution-sensitivity-trend} show that both sensitivity adjustments lower the estimated level of anti-solidarity but do not eliminate the recent years in which anti-solidarity exceeds solidarity. In the original DSL estimates, anti-solidarity exceeds solidarity in 2018, 2019, 2020, and 2025. After either adjustment, the comparatively narrow crossover in 2018 disappears, whereas anti-solidarity continues to exceed solidarity in 2019, 2020, and 2025. Therefore, the recent rise in anti-solidarity remains even after reducing the estimates to account for the observed false-positive errors.
}

\clearpage 

\section{\ak{Party-conditioned model performance}} \label{app:party-conditioned-model-performance} 

\subsection{\ak{Overall classification performance per party}}
\label{app:per-party-classification}

\begin{figure*}[p]
    \centering
    \includegraphics[
        width=\textwidth,
        height=0.90\textheight,
        keepaspectratio
    ]{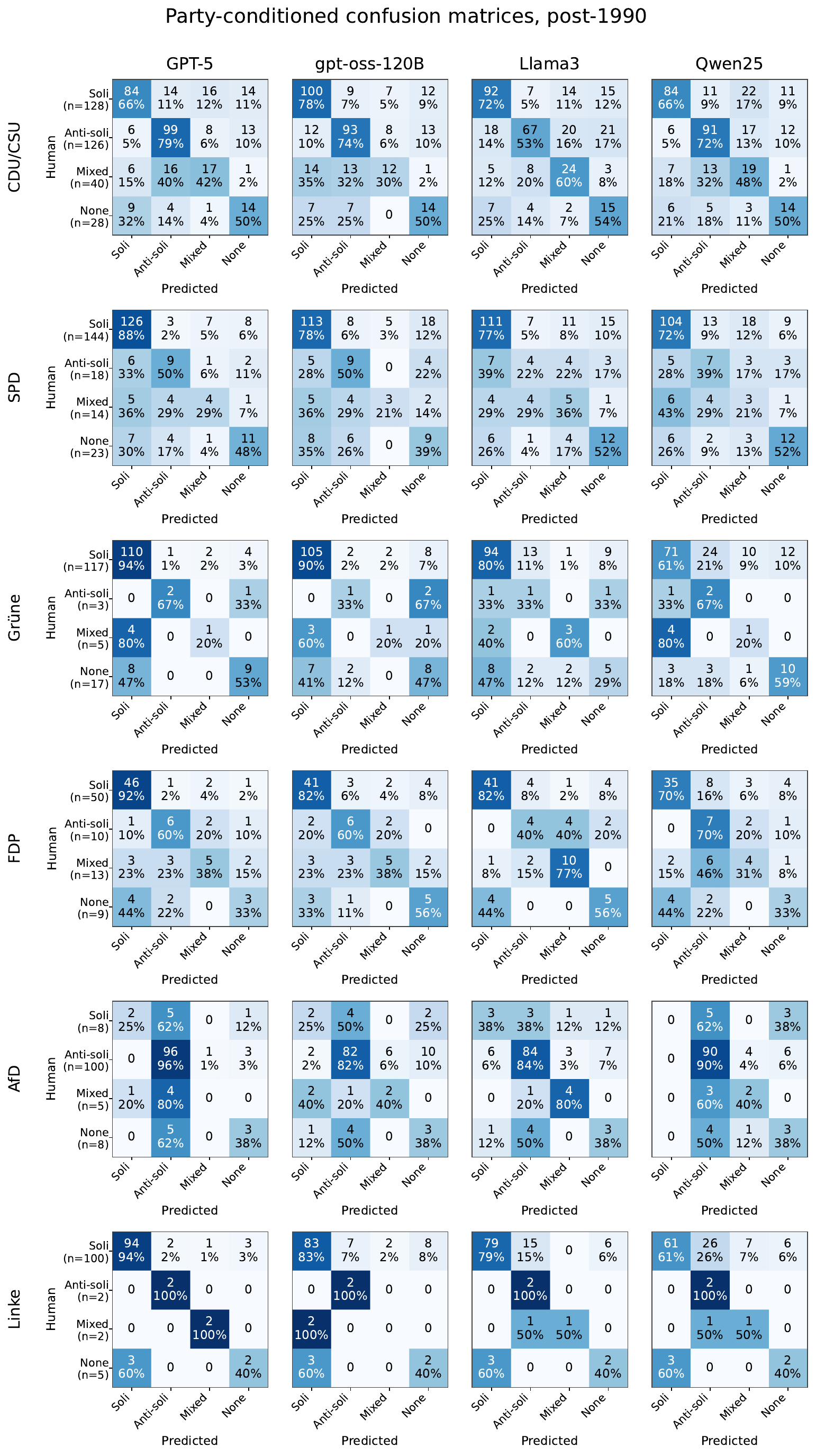
    }
    \caption{\ak{High-level confusion matrices per party on the human-annotated \textit{Migrant} subset from 1990 onward. Cell annotations report the number of instances and the row-normalized percentage for the corresponding human-reference label. The support of each human-reference label is shown as $n$ next to the row label.
    }}
    \label{fig:party-conditioned-confusion-matrices}
\end{figure*}

\ak{To examine whether model errors vary systematically across political parties, we evaluate party-conditioned performance on the human-annotated \emph{Migrant} subset from 1990 onward. We report macro and, because some party-label combinations are very sparse (for example, the human reference contains only 3 \emph{anti-solidarity} cases for Grüne, 2 for Linke, and 8 \emph{solidarity} cases for AfD, see confusion matrices in \Cref{fig:party-conditioned-confusion-matrices}), support-weighted F1 in \Cref{tab:party-conditioned-f1}. We also compare model-predicted and human \emph{solidarity} and \emph{anti-solidarity} shares in \Cref{tab:party-conditioned-deltas}.}

\begin{table*}[htb]
\centering
\small
\renewcommand{\arraystretch}{1.15}
\setlength{\tabcolsep}{4pt}
\begin{tabular}{@{}lrcccccccc@{}}
\toprule
\ak{\textbf{Party}} & \ak{\textbf{$n$}} & \multicolumn{2}{c}{\ak{\textbf{GPT-5}}} & \multicolumn{2}{c}{\ak{\textbf{gpt-oss-120B}}} & \multicolumn{2}{c}{\ak{\textbf{Llama-3.3-70B}}} & \multicolumn{2}{c}{\ak{\textbf{Qwen-2.5}}} \\
\cmidrule(lr){3-4}\cmidrule(lr){5-6}\cmidrule(lr){7-8}\cmidrule(lr){9-10}
 & & \ak{\textbf{Macro}} & \ak{\textbf{Weight.}} & \ak{\textbf{Macro}} & \ak{\textbf{Weight.}} & \ak{\textbf{Macro}} & \ak{\textbf{Weight.}} & \ak{\textbf{Macro}} & \ak{\textbf{Weight.}} \\
\midrule
\ak{CDU/CSU} & \ak{322} & \ak{0.575} & \ak{0.672} & \ak{\textbf{0.572}} & \ak{\textbf{0.678}} & \ak{0.553} & \ak{0.631} & \ak{0.567} & \ak{0.662} \\
\ak{SPD} & \ak{199} & \ak{0.533} & \ak{0.753} & \ak{\textbf{0.454}} & \ak{\textbf{0.687}} & \ak{0.440} & \ak{0.682} & \ak{0.437} & \ak{0.665} \\
\ak{Grüne} & \ak{142} & \ak{0.604} & \ak{0.851} & \ak{\textbf{0.462}} & \ak{\textbf{0.813}} & \ak{0.453} & \ak{0.757} & \ak{0.370} & \ak{0.665} \\
\ak{FDP} & \ak{82} & \ak{0.565} & \ak{0.719} & \ak{0.576} & \ak{0.696} & \ak{\textbf{0.617}} & \ak{\textbf{0.738}} & \ak{0.473} & \ak{0.615} \\
\ak{AfD} & \ak{121} & \ak{0.419} & \ak{0.806} & \ak{0.423} & \ak{0.757} & \ak{\textbf{0.535}} & \ak{\textbf{0.791}} & \ak{0.381} & \ak{0.770} \\
\ak{Linke} & \ak{109} & \ak{0.705} & \ak{0.921} & \ak{0.378} & \ak{\textbf{0.829}} & \ak{\textbf{0.511}} & \ak{0.826} & \ak{0.345} & \ak{0.703} \\
\bottomrule
\end{tabular}
\caption{\ak{Party-conditioned high-level classification performance on the human-annotated \textit{Migrant} subset from 1990 onward. Macro F1 assigns equal weight to each evaluable high-level label, whereas weighted F1 weights each label-specific F1 score by its support in the human reference within the corresponding party. %Weighted F1 therefore reflects the empirical label distribution and is less strongly affected by very rare labels. 
The best score \textbf{among the open models} for each party and metric is shown in bold.
}}
\label{tab:party-conditioned-f1}
\end{table*}

\begin{table}[htb]
\centering
\small
\renewcommand{\arraystretch}{1.15}
\setlength{\tabcolsep}{5pt}
\begin{tabular}{@{}lrrrr@{}}
\toprule
\ak{\textbf{Party}} & \ak{\textbf{GPT-5}} & \ak{\textbf{gpt-oss-120B}} & \ak{\textbf{Llama-3.3-70B}} & \ak{\textbf{Qwen-2.5}} \\
\midrule
\multicolumn{5}{@{}l}{\ak{\textbf{Solidarity ($\Delta S$, pp)}}} \\
\addlinespace[2pt]
\ak{CDU/CSU} & \ak{-7.1} & \ak{+1.6} & \ak{-1.9} & \ak{-7.8} \\
\ak{SPD} & \ak{+0.0} & \ak{-6.5} & \ak{-8.0} & \ak{-11.6} \\
\ak{Grüne} & \ak{+3.5} & \ak{-1.4} & \ak{-8.5} & \ak{-26.8} \\
\ak{FDP} & \ak{+4.9} & \ak{-1.2} & \ak{-4.9} & \ak{-11.0} \\
\ak{AfD} & \ak{-4.1} & \ak{-0.8} & \ak{+1.7} & \ak{-6.6} \\
\ak{Linke} & \ak{-2.8} & \ak{-11.0} & \ak{-16.5} & \ak{-33.0} \\
\midrule
\multicolumn{5}{@{}l}{\ak{\textbf{Anti-solidarity ($\Delta AS$, pp)}}} \\
\addlinespace[2pt]
\ak{CDU/CSU} & \ak{+2.2} & \ak{-1.2} & \ak{-12.4} & \ak{-1.9} \\
\ak{SPD} & \ak{+1.0} & \ak{+4.5} & \ak{-1.0} & \ak{+4.0} \\
\ak{Grüne} & \ak{+0.0} & \ak{+1.4} & \ak{+9.2} & \ak{+18.3} \\
\ak{FDP} & \ak{+2.4} & \ak{+3.7} & \ak{+0.0} & \ak{+15.9} \\
\ak{AfD} & \ak{+8.3} & \ak{-7.4} & \ak{-6.6} & \ak{+1.7} \\
\ak{Linke} & \ak{+1.8} & \ak{+6.4} & \ak{+14.7} & \ak{+24.8} \\
\bottomrule
\end{tabular}
\caption{\ak{Signed differences between model-predicted and human-reference label shares ($p_{\mathrm{model}}$ and $p_{\mathrm{human}}$) by party. We define $\Delta S = p_{\mathrm{model}}(S) - p_{\mathrm{human}}(S)$ and $\Delta AS = p_{\mathrm{model}}(AS) - p_{\mathrm{human}}(AS)$. Positive values indicate overprediction and negative values indicate underprediction. Differences are reported in percentage points (pp), i.e., the arithmetic difference between two percentages. %For example, a human solidarity share of 15.0\% and a model solidarity share of 9.5\% correspond to $\Delta S=-5.5$ pp.
}}
\label{tab:party-conditioned-deltas}
\end{table}

\ak{The largest discrepancies occur for Qwen-2.5 on Grüne and Linke speakers: \emph{anti-solidarity} is overpredicted by 18.3 and 24.8 percentage points, respectively, while \emph{solidarity} is underpredicted by 26.8 and 33.0 points. Llama-3.3-70B shows the same direction of error for both parties, though less strongly. Additionally, Qwen-2.5 shows a notable \emph{anti-solidarity} overprediction for FDP (+15.9 percentage points).}

\ak{By contrast, the corresponding errors for CDU/CSU and AfD are smaller or vary in direction across models (e.g. Llama-3.3-70B and gpt-oss-120B underpredict \emph{anti-solidarity} for AfD by around 7 points, but GPT-5 overpredicts it by 8.3). The confusion matrices in \Cref{fig:party-conditioned-confusion-matrices} show the specific error patterns behind these differences. %individual rates for very small party-label cells should be interpreted cautiously.
} 

\subsection{\ak{Anti-solidarity errors due to misattribution by party}}
\label{app:party-conditioned-misattribution}

\ak{To determine whether the \emph{anti-solidarity} overprediction by Llama-3.3-70B and Qwen-2.5 for Grüne and Linke reflects the misattribution error quantified in \Cref{app:appendix-attribution-audit}, we manually inspect all 280 candidate predictions from 1990 onward in which the human reference is not \emph{anti-solidarity} but gpt-oss-120B, Llama-3.3-70B, or Qwen-2.5 predicts \emph{anti-solidarity} across the six parties.}

\begin{table}[htb]
\centering
\small
\renewcommand{\arraystretch}{1.15}
\setlength{\tabcolsep}{5pt}
\begin{tabular}{@{}lccc@{}}
\toprule
\ak{\textbf{Party}} &
\ak{\textbf{gpt-oss-120B}} &
\ak{\textbf{Llama-3.3-70B}} &
\ak{\textbf{Qwen-2.5}} \\
\midrule
\ak{CDU/CSU} & \ak{0/29 (0.0\%)} & \ak{0/19 (0.0\%)} & \ak{2/29 (6.9\%)} \\
\ak{SPD} & \ak{2/18 (11.1\%)} & \ak{5/12 (41.7\%)} & \ak{8/19 (42.1\%)} \\
\ak{Grüne} & \ak{2/4 (50.0\%)} & \ak{14/15 (93.3\%)} & \ak{23/27 (85.2\%)} \\
\ak{FDP} & \ak{2/7 (28.6\%)} & \ak{4/6 (66.7\%)} & \ak{7/16 (43.8\%)} \\
\ak{AfD} & \ak{1/9 (11.1\%)} & \ak{1/8 (12.5\%)} & \ak{1/12 (8.3\%)} \\
\ak{Linke} & \ak{7/7 (100.0\%)} & \ak{15/16 (93.8\%)} & \ak{24/27 (88.9\%)} \\
\bottomrule
\end{tabular}
\caption{\ak{Misattribution errors among manually audited post-1990 cases in which the human reference is not \emph{anti-solidarity} but the model predicts \emph{anti-solidarity}. Entries report the number of misattribution errors divided by the number of audited candidate predictions, with percentages in parentheses.}}
\label{tab:party-misattribution-audit}
\end{table}

\ak{The audit shows that the excess \emph{anti-solidarity} predictions for Grüne and Linke are closely associated with this misattribution failure. For Llama-3.3-70B and Qwen-2.5, misattribution accounts for 93.3\% and 85.2\% of the corresponding audited candidate predictions for Grüne, and 93.8\% and 88.9\% for Linke, respectively. The corresponding shares are much lower for CDU/CSU and AfD. For gpt-oss-120B, all seven Linke candidate predictions are misattribution errors, although the number of corresponding Grüne cases is very small ($n=4$). Qwen-2.5's \emph{anti-solidarity} overprediction for FDP is also partly driven by misattribution: seven of the eight cases in which the human reference is \emph{solidarity} but Qwen-2.5 predicts \emph{anti-solidarity} are misattribution errors. The other main source is confusion of \emph{mixed} with \emph{anti-solidarity} (\Cref{fig:party-conditioned-confusion-matrices}).}

\subsection{\ak{Robustness of the party-level estimates in the recent period}}
\label{app:party-conditioned-dsl-robustness}

\ak{To assess whether the party-conditioned errors identified in \Cref{app:per-party-classification} affect the substantive party-level estimates discussed in \Cref{sec:recent-distribution-across-parties}, we conduct a robustness check on the full 2009-2025 corpus.\footnote{We use the full corpus rather than the mock-DSL setup in \Cref{sec:dsl}, because further conditioning its 20\% human subsamples by party would leave many party-decade-label combinations with very few or no labeled observations.} This is particularly important because Qwen-2.5 and Llama-3.3-70B, which show the clearest \emph{anti-solidarity} overprediction for Grüne and Linke (and Qwen-2.5 also for FDP), are both part of the three-model committee used by soft-label DSL. We compare raw-committee and main soft-label DSL estimates of party-level \emph{solidarity} and \emph{anti-solidarity} shares, and then test whether the DSL estimates depend on the models showing these skews using two alternative configurations: (i) soft-label DSL without Qwen-2.5, which shows the strongest \emph{anti-solidarity} overprediction for Grüne, Linke, and FDP; and (ii) hard-label DSL based only on gpt-oss-120B, thereby removing both Qwen-2.5 and Llama-3.3-70B. Results are reported in \Cref{tab:party-dsl-robustness}.}

\begin{table*}[htb]
\centering
\small
\renewcommand{\arraystretch}{1.15}
\setlength{\tabcolsep}{4pt}
\begin{tabular}{@{}lrcccccccc@{}}
\toprule
\ak{\textbf{Party}} &
\ak{\textbf{$n$}} &
\multicolumn{2}{c}{\ak{\textbf{Raw committee}}} &
\multicolumn{2}{c}{\ak{\textbf{Soft DSL}}} &
\multicolumn{2}{c}{\ak{\textbf{Soft DSL, no Qwen}}} &
\multicolumn{2}{c}{\ak{\textbf{Hard DSL, gpt-oss}}} \\
\cmidrule(lr){3-4}
\cmidrule(lr){5-6}
\cmidrule(lr){7-8}
\cmidrule(lr){9-10}
& &
\ak{\textbf{S}} &
\ak{\textbf{AS}} &
\ak{\textbf{S}} &
\ak{\textbf{AS}} &
\ak{\textbf{S}} &
\ak{\textbf{AS}} &
\ak{\textbf{S}} &
\ak{\textbf{AS}} \\
\midrule
\ak{CDU/CSU} & \ak{6365} & \ak{45.0\%} & \ak{20.8\%} & \ak{48.3\%} & \ak{27.1\%} & \ak{47.7\%} & \ak{27.5\%} & \ak{45.6\%} & \ak{29.1\%} \\
\ak{SPD} & \ak{3956} & \ak{69.1\%} & \ak{6.5\%} & \ak{75.8\%} & \ak{2.4\%} & \ak{75.5\%} & \ak{2.4\%} & \ak{76.0\%} & \ak{3.7\%} \\
\ak{Grüne} & \ak{2335} & \ak{68.7\%} & \ak{9.3\%} & \ak{71.7\%} & \ak{1.4\%} & \ak{71.0\%} & \ak{1.3\%} & \ak{72.5\%} & \ak{1.1\%} \\
\ak{FDP} & \ak{1473} & \ak{53.5\%} & \ak{13.9\%} & \ak{58.9\%} & \ak{6.8\%} & \ak{58.8\%} & \ak{7.6\%} & \ak{52.6\%} & \ak{6.7\%} \\
\ak{AfD} & \ak{2763} & \ak{5.4\%} & \ak{77.6\%} & \ak{4.5\%} & \ak{83.2\%} & \ak{4.5\%} & \ak{83.2\%} & \ak{4.7\%} & \ak{82.5\%} \\
\ak{Linke} & \ak{2812} & \ak{70.3\%} & \ak{12.6\%} & \ak{88.7\%} & \ak{2.1\%} & \ak{88.7\%} & \ak{1.9\%} & \ak{91.6\%} & \ak{2.7\%} \\
\bottomrule
\end{tabular}
\caption{\ak{2009-2025 party-level solidarity and anti-solidarity estimates under the raw model committee and alternative DSL versions. The \textit{raw committee} averages the label votes of gpt-oss-120B, Llama-3.3-70B, and Qwen-2.5 into a soft label distribution without DSL correction. \textit{Soft DSL} is the main soft-label DSL using the three-model committee and soft human labels. \textit{Soft DSL, no Qwen} removes Qwen-2.5 from the model committee. \textit{Hard DSL, gpt-oss} uses one-hot gpt-oss-120B predictions as the surrogate and human-majority one-hot labels for DSL correction. Cells report party-level shares of solidarity (S) and anti-solidarity (AS).}}
\label{tab:party-dsl-robustness}
\end{table*}

\ak{The main DSL correction substantially reduces the raw committee's \emph{anti-solidarity} estimates for the parties showing the clearest model skews: from 9.3\% to 1.4\% for Grüne, from 12.6\% to 2.1\% for Linke, and from 13.9\% to 6.8\% for FDP. Removing Qwen-2.5 or using only gpt-oss-120B changes these DSL-adjusted \emph{anti-solidarity} estimates by less than one percentage point for all three parties. Although human reference labels are unavailable for the full corpus and we therefore cannot determine which configuration is closest to the true party-level distribution, the party-conditioned skews observed for Qwen-2.5 and Llama-3.3-70B do not appear to drive the final DSL estimates.}

\section{\ak{Decoding stability}}
\label{app:decoding-stability}

\ak{We assess within-model decoding stability using gpt-oss-120B in the same few-shot configuration as in the main evaluation. We repeat inference across five decoding seeds for %all three evaluation settings: Woman, Migrant (T1), and Migrant (T2).
both Migrant evaluation settings: Migrant (T1) and Migrant (T2). %For Woman and Migrant (T1), which each consist of three test splits,
Migrant (T1) consists of three test splits, for which we compute macro-F1 separately for each split and average the scores within each seed. Migrant (T2) consists of a single test set, for which macro-F1 is computed directly. We perform this evaluation separately for the fine-grained and high-level label sets. See \Cref{tab:gptoss-decoding-stability} for results.}

\begin{table}[ht]
\centering
\small
\renewcommand{\arraystretch}{1.15}
\setlength{\tabcolsep}{5pt}
\begin{tabular}{lcc}
\toprule
\ak{\textbf{Dataset}}
& \ak{\textbf{Fine-grained macro-F1}}
& \ak{\textbf{High-level macro-F1}} \\
% & \ak{\textbf{Mean pairwise $\kappa$ (fine / high)}} \\
\midrule
%\ak{Woman} & \ak{0.367 $\pm$ 0.011 [0.351--0.377]} & \ak{0.572 $\pm$ 0.006 [0.567--0.581]} \\
% & \ak{0.642 / 0.720} \\
\ak{Migrant (T1)} & \ak{0.477 $\pm$ 0.023 [0.445--0.506]} & \ak{0.692 $\pm$ 0.014 [0.680--0.716]} \\
% & \ak{0.726 / 0.826} \\
\ak{Migrant (T2)} & \ak{0.278 $\pm$ 0.025 [0.254--0.319]} & \ak{0.519 $\pm$ 0.013 [0.504--0.532]} \\
% & \ak{0.681 / 0.752} \\
\bottomrule
\end{tabular}
\caption{\ak{Decoding stability of gpt-oss-120B in the few-shot setting across five decoding seeds. Macro-F1 is reported as mean $\pm$ standard deviation [minimum--maximum] across seeds.}}
% The final column reports mean pairwise Cohen's $\kappa$ between seed-level predictions for the fine-grained and high-level tasks, respectively.
\label{tab:gptoss-decoding-stability}
\end{table}

%\clearpage

\begin{table*}[h]
\centering
\footnotesize
\renewcommand{\arraystretch}{1.15}
\setlength{\tabcolsep}{5pt}

\begin{tabularx}{\textwidth}{
    >{\raggedright\arraybackslash}p{0.13\textwidth}
    >{\raggedright\arraybackslash}X
    >{\raggedright\arraybackslash}X}
    \toprule
    \textbf{\ak{Frame}} & \textbf{\ak{Solidarity example}} & \textbf{\ak{Anti-solidarity example}} \\
    \midrule

    \textbf{\ak{Group-based}} % 19940428-bt-12-225-01994 and 19131206-rtdkr-13.1-184-00227
    &
    \enquote{\ak{Kein Staat kann hinnehmen, daß ein bedeutsamer Teil der Bevölkerung [...] außerhalb der staatlichen Gemeinschaft [...] steht. \textbf{Keinem Zuwanderer, der [...] Lebensweisen und Wertvorstellungen der Einheimischen teilt, ist zuzumuten, nach den jetzigen Gesetzen als Ausländer ausgegrenzt zu werden.} [...]}}
    &
    \enquote{\ak{[...] \textbf{Niemals aber darf der Wunsch, die üblichen Löhne zu drücken, einer öffentlichen Verwaltung Anlaß geben, Ausländer einzustellen.} [...] Wollen die Grubenherren in Wahrheit national handeln, so haben sie die Pflicht, die ausländischen Arbeiter zu entlassen, wenn geeignete deutsche Arbeiter unbeschäftigt bleiben. Die Verwaltung hat ja viele Waffen, zuletzt die Ausweisungsbefugnis, um der nationalen Idee Geltung zu verschaffen.}}
    \\

    \midrule
    \textbf{\ak{Exchange-based}} % 20150129-bt-18-082-00641 and 20180419-bt-19-026-02457
    &
    \enquote{\ak{[...] Aber wir müssen auch deswegen weltoffen sein, weil es sich unser Land ökonomisch nicht leisten kann, national vernagelt zu sein. \textbf{Unsere Wirtschaft ist exportorientiert, und wir brauchen eine qualifizierte Zuwanderung nach Deutschland.} Dies wollen und werden wir organisieren. [...]}}
    &
    \enquote{\ak{\textbf{[...] Migration kann scheitern, vor allem dann, wenn die Qualifikation der Einwanderer niedrig ist.} 2013 [...] hatten 40 Prozent der Zuwanderer aus dem Nicht-EU-Ausland keinen Abschluss. Seit der Flüchtlingswelle haben die Messerstechereien um 20 Prozent zugenommen, [...] Ist das eine hervorragend erfolgreiche Migration?}}
    \\

    \midrule
    \textbf{\ak{Compassion.}} % 19890914-bt-11-158-00776 and 19921104-bt-12-116-00755
    &
    \enquote{\ak{[...] Alle Maßnahmen, die dazu dienen, Menschen in einer Extremsituation zu helfen, können wir nur unterstützen. \textbf{Die von uns in der Bundesrepublik als Aussiedler und Übersiedler bezeichneten Einwanderer befinden sich tatsächlich in einer solchen Situation.} [...]}}
    &
    \enquote{\ak{[...] Dennoch ist kein Ende des Asylmißbrauchs abzusehen. \textbf{Es gibt einen dramatischen Anstieg der Zuwanderung politisch Nichtverfolgter.} [...] Der Asylmißbrauch hat eine Eigendynamik erreicht, die längst zum Teufelskreis geworden ist.}}
    \\

    \midrule
    \textbf{\ak{Empathic}} % 20020124-bt-14-212-00534 and 20220112-bt-20-010-01011
    &
    \enquote{\ak{[...] \textbf{wir befürchten, dass das Einwanderungsgesetz [...] den Zielen unserer auswärtigen Kulturpolitik direkt widerspricht; denn es fördert nicht die Begegnung gleichberechtigter Kulturen, sondern fordert nahezu Assimilationsbereitschaft als Preis für Einwanderung.} Wir sind für Integration, nicht für Assimilation.}}
    &
    \enquote{\ak{[...] \textbf{ein SPD-Politiker warnte: \enquote{Wir haben uns mit dem Zustand von Menschen aus völlig anderen kulturellen Welten übernommen, 7 Millionen Ausländer in Deutschland sind eine fehlerhafte Entwicklung.} Wir müssen eine Zuwanderung aus fremden Kulturen unterbinden, %Zuwanderung von Menschen aus Anatolien und Schwarzafrika 
    [es] schafft nur ein zusätzliches Problem.} -- Vernünftige Einstellungen. [...]}}
    \\

    \bottomrule
\end{tabularx}

\caption{\ak{Illustrative examples of the four (anti-)solidarity frames from the human annotated subset. Bold text is the main sentence, the other sentences are for context.}}
\label{tab:solidarity-frames-examples-german}

\end{table*}

\begin{table}[ht]
\centering
\small
\begin{tabular}{lr}
\toprule
\textbf{Label} & \textbf{Migrant} \\
\midrule
Group-based Solidarity          & 270 (12.4\%) \\
Exchange-based Solidarity       & 79 (3.6\%) \\
Empathic Solidarity             & 32 (1.5\%) \\
Compassionate Solidarity        & 692 (31.7\%) \\
Solidarity (no subtype)         & 55 (2.5\%) \\
\textbf{Total Solidarity}       & \textbf{1128 (51.6\%)} \\
\midrule
Group-based Anti-solidarity     & 288 (13.2\%) \\
Exchange-based Anti-solidarity  & 55 (2.5\%) \\
Empathic Anti-solidarity        & 6 (0.3\%) \\
Compassionate Anti-solidarity   & 97 (4.4\%) \\
Anti-solidarity (no subtype)    & 19 (0.9\%) \\
\textbf{Total Anti-solidarity}  & \textbf{465 (21.3\%)} \\
\midrule
Mixed                           & 149 (6.8\%) \\
None                            & 442 (20.2\%) \\
\midrule
\textbf{Total}                  & \textbf{2184 (100.0\%)} \\
\bottomrule
\end{tabular}
\caption{Distribution of labels in the human-annotated \Migrant{} subset. Percentages are relative to the total number of migrant-related instances.}
\label{tab:instances-distribution}
\end{table}

\begin{figure*}[!htb]
    \centering
    \includegraphics[width=\linewidth]{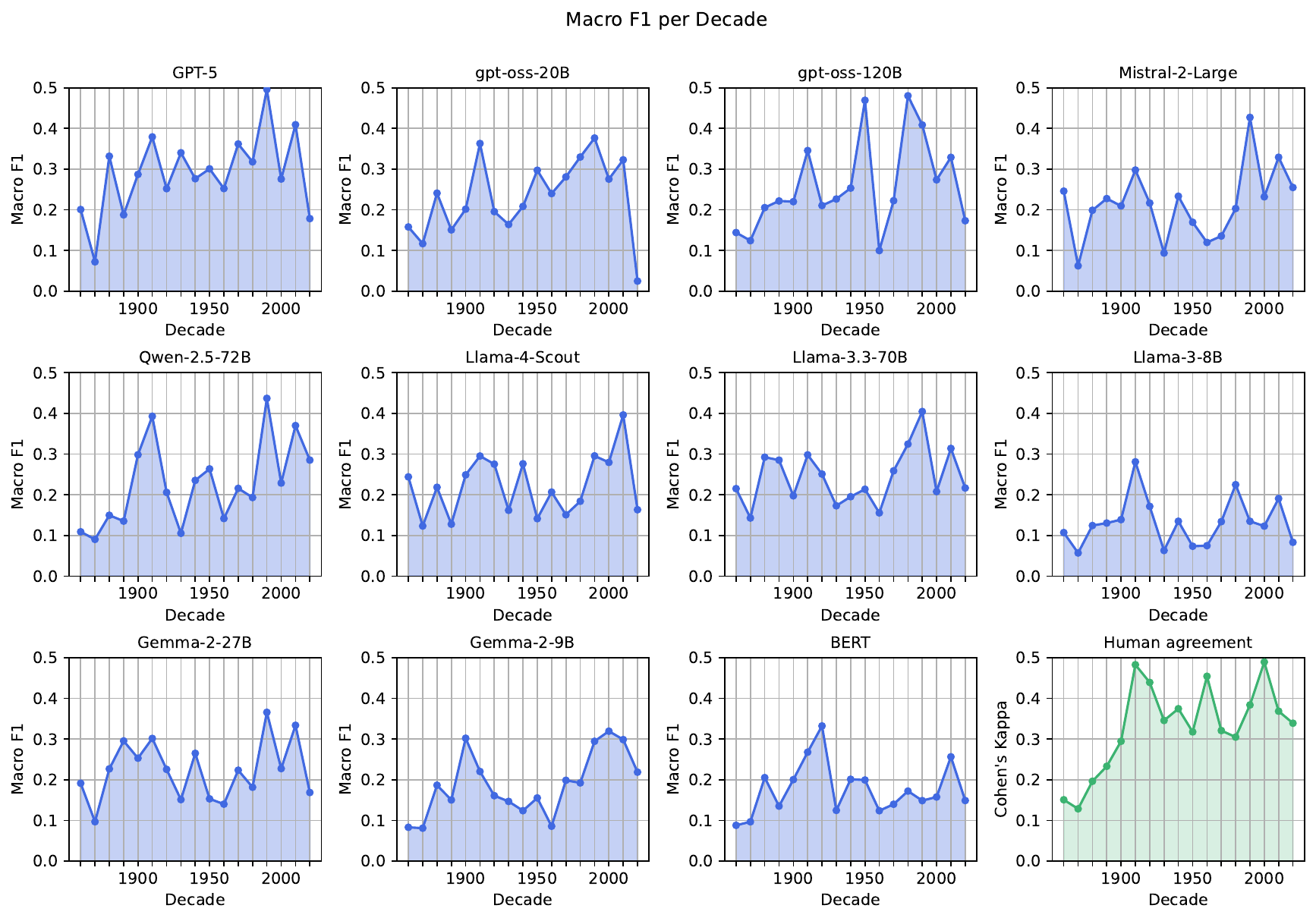}
    \caption{Macro F1 scores over time for each model (best configuration), and average pairwise Cohen's Kappa for human annotators, evaluated on the combined \textit{Migrant} test sets (Test 1 and Test 2) and grouped by decade. %Evaluated on the full test set, including \textit{Frau} and both test sets for \textit{Migrant}, grouped by decade.
    }
    \label{fig:macro-f1-kappa-over-time}
\end{figure*}

\begin{figure*}[!htb]
    \centering
    \includegraphics[width=0.65\linewidth]{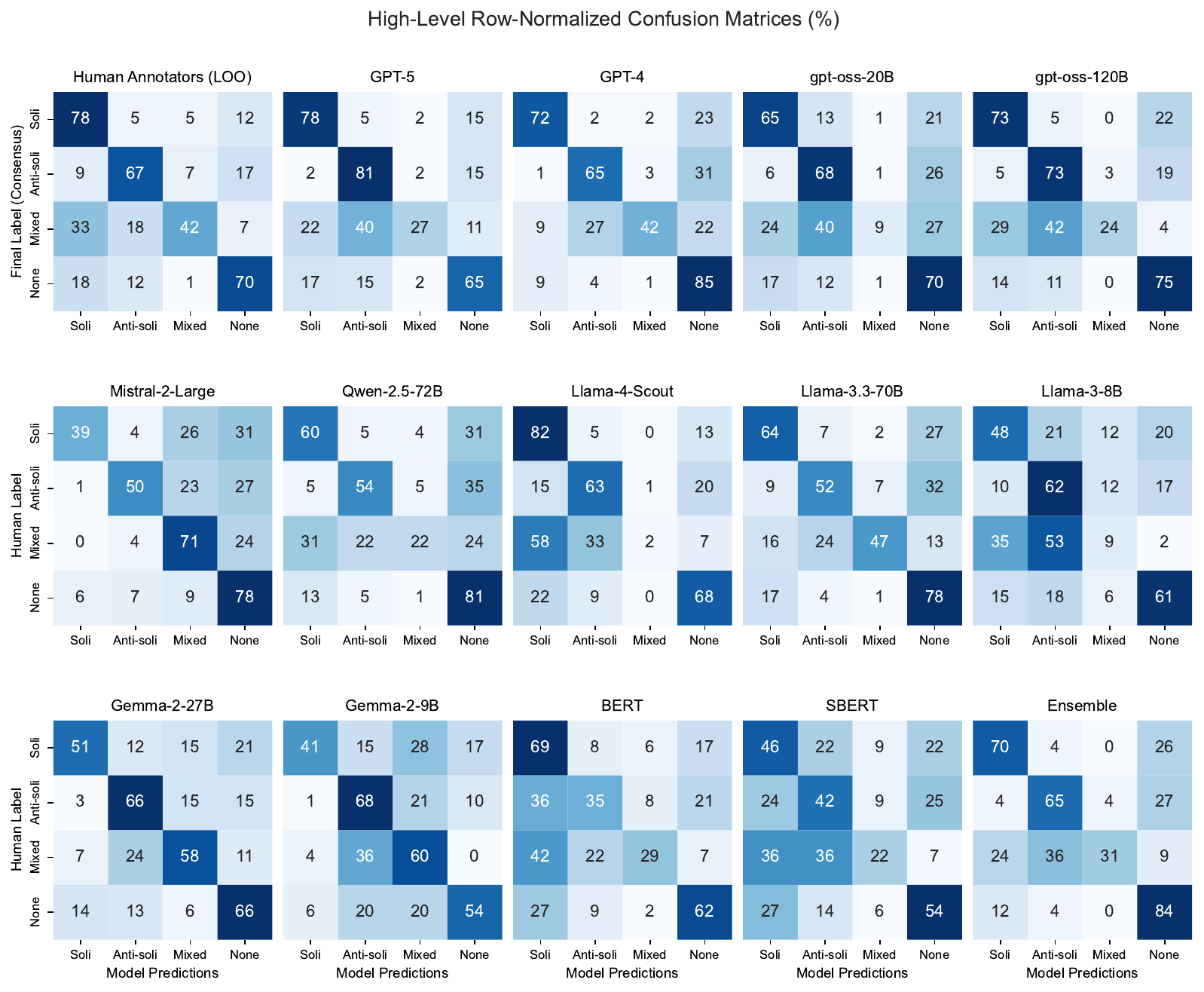}
    \caption{High-level row-normalized confusion matrices (\%) for the best-performing configuration of each model on the two migrant test sets (Test 1 and Test 2, combined). Each row sums to 100, showing how items with a given reference label are distributed across predicted labels. \emph{Human Annotators (LOO)} shows aggregated leave-one-out annotator comparisons, where each annotator is compared to the consensus of the remaining annotators. The \emph{Ensemble} model aggregates predictions from Llama-3.3-70B, Qwen-2.5-72B, and gpt-oss-120B.}
    \label{fig:confusion-matrices-highlevel-full}
\end{figure*}

\begin{figure*}[!htb]
    \centering
    \includegraphics[width=0.90\linewidth]{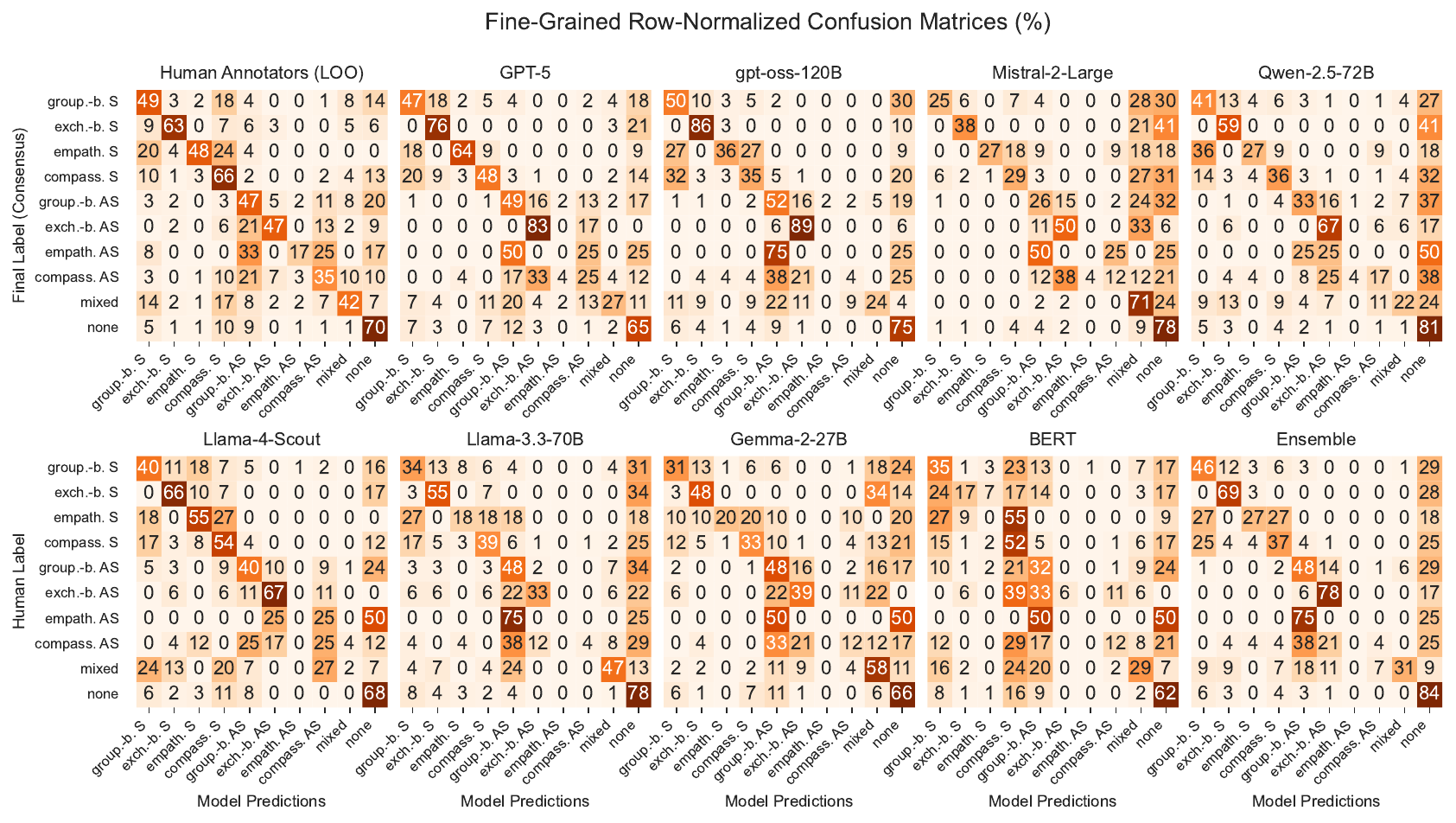}
    \caption{Fine-grained row-normalized confusion matrices (\%) showing the best-performing configuration for each of the selected models on the two migrant test sets (Test 1 and Test 2), combined. Each row sums to 100, showing how items with a given reference label are distributed across predicted labels. \emph{Human Annotators (LOO)} shows aggregated leave-one-out annotator comparisons, where each annotator is compared to the consensus of the remaining annotators. The \emph{Ensemble} model aggregates predictions from Llama-3.3-70B, Qwen-2.5-72B, and gpt-oss-120B.}
    \label{fig:confusion-matrices-finegrained-full}
\end{figure*}

\clearpage
\begin{figure*}%[!htb]
    \centering
    \includegraphics[width=\linewidth]{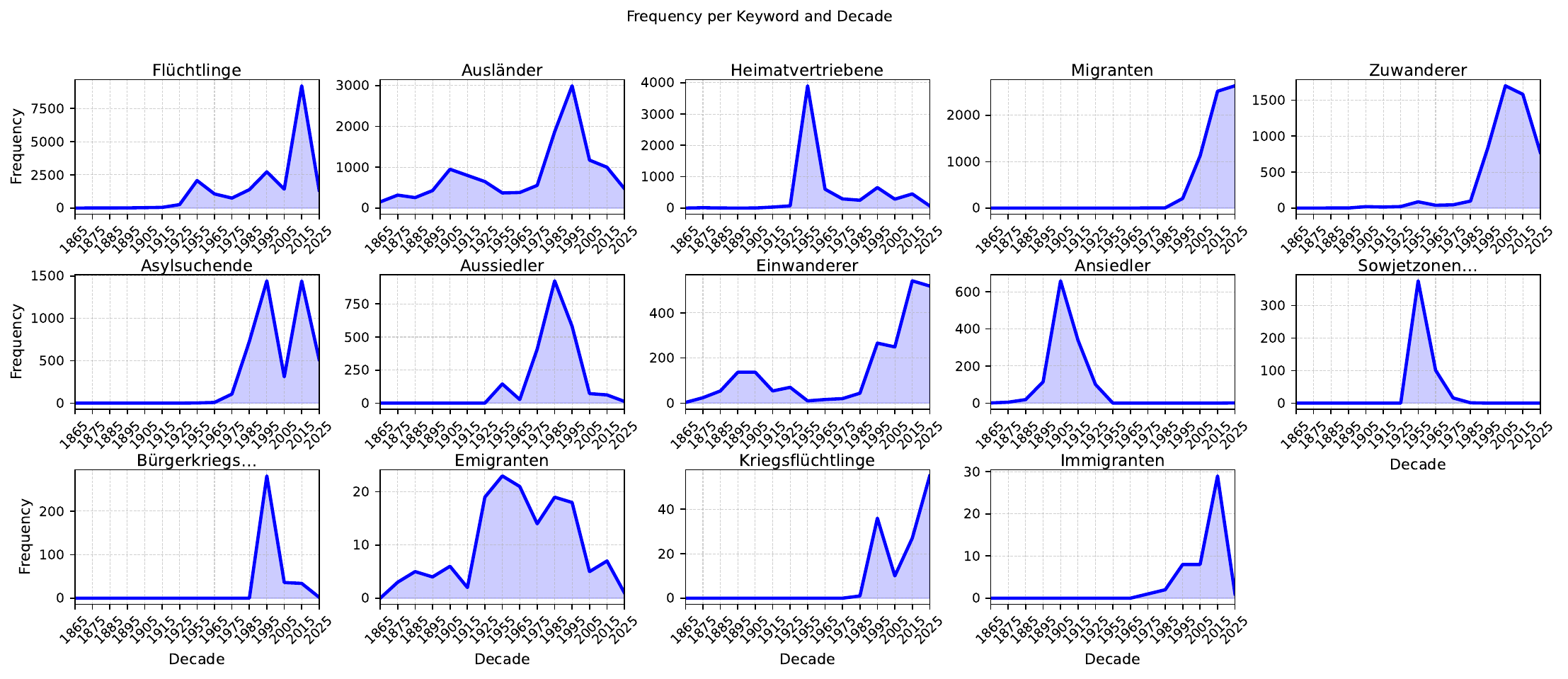}
    \caption{Distribution of all \Migrant{} keywords over the years, normalized per keyword. The keywords are sorted by frequency, which means that the reliability decreases towards the bottom-right.}
    \label{fig:keywords-distrib-over-time}
\end{figure*}

\begin{figure*}%[!htb]
    \centering
    \includegraphics[width=\linewidth]{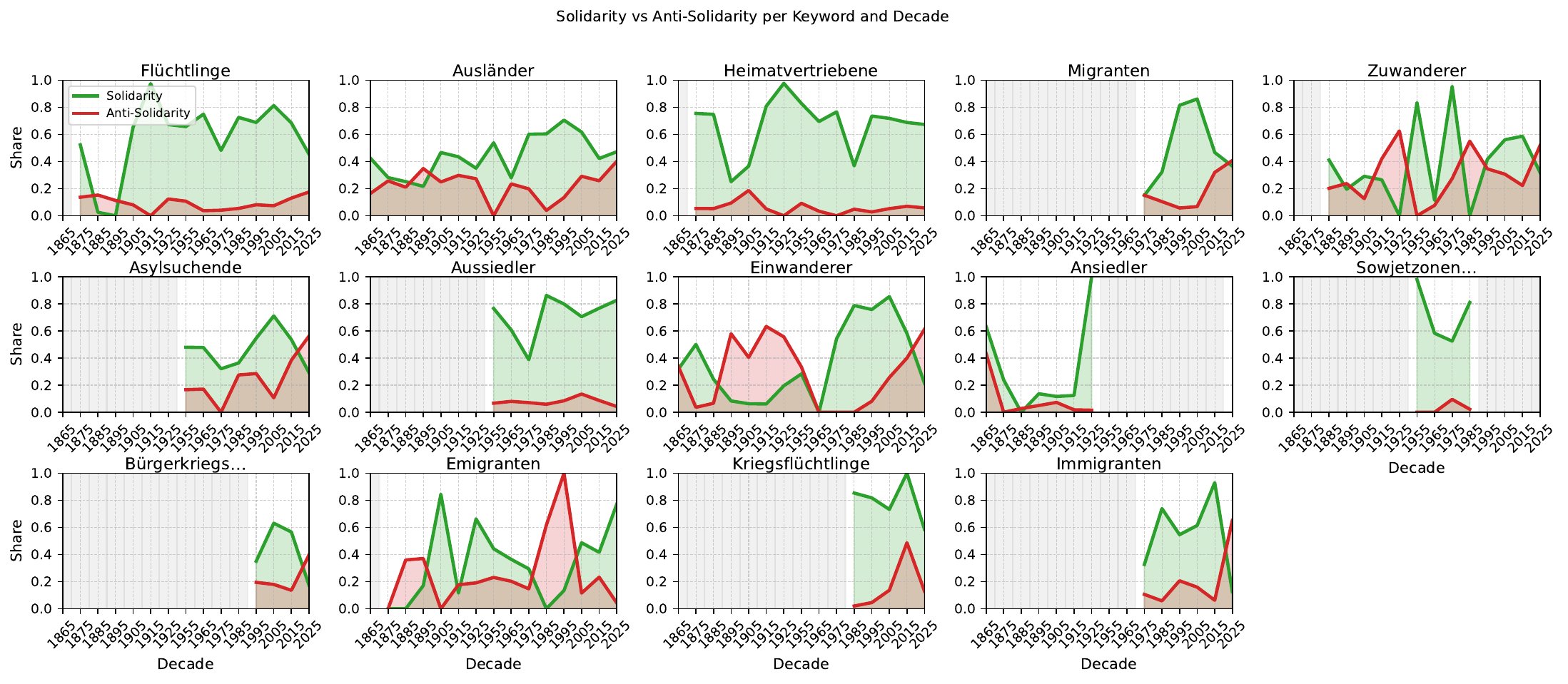}
    \caption{(Anti-)solidarity trends for the \Migrant{} category by keyword and decade, 1867-2025. Each panel shows DSL-adjusted decade-level shares of solidarity and anti-solidarity for one keyword (see definition (c) in \Cref{box:trend-definitions}). The \textit{mixed} and \textit{none} categories are included in the four-class distribution but omitted from the visualization. Keywords are sorted by frequency, so estimates become less reliable toward the bottom right.}
    \label{fig:1867-2025-keywords}
\end{figure*}

%%% Bias Correction with DSL %%%

\begin{figure*}[!htb]
    \centering

    % --- Row 1 ---
    \begin{subfigure}[t]{0.46\linewidth}
        \centering
        \includegraphics[width=0.90\linewidth]{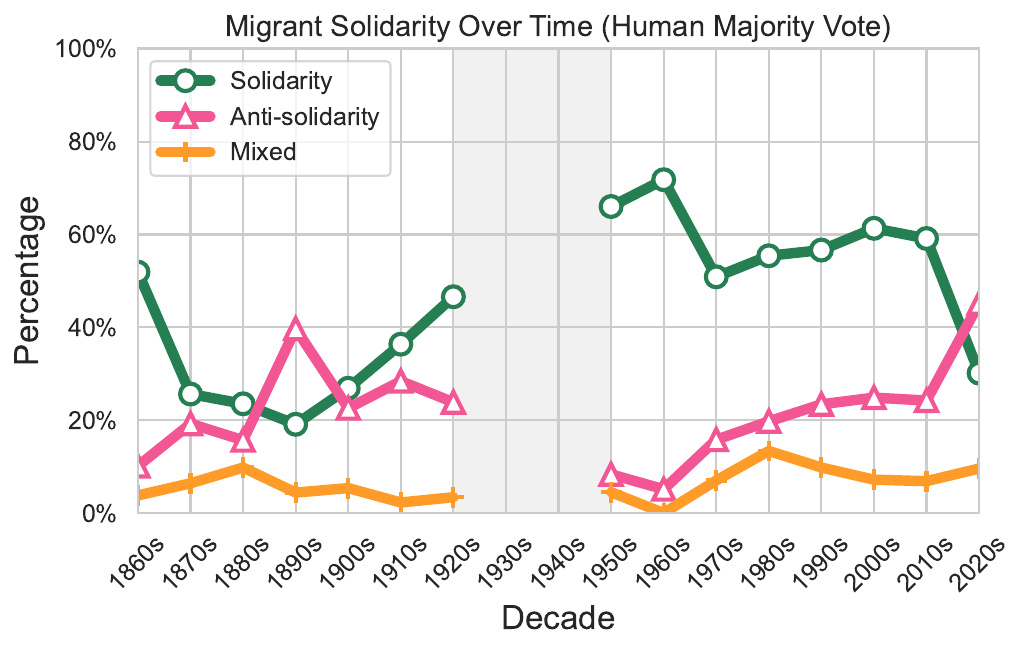}
        \caption{Human majority vote}
        \label{fig:hl-human-maj}
    \end{subfigure}
    \hfill
    \begin{subfigure}[t]{0.46\linewidth}
        \centering
        \includegraphics[width=0.90\linewidth]{plots/analysis/TrendCorrelation/Highlevel/HighLevel_SoftLabelTrends_Humans.pdf}
        \caption{Human soft labels}
        \label{fig:hl-human-soft}
    \end{subfigure}

    \vspace{0.35em}

    % --- Row 2 ---
    \begin{subfigure}[t]{0.46\linewidth}
        \centering
        \includegraphics[width=0.90\linewidth]{plots/analysis/TrendCorrelation/Highlevel/Gptoss120b_Highlevel.pdf}
        \caption{gpt-oss-120B}
        \label{fig:hl-gptoss120b}
    \end{subfigure}
    \hfill
    \begin{subfigure}[t]{0.46\linewidth}
        \centering
        \includegraphics[width=0.90\linewidth]{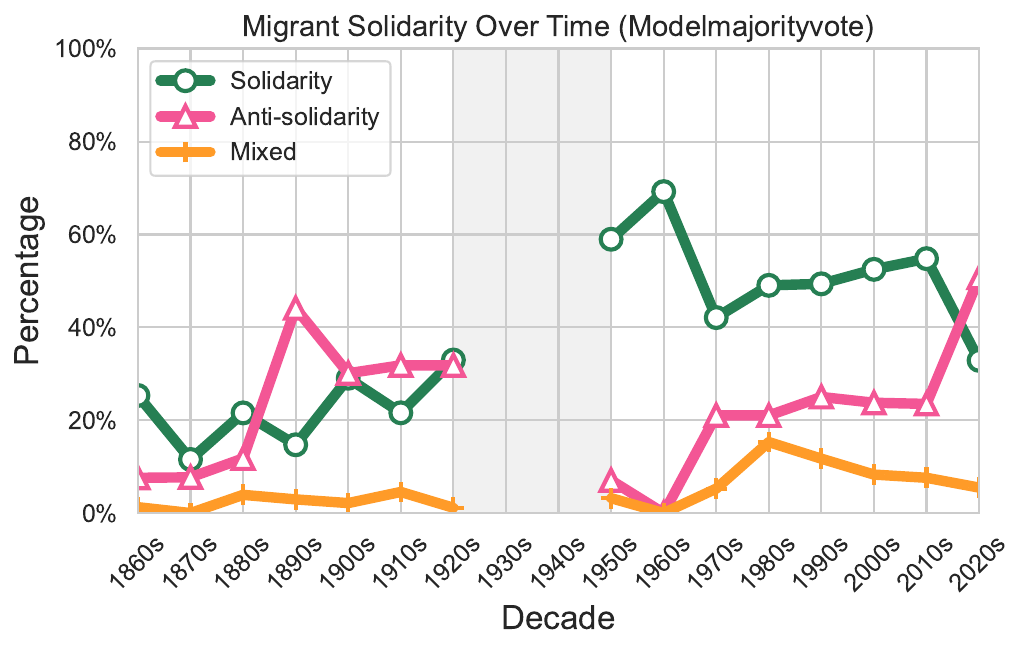}
        \caption{Model majority vote}
        \label{fig:hl-model-maj}
    \end{subfigure}

    \vspace{0.35em}

    % --- Row 3 ---
    \begin{subfigure}[t]{0.46\linewidth}
        \centering
        \includegraphics[width=0.90\linewidth]{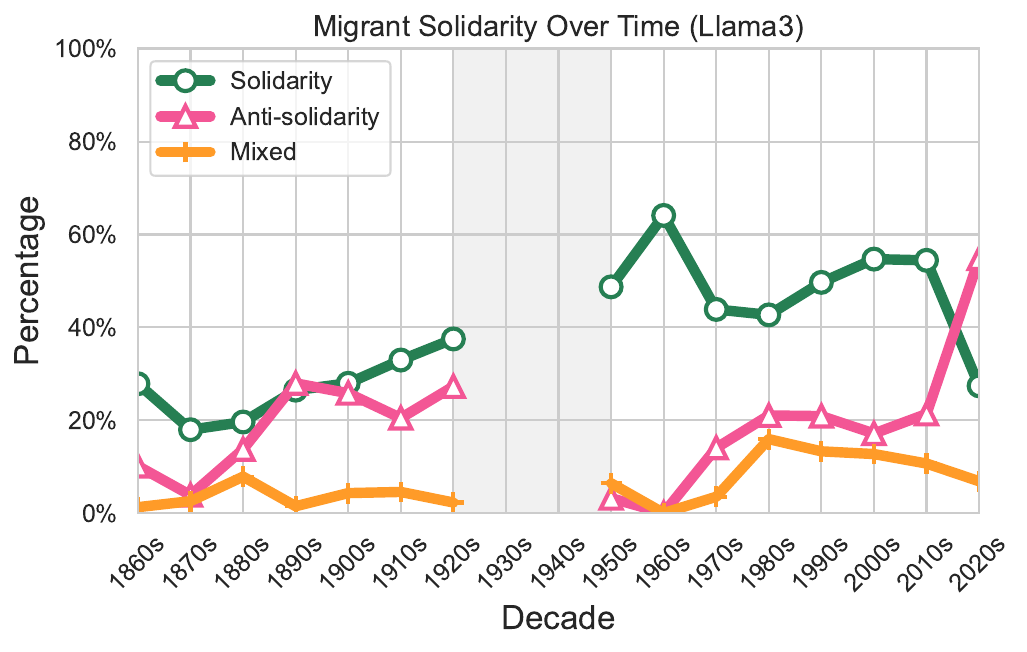}
        \caption{Llama-3.3-70B}
        \label{fig:hl-llama33}
    \end{subfigure}
    \hfill
    \begin{subfigure}[t]{0.46\linewidth}
        \centering
        \includegraphics[width=0.90\linewidth]{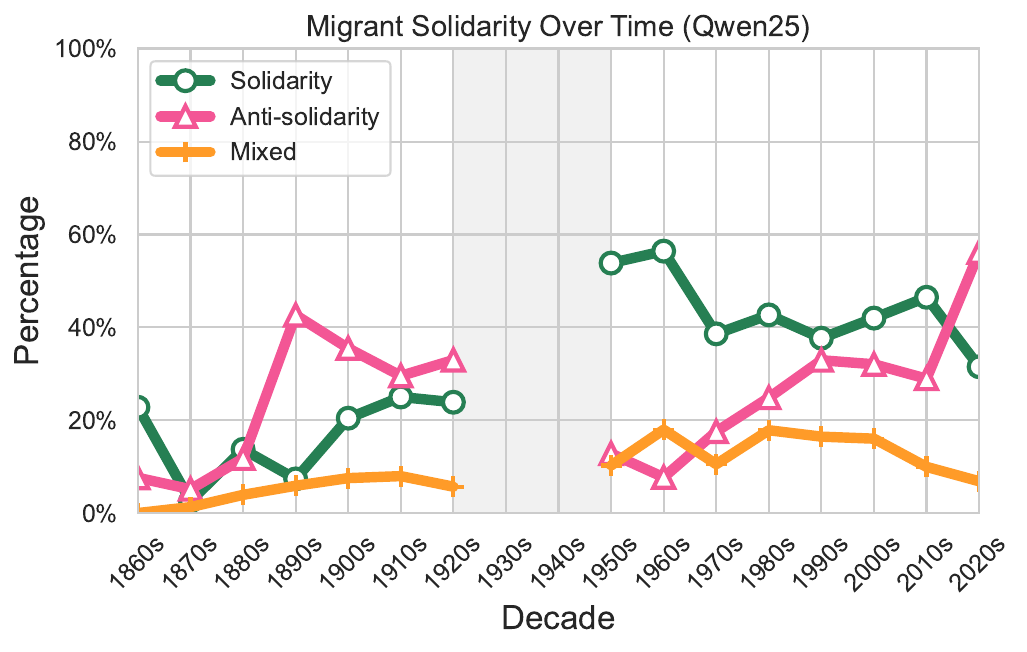}
        \caption{Qwen-2.5-72B}
        \label{fig:hl-qwen25}
    \end{subfigure}

    \vspace{0.35em}

    % --- Row 4 ---
    \begin{subfigure}[t]{0.46\linewidth}
        \centering
        \includegraphics[width=0.90\linewidth]{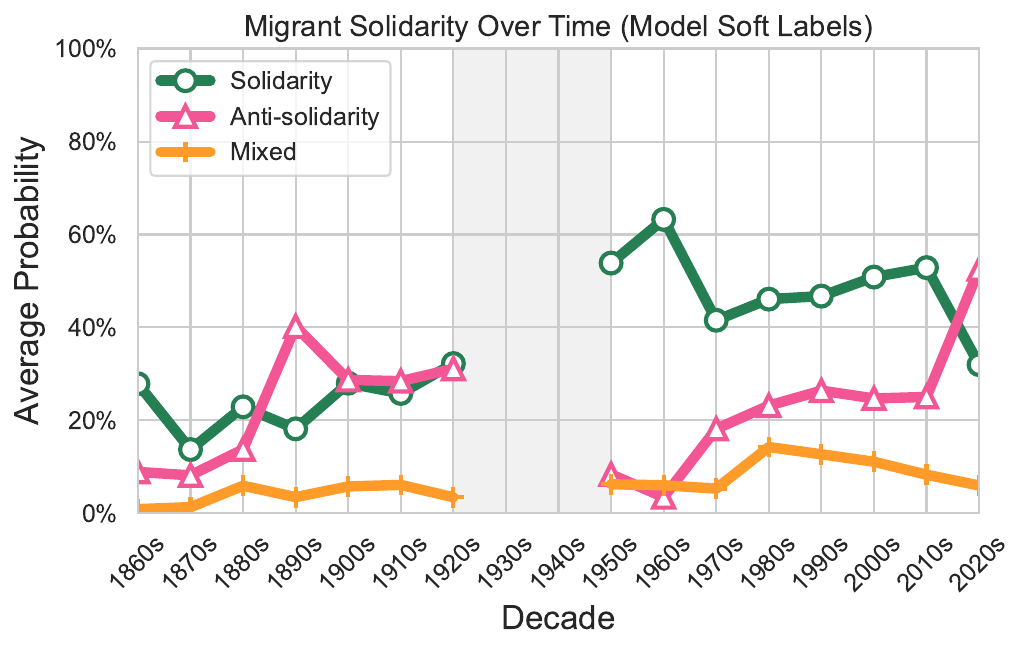}
        \caption{Model soft labels}
        \label{fig:hl-model-soft}
    \end{subfigure}
    \hfill
    \begin{subfigure}[t]{0.46\linewidth}
        \centering
        \includegraphics[width=0.90\linewidth]{plots/analysis/TrendCorrelation/Highlevel/SoftLabelsDSL_Highlevel_300Seeds.pdf}
        \caption{DSL (soft labels)}
        \label{fig:hl-dsl-soft}
    \end{subfigure}

    \caption{
    High-level solidarity trends for the \textit{Migrant} category under selected labeling and correction strategies discussed in \Cref{sec:dsl} based on the %\textbf{2k evaluation subset}. 
    \textbf{2,184-instance human-annotated subset}. Trends are computed as label shares within each decade (see definitions in \Cref{box:trend-definitions}). The period from 1930 to 1949 in each panel (grey shaded area) is excluded due to limited data availability during and immediately after the NS dictatorship. \ak{For DSL, lines show the mean estimate across 300 simulated runs. In each run, approximately 20\% of the human-annotated instances are retained as labeled data, while the remaining human labels are treated as unobserved. Colored shaded areas show the mean $\pm$ the decade-specific SD across runs.}
    }
    \label{fig:highlevel-dsl-all}
\end{figure*}

\begin{figure*}[!htb]
    \centering

    % --- Row 1 ---
    \begin{subfigure}[t]{0.46\linewidth}
        \centering
        \includegraphics[width=0.95\linewidth]{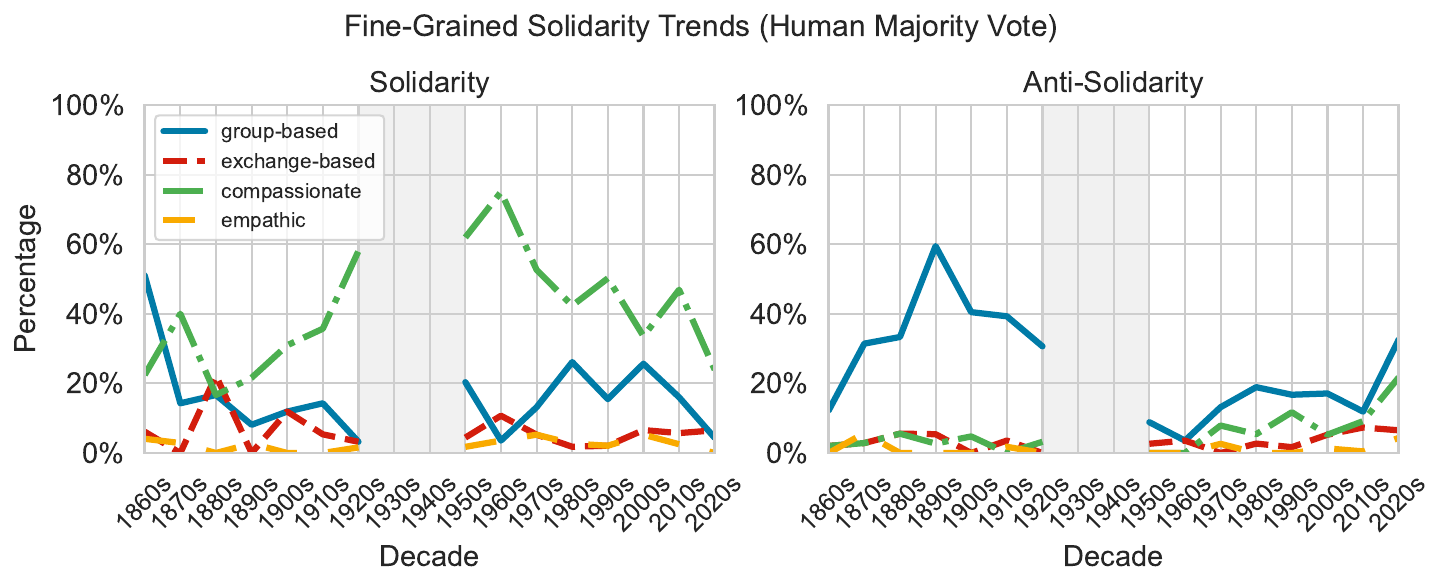}
        \caption{Human majority vote}
        \label{fig:hl-human-maj}
    \end{subfigure}
    \hfill
    \begin{subfigure}[t]{0.46\linewidth}
        \centering
        \includegraphics[width=0.95\linewidth]{plots/analysis/TrendCorrelation/Finegrained/Finegrained_SoftLabelTrends_Human.pdf}
        \caption{Human soft labels}
        \label{fig:hl-human-soft}
    \end{subfigure}

    \vspace{0.35em}

    % --- Row 2 ---
    \begin{subfigure}[t]{0.46\linewidth}
        \centering
        \includegraphics[width=0.95\linewidth]{plots/analysis/TrendCorrelation/Finegrained/Gptoss120b_Finegrained.pdf}
        \caption{gpt-oss-120B}
        \label{fig:hl-gptoss120b}
    \end{subfigure}
    \hfill
    \begin{subfigure}[t]{0.46\linewidth}
        \centering
        \includegraphics[width=0.95\linewidth]{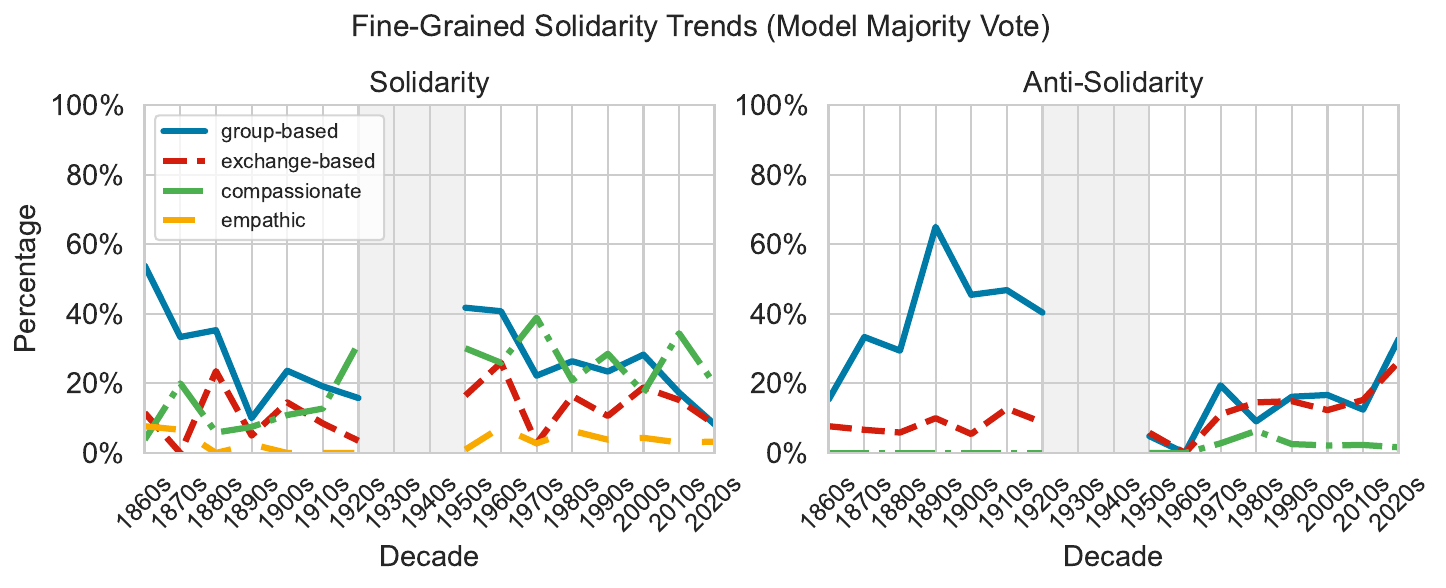}
        \caption{Model majority vote}
        \label{fig:hl-model-maj}
    \end{subfigure}

    \vspace{0.35em}

    % --- Row 3 ---
    \begin{subfigure}[t]{0.46\linewidth}
        \centering
        \includegraphics[width=0.95\linewidth]{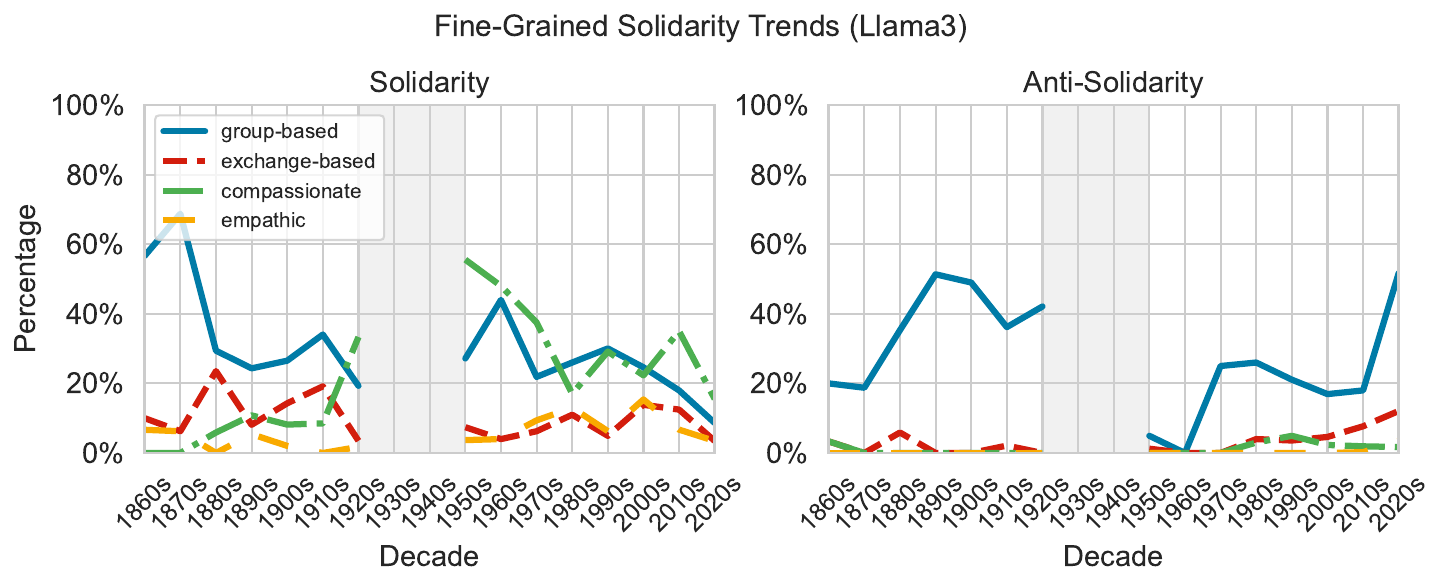}
        \caption{Llama-3.3-70B}
        \label{fig:hl-llama33}
    \end{subfigure}
    \hfill
    \begin{subfigure}[t]{0.46\linewidth}
        \centering
        \includegraphics[width=0.95\linewidth]{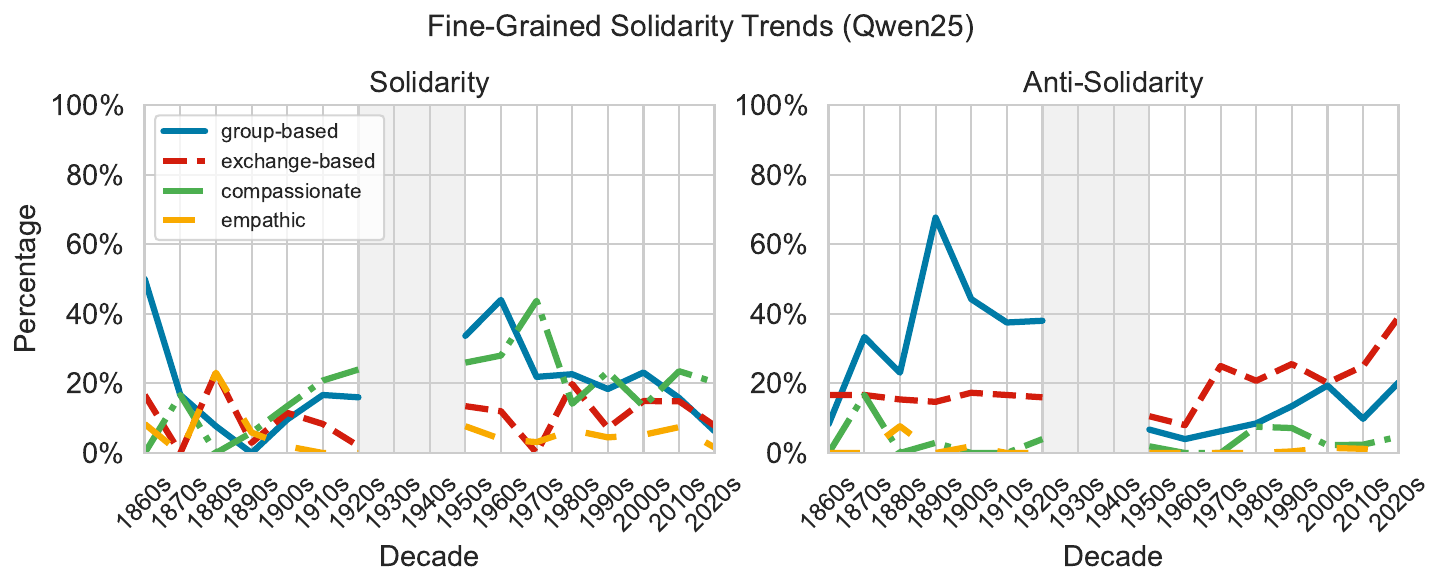}
        \caption{Qwen-2.5-72B}
        \label{fig:hl-qwen25}
    \end{subfigure}

    \vspace{0.35em}

    % --- Row 4 ---
    \begin{subfigure}[t]{0.46\linewidth}
        \centering
        \includegraphics[width=0.95\linewidth]{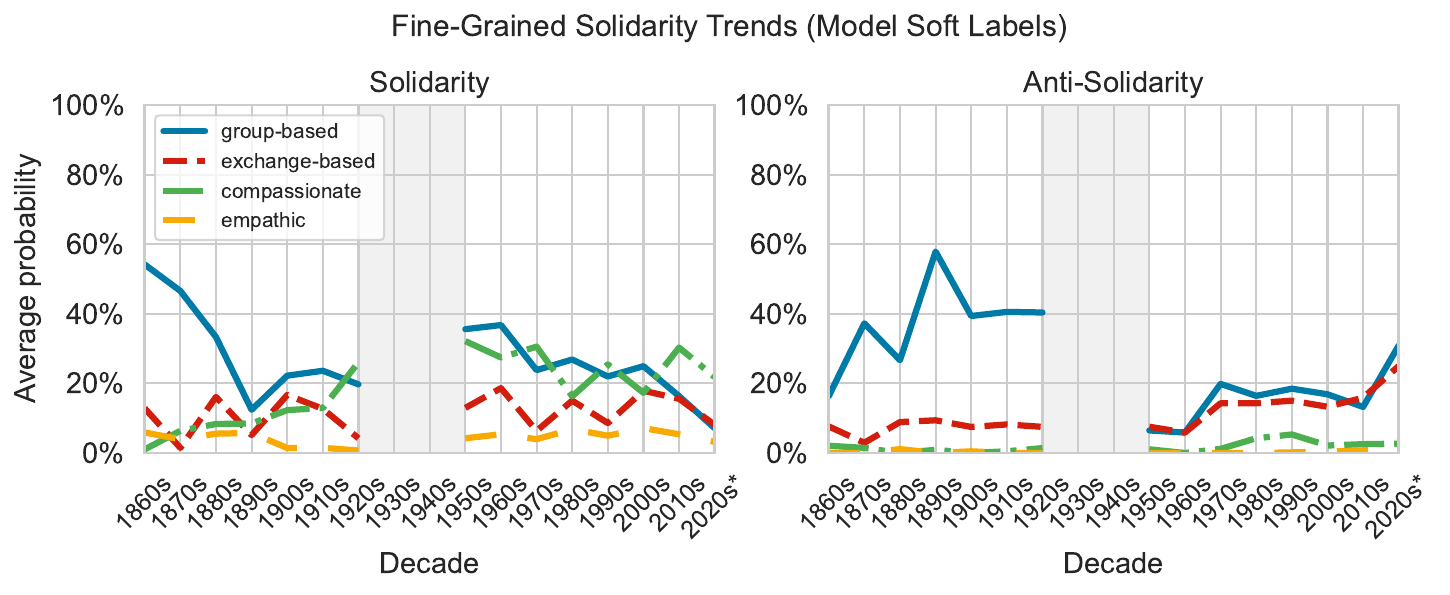}
        \caption{Model soft labels}
        \label{fig:hl-model-soft}
    \end{subfigure}
    \hfill
    \begin{subfigure}[t]{0.46\linewidth}
        \centering
        \includegraphics[width=0.95\linewidth]{plots/analysis/TrendCorrelation/Finegrained/SoftLabelsDSL_Finegrained_300Seeds.pdf}
        \caption{DSL (soft labels)}
        \label{fig:hl-dsl-soft}
    \end{subfigure}

    \caption{
    Fine-grained solidarity trends for the \textit{Migrant} category under selected labeling and correction strategies discussed in \Cref{sec:dsl} based on the %\textbf{2k evaluation subset}. 
    \textbf{2,184-instance human-annotated subset}. Trends are computed as label shares within each decade (see definitions in \Cref{box:trend-definitions}). The period from 1930 to 1949 in each panel (grey shaded area) is excluded due to limited data availability during and immediately after the NS dictatorship. \ak{For DSL, lines show the mean estimate across 300 simulated runs. In each run, approximately 20\% of the human-annotated instances are retained as labeled data, while the remaining human labels are treated as unobserved. Colored shaded areas show the mean $\pm$ the decade-specific SD across runs.}
    }
    \label{fig:finegrained-dsl-all}
\end{figure*}

\begin{table}[t]
\centering
\small
\begin{tabular}{lccc}
\toprule
\ak{\textbf{Model / Estimator}} & \ak{\textbf{Pearson}} & \ak{\textbf{Spearman}} & \ak{\textbf{RMSE}} \\
\midrule
\multicolumn{4}{l}{\ak{\textbf{High-level trends}}} \\
\midrule
\ak{Llama-3.3-70B} & \ak{0.868} & \ak{0.864} & \ak{0.082} \\
\ak{Llama-4-Scout} & \ak{0.738} & \ak{0.757} & \ak{0.084} \\
\ak{Qwen-2.5} & \ak{0.718} & \ak{0.714} & \ak{0.116} \\
\ak{gpt-oss-120B} & \ak{0.847} & \ak{0.810} & \ak{0.068} \\
\ak{GPT-5} & \ak{0.818} & \ak{0.816} & \ak{0.073} \\
\ak{Majority-vote ensemble} & \ak{0.877} & \ak{\textbf{0.874}} & \ak{0.080} \\
\ak{Soft-label ensemble} & \ak{0.839} & \ak{0.800} & \ak{0.075} \\
\ak{DSL (Llama-3.3-70B)} & \ak{0.867 $\pm$ 0.037} & \ak{0.835 $\pm$ 0.045} & \ak{0.053 $\pm$ 0.006} \\
\ak{DSL (gpt-oss-120B)} & \ak{0.864 $\pm$ 0.038} & \ak{0.832 $\pm$ 0.046} & \ak{0.052 $\pm$ 0.006} \\
\ak{DSL (soft-label ensemble)} & \ak{\textbf{0.878 $\pm$ 0.031}} & \ak{0.846 $\pm$ 0.040} & \ak{\textbf{0.046 $\pm$ 0.006}} \\
\midrule
\multicolumn{4}{l}{\ak{\textbf{Fine-grained trends}}} \\
\midrule
\ak{Llama-3.3-70B} & \ak{0.639} & \ak{0.577} & \ak{0.090} \\
\ak{Llama-4-Scout} & \ak{0.578} & \ak{0.580} & \ak{0.070} \\
\ak{Qwen-2.5} & \ak{0.370} & \ak{0.394} & \ak{0.105} \\
\ak{gpt-oss-120B} & \ak{0.445} & \ak{0.517} & \ak{0.100} \\
\ak{GPT-5} & \ak{0.566} & \ak{0.520} & \ak{0.077} \\
\ak{Majority-vote ensemble} & \ak{0.562} & \ak{0.589} & \ak{0.088} \\
\ak{Soft-label ensemble} & \ak{0.576} & \ak{0.579} & \ak{0.088} \\
\ak{DSL (Llama-3.3-70B)} & \ak{0.725 $\pm$ 0.055} & \ak{0.663 $\pm$ 0.058} & \ak{0.050 $\pm$ 0.005} \\
\ak{DSL (gpt-oss-120B)} & \ak{\textbf{0.727 $\pm$ 0.053}} & \ak{\textbf{0.666 $\pm$ 0.058}} & \ak{0.049 $\pm$ 0.005} \\
\ak{DSL (soft-label ensemble)} & \ak{0.668 $\pm$ 0.056} & \ak{0.623 $\pm$ 0.055} & \ak{\textbf{0.047 $\pm$ 0.004}} \\
\bottomrule
\end{tabular}
\caption{\ak{Comparison with human-based decade-level trends across estimators. For DSL estimators, values are reported as mean $\pm$ standard deviation (SD) across 300 random seeds. %uncorrected estimators are fixed and therefore have no seed-to-seed variation.
\textit{Reference:} human majority vote. Results against human soft-label trends are reported in \Cref{sec:dsl}.}}
\label{tab:validation-dsl-full-humanmaj}
\end{table}

\begin{table}[ht]
\centering
\small
\renewcommand{\arraystretch}{1.2}
\setlength{\tabcolsep}{6pt}
\begin{tabular}{llccc}
\toprule
\ak{\textbf{Label}} & \ak{\textbf{Model}} & \ak{\textbf{Pearson}} & \ak{\textbf{Spearman}} & \ak{\textbf{RMSE}} \\
\midrule
\ak{Solidarity} & \ak{Llama-3.3-70B} & \ak{0.921} & \ak{0.929} & \ak{0.099} \\
 & \ak{Model soft raw} & \ak{0.942} & \ak{0.939} & \ak{0.106} \\
 & \ak{DSL soft} & \ak{\textbf{0.955 $\pm$ 0.019}} & \ak{\textbf{0.940 $\pm$ 0.029}} & \ak{\textbf{0.051 $\pm$ 0.010}} \\
\cmidrule(lr){1-5}
\ak{Anti-solidarity} & \ak{Llama-3.3-70B} & \ak{0.872} & \ak{0.836} & \ak{0.069} \\
 & \ak{Model soft raw} & \ak{0.922} & \ak{\textbf{0.896}} & \ak{0.053} \\
 & \ak{DSL soft} & \ak{\textbf{0.935 $\pm$ 0.027}} & \ak{0.869 $\pm$ 0.062} & \ak{\textbf{0.043 $\pm$ 0.009}} \\
\cmidrule(lr){1-5}
\ak{Mixed} & \ak{Llama-3.3-70B} & \ak{\textbf{0.900}} & \ak{\textbf{0.871}} & \ak{\textbf{0.023}} \\
 & \ak{Model soft raw} & \ak{0.757} & \ak{0.607} & \ak{0.024} \\
 & \ak{DSL soft} & \ak{0.747 $\pm$ 0.113} & \ak{0.682 $\pm$ 0.140} & \ak{0.029 $\pm$ 0.006} \\
\cmidrule(lr){1-5}
\ak{None} & \ak{Llama-3.3-70B} & \ak{0.904} & \ak{0.889} & \ak{0.138} \\
 & \ak{Model soft raw} & \ak{0.895} & \ak{0.914} & \ak{0.124} \\
 & \ak{DSL soft} & \ak{\textbf{0.954 $\pm$ 0.021}} & \ak{\textbf{0.942 $\pm$ 0.031}} & \ak{\textbf{0.044 $\pm$ 0.009}} \\
%\midrule
%\ak{\textbf{Average}} & \ak{Llama-3.3-70B} & \ak{\textbf{0.899}} & \ak{\textbf{0.881}} & \ak{0.082} \\
% & \ak{Model soft raw} & \ak{0.879} & \ak{0.839} & \ak{0.077} \\
% & \ak{DSL soft} & \ak{0.898 $\pm$ 0.030} & \ak{0.858 $\pm$ 0.039} & \ak{\textbf{0.042 $\pm$ 0.005}} \\
\bottomrule
\end{tabular}
\caption{\ak{High-level per-label agreement between model-based and human-annotated decade-level trends. For DSL, values are reported as mean $\pm$ standard deviation (SD) across 300 random seeds. Higher is better for Pearson and Spearman; lower is better for RMSE. \textit{Reference:} human soft labels.}}
\label{tab:per-label-high-humansoft}
\end{table}

\begin{table}[ht]
\centering
\small
\renewcommand{\arraystretch}{1.2}
\setlength{\tabcolsep}{6pt}
\begin{tabular}{llccc}
\toprule
\ak{\textbf{Label}} & \ak{\textbf{Model}} & \ak{\textbf{Pearson}} & \ak{\textbf{Spearman}} & \ak{\textbf{RMSE}} \\
\midrule
\ak{Group-based S} & \ak{Llama-3.3-70B} & \ak{0.542} & \ak{0.429} & \ak{0.180} \\
 & \ak{Model soft raw} & \ak{\textbf{0.780}} & \ak{\textbf{0.776}} & \ak{0.117} \\
 & \ak{DSL soft} & \ak{0.771 $\pm$ 0.095} & \ak{0.758 $\pm$ 0.107} & \ak{\textbf{0.055 $\pm$ 0.011}} \\
\cmidrule(lr){1-5}
\ak{Exchange-based S} & \ak{Llama-3.3-70B} & \ak{0.319} & \ak{0.352} & \ak{0.074} \\
 & \ak{Model soft raw} & \ak{\textbf{0.842}} & \ak{\textbf{0.721}} & \ak{0.079} \\
 & \ak{DSL soft} & \ak{0.703 $\pm$ 0.121} & \ak{0.696 $\pm$ 0.130} & \ak{\textbf{0.033 $\pm$ 0.007}} \\
\cmidrule(lr){1-5}
\ak{Empathic S} & \ak{Llama-3.3-70B} & \ak{0.336} & \ak{0.456} & \ak{0.048} \\
 & \ak{Model soft raw} & \ak{0.036} & \ak{0.197} & \ak{0.029} \\
 & \ak{DSL soft} & \ak{\textbf{0.556 $\pm$ 0.195}} & \ak{\textbf{0.520 $\pm$ 0.170}} & \ak{\textbf{0.023 $\pm$ 0.005}} \\
\cmidrule(lr){1-5}
\ak{Compassionate S} & \ak{Llama-3.3-70B} & \ak{0.889} & \ak{0.867} & \ak{0.197} \\
 & \ak{Model soft raw} & \ak{0.846} & \ak{0.844} & \ak{0.219} \\
 & \ak{DSL soft} & \ak{\textbf{0.907 $\pm$ 0.040}} & \ak{\textbf{0.901 $\pm$ 0.051}} & \ak{\textbf{0.073 $\pm$ 0.014}} \\
\cmidrule(lr){1-5}
\ak{Group-based AS} & \ak{Llama-3.3-70B} & \ak{0.833} & \ak{0.779} & \ak{0.099} \\
 & \ak{Model soft raw} & \ak{0.888} & \ak{0.797} & \ak{0.070} \\
 & \ak{DSL soft} & \ak{\textbf{0.921 $\pm$ 0.034}} & \ak{\textbf{0.896 $\pm$ 0.048}} & \ak{\textbf{0.060 $\pm$ 0.012}} \\
\cmidrule(lr){1-5}
\ak{Exchange-based AS} & \ak{Llama-3.3-70B} & \ak{0.485} & \ak{0.468} & \ak{0.031} \\
 & \ak{Model soft raw} & \ak{0.377} & \ak{0.372} & \ak{0.089} \\
 & \ak{DSL soft} & \ak{\textbf{0.685 $\pm$ 0.134}} & \ak{\textbf{0.676 $\pm$ 0.146}} & \ak{\textbf{0.026 $\pm$ 0.007}} \\
\cmidrule(lr){1-5}
\ak{Empathic AS} & \ak{Llama-3.3-70B} & \ak{0.023} & \ak{0.097} & \ak{0.021} \\
 & \ak{Model soft raw} & \ak{-0.049} & \ak{0.184} & \ak{0.019} \\
 & \ak{DSL soft} & \ak{\textbf{0.730 $\pm$ 0.230}} & \ak{\textbf{0.563 $\pm$ 0.192}} & \ak{\textbf{0.013 $\pm$ 0.003}} \\
\cmidrule(lr){1-5}
\ak{Compassionate AS} & \ak{Llama-3.3-70B} & \ak{0.395} & \ak{0.459} & \ak{0.075} \\
 & \ak{Model soft raw} & \ak{0.578} & \ak{0.707} & \ak{0.067} \\
 & \ak{DSL soft} & \ak{\textbf{0.777 $\pm$ 0.095}} & \ak{\textbf{0.749 $\pm$ 0.112}} & \ak{\textbf{0.040 $\pm$ 0.008}} \\
%\midrule
%\ak{\textbf{Average}} & \ak{Llama-3.3-70B} & \ak{0.478} & \ak{0.488} & \ak{0.091} \\
% & \ak{Model soft raw} & \ak{0.537} & \ak{0.575} & \ak{0.086} \\
% & \ak{DSL soft} & \ak{\textbf{0.756 $\pm$ 0.045}} & \ak{\textbf{0.720 $\pm$ 0.047}} & \ak{\textbf{0.040 $\pm$ 0.004}} \\
\bottomrule
\end{tabular}
\caption{\ak{Fine-grained per-label agreement between model-based and human-annotated decade-level trends. For DSL, values are reported as mean $\pm$ standard deviation (SD) across 300 random seeds. Higher is better for Pearson and Spearman; lower is better for RMSE. \textit{Reference:} human soft labels.}}
\label{tab:per-label-fine-humansoft}
\end{table}

\begin{table}[ht]
\centering
\small
\renewcommand{\arraystretch}{1.15}
\setlength{\tabcolsep}{5pt}
\begin{tabular}{lccccc}
\toprule
\ak{\textbf{Label}} & \ak{\shortstack{\textbf{Full}\\\textbf{prevalence}}} & \ak{\shortstack{\textbf{HumanSoft}\\\textbf{trend SD}}} & \ak{\shortstack{\textbf{Seed variation /}\\\textbf{human signal}}} & \ak{\shortstack{\textbf{Pearson}\\\textbf{SD}}} & \ak{\shortstack{\textbf{Sampling-miss}\\\textbf{rate}}} \\
\midrule
\ak{Solidarity} & \ak{52.5\%} & \ak{0.156} & \ak{0.314} & \ak{0.019} & \ak{0.0\%} \\
\ak{None} & \ak{17.8\%} & \ak{0.135} & \ak{0.324} & \ak{0.021} & \ak{0.3\%} \\
\ak{Anti-solidarity} & \ak{22.4\%} & \ak{0.105} & \ak{0.397} & \ak{0.027} & \ak{0.2\%} \\
\ak{Mixed} & \ak{7.4\%} & \ak{0.033} & \ak{0.835} & \ak{0.113} & \ak{1.6\%} \\
\bottomrule
\end{tabular}
\caption{\ak{High-level diagnostics of DSL stability across 300 simulated runs. Labels are ordered from lowest to highest Pearson SD. See the column definitions below.%Full prevalence is the label's HumanSoft share in the complete human-annotated evaluation subset. HumanSoft trend SD measures temporal variation in the human reference trend across decades. Seed variation / human signal is the ratio of the mean across-decade SD of DSL estimates across runs to the HumanSoft trend SD. %Pearson SD measures variation in DSL--HumanSoft Pearson correlation across runs. 
%Sampling-miss rate is the percentage of seed-decade samples in which the label has a HumanSoft share of zero, conditional on that label being present in the complete human-annotated decade.
}}
\label{tab:dsl-coverage-diagnostics-high}
\end{table}

\begin{table}[ht]
\centering
\small
\renewcommand{\arraystretch}{1.15}
\setlength{\tabcolsep}{5pt}

\begin{tabular}{lccccc}
\toprule
\ak{\textbf{Label}} & \ak{\shortstack{\textbf{Full}\\\textbf{prevalence}}} & \ak{\shortstack{\textbf{HumanSoft}\\\textbf{trend SD}}} & \ak{\shortstack{\textbf{Seed variation /}\\\textbf{human signal}}} & \ak{\shortstack{\textbf{Pearson}\\\textbf{SD}}} & \ak{\shortstack{\textbf{Sampling-miss}\\\textbf{rate}}} \\
\midrule
\ak{Group-based AS} & \ak{13.4\%} & \ak{0.136} & \ak{0.417} & \ak{0.034} & \ak{2.3\%} \\
\ak{Compassionate S} & \ak{32.7\%} & \ak{0.134} & \ak{0.509} & \ak{0.040} & \ak{0.0\%} \\
\ak{Group-based S} & \ak{12.5\%} & \ak{0.069} & \ak{0.766} & \ak{0.095} & \ak{0.8\%} \\
\ak{Compassionate AS} & \ak{5.5\%} & \ak{0.049} & \ak{0.685} & \ak{0.095} & \ak{13.3\%} \\
\ak{Exchange-based S} & \ak{3.6\%} & \ak{0.031} & \ak{0.972} & \ak{0.121} & \ak{16.7\%} \\
\ak{Exchange-based AS} & \ak{3.0\%} & \ak{0.022} & \ak{1.054} & \ak{0.134} & \ak{21.2\%} \\
\ak{Empathic S} & \ak{1.9\%} & \ak{0.019} & \ak{1.041} & \ak{0.195} & \ak{27.7\%} \\
\ak{Empathic AS} & \ak{0.5\%} & \ak{0.016} & \ak{0.575} & \ak{0.230} & \ak{47.1\%} \\
\bottomrule
\end{tabular}

\caption{\ak{Fine-grained diagnostics of DSL stability across 300 simulated runs. Labels are ordered from lowest to highest Pearson SD. %Full prevalence is the label's HumanSoft share in the complete human-annotated evaluation subset. HumanSoft trend SD measures temporal variation in the human reference trend across decades. Seed variation / human signal is the ratio of the mean across-decade SD of DSL estimates across runs to the HumanSoft trend SD. Sampling-miss rate is the percentage of seed-decade samples in which the label has a HumanSoft share of zero, conditional on that label being present in the complete human-annotated decade.
See the column definitions below.}}
\label{tab:dsl-coverage-diagnostics-fine}

\vspace{0.5em}

\begin{minipage}{0.95\linewidth}
\small
\raggedright
\ak{\textit{Column definitions.}
\begin{itemize}
\item Full prevalence is the HumanSoft share of that label in the complete human-annotated subset.
\item HumanSoft trend SD is the standard deviation of the label's HumanSoft share across the 14 decades; higher values mean that the human trend changes more over time.
\item  Seed variation / human signal is
\[
\frac{\text{average DSL SD across seeds within each decade}}
{\text{HumanSoft trend SD}}.
\]
It therefore shows how large the seed-to-seed variation is relative to the actual variation of the human trend.
\item Pearson SD is the SD of the DSL-HumanSoft Pearson correlation across the 300 seeds.
\item Sampling-miss rate is the percentage of seed-decade samples in which the label has a HumanSoft share of zero, conditional on that label being present in the complete human-annotated decade.
\end{itemize}
}
\end{minipage}

\end{table}

\begin{figure*}[p]
\centering

\begin{subfigure}[t]{0.95\linewidth}
    \centering
    \includegraphics[width=\linewidth]{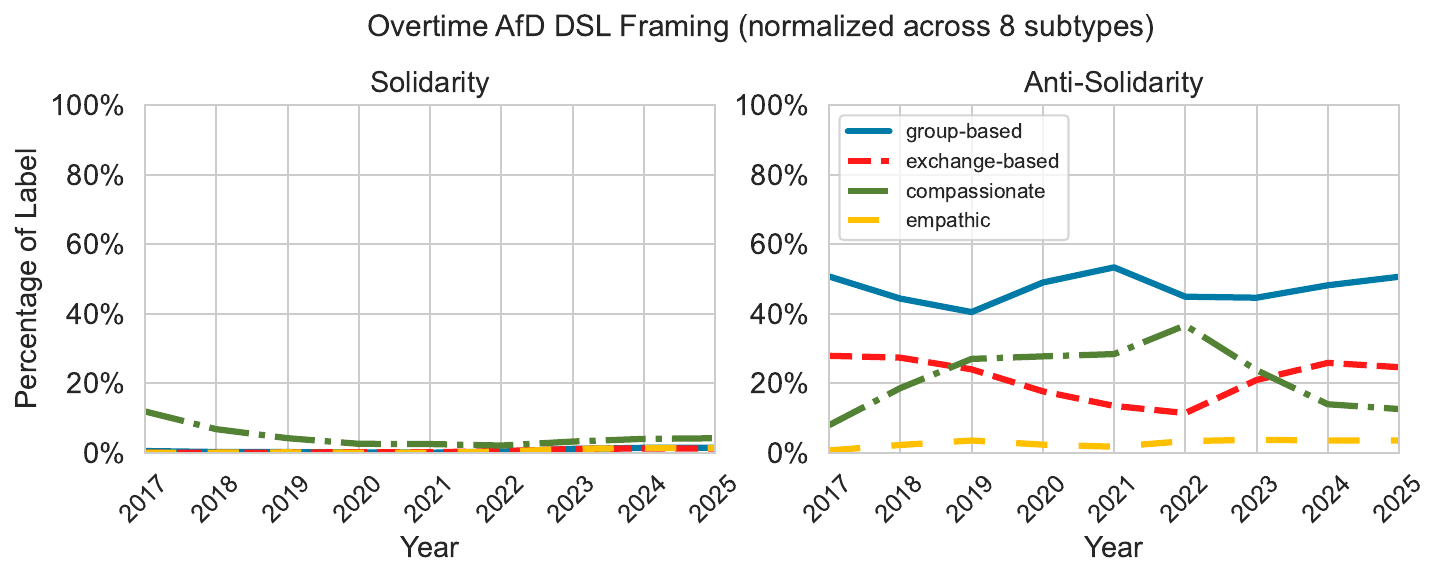}
    \caption{AfD}
    \label{fig:finegrained-afd-app}
\end{subfigure}

\vspace{1em}

\begin{subfigure}[t]{0.95\linewidth}
    \centering
    \includegraphics[width=\linewidth]{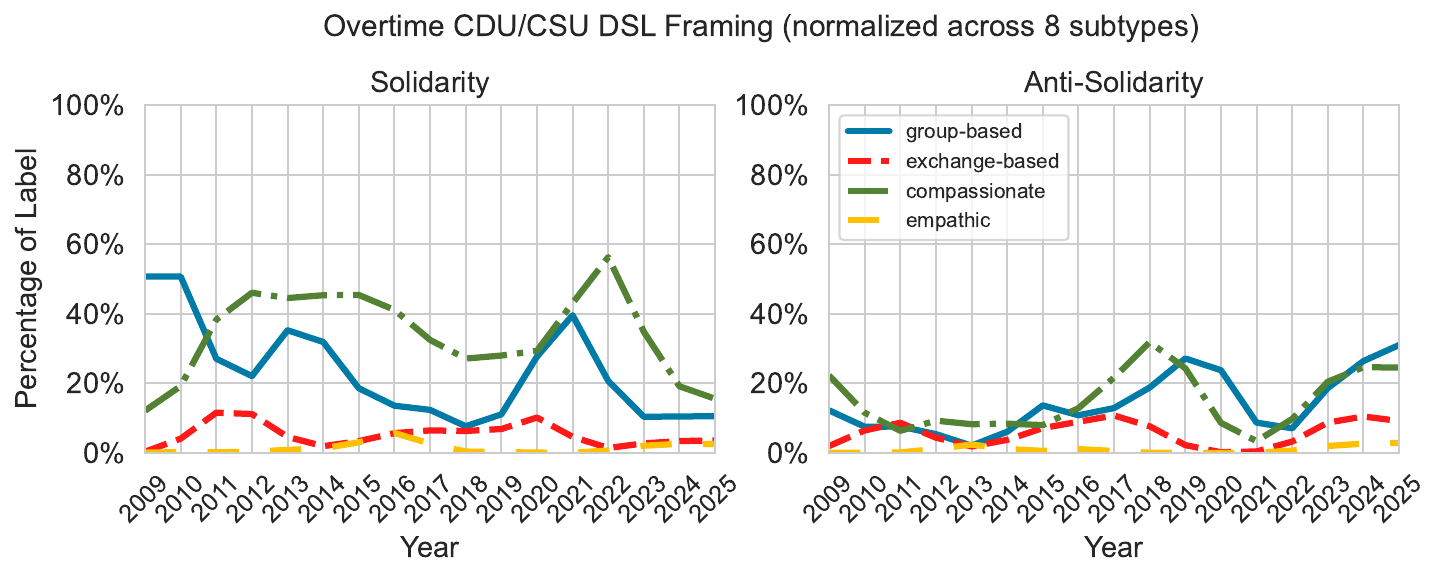}
    \caption{CDU/CSU}
    \label{fig:finegrained-cdu-app}
\end{subfigure}

\vspace{1em}

\begin{subfigure}[t]{0.95\linewidth}
    \centering
    \includegraphics[width=\linewidth]{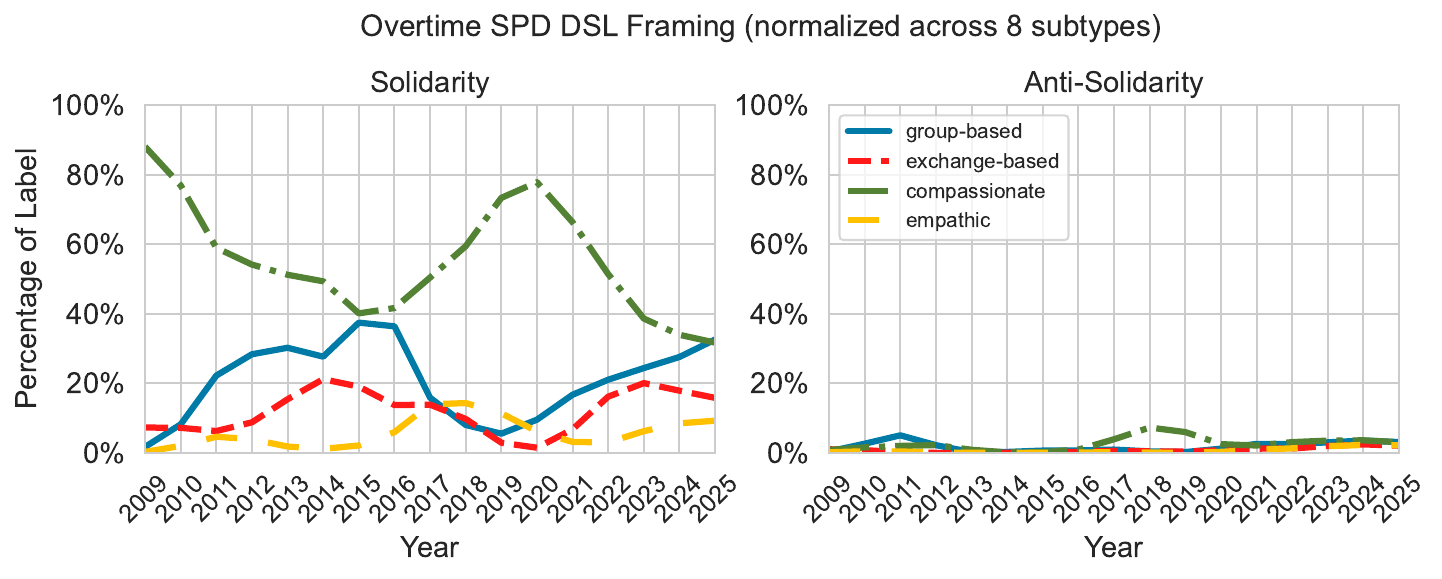}
    \caption{SPD}
    \label{fig:finegrained-spd-app}
\end{subfigure}

\caption{Party-specific DSL-adjusted yearly shares of (anti-)solidarity subtypes, 2009-2025. This figure provides a larger-format version of the party-specific panels shown in \Cref{fig:finegrained-per-party-2x2}. Within each year and party, subtype shares are renormalized across the eight solidarity and anti-solidarity categories so that they sum to 100\%.}
\label{fig:finegrained-per-party-enlarged}
\end{figure*}

\begin{figure*}[p]
\ContinuedFloat
\centering
\setcounter{subfigure}{3}

\begin{subfigure}[t]{0.95\linewidth}
    \centering
    \includegraphics[width=\linewidth]{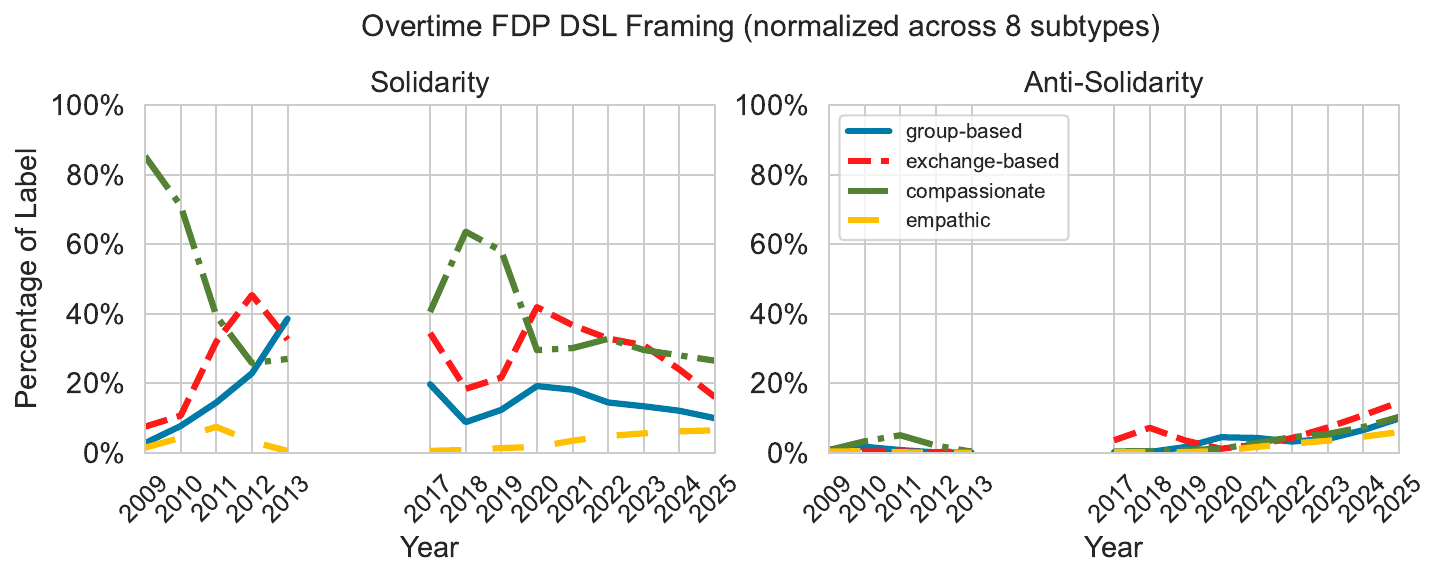}
    \caption{FDP}
    \label{fig:finegrained-fdp-app}
\end{subfigure}

\vspace{1em}

\begin{subfigure}[t]{0.95\linewidth}
    \centering
    \includegraphics[width=\linewidth]{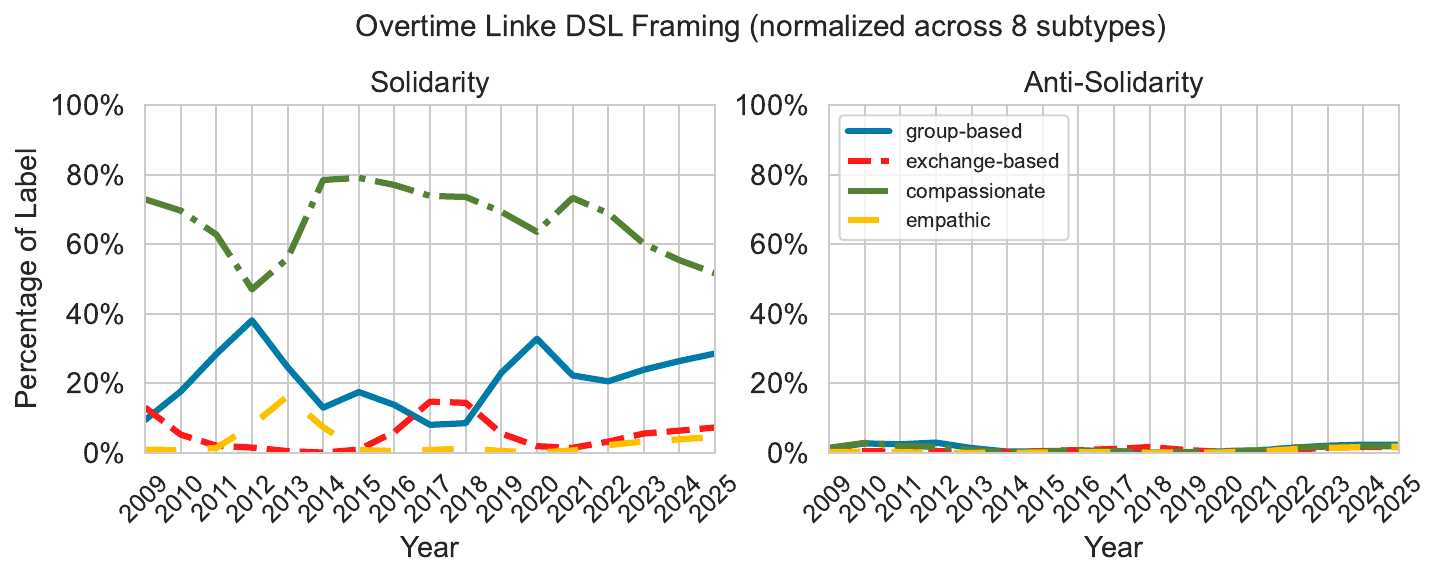}
    \caption{Die Linke}
    \label{fig:finegrained-linke-app}
\end{subfigure}

\vspace{1em}

\begin{subfigure}[t]{0.95\linewidth}
    \centering
    \includegraphics[width=\linewidth]{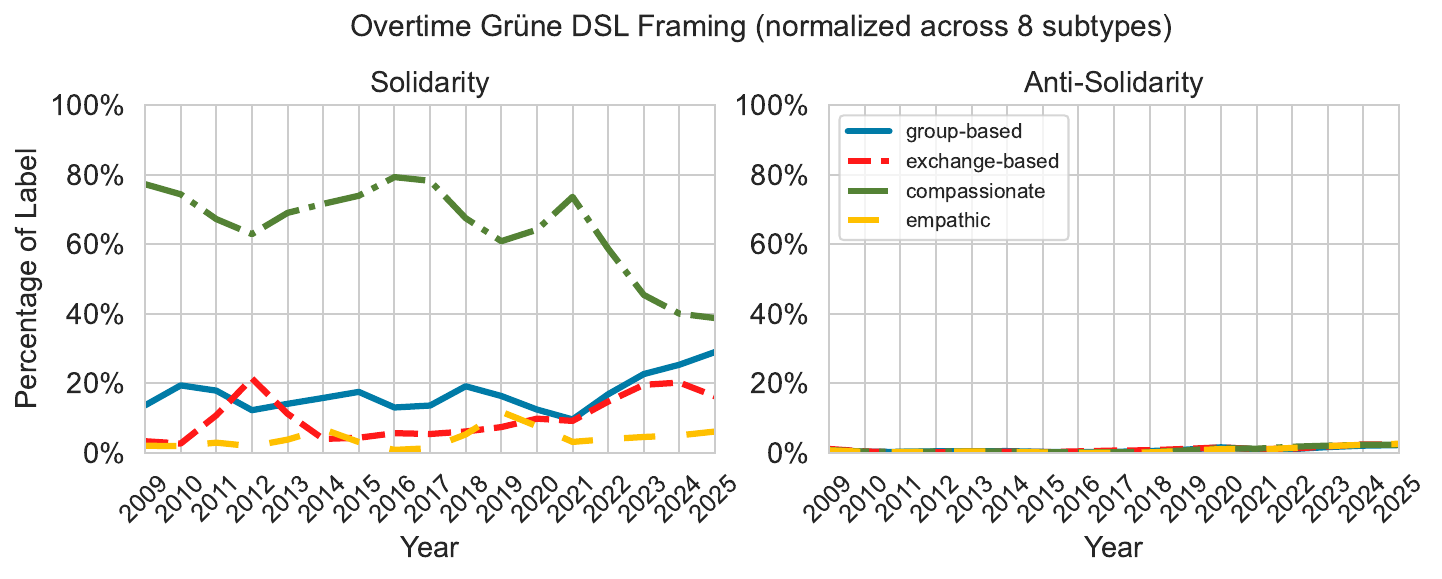}
    \caption{Die Grünen}
    \label{fig:finegrained-gruene-app}
\end{subfigure}

\caption[]{Party-specific DSL-adjusted yearly shares of (anti-)solidarity subtypes,
2009--2025 (continued).}
\end{figure*}

%%% Prompts %%%

\clearpage 

\begin{comment}
\begin{figure*}[t]
  \centering
  \includegraphics[width=0.8\linewidth]{plots/analysis/Prompt/Frau_Onestep_prompt.pdf}
  \caption{One-step zero-shot prompt set used for the evaluation of GPT-4 on the \emph{Frau} dataset.}
  \label{fig:one-step-prompt-frau}
\end{figure*}
\end{comment}

% Fig. X(a) - Migrant
\begin{figure*}
  \centering
  \begin{subfigure}{0.8\linewidth}
    \includegraphics[width=\linewidth]{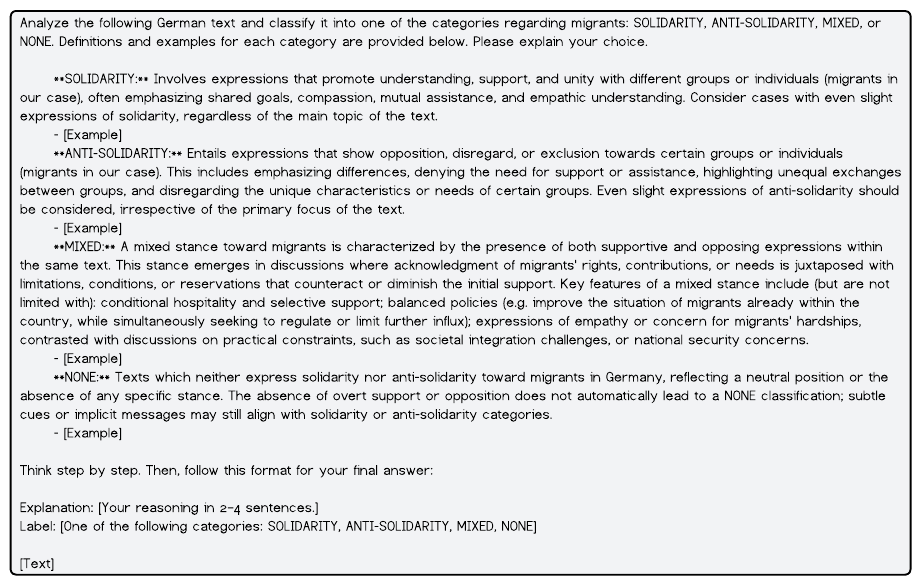}
    \caption{High-level prompt (Step 1).}
    \label{fig:two-step-prompt-a}
  \end{subfigure}
  \caption{Two-step few-shot prompt set used for the evaluation of open-weight models on the \emph{Migrant} dataset and for the final large-scale annotation with Llama-3.3-70B. 
  The model is first given the high-level classification prompt in \Cref{fig:two-step-prompt-a}. 
  If the output is Solidarity or Anti-solidarity, the corresponding subtype prompt in \Cref{fig:two-step-prompt-b} or \Cref{fig:two-step-prompt-c} is used.}
  \label{fig:two-step-prompt-migrant}
\end{figure*}

% Fig. X(b) - Migrant
\begin{figure*}\ContinuedFloat
  \centering
  \begin{subfigure}{0.8\linewidth}
    \includegraphics[width=\linewidth]{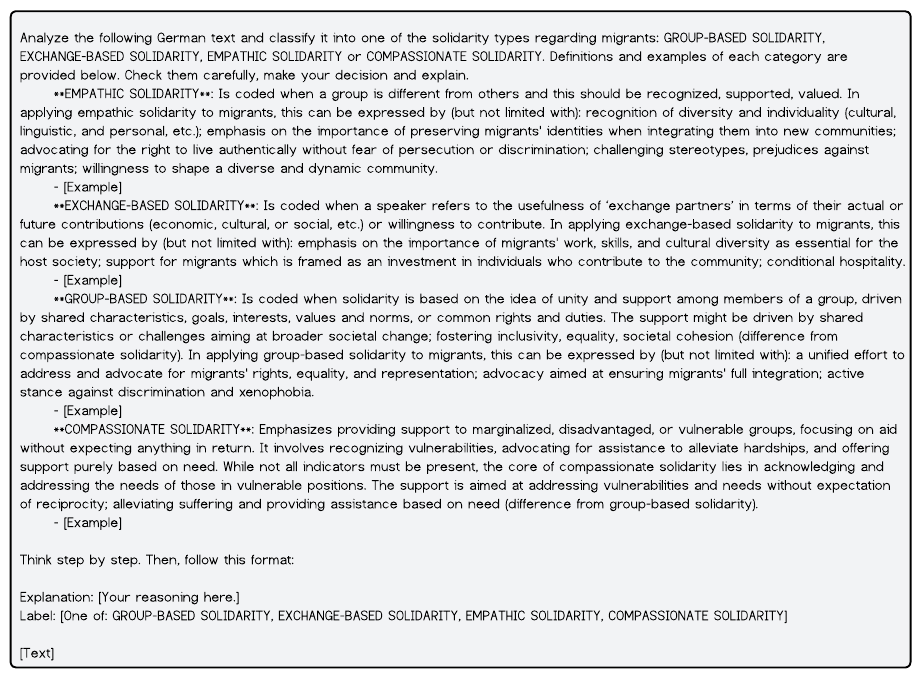}
    \caption{Subtype prompt for Solidarity (Step 2a).}
    \label{fig:two-step-prompt-b}
  \end{subfigure}
  \caption{Two-step few-shot prompt set used for the evaluation of open-weight models on the \emph{Migrant} dataset and for the final large-scale annotation with Llama-3.3-70B (continued from \Cref{fig:two-step-prompt-migrant}).}
\end{figure*}

% Fig. X(c) — Migrant
\begin{figure*}\ContinuedFloat
  \centering
  \begin{subfigure}{0.8\linewidth}
    \includegraphics[width=\linewidth]{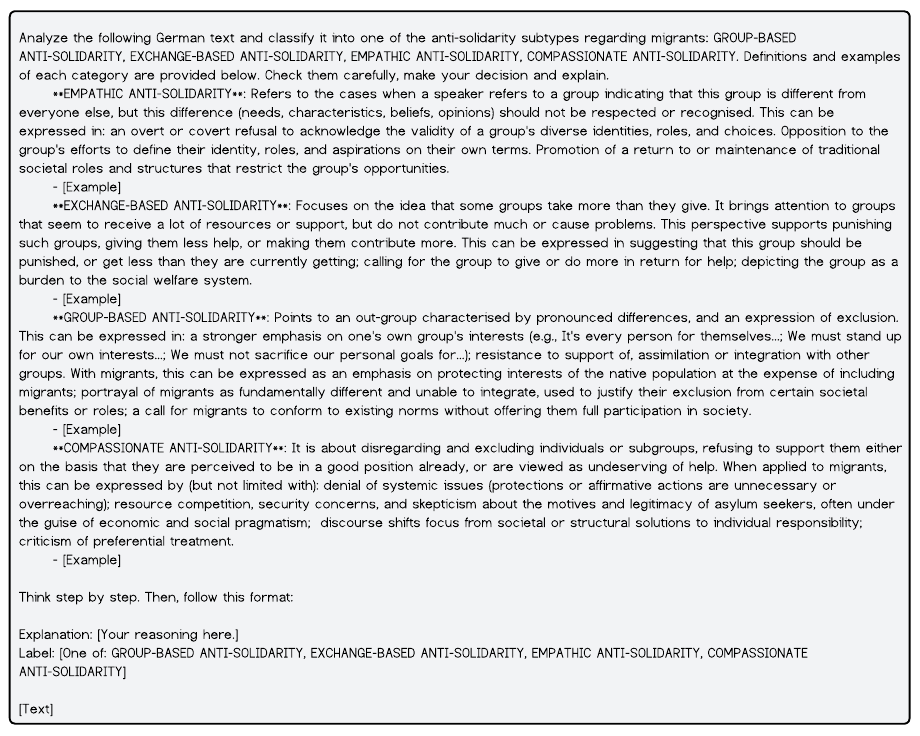}
    \caption{Subtype prompt for Anti-solidarity (Step 2b).}
    \label{fig:two-step-prompt-c}
  \end{subfigure}
  \caption{Two-step few-shot prompt set used for the evaluation of open-weight models on the \emph{Migrant} dataset and for the final large-scale annotation with Llama-3.3-70B (continued from \Cref{fig:two-step-prompt-migrant}).}
\end{figure*}

\begin{figure*}[!htb]
    \centering
    \subfloat[Solidarity and anti-solidarity frames.]{
        \includegraphics[width=0.57\linewidth]{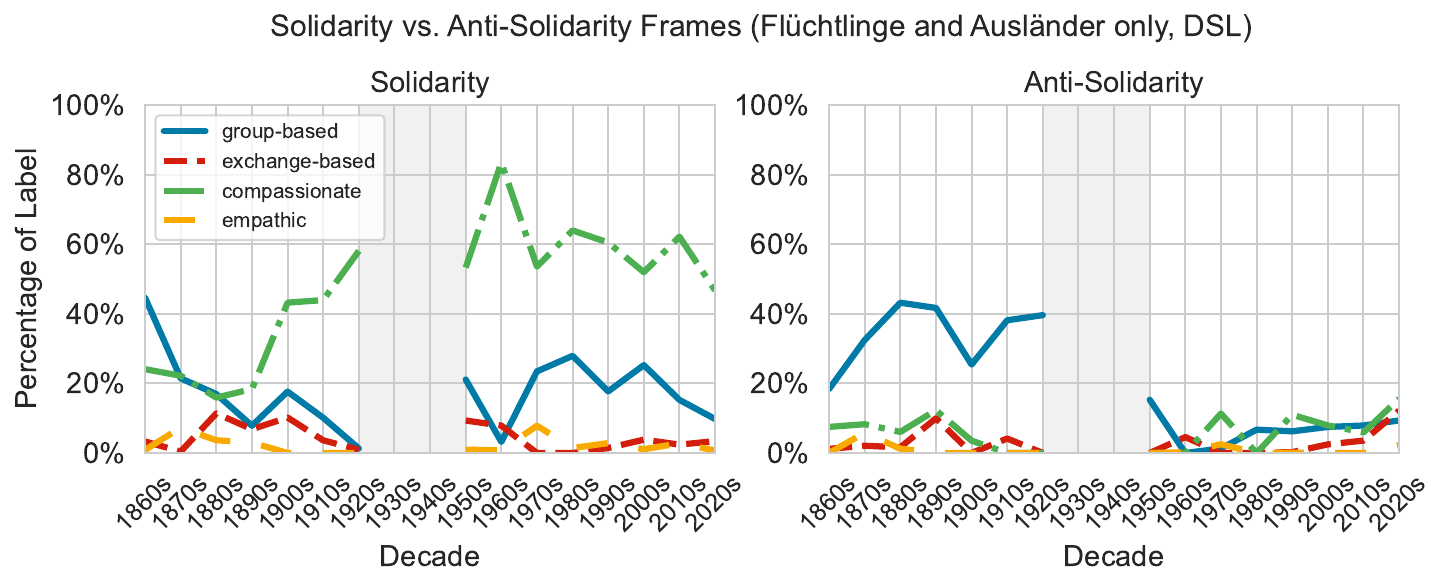}
        \label{fig:solidarity_antisolidarity-general-keywords}
    }
    \subfloat[Solidarity and anti-solidarity.]{
        \includegraphics[width=0.37\linewidth]{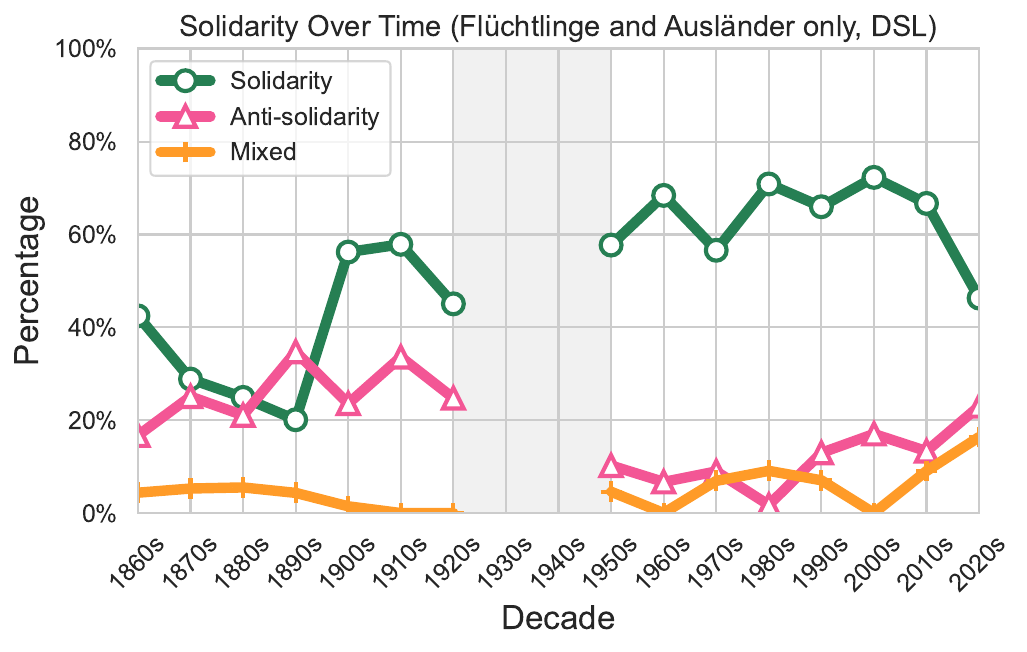}
        \label{fig:solidarity-per-decade-general-keywords}
    }
    %\vspace{-0.1cm}
    \caption{Solidarity trends for \textit{Migrant} speeches computed using DSL-corrected soft labels (see definition (c) in \Cref{box:trend-definitions}) on the restricted corpus containing only the keyword groups \emph{Flüchtlinge} and \emph{Ausländer}. \Cref{fig:solidarity-per-decade-general-keywords} shows the high-level (anti-)solidarity trends, while \Cref{fig:solidarity_antisolidarity-general-keywords} shows the distribution of fine-grained (anti-)solidarity subtypes over time. In the fine-grained panel, percentages are normalized within each decade across the eight solidarity and anti-solidarity subtypes. The period 1930-1949 (grey area) is excluded due to limited data during and after the Nazi dictatorship.}
    \label{fig:1867-2022-general-keywords}
\end{figure*}

\begin{table}[ht]
\centering
\scriptsize
\setlength{\tabcolsep}{3pt}

% \subfloat[Migrant]{
\begin{tabular}{p{2.5cm}ccccccccc}
\toprule
\textbf{Label} & \textbf{GPT-5} & \textbf{GPT-4} & \makecell{\textbf{gpt-oss}\\\textbf{120B}} &
\makecell{\textbf{Qwen}\\\textbf{2.5-72B}} &
\makecell{\textbf{Llama}\\\textbf{3.3-70B}} &
\makecell{\textbf{Llama}\\\textbf{4-Scout}} &
\makecell{\textbf{Mistral}\\\textbf{2-Large}} &
\makecell{\textbf{Gemma}\\\textbf{2-27B}} &
\textbf{BERT} \\
\midrule
Group-based S         & 0.42 & 0.33 & 0.39 & 0.41 & 0.32 & 0.36 & 0.34 & 0.34 & 0.31 \\
Exchange-based S      & 0.43 & 0.60 & 0.56 & 0.40 & 0.36 & 0.46 & 0.41 & 0.37 & 0.22 \\
Empathic S            & 0.48 & 0.05 & 0.28 & 0.19 & 0.11 & 0.17 & 0.38 & 0.24 & 0.00 \\
Compassionate S       & 0.60 & 0.50 & 0.48 & 0.50 & 0.53 & 0.61 & 0.43 & 0.45 & 0.53 \\
\midrule
\textbf{Solidarity (high-level)}
& \textbf{0.81} & \textbf{0.80} & \textbf{0.78} & \textbf{0.68} & \textbf{0.71} & \textbf{0.78} & \textbf{0.54} & \textbf{0.63} & \textbf{0.67} \\
\midrule
Group-based AS        & 0.47 & 0.51 & 0.49 & 0.42 & 0.46 & 0.43 & 0.33 & 0.42 & 0.32 \\
Exchange-based AS     & 0.43 & 0.49 & 0.48 & 0.38 & 0.39 & 0.52 & 0.31 & 0.25 & 0.10 \\
Empathic AS           & 0.00 & 0.00 & 0.00 & 0.00 & 0.00 & 0.00 & 0.00 & 0.00 & 0.00 \\
Compassionate AS      & 0.20 & 0.00 & 0.06 & 0.18 & 0.07 & 0.21 & 0.18 & 0.13 & 0.18 \\
\midrule
\textbf{Anti-solidarity (high-level)}
& \textbf{0.70} & \textbf{0.69} & \textbf{0.67} & \textbf{0.59} & \textbf{0.56} & \textbf{0.62} & \textbf{0.57} & \textbf{0.57} & \textbf{0.39} \\
\midrule
Mixed                 & 0.32 & 0.46 & 0.35 & 0.24 & 0.48 & 0.04 & 0.26 & 0.30 & 0.26 \\
None                  & 0.66 & 0.71 & 0.69 & 0.64 & 0.66 & 0.68 & 0.64 & 0.64 & 0.62 \\
\bottomrule
\end{tabular}
\caption{F1 scores for selected models on fine-grained and high-level solidarity classification %for \textit{Migrant} (both test sets combined) and \textit{Woman}
on the combined \textit{Migrant} test sets (Test 1 and Test 2). The top section contains fine-grained categories, and the bold rows show performance on high-level classification. Scores are chosen for the best performing configuration for the model in terms of Macro F1.}
\label{tab:f1_solidarity}
\end{table}

%\clearpage
\begin{table}[htbp]
\centering
\footnotesize
\begin{tabular}{lrrrrr}
\toprule
\textbf{Keyword} & \textbf{Solidarity} & \textbf{Anti-Solidarity} & \textbf{Mixed} & \textbf{None} & \textbf{N (statements)} \\
\midrule
Asylsuchende           & 4.11   & 2.06   & 0.48   & 3.35    & \textbf{10} \\
Ausländer              & 291.43 & 118.86 & 21.43  & 326.28  & \textbf{758} \\
Aussiedler             & 118.43 & 12.48  & 15.22  & 25.87   & \textbf{172} \\
Einwanderer            & 2.84   & 3.21   & 0.20   & 19.75   & \textbf{26} \\
Emigranten             & 21.14  & 11.08  & 2.61   & 10.18   & \textbf{45} \\
Flüchtlinge            & 1917.08& 321.69 & 174.29 & 859.94  & \textbf{3273} \\
Heimatvertriebene      & 3158.67& 422.75 & 167.72 & 1006.85 & \textbf{4756} \\
Sowjetzonenflüchtlinge & 308.99 & 27.13  & 2.63   & 138.25  & \textbf{477} \\
Zuwanderer             & 58.24  & 6.54   & 0.86   & 60.36   & \textbf{126} \\
\bottomrule
\end{tabular}
\caption{Distribution of DSL-adjusted expected probability mass for solidarity, anti-solidarity, mixed, and none per keyword included in the post-war trends analysis \Cref{sec:postwar}. Values reflect summed DSL-adjusted expected counts. Totals correspond to the number of keyword-coded statements in the analyzed subset.}
\label{tab:keyword_distribution_postwar}
\end{table}

\begin{table}[htb]
\centering
\footnotesize
\begin{tabular}{lrrrrr}
\toprule
\textbf{Party} & \textbf{Solidarity} & \textbf{Anti-Solidarity} & \textbf{Mixed} & \textbf{None} & \textbf{N (statements)} \\
\midrule
BP       & 134.13 & 54.77  & 67.51  & 46.60   & \textbf{303} \\
CDU/CSU  & 2040.48& 229.98 & 94.93  & 1050.62 & \textbf{3416} \\
DP       & 138.21 & 40.81  & -0.54  & 26.52   & \textbf{205} \\
DRP      & 34.53  & 1.47   & 1.74   & 3.26    & \textbf{41} \\
FDP      & 539.79 & 38.02  & 56.78  & 251.41  & \textbf{886} \\
GB/BHE   & 318.75 & 27.61  & 9.03   & 17.62   & \textbf{373} \\
KPD      & 151.75 & 57.43  & -1.14  & 18.96   & \textbf{227} \\
SPD      & 1595.06& 153.56 & 37.13  & 478.25  & \textbf{2264} \\
WAV      & 191.81 & 19.72  & 37.12  & 15.35   & \textbf{264} \\
Zentrum  & 53.61  & 3.02   & 1.35   & 5.01    & \textbf{63} \\
\bottomrule
\end{tabular}
\caption{Distribution of DSL-adjusted expected probability mass for solidarity, anti-solidarity, mixed, and none per party included in the post-war trends analysis \Cref{sec:postwar}. Values reflect summed DSL-corrected expected counts. Totals correspond to the number of party-coded statements in the analyzed subset.}
\label{tab:party_distribution_postwar}
\end{table}

\begin{table*}[h!]
\centering
\footnotesize
\begin{tabular}{p{0.48\textwidth} p{0.48\textwidth}}
    \toprule
    \textbf{Original Text} & \textbf{Translation} \\
    \specialrule{\lightrulewidth}{0pt}{0pt}

    \errorexamplehead
    {19131205-rtdkr-13.1-183-00456} % example 1 before
    {\textbf{Predictions:}
    \none{none} (\texttt{gpt-oss}, \texttt{Qwen}, ensemble);
    \solidarity{exchange-based solidarity} (\texttt{GPT-5}, \texttt{Llama}).}
    %{\solidarity{exchange-based solidarity}}
    {\solidarity{exchange-based solidarity}; \textbf{human votes:} 3$\times$ \solidarity{exchange-based solidarity}, 1$\times$ \mixed{mixed}}
    {(5 Dec 1913)}
    % humans: 3x s.exchange-based, 1x mixed.none

    \cmidrule(lr){1-2}
    \enquote{
    [...] das Interessante [...] ist, daß nicht nur die östliche Landwirtschaft sondern auch die Industrie Ausländer braucht, daß die Zahlen der industrietätigen Ausländer im Lause der Jahre viel mehr gewachsen sind als die Zahlen der von der Landwirtschaft gebrauchten Arbeiter. [...] \textbf{Das Maß des Bedürfnisses der Beschäftigung von Ausländern ist natürlich auch hier örtlich und nach Betrieben verschieden.} Es gibt gewisse Betriebe, für die inländische Arbeiter kaum noch zu bekommen sind. [...] Es gibt aber auch industrielle Gebiete, die ohne diese Ausländer gar nicht weiter existieren können.
    }
    &
    \enquote{
    [...] what is interesting [...] is that not only eastern agriculture but also industry needs foreigners, and that the number of foreigners employed in industry has grown over the years much more than the number of workers needed in agriculture. [...] \textbf{The extent of the need for the employment of foreigners naturally also varies here by locality and by sector.} There are certain sectors in which domestic workers can scarcely be found anymore. [...] There are also industrial regions that simply could not continue to exist without these foreigners.
    } \\

    %\midrule
    \specialrule{\lightrulewidth}{0pt}{0pt}
    \errorexamplehead
    {20180615-bt-19-040-03954} %
    {\textbf{Predictions:} \mixed{mixed} (\texttt{GPT-5}, \texttt{Llama}), \antisolidarity{group-based anti-solidarity} (\texttt{gpt-oss}, ensemble), \solidarity{exchange-based solidarity} (\texttt{Qwen}).}
    {\mixed{mixed}; \textbf{human votes:} 3$\times$ \mixed{mixed}, 1$\times$ \solidarity{group-based solidarity}}
    {(15 June 2018)}

    %\multicolumn{2}{>{\raggedright\arraybackslash}p{\dimexpr\textwidth-2\tabcolsep\relax}}{\textit{Example (2)%
    %    }: \textbf{Predictions:} \mixed{mixed} (\texttt{GPT-5}, \texttt{Llama}), \antisolidarity{group-based anti-solidarity} (\texttt{gpt-oss}, ensemble), \solidarity{exchange-based solidarity} (\texttt{Qwen}). \textbf{True:} \mixed{mixed} (15 June 2018)} \\ %%20180615-bt-19-040-03954; 
    % humans: 3x s.mixed.none, 1s.group-based
    \cmidrule(lr){1-2}

    \enquote{[...] Die Beschleunigung der Asylverfahren ist überfällig, und eine konsequente Rückführung [...], und [...] eine konsequente Abschiebung. Übrigens sind zügige Verfahren auch und gerade im Interesse der Menschen mit Bleibeperspektive, die wir dann umso besser integrieren können, weil wir ihnen eine Perspektive geben können. Zu guter Letzt müssen wir die Einwanderung nach Deutschland insgesamt neu ordnen, mit einem Einwanderungsgesetz, das diesen Namen verdient. \textbf{Herr Kollege Castellucci, möglicherweise wird umgekehrt ein Schuh daraus: Wer Akzeptanz für legale Migration erhöhen will, der muss auch entschlossen illegale Migration bekämpfen.} [...]}
    &
    \enquote{Speeding up asylum procedures is overdue, as are consistent returns [...] and [...] consistent deportations. Swift procedures are also, and especially, in the interest of people who have a prospect of staying, because we can then integrate them all the better by giving them a perspective. Finally, we must reorganize immigration to Germany as a whole, with an immigration law worthy of the name. \textbf{Mr Castellucci, perhaps the reverse is true: anyone who wants to increase acceptance of legal migration must also resolutely combat illegal migration.} [...]} \\

    \specialrule{\lightrulewidth}{0pt}{0pt}

    \errorexamplehead
    {19950119-bt-13-012-04686} %
    {\textbf{Predictions:}
    \solidarity{compassionate solidarity} (\texttt{gpt-oss}, \texttt{GPT-5}, ensemble), \antisolidarity{group-based anti-solidarity} (\texttt{Llama}%split 3
    ), \antisolidarity{exchange-based anti-solidarity} (\texttt{Qwen}).
    }
    {\solidarity{compassionate solidarity}; \textbf{human votes:} 2$\times$ \solidarity{compassionate solidarity}}
    {(19 Jan 1995)}

    %\multicolumn{2}{>{\raggedright\arraybackslash}p{\dimexpr\textwidth-2\tabcolsep\relax}}{\textit{Example (3)% (10)
    %}: \textbf{Predictions:} \solidarity{compassionate solidarity} (\texttt{gpt-oss}, \texttt{GPT-5}, ensemble), \antisolidarity{group-based anti-solidarity} (\texttt{Llama} %split 3
    %), \antisolidarity{exchange-based anti-solidarity} (\texttt{Qwen}). \textbf{True:} \solidarity{compassionate solidarity} (19 Jan 1995)} \\ %19950119-bt-13-012-04686
    % humans: 2x s.compassionate
    \cmidrule(lr){1-2}

    \enquote{[...] \textbf{Sie [die europäischen Staaten] sind aber nicht bereit, Flüchtlinge vor Menschenrechtsverletzungen zu schützen, sondern diffamieren sie als Schmarotzer und Bedrohung unserer Sicherheit, wie dies leider auch gerade durch den Kollegen der SPD wieder geschehen ist.} Der Vertrag von Maastricht sieht die gemeinsame Einführung eines Visumzwangs vor, wenn auf Grund einer Notlage in einem Land eine größere Fluchtbewegung zu erwarten ist. So wird ganz bewußt verfolgten oder unter sonstigen Notlagen leidenden Menschen die Flucht erheblich erschwert, oft sogar unmöglich gemacht. Flüchtlinge werden damit geradezu in die Arme von Schlepperorganisationen getrieben, über die man sich hier so gerne beschwert.}
    &
    \enquote{[...] \textbf{But they [the European states] are not willing to protect refugees from human rights violations; instead, they defame them as parasites and a threat to our security, as unfortunately just happened again in the remarks of the SPD colleague.} The Maastricht Treaty provides for the joint introduction of a visa requirement if, because of an emergency in a country, a larger refugee movement is to be expected. In this way, it is deliberately made much harder, often even impossible, for people who are persecuted or otherwise suffering hardship to flee. Refugees are thereby virtually driven into the arms of smuggling organizations, about which people here so readily complain.} \\

    \bottomrule
\end{tabular}
\caption{Error analysis examples. Bold text is the main sentence, the other sentences are for context. The reference label is the final human consensus label. Where relevant, we also report the distribution of individual human annotations.}
\label{tab:error-analysis-examples}
\end{table*}

\begin{table*}[h!]
\ContinuedFloat
\centering
\footnotesize
\begin{tabular}{p{0.48\textwidth} p{0.48\textwidth}}
    \toprule
    \textbf{Original Text} & \textbf{Translation} \\
        
        \specialrule{\lightrulewidth}{0pt}{0pt}
        \errorexamplehead
        {20240410-bt-20-162-03269} %
        {\textbf{Predictions:}
       \solidarity{group-based solidarity} (\texttt{gpt-oss}, \texttt{Qwen}), \antisolidarity{group-based anti-solidarity} (\texttt{Llama}).}
        {\solidarity{group-based solidarity}}
        {(10 Apr 2024)}
        %\multicolumn{2}{>{\raggedright\arraybackslash}p{\dimexpr\textwidth-2\tabcolsep\relax}}{\textit{Example (4)% (11)
        %}: \textbf{Predictions:} \solidarity{group-based solidarity} (\texttt{gpt-oss}, \texttt{Qwen}), \antisolidarity{group-based anti-solidarity} (\texttt{Llama}). \textbf{True:} \solidarity{group-based solidarity} (10 Apr 2024)} \\ %20240410-bt-20-162-03269
        % 
        \cmidrule(lr){1-2}

       \enquote{Kommen wir zum Entwurf der AfD und zu dem, was ihn einfach grottenschlecht macht. Sie fordern, dass in allen Fällen, in denen ein Elternteil nichtdeutscher Herkunft ist, die Ausländerbehörde beteiligt wird. Nicht nur stellen Sie damit alle Ausländer/-innen unter Generalverdacht [...], sondern Sie verstoßen damit auch gegen EU-Recht [...]. \textbf{Sie \enquote{übersehen} in Ihrem Entwurf, dass ein Großteil der Ausländer, die hier die Vaterschaftsanerkennung beantragen, längst eine Niederlassungserlaubnis haben.} Aber gut, dass Sie gegen unsere ausländischen Mitbürgerinnen und Mitbürger hetzen. Ist ja wirklich Ihr politisches Tagesgeschäft, so bedauerlich das auch ist. [...]
        }
        &
       \enquote{[...] Let us turn to the AfD's draft and to what makes it simply atrocious. You demand that in all cases where one parent is of non-German origin, the immigration authority must be involved. Not only does this place all foreigners under general suspicion [...], but it also violates EU law [...]. \textbf{You overlook in your draft that a large proportion of the foreigners who apply here for recognition of paternity already have permanent residency.} But of course, it fits your agenda to agitate against our foreign fellow citizens — that is really your daily political business, as regrettable as it is. [...]} \\

       %\midrule
        \specialrule{\lightrulewidth}{0pt}{0pt}
       \errorexamplehead
        {19600505-bt-03-112-00165} %
        {\textbf{Predictions:}
       \solidarity{group-based solidarity} (\texttt{GPT-5}, \texttt{Qwen}); \solidarity{compassionate solidarity} (\texttt{gpt-oss}, \texttt{Llama}, ensemble).}
        {\solidarity{compassionate solidarity}; \textbf{human votes:} 2$\times$ \solidarity{group-based solidarity}, 2$\times$ \solidarity{compassionate solidarity}}
        {(5 May 1960)}

        %\multicolumn{2}{>{\raggedright\arraybackslash}p{\dimexpr\textwidth-2\tabcolsep\relax}}{%
        %\textit{Example (5)% (1)
        %}: \textbf{Predictions:}
        %\solidarity{group-based solidarity} (\texttt{GPT-5}, \texttt{Qwen});
        %\solidarity{compassionate solidarity} (\texttt{gpt-oss-120B}, \texttt{Llama}, ensemble).
        %\textbf{True:} \solidarity{compassionate solidarity} (5 May 1960)
        %} \\ %19600505-bt-03-112-00165
        % humans: 2x s.compassionate; 2x s.group-based, example of a borderline case
        \cmidrule(lr){1-2}
        \enquote{
        [...] Diese Rückführung [der Evakuierten] muß beschleunigt werden, und wir müssen gesetzgeberische Maßnahmen ergreifen, damit die Evakuierten nicht immer wieder auf lange Zeit vertröstet werden, sondern in absehbarer Zeit zu ihren Rechten, zu einer menschenwürdigen Behandlung kommen. \textbf{[...] Gerade das Vertrauen dieser Millionen von Menschen verpflichtet uns, sie auch in den nächsten Jahren nicht zu enttäuschen, [...] diesen Eingliederungsprozeß auf allen Gebieten mit dem größten Nachdruck zu fördern [...].} In der politischen Diskussion hat zwar das Problem der sachlichen, wirtschaftlichen und sozialen Interessen der Vertriebenen und aller Kriegsgeschädigten kaum noch eine Rolle gespielt [...].
        }
        &
        \enquote{
        [...] This return [of evacuees] must be accelerated, and we must take legislative measures so that evacuees are not kept waiting indefinitely time and again, but instead obtain their rights and humane treatment within a foreseeable period of time. \textbf{[...] It is precisely the trust of these millions of people that obliges us not to disappoint them in the years ahead either, [...] and to promote this process of integration in every sphere with the greatest possible determination [...].} In political debate, the problem of the material, economic, and social interests of the expellees and all other victims of war has in fact hardly played any role any longer [...].
        } \\
        
        \specialrule{\lightrulewidth}{0pt}{0pt}
        \errorexamplehead
        {20101028-bt-17-068-06571} %
        {\textbf{Predictions:}
       \solidarity{group-based solidarity} (\texttt{GPT-5}, \texttt{Qwen}, \texttt{gpt-oss}, \texttt{Llama}, ensemble).}
        {\solidarity{compassionate solidarity}; \textbf{human votes}: 4$\times$ \solidarity{compassionate solidarity}}
        {(28 Oct 2010)}
        %\multicolumn{2}{>{\raggedright\arraybackslash}p{\dimexpr\textwidth-2\tabcolsep\relax}}{\textit{Example (6)% (2)
        %}: \textbf{Predictions:} \solidarity{group-based solidarity} (\texttt{GPT-5}, \texttt{Qwen}, \texttt{gpt-oss-120B}, \texttt{Llama}, ensemble). \newline\textbf{True:} \solidarity{compassionate solidarity} (28 Oct 2010)} \\ %20101028-bt-17-068-06571
        % humans: 4x s.compassionate
        \cmidrule(lr){1-2}
        \enquote{Die Bundesrepublik hat [...] einen Vorbehalt zur UN-Kinderrechtskonvention zurückgenommen, der ihr zusichern sollte, dass sie ausländische Kinder schlechter behandeln kann als inländische. [...] \textbf{In den vorliegenden Anträgen [...] ist dargelegt, was alles zu tun wäre: Heraufsetzung der Verfahrensmündigkeit von minderjährigen Flüchtlingen von 16 auf 18 Jahre, Ausbau und Verbesserung der Betreuung unbegleiteter minderjähriger Flüchtlinge durch die Jugendfürsorge, ihre Herausnahme aus dem Flughafenverfahren, keine Abschiebehaft für Minderjährige, keine Unterbringung von Kindern und Jugendlichen in Sammelunterkünften für Asylsuchende und Flüchtlinge.} Unbegleitete Kinder und Jugendliche, deren Eltern nicht ausfindig gemacht werden können, sollen genauso behandelt werden wie elternlose deutsche Kinder. [...]
        }
        &
        \enquote{The Federal Republic has [...] withdrawn a reservation to the UN Convention on the Rights of the Child that had ensured it could treat foreign children worse than German children. [...] \textbf{The motions submitted [...] outline what still needs to be done: raising the age of legal capacity in proceedings for minor refugees from 16 to 18; expanding and improving youth welfare support for unaccompanied minor refugees; removing them from the airport procedure; no detention pending deportation for minors; no accommodation of children and adolescents in collective facilities for asylum seekers and refugees.} Unaccompanied children and adolescents whose parents cannot be located should be treated in the same way as German children without parents. [...]
        } \\
        
        \bottomrule
\end{tabular}
\caption{Error analysis examples (continued). Bold text is the main sentence, the other sentences are for context. The reference label is the final human consensus label. Where relevant, we also report the distribution of individual human annotations.}
\label{tab:error-analysis-examples}
\end{table*}

\begin{table*}[h!]
\ContinuedFloat
\centering
\footnotesize
\begin{tabular}{p{0.48\textwidth} p{0.48\textwidth}}
    \toprule
    \textbf{Original Text} & \textbf{Translation} \\
        
        \specialrule{\lightrulewidth}{0pt}{0pt}

        \errorexamplehead
        {20200116-bt-19-140-01684} %
        {\textbf{Predictions:}
       \solidarity{group-based} (\texttt{gpt-oss}, ensemble), \solidarity{exchange-based} (\texttt{Qwen}, \texttt{GPT-5}) or \solidarity{compassionate solidarity} (\texttt{Llama}).}
        {\solidarity{compassionate solidarity}; \textbf{human votes}: 3$\times$ \solidarity{compassionate solidarity}, 1$\times$ \solidarity{group-based solidarity}}
        {(16 Jan 2020)}

        %\multicolumn{2}{>{\raggedright\arraybackslash}p{\dimexpr\textwidth-2\tabcolsep\relax}}{\textit{Example (7)% (3)
        %}: \textbf{Predictions:} \solidarity{group-based} (\texttt{gpt-oss}, ensemble), \solidarity{exchange-based} (\texttt{Qwen}, \texttt{GPT-5}) or \solidarity{compassionate solidarity} (\texttt{Llama}). \textbf{True:} \solidarity{compassionate solidarity} (16 Jan 2020)} \\ %20200116-bt-19-140-01684; illustrates why humans choose compassionate, but models choose group-based or exchange-based
        % humans: 3x s.compassionate, 1x s.group-based
        \cmidrule(lr){1-2}

        \enquote{Das Hamburger Modell ist geradezu vorbildlich; denn hier werden nicht Menschen mit geringem Einkommen und Geflüchtete gegeneinandergestellt, sondern alle haben den gleichen Anspruch auf Mobilität, mit dem Unterschied, dass die Geflüchteten das von ihrem Taschengeld zwangsweise abgezogen bekommen. [...] \textbf{\enquote{Wir sind auf Migration angewiesen, aber die Flüchtlinge...}, könnte man sagen}. [...] Ich bin so erzogen: Wenn jemand Hilfe braucht und du kannst helfen, dann hilf. Ich bin stolz und froh, dass unser Land helfen kann [...].}
        &
        \enquote{The Hamburg model is virtually exemplary, because here people with low incomes and refugees are not being pitted against one another; rather, everyone has the same entitlement to mobility, with the difference that refugees have it forcibly deducted from their allowance. [...] \textbf{One might say, \enquote{We depend on migration, but the refugees...}.} [...] I was raised to believe this: if someone needs help and you can help, then help. I am proud and glad that our country is able to help [...].} \\

        \specialrule{\lightrulewidth}{0pt}{0pt}

        \errorexamplehead
        {19650212-bt-04-163-01657} %
        {\textbf{Predictions:}
       \solidarity{group-based solidarity} (\texttt{gpt-oss}, \texttt{Qwen}, \texttt{Llama}, ensemble), \solidarity{empathic solidarity} (\texttt{GPT-5}).}
        {\solidarity{empathic solidarity}; \textbf{human votes}: 3$\times$ \solidarity{empathic solidarity}, 1$\times$ \solidarity{group-based solidarity}}
        {(12 Feb 1965)}

        %\multicolumn{2}{>{\raggedright\arraybackslash}p{\dimexpr\textwidth-2\tabcolsep\relax}}{\textit{Example (8)% (5)
        %}: \textbf{Predictions:} \solidarity{group-based solidarity} (\texttt{gpt-oss}, \texttt{Qwen}, \texttt{Llama}, ensemble), \solidarity{empathic solidarity} (\texttt{GPT-5}). \textbf{True:} \solidarity{empathic solidarity} (12 Feb 1965)} \\ %19650212-bt-04-163-01657; illustrates why humans choose s.empathic but models choose s.group-based
        % humans: 3x s.empathic, 1x s.group-based
        \cmidrule(lr){1-2}

        \enquote{[...] Die große Aufgabe, die uns hier gesetzt war, war der Versuch, einen Kompromiß [...] zwischen den Interessen der Ausländer in der Bundesrepublik Deutschland und den berechtigten Ansprüchen der deutschen Bevölkerung auf Aufrechterhaltung der öffentlichen Ordnung und Sicherheit zu finden. \textbf{Das Gesetz ist ein Ausdruck unserer freiheitlichen demokratischen Grundordnung, indem es den Ausländern in der Bundesrepublik ein sehr hohes Maß von Entfaltungsmöglichkeiten gewährt.} [...] Das Gesetz nimmt auch besonders Bedacht auf eine Verbesserung der Rechtslage der ausländischen Flüchtlinge [...], die ihre Heimat unter dem Druck totalitärer Systeme verlassen mußten. [...]}
        &
        \enquote{[...] The major task before us was to try to find a compromise [...] between the interests of foreigners in the Federal Republic of Germany and the legitimate claims of the German population to the maintenance of public order and security. \textbf{This law is an expression of our liberal democratic constitutional order in that it grants foreigners in the Federal Republic a very high degree of scope for self-development.} [...] The law also gives special consideration to improving the legal position of foreign refugees [...] who were forced to leave their homeland under pressure from totalitarian systems. [...]} \\

        \specialrule{\lightrulewidth}{0pt}{0pt}

        \errorexamplehead
        {20170601-bt-18-237-02406} %
        {\textbf{Predictions:}
       \antisolidarity{exchange-based anti-solidarity} (\texttt{gpt-oss}, \texttt{Qwen}, \texttt{GPT-5}, ensemble), \antisolidarity{group-based anti-solidarity} (\texttt{Llama}).}
        {\antisolidarity{group-based anti-solidarity}; \textbf{human votes}: 2$\times$ \antisolidarity{group-based anti-solidarity}, 2$\times$ \antisolidarity{exchange-based anti-solidarity}}
        {(1 June 2017)}

        %\multicolumn{2}{>{\raggedright\arraybackslash}p{\dimexpr\textwidth-2\tabcolsep\relax}}{\textit{Example (10)%(6)
        %}: \textbf{Predictions:} \antisolidarity{exchange-based anti-solidarity} (\texttt{gpt-oss}, \texttt{Qwen}, \texttt{GPT-5}, ensemble), \antisolidarity{group-based anti-solidarity} (\texttt{Llama}). \textbf{True:} \antisolidarity{group-based anti-solidarity} (1 June 2017)} \\ %20170601-bt-18-237-02406; illustrates group-based vs exchange-basedas confusion
        % humans: 2x as.group-based, 2x as.exchange-based
        \cmidrule(lr){1-2}

        \enquote{[...] \textbf{Mit diesem auf Angebotsorientierung im Migrationsrecht zielenden Gesetzentwurf wollen die Grünen im Kern nichts anderes, als die Einwanderung massiv vereinfachen und die Tore weit aufstoßen.} Wir als Unionsfraktion stellen uns gegen die Beliebigkeit im Einwanderungsrecht und stehen daher auch klar für eine gezielte Auswahl bei der Migration ein. Der Staat und die Politik haben die Pflicht, bewusst darauf zu achten, dass die Migration für unser Land vor allem einen Mehrwert und keine Belastung darstellt, und genau darin liegt der Unterschied zum Asyl- und Flüchtlingsrecht. [...] }
        &
        \enquote{[...] \textbf{With this bill, which is aimed at a supply-oriented approach to migration law, the Greens essentially want nothing other than to make immigration much easier and throw the gates wide open.} As the CDU/CSU parliamentary group, we oppose arbitrariness in immigration law and therefore clearly support a targeted selection in migration. The state and policymakers have a duty to ensure consciously that migration represents above all an added value for our country and not a burden, and that is precisely the difference from asylum and refugee law. [...]} \\

        \bottomrule
\end{tabular}
\caption{Error analysis examples (continued). Bold text is the main sentence, the other sentences are for context. The reference label is the final human consensus label. Where relevant, we also report the distribution of individual human annotations.}
\label{tab:error-analysis-examples}
\end{table*}

\begin{table*}[h!]
\ContinuedFloat
\centering
\footnotesize
\begin{tabular}{p{0.50\textwidth} p{0.46\textwidth}}
    \toprule
    \textbf{Original Text} & \textbf{Translation} \\

    \specialrule{\lightrulewidth}{0pt}{0pt}

        \errorexamplehead
        {20190517-bt-19-102-03948} %
        {\textbf{Predictions:}
       \antisolidarity{group-based anti-solidarity} (\texttt{gpt-oss}, \texttt{Qwen}, \texttt{GPT-5}, \texttt{Llama}, ensemble).}
        {\antisolidarity{empathic anti-solidarity}; \textbf{human votes}: 2$\times$ \antisolidarity{group-based anti-solidarity}, 2$\times$ \antisolidarity{empathic anti-solidarity}}
        {(17 May 2019)}

        %\multicolumn{2}{>{\raggedright\arraybackslash}p{\dimexpr\textwidth-2\tabcolsep\relax}}{\textit{Example (9)% (7)
        %}: \textbf{Predictions:} \antisolidarity{group-based anti-solidarity} (\texttt{gpt-oss}, \texttt{Qwen}, \texttt{GPT-5}, \texttt{Llama}, ensemble). \textbf{True:} \antisolidarity{empathic anti-solidarity} (17 May 2019)} \\ %20190517-bt-19-102-03948; illustrates as.empathic vs as.group-based confusion
        % humans: 2x as.group-based, 2x as.empathic
        \cmidrule(lr){1-2}

        \enquote{52 Prozent der britischen Muslime sagen nicht nur: \enquote{Homosexualität ist moralisch falsch}, sondern: Homosexualität sollte verboten werden. [...] %London ist heute schon da, wo Berlin, das Ruhrgebiet und Köln bald sein werden. 
        \textbf{%Nach Berechnungen des Pew Research Centers wird sich 
        bei fortgesetzter Einwanderung [wird sich] der Anteil der Muslime in Deutschland in den kommenden Jahren auf 20 Prozent erhöhen.} In vielen Städten und Gemeinden sind sie dann die Mehrheit. Der Islam kennt keine Minderheitenrechte. Das sind die dunklen Wolken, die über den Homosexuellen in Westeuropa aufziehen.}
        &
        \enquote{52 percent of British Muslims say not only, \enquote{Homosexuality is morally wrong}, but also: homosexuality should be banned. [...] %London is already today where Berlin, the Ruhr area, and Cologne will soon be. 
        \textbf{if immigration continues, the share of Muslims in Germany will rise to 20 percent in the coming years.} In many cities and municipalities, they will then be the majority. Islam recognizes no minority rights. These are the dark clouds gathering over homosexuals in Western Europe.} \\

        \specialrule{\lightrulewidth}{0pt}{0pt}
        \errorexamplehead
        {18680416-rtndb-02-008-00361} %
        {\textbf{Predictions:}
       \none{none} (\texttt{gpt-oss}, \texttt{Qwen}, \texttt{GPT-5}, ensemble), \antisolidarity{group-based anti-solidarity} (\texttt{Llama}).}
        {\solidarity{group-based solidarity}; \textbf{human votes}: 2$\times$ \solidarity{group-based solidarity}, 1$\times$ \antisolidarity{group-based anti-solidarity}}
        {(16 Apr 1868)}
        %\multicolumn{2}{>{\raggedright\arraybackslash}p{\dimexpr\textwidth-2\tabcolsep\relax}}{\textit{Example (11)% (8)
        %}: \textbf{Predictions:} \none{none} (\texttt{gpt-oss}, \texttt{Qwen}, \texttt{GPT-5}, ensemble), \antisolidarity{group-based anti-solidarity} (\texttt{Llama}). \textbf{True:} \solidarity{group-based solidarity} (16 Apr 1868)} \\ %18680416-rtndb-02-008-00361
        % humans: 2x s.group-based 1x as.group-based. Confusing case because the speaker criticizes the law, but more subtly than in modern speeches, with less emotional undertone
        \cmidrule(lr){1-2}

        \enquote{Ausländer bleibt unerachtet solchen Aufenthalts innerhalb der Mecklenburgischen Lande immer Ausländer, und einem solchen die Trauung innerhalb der diesseitigen Lande zu gestatten, ist allein das Großherzogliche Ministerium des Innern competent. [...] \textbf{Vielmehr muß in solchen Fällen der Wiederverheirathung der Mecklenburger sein Recht zur Verheiratung durch einen gesetzlichen Domizilschein über das ihm für einen Mecklenburgischen Ort ertheilte Recht der Niederlassung und Verheiratung nachweisen und der Ausländer den Konsens des Großherzoglichen Ministeriums des Innern beibringen, bevor die Trauung geschehen darf.} Ja, meine Herren, da sehen Sie nun, was in Mecklenburg das Indigenat aus dem Artikel 2 der Bundesverfassung für eine Bedeutung hat, und wie nothwendig es ist, daß irgend eine Ergänzung des Freizügigkeitsgesetzes [...] erfolge, damit das Gesetz dort nicht blos eine Schale ohne Kern bleibe.}
        &
        \enquote{A foreigner remains a foreigner within the lands of Mecklenburg despite such residence, and only the Grand Ducal Ministry of the Interior is competent to permit such a person to marry within these territories. [...] \textbf{Rather, in such cases of remarriage, the Mecklenburger must prove his right to marry by means of a lawful certificate of domicile showing that he has been granted the right of residence and marriage in a Mecklenburg locality, and the foreigner must produce the consent of the Grand Ducal Ministry of the Interior before the marriage may take place.} Gentlemen, there you can see what significance indigeneity under Article 2 of the federal constitution has in Mecklenburg, and how necessary it is that some supplementation of the Freedom of Movement Act [...] should be made, so that the law there may not remain merely a shell without a kernel.} \\

        \specialrule{\lightrulewidth}{0pt}{0pt}
        \errorexamplehead
        {18860115-rtdkr-06.2-025-00082} %
        {\textbf{Predictions:}
       \none{none} (\texttt{gpt-oss}, \texttt{Qwen}, \texttt{Llama}, ensemble), \mixed{mixed} (\texttt{GPT-5}).}
        {\solidarity{group-based solidarity}; \textbf{human votes}: 2$\times$ \solidarity{group-based solidarity}, 1$\times$ \none{none}, 1$\times$ \mixed{mixed}}
        {(15 Jan 1886)}
        %\multicolumn{2}{>{\raggedright\arraybackslash}p{\dimexpr\textwidth-2\tabcolsep\relax}}{\textit{Example (12)% (9)
        %}: \textbf{Predictions:} \none{none} (\texttt{gpt-oss}, \texttt{Qwen}, \texttt{Llama}, ensemble), \mixed{mixed} (\texttt{GPT-5}). \newline\textbf{True:} \solidarity{group-based solidarity} (15 Jan 1886)} \\ %18860115-rtdkr-06.2-025-00082
        % humans: 2x s.group-based 1x none 1x mixed
        \cmidrule(lr){1-2}

        \enquote{Wann und warum ein Ausländer ohne ein vorangegangenes gerichtliches Erkenntniß aus dem Reich oder einem Partikularstaat ausgewiesen werden darf, darüber existiren zwar keine gesetzlichen Normen und Bestimmungen [...]. Rechtfertigende Gründe der Ausweisung eines Ausländers nach dem allgemein anerkannten Völkerrechte sind folgende: erstens Gefährdung der öffentlichen Sicherheit durch ein feindliches Auftreten des Ausgewiesenen [...]; zweitens [...]; drittens[...], und viertens [...]. \textbf{Damit sind alle Fälle erschöpft, in welchen nach dem modernen Völkerrecht Ausweisungen von einzelnen Personen stattfinden dürfen, und alle Lehrer des Völkerrechts sind um so vorsichtiger in der Empfehlung auch nur eines dieser Mittel, als sich das moderne Völkerrecht immer mehr der Theorie zuwendet, daß ein domizilirter Ausländer zu den wirklichen Staatsangehörigen gehört und als solcher zu behandeln ist.} [...]}
        &
        \enquote{Although there are no legal norms or provisions governing when and why a foreigner may be expelled from the Empire or from an individual state without prior judicial judgment [...], the reasons justifying the expulsion of a foreigner under generally recognized international law are the following: first, endangerment of public security through hostile conduct on the part of the person expelled [...]; second [...]; third [...]; and fourth [...]. \textbf{These exhaust all the cases in which, under modern international law, expulsions of individual persons may take place, and all teachers of international law are all the more cautious in recommending even one of these measures because modern international law is increasingly turning toward the theory that a domiciled foreigner belongs among the true members of the state and is to be treated as such.} [...]} \\

        \bottomrule
\end{tabular}
\caption{Error analysis examples (continued). Bold text is the main sentence, the other sentences are for context. The reference label is the final human consensus label. Where relevant, we also report the distribution of individual human annotations.}
\label{tab:error-analysis-examples}
\end{table*}

\begin{table*}[t]
    \centering
    {\footnotesize
    \begin{tabularx}{\textwidth}{>{\hsize=.4\hsize\raggedright\arraybackslash}X >{\hsize=1.43\hsize}X >{\hsize=1.17\hsize}X}
        \toprule
        \textbf{} & \textbf{Original Text} & \textbf{Translation} \\
        \midrule
        \setcounter{examplectr}{0}
        \examplerow
        {19520514-bt-01-211-00608}
        {Dominant DSL subtype score: group-based solidarity towards expellees}
        {(May 14, 1952)}
        %(1) \textbf{Dominant DSL subtype score: group-based solidarity towards expellees}\newline{}(May 14, 1952) %19520514-bt-01-211-00608
        {\enquote{[...] das Lastenausgleichsgesetz und die Eingliederung der Vertriebenen [steht] in keinem Widerspruch zu unserm Anspruch und unserm Recht auf die Heimat. \textbf{Im Gegenteil, %wir sind der Überzeugung, 
        [...] daß die Rücksiedlung [...] nur durchgeführt werden kann, wenn die körperliche, seelische und auch die wirtschaftliche Substanz der Vertriebenen erhalten geblieben ist, und das kann nur geschehen, wenn sie in die westdeutsche Wirtschaft eingegliedert werden.} [...] Wir sind durchaus für die Ausgestaltung des Lastenausgleichs nach sozialen Gesichtspunkten [...]. %aber der Rechtsanspruch darf nicht so behandelt werden, als sei er nur etwas Zweitrangiges, angewiesen auf das, was übrigbleibt.
        }}
        {\enquote{[...] the Equalization of Burdens Act and the integration of the expellees does not contradict our right to the homeland. \textbf{On the contrary, resettlement [...] can only be achieved if the physical, mental, and economic substance of the expellees has been preserved, and that can only happen if they are integrated into the West German economy.} [...] We are fully in favor of shaping the Equalization of Burdens in accordance with social considerations [...]. %but the legal entitlement should not be treated as %something 
        %secondary, dependent only on what is left over.
        }}
        \midrule
        \examplerow
        {19530225-bt-01-250-01231}
        {Dominant DSL subtype score: group-based solidarity towards expellees}
        {(Feb 25, 1953)}
        %(2) \textbf{\ak{Dominant DSL subtype score: group-based solidarity} towards expellees}\newline{}(Feb 25, 1953) %19530225-bt-01-250-01231
       {\enquote{[...] \textbf{Der Ihnen vorliegende Entwurf [zum Bundesvertriebenengesetz] geht [...] von der Zielsetzung aus, die [...] Leistungskraft der Vertriebenen und Flüchtlinge [...] im Rahmen der Gesamtwirtschaft zur Geltung zu bringen.} Die Eingliederungsbestimmungen sollen der Erfüllung berechtigter Ausgleichsansprüche hinsichtlich der verlorenen Existenz dienen [...] mit dem Ziel, daß sich das gesamte Arbeitspotential, das der Einheimischen und der Vertriebenen, so betätigen kann, daß eine größtmögliche Wirkung erzielt wird.}}
        {\enquote{[...] \textbf{The draft [on the Federal Expellees Act] before you starts from the objective [...] of making [...] productive capacities of the expellees and refugees [...] effective within the overall economy.} The integration provisions are intended to meet legitimate compensation claims with regard to lost livelihoods [...] with the goal that the entire labor potential, that of natives and of expellees alike, can be deployed in such a way that the greatest possible effect is achieved.}}
        \midrule
        \examplerow
        {19550615-bt-02-087-02450}
        {Dominant DSL subtype score: compassionate solidarity towards Soviet-zone refugees}
        {(June 15, 1955)}
        %(3) \textbf{Dominant DSL subtype score: compassionate solidarity towards Soviet-zone refugees}\newline{}(June 15, 1955) %19550615-bt-02-087-02450
        {\enquote{[...] \textbf{%im Rahmen der Haushaltsdebatte 
        es ist notwendig, zu den noch immer vielseitigen Problemen der Sowjetzonenflüchtlinge Stellung zu nehmen [...].} Wir erkennen selbstverständlich an, daß durch die Neuregelung des Härtefonds im Lastenausgleich eine Verbesserung für die Sowjetzonenflüchtlinge erfolgte, weil die Ermessensfragen, unter denen dieser Personenkreis bislang besonders zu leiden hatte, dadurch eingeschränkt werden. [...] Wenn wir aber die Sicherheit haben wollen, daß sich diese Neuregelung auch wirklich positiv für den Flüchtling auswirkt, so ist es notwendig, daß die vorgesehene Erhöhung des Härtefonds %auf 150 Millionen DM 
        [...] wirklich realisiert wird.}}
        {\enquote{[...] \textbf{it is necessary to address the still manifold problems of the Soviet-zone refugees %in the context of the budget debate 
        [...].} We naturally recognize that the reform of the hardship fund in the Equalization of Burdens has brought improvements [...] because the discretionary decisions that this group had particularly suffered from have been curtailed. But if we want to be sure that this reform really has a positive effect for the refugee, then [...] the planned increase of the hardship fund %to 150 million DM 
        [...] must actually be implemented.}}
        \midrule
        \examplerow
        {19501213-bt-01-106-00058}
        {Dominant DSL subtype score: group-based anti-solidarity towards expellees}
        {(Dec 13, 1950)}
        %(4) \textbf{Dominant DSL subtype score: group-based anti-solidarity towards expellees}\newline{}(Dec 13, 1950) %19501213-bt-01-106-00058
        {\enquote{In unseren Landkreisverwaltungen sind heute 42,2\% Heimatvertriebene. [...] Auch die einheimische Bevölkerung beklagt sich [...] man höre auf den Landratsämtern überwiegend ostpreußisch, und das Plattdeutsche sei zur Seite gedrängt. Unter den Lehrern [...] 51\% Heimatvertriebene, unter den vollbeschäftigten Richtern [...] mehr als 50\%, unter der Polizei sogar 61\%; zweifellos eine Überfremdung, die Besorgnis erregt! \textbf{Aber all diese ehrlichen Bemühungen um Eingliederung der Heimatvertriebenen sind angesichts der ungeheuren Zahl vergeblich.} Arbeitslosenzahl [...] beträgt unter den Heimatvertriebenen fast 60\%. Die Lasten sind unerschwinglich.}}
        {\enquote{In our district administrations today, 42.2\% are expellees. The native population [...] complains [...] that in the district offices one mostly hears East Prussian, and that Low German has been pushed aside. Among the teachers 51\% are expellees, among the full-time judges also more than 50\%, among the police even 61\% -- undoubtedly an over-foreignization that arouses concern! \textbf{But all these honest efforts to integrate the expellees are in vain in view of their immense numbers.} Unemployment [...] almost 60\% among the expellees [...]. The burdens are unbearable.}}
        \bottomrule
    \end{tabularx}}
    \caption{Examples for postwar period discussed in \Cref{sec:postwar}. Bold text is the main sentence, the other sentences are for context. The label shown is the instance's dominant DSL subtype, defined as the subtype with the highest DSL-adjusted score.}
    \label{tab:postwar-predictions-examples}
\end{table*}

\begin{table*}[t]
    \ContinuedFloat
    \centering
    {\footnotesize
    \begin{tabularx}{\textwidth}{>{\hsize=.43\hsize\raggedright\arraybackslash}X >{\hsize=1.41\hsize}X >{\hsize=1.15\hsize}X}
        \toprule
        \textbf{} & \textbf{Original Text} & \textbf{Translation} \\
        \midrule
        \examplerow
        {19540923-bt-02-044-01194}
        {Dominant DSL subtype score: group-based/compass. solidarity towards Soviet-zone refugees}
        {(Sept 23, 1954)}
        %(5) \textbf{\ak{Dominant DSL subtype score: group-based/\newline compassionate} solidarity towards Soviet-zone refugees}\newline{}(Sept 23, 1954) %19540923-bt-02-044-01194
        {\enquote{[...] [Sowjetzonenflüchtlinge] meinen, es könnte ihnen sonst bei mangelnder Bereitwilligkeit der Bevölkerung zu neuen Opfern vielleicht nicht einmal mehr das zur Verfügung stehen, was ihnen heute aus dem Härtefonds zur Verfügung steht. [...] \textbf{Wir halten es für unvertretbar, daß [...] diese nach wie vor auf den genannten, im Rahmen des Gesamtlastenausgleichs verhältnismäßig geringen Betrag ohne Rechtsanspruch angewiesen sein sollen.} [Die], deren Urstatus dem der Heimatvertriebenen in Westdeutschland doch gleich ist, im Lastenausgleich diesen Heimatvertriebenen nicht gleichgestellt sind.}}
        {\enquote{[...] [Soviet-zone refugees] fear that, if public willingness for further sacrifices declines, they may no longer receive even the support currently provided through the hardship fund [...]. \textbf{[It is] indefensible that [...] they should still be dependent on the aforementioned, comparatively small amount within the overall equalization scheme without any legal entitlement.} [...] those whose original legal status is equivalent to that of expellees in West Germany are not treated equally under the Equalization of Burdens system.}}
        \midrule
        \examplerow
        {19560621-bt-02-151-01596}
        {Dominant DSL subtype score: group-based anti-solidarity towards foreigners}
        {(June 21, 1956)}
        %(6) \textbf{Dominant DSL subtype score: group-based anti-solidarity towards foreigners}\newline{}(June 21, 1956) %19560621-bt-02-151-01596
        {\enquote{[...] Wir müssen das Bundesgebiet nicht nur gegen politisch subversive Elemente, \textbf{wir müssen uns auch gegen kriminelle Elemente abschirmen}. [...] \textbf{die Bundesrepublik [...] [hat] Anlaß, sich gegen Zehntausende von heimatlosen Ausländern abzuschirmen, die [...] in der Absicht [kommen], bei uns wirtschaftliche Versorgung oder politisches Asyl zu finden.} [...] alle internationalen Verpflichtungen, die wir [...] erfüllen, [gehören] selbstverständlich nicht unter das, was ich hier ausgeführt habe.}}
        {\enquote{[...] We must shield the Federal territory not only against politically subversive elements, \textbf{we but also against criminal elements}. [...] \textbf{the Federal Republic [...] has reason to shield itself against [...] stateless foreigners who [...] come with the intention of seeking economic support or political asylum here.} [...] all international obligations which we [...] fulfill, of course, do not fall under what I have set out here.}}
        \midrule
        \examplerow
        {19501018-bt-01-092-00330}
        {Dominant DSL subtype score: group-based solidarity towards foreigners}
        {(Oct 18, 1950)}
        %(7) \textbf{Dominant DSL subtype score: group-based solidarity towards foreigners}\newline{}(Oct 18, 1950) %19501018-bt-01-092-00330
        {\enquote{[...] \textbf{[HAuslG] [...] [wird] den Besatzungsmächten den Entschluß erleichtern, der Bundesrepublik die volle Verantwortung für die heimatlosen Ausländer zu übertragen.} Wenn [...] das deutsche Volk seine Bereitschaft bekundet, die Eingliederung der heimatlosen Ausländer in die deutsche Betreuung [...] zu erleichtern, so gehen wir dabei von der Erwartung aus, daß auch die heimatlosen Ausländer bemüht sein werden, sich in die nun einmal gegebene Lage einzufügen. Sie werden sich dabei bewußt sein müssen, daß auch die Lage der deutschen Bevölkerung schwierig ist und daß viele Zehntausende insbesondere von Heimatvertriebenen noch immer arbeitslos sind und über keinen angemessenen Wohnraum verfügen. [...] der heimatlose Ausländer [wird] viele der Härten dieser Nachkriegszeit mit den Deutschen im Gebiete der Bundesrepublik teilen müssen.
        }}
        {\enquote{
        [...] \textbf{[HAuslG] [...] will make it easier for the occupying powers to decide to transfer full responsibility for stateless foreigners to the Federal Republic.} If [...] the German people declare their willingness to facilitate [...] the integration of stateless foreigners into German administration and care, we proceed on the assumption that the stateless foreigners will also strive to adapt to the situation as it now stands. They must remain aware that the situation of the German population is also difficult and that many tens of thousands, especially expellees, remain unemployed and lack adequate housing. [...] stateless foreigners will have to share many of the hardships of this postwar period with the Germans in the territory of the Federal Republic.
        }}
        \midrule
        \examplerow
        {19501018-bt-01-092-00327}
        {Dominant DSL subtype score: group-based solidarity towards foreigners}
        {(Oct 18, 1950)}
        %(7.1) \textbf{Dominant DSL subtype score: group-based solidarity towards foreigners}\newline{}(Oct 18, 1950) %19501018-bt-01-092-00327
        {\enquote{[...] Auch in der öffentlichen Fürsorge sollen sie Leistungen in gleicher Höhe erhalten wie deutsche Staatsangehörige. Es soll ihnen Zutritt zu Schulen aller Art zu den gleichen Bedingungen... gewährt werden. [...] die Bundesrepublik [gewährt] ihnen einen Rechtsschutz, dessen sie bedürfen, weil sie nicht mehr den Schutz ihres Heimatstaates genießen [...]. \textbf{Der Gesetzentwurf geht [...] über den Konventionsentwurf, dessen Grundsätze Mindestforderungen für die Behandlung heimatloser Ausländer darstellen, hinaus. [...]}}}
        {\enquote{[...] In public welfare too, they are to receive benefits equal to those of German citizens. They are to be granted access to schools of all kinds under the same conditions [...]. the Federal Republic shall grant them legal protection, which they need because they no longer enjoy the protection of their home state. \textbf{The law goes [...] beyond the draft convention, whose principles set minimum standards for the treatment of stateless foreigners. [...]}}}
        \bottomrule
    \end{tabularx}}
    \caption{Examples for postwar period discussed in \Cref{sec:postwar} (continued). Bold text is the main sentence, the other sentences are for context. The label shown is the instance's dominant DSL subtype, defined as the subtype with the highest DSL-adjusted score.}
    \label{tab:postwar-predictions-examples}
\end{table*}

\begin{table*}[t]
    \ContinuedFloat
    \centering
    {\footnotesize
    \begin{tabularx}{\textwidth}{>{\hsize=.5\hsize\raggedright\arraybackslash}X >{\hsize=1.3\hsize}X >{\hsize=1.2\hsize}X}
        \toprule
        \textbf{} & \textbf{Original Text} & \textbf{Translation} \\
        \midrule
        \examplerow
        {19510419-bt-01-136-01759}
        {Dominant DSL subtype score: exchange-based solidarity towards foreigners}
        {(April 19, 1951)}
        %(8) \textbf{Dominant DSL subtype score: exchange-based solidarity towards foreigners}\newline{}(April 19, 1951) %19510419-bt-01-136-01759
        {\enquote{[...] \textbf{Wir begrüßen die Verabschiedung des Gesetzes für heimatlose Ausländer, weil uns damit die Möglichkeit geboten war, dem Ausland zu beweisen, wie wir Angehörige anderer Nationen in unserem eigenen Staate zu behandeln gewillt sind, und weil wir uns [...] erhoffen, daß sie [...] dazu beitragen wird, die Frage der Internationalisierung unseres Flüchtlingsproblems und des Zuflusses von Mitteln aus dem Ausland zu fördern.} [...]}}
        {\enquote{[...] \textbf{We welcome the adoption of the law for stateless foreigners, because it gave us the opportunity to demonstrate to foreign countries how we intend to treat nationals of other nations within our own state, and because we [...] hope that it will make a contribution to promoting the internationalization of our refugee problem and the inflow of funds from abroad.} [...]}}
        \midrule
        \examplerow
        {19550615-bt-02-087-03279}
        {Dominant DSL subtype score: group-based solidarity towards foreigners}
        {(June 15, 1955)}
        %(9) \textbf{\ak{Dominant DSL subtype score: group-based} solidarity towards foreigners}\newline{}(June 15, 1955) %19550615-bt-02-087-03279
        {\enquote{[...] \textbf{Es ist kein Zweifel, daß die heimatlosen Ausländer natürlich vom Arbeitsministerium und vom Innenministerium [...] eine rein materielle Unterstützung bekommen.} Aber diese materielle Unterstützung gibt keine Möglichkeit, ihnen kulturell oder sonstwie unter die Arme zu greifen. [...] [Es ist sehr schwer] diese Dinge heute den Ländern nahezulegen.}}
        {\enquote{[...] \textbf{There is no doubt that the stateless foreigners, of course, receive [...] purely material support from the Ministry of Labor and the Ministry of the Interior.} But this material support provides no opportunity to assist them culturally or in other ways. [...] [It is very difficult] today to persuade the Länder on these matters. [...]}}
        \midrule
        \examplerow
        {19491020-bt-01-012-01359}
        {Dominant DSL subtype score: exchange-based solidarity towards expellees,\newline\newline Party: FDP}
        {(Oct 20, 1949)}
        %(10) \textbf{Dominant DSL subtype score: exchange-based solidarity towards expellees,\newline\newline Party: FDP}\newline{}(Oct 20, 1949) %19491020-bt-01-012-01359
        {\enquote{Die werden [...] von früh morgens bis spät abends arbeiten, wenn man ihnen das nötige Gelände, [Baumaterialien oder das Geld] zur Verfügung stellt. So werden sie in Gemeinschaftsarbeit wesentlich mehr leisten, als wir das sonst heute bei diesen Bauten sehen. In diesem Falle wäre das Geld, das wir dafür aufnehmen, praktisch nur als Kredit anzusehen. \textbf{Wir hätten die Möglichkeit, daß diese Heimatvertriebenen sich ihr Heim selbst bauen, daß es allmählich ihr Eigentum wird.} Wenn wir nur normale Mietsätze verrechnen und das Kapital ohne Zinsen oder zu geringen Zinssätzen verzinsen lassen, so wäre in wenigen, spätestens 15 Jahren, dieses Haus [...] Eigentum des Betreffenden. %Wir würden dadurch noch einen Vorteil haben. Die neu erstandenen Häuser könnten vorerst Eigentum der Genossenschaft bleiben und zur Sicherung des Kredites gegenüber dem Geldgeber hypothekarisch belastet werden.
        }}
        {\enquote{They would [...] work from early morning until late at night if they were provided with the necessary land, [building materials or the money]. In cooperative work they would achieve far more than we otherwise see with such construction today. \textbf{In this case the money we raise would practically only be regarded as a loan.} We would have the possibility that these expellees build their homes themselves, gradually becoming their property. If we only charged normal rents and let the capital bear no or low interest, then within at most fifteen years this house [...] would become the property of the person concerned.}}
        \midrule
        \examplerow
        {19531029-bt-02-005-01150}
        {Dominant DSL subtype score: group-based/\newline compassionate solidarity towards expellees,\newline\newline Party: GB/BHE}
        {(Oct 29, 1953)}
        %(11) \textbf{Dominant DSL subtype score: group-based/\newline compassionate solidarity towards expellees,\newline\newline Party: GB/BHE}\newline{}(Oct 29, 1953) %19531029-bt-02-005-01150
        {\enquote{%Wir können hier nicht allein nach fiskalischen Gesichtspunkten vorgehen. Wir müssen dieses Problem auch von einem allgemein menschlichen Standpunkt aus sehen. Es ist unmöglich, ein Volk sozial zu einen und zu befrieden, wenn man einen großen Teil des Volkes von der Entwicklung ausnimmt. 
        [...] \textbf{Wir dürfen die Augen auch nicht vor der Tatsache verschließen, daß Millionen von Vertriebenen und Kriegsgeschädigten noch nicht den Anschluß an unsere soziale Entwicklung gefunden haben.} Auch hier mußten bisher Zehntausende arbeitswilliger tüchtiger Mitbürger abseitsstehen. Bei unserem Verlangen nach beschleunigter Eingliederung dieser Bevölkerungskreise geht es nicht nur um die Ablösung einer unproduktiven Fürsorge durch aufbauende Mitarbeit. Es geht auch nicht nur um die Erfüllung von \enquote{Rechtsansprüchen}, und ich darf diesen Ausdruck Rechtsansprüche hier noch einmal betonen.
        }}
        {\enquote{%We cannot approach this problem solely from fiscal considerations. We must also look at it from a general human standpoint. It is impossible to unite and reconcile a people socially if a large part of the population is excluded from development. 
        [...] \textbf{We must not close our eyes to the fact that millions of expellees and war victims have still not been integrated into our social development.} Here too, tens of thousands of willing and capable fellow citizens have so far had to stand aside. In our demand for the accelerated integration of these groups, it is not only a matter of replacing unproductive welfare with constructive participation. It is also not only a matter of fulfilling legal claims — and I would like to emphasize this expression \enquote{legal claims} once again.}}
        \bottomrule
    \end{tabularx}}
    \caption{Examples for postwar period discussed in \Cref{sec:postwar} (continued). Bold text is the main sentence, the other sentences are for context. The label shown is the instance's dominant DSL subtype, defined as the subtype with the highest DSL-adjusted score.}
    \label{tab:postwar-predictions-examples}
\end{table*}

\begin{table*}[t]
    \ContinuedFloat
    \centering
    {\footnotesize
    \begin{tabularx}{\textwidth}{>{\hsize=.4\hsize\raggedright\arraybackslash}X >{\hsize=1.43\hsize}X >{\hsize=1.17\hsize}X}
        \toprule
        \textbf{} & \textbf{Original Text} & \textbf{Translation} \\
        \midrule
        \examplerow
        {19500323-bt-01-050-00140}
        {Dominant DSL subtype score: compassionate solidarity towards refugees,\newline\newline Party: WAV}
        {(March 23, 1950)}
        %(12) \textbf{Dominant DSL subtype score: compassionate solidarity towards refugees,\newline\newline Party: WAV}\newline{}(March 23, 1950) %19500323-bt-01-050-00140
        {\enquote{Die Aufnahme dieser Flüchtlinge wird in manchen Ländern bewußt erschwert, sabotiert und unmöglich gemacht. [...] \textbf{Seitdem die Spannung zwischen Osten und Westen zunimmt, hat ein neuer Strom von Flüchtlingen eingesetzt.} [...] Polen und die Tschechei [durchführen] die neue Ausweisungsaktion auf Befehl Moskaus, [...] um das Flüchtlingsheer in Westdeutschland zu vermehren und noch mehr Arbeitslose zu schaffen, den Staat zu belasten und seine Wirtschaft zu zerstören. [...]%Es ist deshalb unverständlich, daß die Besatzungsmächte gegen die Aufnahme dieser armen Menschen Stellung nehmen und sie verhindern wollen. 
        %Für uns als deutsche Volksvertreter ist es untragbar, daß man im Auftrage der Besatzungsmächte die Grenzen sperrt und die Transporte der Elenden und Bedrängten in der Sowjetzone stehen läßt.
        }}
        {\enquote{The admission of these refugees is being deliberately obstructed, sabotaged, and made impossible in some states. [...] \textbf{Since the tension between East and West has increased, a new stream of refugees has begun.} [...] Poland and Czechoslovakia are carrying out the new expulsions on orders from Moscow, [...] in order to enlarge the refugee army in West Germany, to create even more unemployed, to burden the state, and to destroy its economy. %It is therefore incomprehensible that the occupying powers should oppose the admission of these poor people and seek to prevent it. 
        [...]}} %For us as German representatives of the people, it is intolerable that, on the orders of the occupying powers, the borders are closed and the transports of the wretched and afflicted are left standing in the Soviet zone.} \\
        \midrule
        \examplerow
        {19491020-bt-01-012-01314}
        {Dominant DSL subtype score: group-based anti-solidarity towards expellees,\newline\newline Party: BP}
        {(Oct 20, 1949)}
        %(13) \textbf{Dominant DSL subtype score: group-based anti-solidarity towards expellees,\newline\newline Party: BP}\newline{}(Oct 20, 1949) %19491020-bt-01-012-01314
        {\enquote{Die Sudetendeutschen [...] und die Südostdeutschen [stehen] dem süddeutschen, also dem bayerischen Kulturkreis am nächsten. [...] Die aus dem Sudetenland bzw. aus Südosteuropa kommenden Deutschen [sollen] zunächst in Süddeutschland zusammen[gefasst werden], [...] damit die Eingliederung [...] schneller und reibungsloser vor sich gehen kann. %Wir wissen ja nicht, wie weit die Eingliederung vor sich gehen muß und ob sie nicht [...] eine ewige sein wird. 
        \textbf{[Wir] wollen [...] nicht Schwierigkeiten schaffen, wo Schwierigkeiten nicht überwunden werden müssen, und nicht Gräben aufreißen, wo solche nicht vorhanden sind.} Das hat mit Rassentheorie [...] gar nichts zu tun. [...] %sondern ist eine [...] Überlegung der Möglichkeit der Verschmelzung und der Eingliederung, die sich aus der Erkenntnis des Volkes und der Volksstämme selbst ergibt.
        }}
        {\enquote{The Sudeten Germans [...] and the Southeast Germans are closest to the South German, that is, the Bavarian cultural sphere. [...] The Germans coming from the Sudetenland and from Southeastern Europe [should] initially be concentrated in southern Germany, [...] so that integration [...] can proceed more quickly and smoothly. %We do not know how far integration must go, and whether it might not [...] be permanent. 
        \textbf{[...] we do not want to create difficulties where difficulties need not arise, and we do not want to open up rifts where none exist.} This has nothing to do with racial theory [...]. %but is a [...] consideration of the possibility of merging and integration, arising from the awareness of the people and the tribes themselves.
        }}
        \midrule
        \examplerow
        {19501213-bt-01-106-00457}
        {Dominant DSL subtype score: group-based solidarity towards refugees,\newline\newline Party: KPD}
        {(Dec 13, 1950)}
        %(14) \textbf{Dominant DSL subtype score: group-based solidarity towards refugees,\newline\newline Party: KPD}\newline{}(Dec 13, 1950) %19501213-bt-01-106-00457
        {\enquote{Wir glauben, daß das Problem der Beschaffung von Arbeit für [Umsiedler] unter dem kapitalistischen System nicht in der Form gelöst werden kann [...]. Wir sind also der Auffassung, daß die Umsiedlung eine Notwendigkeit ist [...]. \textbf{Wenn hier der Sprecher der CDU das schöne, stolze und richtige Wort geprägt hat \enquote{Unser Haus ist deutsch}, wenn er am Schluß gesagt hat \enquote{Das ganze Deutschland soll es sein}, wie kann man dann [...] gleichzeitig die Tatsache, daß sich die Vertriebenen, Flüchtlinge und Umsiedler, wie Sie sie bezeichnen, politisch zu Parteien zusammengeschlossen haben, als die Gefahr einer Überfremdung bezeichnen?}
        }}
        {\enquote{We believe that the problem of providing work for [resettlers] cannot be solved under the capitalist system [...]. We therefore take the view that resettlement is a necessity [...]. \textbf{If the CDU speaker here has coined the fine, proud, and correct phrase \enquote{Our house is German}, and if he concluded by saying \enquote{It shall be the whole of Germany}, how then can one at the same time describe the fact that the expellees, refugees, and resettlers, as you call them, have politically organized themselves into parties as a danger of being overrun by foreigners?}}}
        \midrule
        \examplerow
        {19510131-bt-01-115-00271}
        {Dominant DSL subtype score: exchange-based solidarity towards refugees \newline\newline Party: CDU/CSU}
        {(Jan 31, 1951)}
        %(15) \textbf{\ak{Predicted label: exchange-based solidarity towards refugees} \newline\newline Party: CDU/CSU}
        %\newline{}(Jan 31, 1951) %19510131-bt-01-115-00271
        {\enquote{[...] \textbf{Es ist Voraussetzung, dass das deutsche Volk für das System des Lastenausgleichs einen Weg findet, damit das, was aus der deutschen Produktion und Wirtschaftskraft herausgenommen werden muss, [...] dafür verwendet wird, die Kräfte der Kriegsgeschädigten, der Flüchtlinge und der Heimatvertriebenen wieder in die deutsche Produktion hineinzustellen und diese damit zu stärken.} [...] Was wir an Eingliederungshilfe [...], an Wohnraumhilfe [...] und an Förderungshilfe aller Art [...] verwenden, ist [...] produktive Ausgabe. [...] das ist rund die Hälfte des gesamten Aufkommens.
        }}
        {\enquote{[...] \textbf{It is a prerequisite that the German people find a way, within the system of burden sharing, to ensure that what must be taken from German production and economic strength is used [...] to restore the productive capacity of war victims, refugees, and expellees and to reintegrate them into German production, thereby strengthening it.} [...] What we allocate for integration assistance [...], for housing assistance [...], for all kinds of development assistance is [...] productive expenditure. [...] that is roughly half of the total revenue.}} 
        \bottomrule
    \end{tabularx}}
    \caption{Examples for postwar period discussed in \Cref{sec:postwar} (continued). Bold text is the main sentence, the other sentences are for context. The label shown is the instance's dominant DSL subtype, defined as the subtype with the highest DSL-adjusted score.}
    \label{tab:postwar-predictions-examples}
\end{table*}

\begin{table*}[t]
    \ContinuedFloat
    \centering
    {\footnotesize
    \begin{tabularx}{\textwidth}{>{\hsize=.4\hsize\raggedright\arraybackslash}X >{\hsize=1.43\hsize}X >{\hsize=1.17\hsize}X}
        \toprule
        \textbf{} & \textbf{Original Text} & \textbf{Translation} \\
        \midrule
        \examplerow
        {19530318-bt-01-254-02259}
        {Dominant DSL subtype score: group-based anti-solidarity towards expellees,\newline\newline Party: DP}
        {(March 18, 1953)}
        %(A) \textbf{Dominant DSL subtype score: group-based anti-solidarity towards expellees,\newline\newline Party: DP}\newline{}(March 18, 1953) %19530318-bt-01-254-02259
        {\enquote{Also auch der dritte, vierte und fünfte Sohn eines ostdeutschen Bauern [...] kann alle [...] Vergünstigungen erhalten. [...] Welcher einheimische Bauer kann [...] seinen drei oder vier Jungen eine solche Mitgift erarbeiten? Wir schaffen, wenn wir die alte Vorlage wiederherstellen, zweierlei Recht. \textbf{Das dürfte so viel böses Blut zur Folge haben, dass das auch nicht im Interesse der Vertriebenen liegen kann.} Herr Kollege Dr. Kather hat vorhin gesagt, dass 300 000 bäuerliche Familien aus dem Osten geflüchtet sind. Infolge der Formulierung des § 35 ist die Zahl der Bewerber um einen Betrieb weitaus größer geworden. Ich meine, wir sind unehrlich, wenn wir den Leuten Hoffnungen machen, die doch nicht zu erfüllen sind.
        }}
        {\enquote{Thus even the third, fourth, and fifth son of an East German farmer [...] can receive all the benefits. […] Which local farmer [...] can earn such a dowry for three or four of his sons? If we restore the old version of the bill, we create two kinds of law. \textbf{That would cause so much bad blood that it cannot possibly be in the interest of the expellees themselves.} [...] Dr. Kather said that 300,000 farming families fled from the East. Because of the wording of § 35, the number of applicants for a farm has become far greater. I believe we are being dishonest if we give people hopes that cannot in fact be fulfilled.}}
        \bottomrule
    \end{tabularx}}
    \caption{Examples for postwar period discussed in \Cref{sec:postwar} (continued). Bold text is the main sentence, the other sentences are for context. The label shown is the instance's dominant DSL subtype, defined as the subtype with the highest DSL-adjusted score.}
    \label{tab:postwar-predictions-examples}
\end{table*}

\begin{table*}[t]
    %\ContinuedFloat
    \centering
    {\footnotesize
    \begin{tabularx}{\textwidth}{>{\hsize=.5\hsize\raggedright\arraybackslash}X >{\hsize=1.3\hsize}X >{\hsize=1.2\hsize}X}
        \toprule
        \textbf{} & \textbf{Original Text} & \textbf{Translation} \\
        \midrule
        \examplerow
        {20141106-bt-18-063-04755}
        {Dominant DSL subtype score: compassionate solidarity towards refugees \newline\newline Party: Die Grünen}
        {(Nov 6, 2014)}
        %(15) \textbf{Dominant DSL subtype score: compassionate solidarity towards refugees \newline\newline Party: Die Grünen}
        %\newline{}(Nov 6, 2014) %20141106-bt-18-063-04755
        {\enquote{\textbf{Ich bin froh, dass es heute bei uns in Deutschland eine andere Willkommenskultur gibt und dass sich viele Menschen mit hohem Engagement für das Wohl der Flüchtlinge, die im Augenblick zu uns kommen, einsetzen.} [...] Ich will ganz klar sagen: Das Boot ist nicht voll [...]. Heute geht es um Grundrechte von Menschen, denen die Grundrechte in ihren Heimatländern versagt wurden [...].
        }}
        {\enquote{\textbf{I am glad that today in Germany we have a different welcoming culture, and that many people are working with great commitment for the well-being of the refugees who are currently coming to us.} [...] I want to state very clearly: the boat is not full [...]. Today it is about the fundamental rights of people who have been denied those rights in their home countries [...].}} 
        \midrule
        \examplerow
        {20151001-bt-18-127-00708}
        {Dominant DSL subtype score: group-based anti-solidarity towards refugees}
        {(Oct 1, 2015)}
        %(16) \textbf{Dominant DSL subtype score: group-based anti-solidarity towards refugees}\newline{}(Oct 1, 2015) %20151001-bt-18-127-00708
        {\enquote{[...] Deutschland [wird] über kurz oder lang überfordert sein. Was die Registrierung, Antragstellung und Antragsbearbeitung anbelangt, ist dies eine riesige Herausforderung, und die Unterbringung sowie die Integration in unsere Gesellschaft und in den Arbeitsmarkt sind [...] auf die Dauer nicht zu leisten [...]. \textbf{allein im September sind zehntausend Flüchtlinge und Asylbewerber nach Deutschland gekommen.} Sie wurden nicht registriert. Sie wurden nicht kontrolliert. [...] Damit besteht auch ein Sicherheitsrisiko.
        }}
        {\enquote{[...] Germany will sooner or later be overburdened. Registration, application processing and decision-making are a huge challenge, and accommodation as well as integration into our society and the labor market will [...] not be sustainable in the long run [...]. \textbf{In September alone, several tens of thousands of refugees and asylum seekers came to Germany.} They were not registered. They were not controlled. [...] This poses a major security risk.}}
        \bottomrule
    \end{tabularx}}
    \caption{Examples for the 2009-2025 period discussed in \Cref{sec:recent}. Bold text is the main sentence, the other sentences are for context. The label shown is the instance's dominant DSL subtype, defined as the subtype with the highest DSL-adjusted score.}
    \label{tab:recent-predictions-examples}
\end{table*}

\begin{table*}[t]
    \ContinuedFloat
    \centering
    {\footnotesize
    \begin{tabularx}{\textwidth}{>{\hsize=.5\hsize\raggedright\arraybackslash}X >{\hsize=1.35\hsize}X >{\hsize=1.15\hsize}X}
        \toprule
        \textbf{} & \textbf{Original Text} & \textbf{Translation} \\
        \midrule
        \examplerow
        {20180516-bt-19-032-01219}
        {Dominant DSL subtype score: exchange-based anti-solidarity towards refugees\newline\newline Party: AfD}
        {(May 16, 2018)}
        %(17) \textbf{Dominant DSL subtype score: exchange-based anti-solidarity towards refugees\newline\newline Party: AfD}\newline{}(May 16, 2018) %20180516-bt-19-032-01219, AfD
        {\enquote{\textbf{[...] jährlich [verlassen] etwa 140 000 Hochkompetente [...] dieses Land und im Gegenzug [...] um die 200 000 sogenannte Flüchtlinge hereinkommen, von denen nicht nur kaum einer hochkompetent ist, sondern viele Analphabeten sind.} In Deutschland lebt bereits eine große soziale Unterschicht [...]. Diese Zahl wächst seit 2015 durch die Zuwanderung von Gering- und Unqualifizierten rasant. [...] Deutschland ist kein reiches Land, sondern ein leistungsfähiges.
        }}
        {\enquote{\textbf{[...] each year about 140,000 highly qualified people leave this country, while in return, [...] about 200,000 so-called refugees enter, of whom hardly any are highly qualified, but many are illiterate.} In Germany, there already lives a large social underclass [...]. Since 2015, this number has been growing rapidly through the immigration of low- and unqualified people. [...] Germany is not a rich country, but a capable one.}}
        \midrule
        \examplerow
        {20221019-bt-20-062-02988}
        {Dominant DSL subtype score: group-based anti-solidarity towards refugees\newline\newline Party: AfD}
        {(Oct 19, 2022)}
        %(18) \textbf{Dominant DSL subtype score: group-based anti-solidarity towards refugees\newline\newline Party: AfD}\newline{}(Oct 19, 2022) %20221019-bt-20-062-02988, AfD
        {\enquote{[...] Echte Kriegsflüchtlinge aus der Ukraine brauchen Schutz – keine Frage. Wir sagen aber Nein zu Fluchttourismus durch halb Europa, und wir sagen Nein zu bevorzugter Behandlung auf Kosten der deutschen Steuerzahler. Wir fordern daher die sofortige Abschaltung dieses Sozialstaatsmagneten, [...] \textbf{Wir fordern außerdem die Umstellung von Geldleistungen auf Sachleistungen für Asylbewerber, damit Migrationsanreize gesenkt werden und der Missbrauch effektiv unterbunden wird.} [...] Und wer sich wie Frau Baerbock in erster Linie der ukrainischen Bevölkerung verpflichtet fühlt, der soll sich in der Ukraine um ein Amt bewerben, aber nicht in Deutschland.
        }}
        {\enquote{[...] Genuine war refugees from Ukraine need protection — no question. But we say no to flight tourism across half of Europe, and we say no to preferential treatment at the expense of German taxpayers. We therefore demand the immediate shutdown of this welfare magnet, [...] \textbf{We also demand the switch from cash benefits to benefits in kind for asylum seekers, so that incentives for migration are reduced and abuse is effectively curbed.} [...] And whoever, like Ms. Baerbock, feels primarily obligated to the Ukrainian population should run for office in Ukraine, but not in Germany.}}
        \midrule
        \examplerow
        {20230512-bt-20-104-02937}
        {Dominant DSL subtype score: group-based anti-solidarity towards refugees\newline\newline Party: AfD}
        {(12 May, 2023)}
        %(19) \textbf{Dominant DSL subtype score: group-based anti-solidarity towards refugees\newline\newline Party: AfD}\newline{}(12 May, 2023) %20230512-bt-20-104-02937
        {\enquote{[...] alte Leute sollen lieber aus ihren großen Wohnungen ausziehen, weil grün-linke Politik Millionen Gäste aus Afrika und dem Nahen Osten einlädt und gleichzeitig den Wohnungsbau abwürgt. [...] \textbf{Dafür werden bereits Zimmer in angeblich christlichen Altenheimen gekündigt, weil angebliche Flüchtlinge lukrativer sind, Messerzuwanderung inklusive.} [...] Bei diesem Recht auf Wohnungstausch der Linken handelt es sich um eine Mischung aus Neid zulasten alter Leute und Zwang zulasten von Vermietern. Es ist die Fortsetzung einer inländerfeindlichen...
        }}
        {\enquote{[...] elderly people are expected to leave their large apartments because green-left politics invites millions of guests from Africa and the Middle East while choking off housing construction. [...] \textbf{Rooms in supposedly Christian nursing homes are already being canceled because so-called refugees are more lucrative, knife migration included.} [...] This \enquote{right to apartment swaps} proposed by the Left is a mixture of envy at the expense of the elderly and coercion at the expense of landlords. It is the continuation of an anti-native [policy]…}}
        \midrule
        \examplerow
        {20230525-bt-20-106-05636}
        {Dominant DSL subtype score: exchange-based solidarity towards immigrants\newline\newline Party: FDP}
        {(May 25, 2023)}
        %(20) \textbf{Dominant DSL subtype score: exchange-based solidarity towards immigrants\newline\newline Party: FDP}\newline{}(May 25, 2023) %20230525-bt-20-106-05636
        {\enquote{[...] Menschen, die bei uns arbeiten wollen, sollten wir eine echte Chance geben; denn viele, die an unserer Grenze ankommen, wollen ja gerade arbeiten. Wir arbeiten in der Ampel an der Reform des Fachkräfteeinwanderungsgesetzes. \textbf{Da geht es um mehr Regulierung, gezielte Einwanderung in unseren Arbeitsmarkt.} Menschen aus dem Ausland sollten eine realistische Chance bekommen, zu uns zu kommen und arbeiten zu können. Das Fachkräfteeinwanderungsgesetz wird das jetzt endlich besser machen und mehr Menschen die Chance geben, die unsere Wirtschaft auch dringend braucht. [...]
        }}
        {\enquote{[...] People who want to work here should be given a real chance; because many who arrive at our border actually want to work. Within the coalition we are working on reforming the Skilled Immigration Act. \textbf{This is about more regulation and targeted immigration into our labor market.} People from abroad should have a realistic chance to come to us and be able to work. The Skilled Immigration Act will now finally improve this and give more people the opportunity that our economy urgently needs. [...]}}
        \bottomrule
    \end{tabularx}}
    \caption{Examples for the 2009-2025 period discussed in \Cref{sec:recent} (continued). Bold text is the main sentence, the other sentences are for context. The label shown is the instance's dominant DSL subtype, defined as the subtype with the highest DSL-adjusted score.}
    \label{tab:recent-predictions-examples}
\end{table*}

\begin{table*}[t]
    \ContinuedFloat
    \centering
    {\footnotesize
    \begin{tabularx}{\textwidth}{>{\hsize=.5\hsize\raggedright\arraybackslash}X >{\hsize=1.3\hsize}X >{\hsize=1.2\hsize}X}
        \toprule
        \textbf{} & \textbf{Original Text} & \textbf{Translation} \\
        \midrule
        \examplerow
        {20250514-bt-21-003-01509}
        {Dominant DSL subtype score: group-based anti-solidarity towards migrants\newline\newline Party: CDU/CSU}
        {(May 14, 2025)}
        %(21) \textbf{Dominant DSL subtype score: group-based anti-solidarity towards migrants\newline\newline Party: CDU/CSU}\newline{}(May 14, 2025) %20250514-bt-21-003-01509
        {\enquote{%Vertrauen durch Verantwortung: Das ist das gemeinsame Ziel.
        [...] Die gute Nachricht ist: Der Politikwechsel hat begonnen. %Er hat auch begonnen, ja, an den deutschen Grenzen. 
        [...] \textbf{Die illegale Migration der letzten zehn Jahre gefährdet die politische Stabilität Deutschlands und Europas.} %Wer mit Bürgermeistern und Landräten spricht, egal von welcher Partei, weiß: 
        [...] Unsere Städte und Gemeinden, die Landkreise: Alle sind über dem Limit. In den Kitas, in den Schulen, auf dem Wohnungsmarkt, bei der Sicherheit an den Bahnhöfen und auf den Marktplätzen wird diese Überforderung sehr konkret. Deswegen ist es gut, dass wir gemeinsam im Koalitionsvertrag die Migrationswende vereinbart haben: mehr Rückführungen, schnellere Verfahren und, ja, ebenso die Zurückweisung an den deutschen Grenzen auch bei Asylgesuchen.
        }}
        {\enquote{%Trust through responsibility: That is the common goal. 
        [...] The good news is: the policy shift has begun. %It has also begun, yes, at the German borders. 
        [...] \textbf{The illegal migration of the last ten years endangers the political stability of Germany and Europe.} [...] %Anyone who speaks with mayors and district administrators, regardless of party, knows: 
        Our towns and municipalities, the districts—everyone is over the limit. In day-care centers, in schools, in the housing market, in security at train stations and in marketplaces, this overburdening becomes very concrete. That is why it is good that we have agreed together in the coalition agreement on a migration turnaround: more deportations, faster procedures, and, yes, also rejections at the German borders, even in the case of asylum applications.}}
        \midrule
        \examplerow
        {20250130-bt-20-210-02826}
        {Dominant DSL subtype score: exchange-based anti-solidarity towards immigrants\newline\newline Party: FDP}
        {(Jan 30, 2025)}
        %(22) \textbf{Dominant DSL subtype score: exchange-based anti-solidarity towards immigrants\newline\newline Party: FDP}\newline{}(Jan 30, 2025) %20250130-bt-20-210-02826
        {\enquote{Das Grundrecht auf Asyl darf nicht dazu führen, dass allein ein Antrag auf Asyl zu mehreren Jahren freiem und staatlich subventioniertem Aufenthalt in Deutschland führt, wenn er völlig ohne Grundlage erfolgt. Solche Missstände müssen wir beseitigen, damit wir uns auf jene konzentrieren können, die unsere Hilfe wirklich brauchen [...]. \textbf{SPD und Grüne [...] wollen die illegale Einwanderung nach Deutschland kaum begrenzen und fordern lieber weiteren Familiennachzug.} Diesen falschen Weg ist Deutschland leider viel zu lange gegangen. [...] Eine immer linkere Politik wird [...] dazu führen, dass die AfD ihr teuflisches Werk immer weiter fortsetzen kann. [...]
        }}
        {\enquote{The fundamental right to asylum must not result in a situation where merely submitting an asylum application leads to several years of free and state-subsidized residence in Germany when it is entirely unfounded. Such abuses must be eliminated so that we can focus on those who truly need our help [...]. \textbf{The SPD and the Greens [...] hardly want to limit illegal immigration to Germany and instead prefer to demand further family reunification.} Unfortunately, Germany has followed this wrong path for far too long. [...] Ever more left-leaning policies [...] will only enable the AfD to continue its diabolical work [...].}}
        \midrule
        \examplerow
        {20240912-bt-20-185-00868}
        {Dominant DSL subtype score: group-based anti-solidarity towards migrants\newline\newline Party: SPD}
        {(12 Sept, 2024)}
        %(24) \textbf{Dominant DSL subtype score: group-based anti-solidarity towards migrants\newline\newline Party: SPD}\newline{}(12 Sept, 2024) %20240912-bt-20-185-00868
        {\enquote{Das Sicherheitspaket, das heute hier zur Debatte steht, ist der erste Schritt, um Konsequenzen aus der furchtbaren Katastrophe in Solingen zu ziehen. Waffenverbotszonen [...] helfen [aber] Waffenverbotszonen nicht. \textbf{Deshalb brauchen wir mehr Befugnisse für das Bundeskriminalamt und das Bundesamt für Migration.} Sie müssen in die Lage versetzt werden, mit den öffentlich zugänglichen Informationen Straftaten zu verhindern. Das ist dringend erforderlich, und deshalb werden wir das jetzt umsetzen. Wir werden weitere Schritte veranlassen, damit ausreisepflichtige Personen unser Land verlassen.
        }}
        {\enquote{The security package being debated here today is the first step toward drawing consequences from the terrible catastrophe in Solingen. Weapons prohibition zones [...] do not, however, help against terrorists. \textbf{That is why we need more powers for the Federal Criminal Police Office and the Federal Office for Migration.} They must be put in a position to use publicly available information to prevent crimes. That is urgently necessary, and that is why we will now implement it. We will take further steps to ensure that persons obliged to leave the country actually leave our country.}}      
        \bottomrule
    \end{tabularx}}
    \caption{Examples for the 2009-2025 period discussed in \Cref{sec:recent} (continued). Bold text is the main sentence, the other sentences are for context. The label shown is the instance's dominant DSL subtype, defined as the subtype with the highest DSL-adjusted score.}
    \label{tab:recent-predictions-examples}
\end{table*}

\begin{table*}[t]
    \ContinuedFloat
    \centering
    {\footnotesize
    \begin{tabularx}{\textwidth}{>{\hsize=.5\hsize\raggedright\arraybackslash}X >{\hsize=1.3\hsize}X >{\hsize=1.2\hsize}X}
        \toprule
        \textbf{} & \textbf{Original Text} & \textbf{Translation} \\
        \midrule
        \examplerow
        {20240119-bt-20-148-01053}
        {Dominant DSL subtype score: exchange-based solidarity towards immigrants\newline\newline Party: Die Grünen}
        {(Jan 19, 2024)}
        %(26) \textbf{Dominant DSL subtype score: exchange-based solidarity towards immigrants\newline\newline Party: Die Grünen}\newline{}(Jan 19, 2024) %20240119-bt-20-148-01053
        {\enquote{Wissenschaftliche Institute sprechen von 400 000 Menschen jedes Jahr. Darauf müssen wir die Gesellschaft und die Politik vorbereiten. Wir müssen also ohnehin die entsprechende Infrastruktur ausbauen – Wohnungen, Schulen, Kinderbetreuung. \textbf{Wir brauchen gezielte Zuwanderung, und dafür haben wir das Fachkräfteeinwanderungsgesetz gemacht.} Aber wir wären doch mit dem Klammerbeutel gepudert, wenn wir nicht auch das Potenzial der Menschen, die aus humanitären Gründen zu uns kommen, nutzten und dafür sorgten, dass sie aus- und weitergebildet werden und Teil unserer Gesellschaft werden. Das wäre die Hauptaufgabe. Und weil wir Zuwanderung brauchen, brauchen wir eine Willkommenskultur.
        }}
        {\enquote{Scientific institutes speak of 400,000 people every year. Society and politics must be prepared for that. We therefore have to expand the relevant infrastructure anyway — housing, schools, childcare. \textbf{We need targeted immigration, and for that we created the Skilled Workers Immigration Act.} But we would have to be out of our minds not to also make use of the potential of people who come to us for humanitarian reasons, and ensure that they receive education and training and become part of our society. That would be the main task. And because we need immigration, we need a culture of welcome.}}
        \midrule
        \examplerow
        {20250130-bt-20-210-02938}
        {Dominant DSL subtype score: compassionate solidarity towards migrants\newline\newline Party: Die Linke}
        {(Jan 30, 2025)}
        %(27) \textbf{Dominant DSL subtype score: \ak{compassionate} solidarity towards migrants\newline\newline Party: Die Linke}\newline{}(Jan 30, 2025) %20250130-bt-20-210-02938
        {\enquote{Sie sagen, Sie wollen Frauen und Kinder schützen. Sie wollen Kinder und Frauen in Kriegsgebiete abschieben. So ist es. \textbf{Wir führen eine Debatte über Sicherheit und Migration.} Und leider erklären alle Parteien außer der Linken, dass sie mit Abschottung dieses Land sicherer machen wollen, und reden von \enquote{Abschiebungen im großen Stil}, als seien Migranten grundsätzlich eine Gefahr. So werden Menschen unter Generalverdacht gestellt. Weniger Migration mache das Land sicherer, das ist doch Unsinn.
        }}
        {\enquote{You say you want to protect women and children. You want to deport children and women to war zones. That is the reality. \textbf{We are having a debate about security and migration.} And unfortunately, all parties except the Left declare that they want to make this country safer through isolation, and talk about \enquote{deportations on a large scale,} as if migrants were inherently a danger. In this way, people are placed under general suspicion. Less migration makes the country safer — that is nonsense.}}
        \midrule
        \examplerow
        {20160219-bt-18-156-00811}
        {Dominant DSL subtype score: compassionate anti-solidarity towards asylum seekers\newline\newline Party: CDU/CSU}
        {(19 Feb 2016)}
        %(28) \textbf{Dominant DSL subtype score: \ak{compassionate anti-solidarity} towards asylum seekers\newline\newline Party: CDU/CSU}\newline{}(19 Feb 2016) %20160219-bt-18-156-00811
        {\enquote{Das sind aber nicht nur Flüchtlinge, die Anspruch auf Schutz haben. [...] bei mehr als der Hälfte [konnte] keine Schutzbedürftigkeit festgestellt werden. \textbf{Wenn wir bei Asylbewerbern, bei denen schon bei der Ankunft absehbar ist, dass sie keine Bleibeaussicht haben, nicht in kürzester Zeit entscheiden und sie in ihre Herkunftsländer zurückführen, werden wir noch mehr Anreize schaffen, ohne berechtigten Grund nach Deutschland zu kommen.} [...] Das ist nicht nur ein klares Signal im Sinne von \enquote{Wir nehmen nur wirklich Schutzbedürftige auf}, sondern entlastet auch deutlich unsere Kommunen bei der Unterbringung. Man kann dabei auch nicht von einer Einschränkung oder Aushöhlung des Asylrechts sprechen.
        }}
        {\enquote{These are not only refugees who are entitled to protection. [...] in more than half of the cases no need for protection could be established. \textbf{If, in the case of asylum seekers for whom it is already apparent upon arrival that they have no prospect of remaining, we do not make a decision as quickly as possible and return them to their countries of origin, we will create even more incentives to come to Germany without justified reason.} [...] This is not only a clear signal in the sense of \enquote{We only take in those who genuinely need protection,} but it also significantly relieves our municipalities with regard to accommodation. Nor can one speak of a restriction or hollowing-out of the right to asylum in this context.}}
        \bottomrule
    \end{tabularx}}
    \caption{Examples for the 2009-2025 period discussed in \Cref{sec:recent} (continued). Bold text is the main sentence, the other sentences are for context. The label shown is the instance's dominant DSL subtype, defined as the subtype with the highest DSL-adjusted score.}
    \label{tab:recent-predictions-examples}
\end{table*}

\clearpage
\section{Annotation Process Details}
\label{app:annotation-details}

Annotators first assigned a high-level stance label and, where applicable, an (anti-)solidarity subtype. To document their decisions, they also identified the specific resource at issue (e.g.\, time, money, rights, or access to education), selected an indicator associated with the chosen category, highlighted supporting text spans, and provided a brief free-text explanation (1-2 sentences). Where relevant, annotators also marked expressions of the speaker's own position and references to others' positions. These additional annotations supported interpretation and documentation during the annotation process, but were not used in the analyses reported in this paper. The full annotation guidelines are provided in \Cref{app:annotation-guidelines} in the Appendix, and \Cref{fig:annotation-process} shows an example from the annotation file.

\begin{figure}[H]
    \centering

    \begin{subfigure}[t]{\linewidth}
        \centering
        \includegraphics[width=0.9\linewidth]{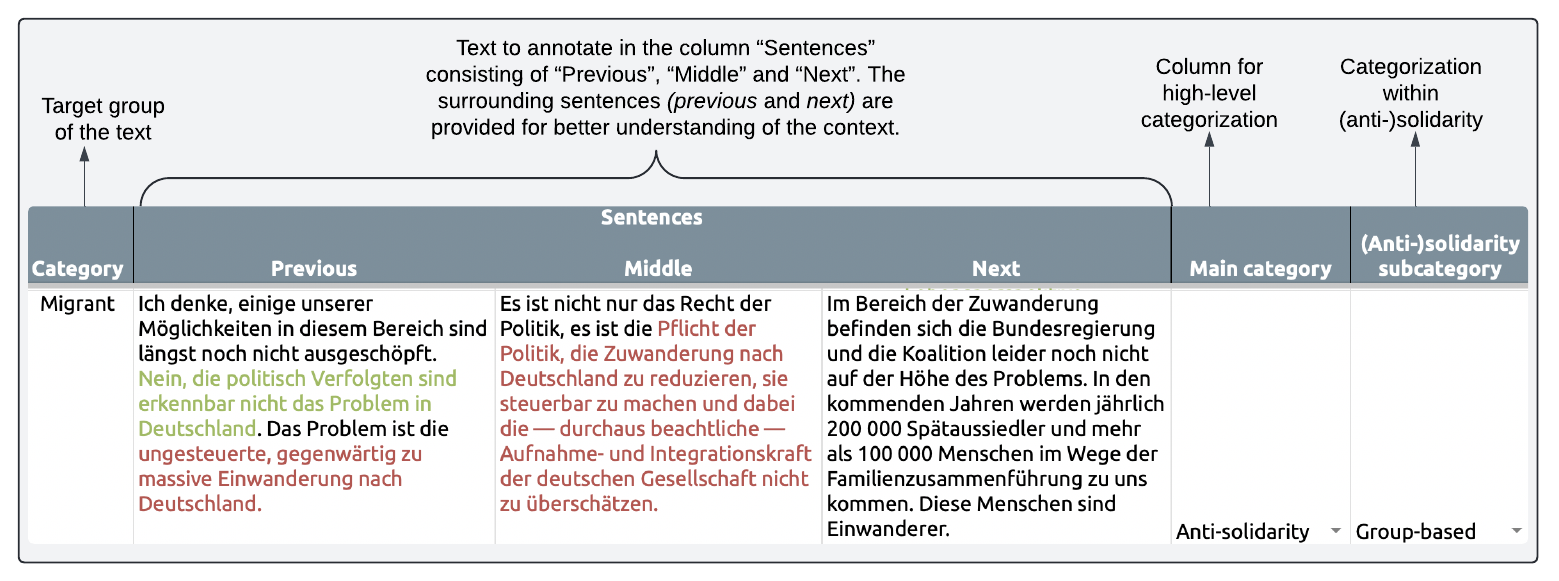}
        \caption{Columns for high-level and (anti-)solidarity categorizations.}
        \label{fig:columns-categorization}
    \end{subfigure}

    \vspace{0.8em}

    \begin{subfigure}[t]{0.9\linewidth}
        \centering
        \includegraphics[width=\linewidth]{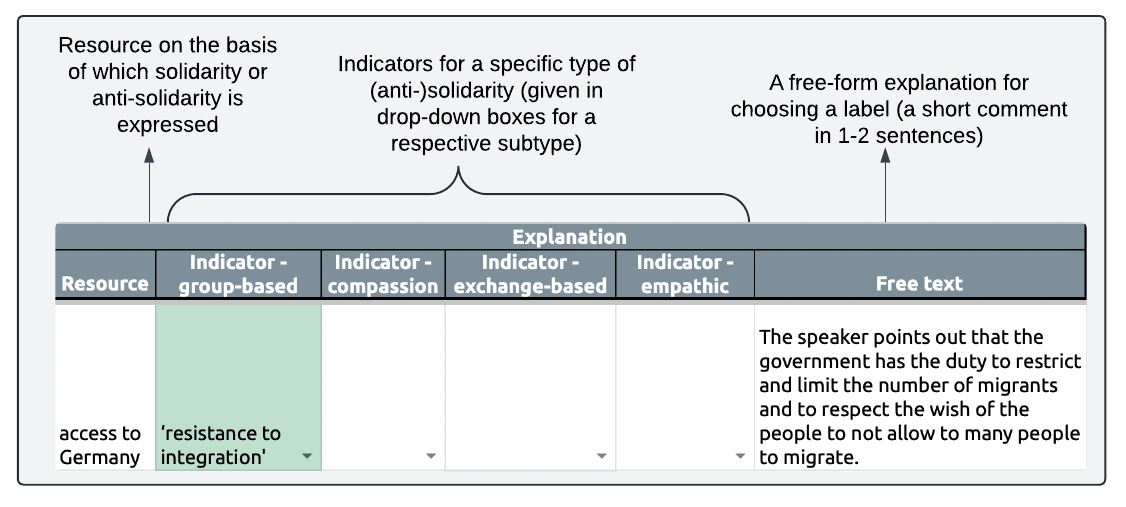}
        \caption{Columns for providing explanations.}
        \label{fig:columns-explanations}
    \end{subfigure}

    \caption{
    \ak{Example of the annotation process from the annotation file. \Cref{fig:columns-categorization} illustrates the step where annotators choose a high-level label and an (anti-)solidarity subcategory, if applicable. \Cref{fig:columns-explanations} shows columns for detailed explanations, including the choice of a resource, an indicator, and providing free-text commentary}
    }
    \label{fig:annotation-process}
\end{figure}

\clearpage
\section{Annotation Guidelines Excerpt}\label{app:annotation-guidelines}

\begin{guidelinebox}
This appendix reproduces the annotation guidelines used by the annotators, with minor formatting edits.

\subsection*{General Information}

\paragraph{Goal of the study and annotation}
The purpose of this study is to examine expressions of solidarity and anti-solidarity towards two specific vulnerable groups---women and migrants---within German parliamentary proceedings. The definitions of (anti-)solidarity are derived from the work \citet{TwitterDataset}, and \citet{eger2022measuring}. We distinguish four subtypes of solidarity and anti-solidarity based on \citet{thijssen2022s}: \emph{group-based} (shared identity or exclusion), \emph{compassionate} (protection or dismissal of vulnerable groups), \emph{exchange-based} (contribution, reciprocity, or burden), and \emph{empathic} (respect for or rejection of diversity).

\subsection*{Details of the Annotation Process}

\paragraph{Annotation instructions}
In the next section (\textit{Annotation Categories}), we provide general theoretical information about the categories used in the annotation. This material is intended to support a better understanding of the concepts underlying the task.

It is critical that annotators provide a clear explanation for the chosen category or categories in the \emph{Explanation} column, while keeping comments short. The \emph{Other comments} column can be used for any additional observations.

\paragraph{Contextual evaluation}
Pay close attention to the middle text, which is the sentence whose stance is to be assessed. The surrounding sentences (previous and next) are provided to support contextual interpretation.

\paragraph{Highlighting sentences}
When labeling the instances, annotators highlight parts of or full sentences that guided the label choice. Green is used for solidarity categories and red for anti-solidarity categories. In cases of mixed stance, solidarity-expressing parts are highlighted in green and anti-solidarity-expressing parts in red.

\paragraph{Pay attention to the target group}
If the target group in the \emph{Category} column is \emph{women}, but the instance is about \emph{migrants}, the instance should be marked as \emph{None}. For example, if the target group is \emph{women}, then a sentence such as \enquote{all refugees must now leave this country} should also be marked as \emph{None}.

%\paragraph{Feedback, suggestions, and questions.}
%Annotators are encouraged to provide feedback, suggestions, or questions at any stage of the process. This feedback may concern the clarity of the guidelines, the applicability of the labels, the handling of multi-label annotations, or any other aspect of the annotation process.

\subsection*{Structure of the Annotation File}

The annotation file contains the following fields:

\begin{itemize}
    \item \textbf{ID:} ID of the sentence. The first four numbers after the word \texttt{debates} indicate the year.
    
    \item \textbf{Category:} Indicates the target group (\emph{woman} or \emph{migrant}). Pay close attention to the target group. If the target group in the \emph{Category} column is \emph{women}, but the instance is about \emph{migrants}, the instance should be marked as \emph{None}.
    
    \item \textbf{Sentences:} The text to annotate, consisting of \emph{Previous}, \emph{Middle}, and \emph{Next}. The surrounding sentences are provided for contextual interpretation. However, the central focus of the decision should be the middle text.
    
    \item \textbf{Main category:} One of \emph{Solidarity}, \emph{Anti-solidarity}, \emph{Mixed stance}, or \emph{None}.
    
    \item \textbf{(Anti-)solidarity subcategory:} One of the subcategories, selected only for instances labeled as \emph{Solidarity} or \emph{Anti-solidarity}.
    
    \item \textbf{Explanation:}
    \begin{itemize}
        \item \textbf{Resource:} The resource on the basis of which solidarity or anti-solidarity is expressed (e.g., access to Germany, rights and well-being).
        \item \textbf{Indicators:} An indicator motivating the selected anti-solidarity or solidarity type. Indicators are given in drop-down boxes for the respective (anti-)solidarity type and are also listed in the \emph{Annotation Categories} section.
        \item \textbf{Free text:} A short explanation for the chosen label (1--2 sentences). Annotators should provide a clear explanation for the selected category.
    \end{itemize}
    
    \item \textbf{Other comments:} Any additional observations.
\end{itemize}

\subsection*{Annotation Categories}\label{app:annotation-categories}

Definition of solidarity (based on \citet{TwitterDataset}): solidarity is the readiness to share one’s own resources with others (be it time, financial means, rights, access to education etc.). 

\begin{center}
\textbf{\large 1. Solidarity}
\end{center}

\begin{center}
\textbf{\solidarity{Group-based solidarity}}
\end{center}

\textbf{Definition.} Group-based solidarity stems from a perceived similarity among members of the same group, leading to a shared identity. The group-based solidarity frames often refer to a certain (desired) commonality and a sense of togetherness due to common interests and goals, shared values and norms, or common rights and duties.

\textbf{Characteristics.} To identify group-based solidarity, look for:
\begin{enumerate}
    \item References to a shared identity, common interests, goals, values, or norms within a group (e.g.\ \textit{It's our collective responsibility to...; we stand together in the fight...; We must unite to...}).
    \item The existence of mutual obligations or shared rights and duties.
    \item Indication that the group must be integrated or even assimilated.
    \item Possibly, the identification of an outgroup that contrasts with the in-group.
\end{enumerate}

Emphasis on unity and shared goals/duties/rights among the whole group.

\begin{center}
\textbf{\solidarity{Compassionate solidarity}}
\end{center}

\textbf{Definition.} Compassionate solidarity is primarily about recognising and supporting individuals or subgroups who are marginalised, disadvantaged, or vulnerable. It extends assistance to these individuals or subgroups, emphasising their need for protection or help.

\textbf{Characteristics.} To identify compassionate solidarity, look for:
\begin{enumerate}
    \item Acknowledgement of the challenges, hardships, or vulnerabilities faced by individuals or subgroups (e.g.\ \textit{We must pay special attention to the needs...; We need to recognize the difficulties...}).
    \item Emphasis on the need for their protection, assistance, or special consideration due to their marginalised or disadvantaged status (e.g.\ \textit{It is our duty to support...}).
\end{enumerate}

Emphasis on extra care, support, and protection for those most vulnerable (both in-group and out-group).

\begin{center}
\textbf{\solidarity{Exchange-based solidarity}}
\end{center}

\textbf{Definition.} Exchange-based solidarity is linked to the principle of the role of exchange between partners with complementary qualities. Exchange-based solidarity is coded when a speaker refers to the usefulness of `exchange partners' in terms of their actual or future contributions or willingness to contribute. These exchange partners can be rewarded or stimulated, and may also be required to contribute more to receive support.

\textbf{Characteristics.} To identify exchange-based solidarity, look for references to a group:
\begin{enumerate}
    \item That was actively doing, is doing and / or will be doing something useful.
    \item Should either be rewarded for past, present or future behavior or that it should be supported to continue that useful behavior.
    \item Does not receive enough and needs more support for its contribution (do not confuse it with compassionate solidarity).
\end{enumerate}

\begin{center}
\textbf{\solidarity{Empathic solidarity}}
\end{center}

\textbf{Definition.} Empathic solidarity is a concept grounded in the intersubjective valuation of differences among individuals, recognising and respecting their unique characteristics. In contrast to compassionate solidarity, empathic solidarity is coded when a difference is not something that is bad or something that should disappear but is rather a strength or something positive. Empathic solidarity is coded when a speaker acknowledges diversity, difference, or unique characteristics as aspects that should be respected and taken into account.

\textbf{Characteristics.} To identify empathic solidarity, look for:
\begin{enumerate}
    \item Acknowledgement of a group's diversity and that this diversity must be respected;
    \item Recognition of the group's unique and distinct characteristics;
    \item Affirmation of the group's right to self-expression, which can relate to their specific needs, traits, views, or belief choices.
    \item Advocacy for the group's freedom to develop themselves the way they want.
\end{enumerate}

\vspace{2pt}

\begin{center}
\textbf{\large 2. Anti-solidarity}
\end{center}

\begin{center}
\textbf{\antisolidarity{Group-based anti-solidarity}}
\end{center}

\textbf{Definition.} The group-based anti-solidarity often points to an out-group characterised by pronounced differences (concerning interests and goals, values and norms, individual rights and responsibilities), and an expression of exclusion and lack of support for that group based on these differences.

\textbf{Characteristics.} To identify group-based anti-solidarity, look for:
\begin{enumerate}
    \item References to differing interests, goals, responsibilities, values, or norms between the groups and at the same time not willing to provide support based on the differences, sense of belonging to different groups.
    \item A stronger emphasis on one's own group's interests (e.g.\ \textit{It's every person for themselves...; We must stand up for our own interests...; We must not sacrifice our personal goals for...}).
    \item Resistance to assimilation or integration with other groups.
\end{enumerate}

\begin{center}
\textbf{\antisolidarity{Compassionate anti-solidarity}}
\end{center}

\textbf{Definition.} Compassionate anti-solidarity is about disregarding and excluding individuals or subgroups either on the basis that they are perceived to be in a ``good position'' already, or are viewed as undeserving of help. It neglects these individuals or subgroups, downplaying their need for protection or help.

\textbf{Characteristics.} To identify compassionate anti-solidarity, look for:
\begin{enumerate}
    \item Denial of the challenges, hardships, or vulnerabilities faced by individuals or subgroups (e.g.\ \textit{We shouldn't focus on the problems of...; We don't need to pay attention to the struggles of...}).
    \item Minimising the need for their protection, assistance, or special consideration (e.g.\ \textit{It is not our responsibility to aid...}) as they are already in a \enquote{good position} or are in a \enquote{bad position}, but still do not deserve help.
\end{enumerate}

\begin{center}
\textbf{\antisolidarity{Exchange-based anti-solidarity}}
\end{center}

\textbf{Definition.} Exchange-based anti-solidarity focuses on the idea that some groups take more than they give. It brings attention to groups that seem to receive a lot of resources or support, but do not contribute much or cause problems. This perspective supports punishing such groups, giving them less help, or making them contribute more.

\textbf{Characteristics.} To identify exchange-based anti-solidarity, look for references to a group:
\begin{enumerate}
    \item Talk about a group that has received a lot, but has not given much back or has caused harm.
    \item Suggestions that this group should be punished, or get less than they are currently getting.
    \item Calls for the group to give or do more in return for help.
\end{enumerate}

\begin{center}
\textbf{\antisolidarity{Empathic anti-solidarity}}
\end{center}

\textbf{Definition.} Empathic anti-solidarity refers to the cases when a speaker refers to a group indicating that this group is different from everyone else, but this difference (needs, characteristics, beliefs, opinions) should not be respected or recognised. The speaker could argue that this difference should not be acknowledged. The core logic behind this is: \enquote{You're not like me, and that's a shame. I don't understand or respect you or your differences}.

\textbf{Characteristics.} To identify empathic anti-solidarity, look for:
\begin{enumerate}
    \item Disregard for a group's diversity, arguing that this diversity should not be respected.
    \item Ignorance or dismissal of the group's unique and distinct characteristics.
    \item Denial of the group's right to self-expression, which can relate to their specific needs, traits, views, or belief choices.
\end{enumerate}

\begin{center}
\textbf{\large \mixed{3. Mixed stance}}
\end{center}

\textbf{Definition.} Mixed stance refers to instances where the speaker expresses both support and opposition towards the target group within the same discourse or text. It involves contradictions or conflicting statements regarding the speaker's stance towards the target group.

\textbf{Characteristics.}
\begin{enumerate}
    \item Annotate instances where the speaker presents a mixed stance by expressing both support and opposition towards the target group in the same discourse or text.
    \item Look for contradictions or inconsistencies in the level of support or opposition expressed towards the target group.
    \item Note that the middle sentence has more weight.
\end{enumerate}

\begin{center}
\textbf{\large 4. None}
\end{center}

\textbf{Definition.} None refers to instances where the speaker neither expresses explicit support nor opposition towards the target group. The statement is unrelated to either solidarity or anti-solidarity or is unrelated to the group altogether.

\textbf{Characteristics.}
\begin{enumerate}
    \item Annotate instances where the speaker maintains a neutral position and does not express clear support or opposition towards the target group.
    \item Look for statements that provide factual information, or refrain from taking a stance or expressing personal opinion.
    \item Also label the instance as None if you think that it is unrelated to the target group.
\end{enumerate}

\end{guidelinebox}

\end{document}